\documentclass[journal]{IEEEtran}
\usepackage{cite}
\usepackage{amsmath,amssymb,amsfonts}
\usepackage{algorithmic}
\usepackage{graphicx}
\usepackage{textcomp}
\usepackage{xcolor}
\usepackage{subfig}
\usepackage{bm}
\usepackage{pifont}
\usepackage{booktabs}
\usepackage{pifont}
\usepackage{multirow}
\usepackage{pdfpages}
\allowdisplaybreaks[4]

\usepackage[pagebackref=true,breaklinks=true,colorlinks,bookmarks=false,citecolor=green]{hyperref}

\usepackage[font=small]{caption}

\begin{document}


\title{Night-Voyager: Consistent and Efficient Nocturnal Vision-Aided State Estimation in Object Maps}

\author{Tianxiao Gao$^{*}$, Mingle Zhao$^{*}$, Chengzhong Xu, and Hui Kong
\thanks{Code and files: \url{https://github.com/IMRL/Night-Voyager}} 
\thanks{Supplementary video: \url{https://youtu.be/x7XtC_ALz80}}
} 


\maketitle 


\begin{abstract}
Accurate and robust state estimation at nighttime is essential for autonomous robotic navigation to achieve nocturnal or round-the-clock tasks. An intuitive question arises: Can low-cost standard cameras be exploited for nocturnal state estimation? Regrettably, most existing visual methods may fail under adverse illumination conditions, even with active lighting or image enhancement. A pivotal insight, however, is that streetlights in most urban scenarios act as stable and salient prior visual cues at night, reminiscent of stars in deep space aiding spacecraft voyage in interstellar navigation. Inspired by this, we propose Night-Voyager, an object-level nocturnal vision-aided state estimation framework that leverages prior object maps and keypoints for versatile localization. We also find that the primary limitation of conventional visual methods under poor lighting conditions stems from the reliance on pixel-level metrics. In contrast, metric-agnostic, non-pixel-level object detection serves as a bridge between pixel-level and object-level spaces, enabling effective propagation and utilization of object map information within the system. Night-Voyager begins with a fast initialization to solve the global localization problem. By employing an effective two-stage cross-modal data association, the system delivers globally consistent state updates using map-based observations. To address the challenge of significant uncertainties in visual observations at night, a novel matrix Lie group formulation and a feature-decoupled multi-state invariant filter are introduced, ensuring consistent and efficient estimation. Through comprehensive experiments in both simulation and diverse real-world scenarios (spanning approximately 12.3 km), Night-Voyager showcases its efficacy, robustness, and efficiency, filling a critical gap in nocturnal vision-aided state estimation. 
\end{abstract}







\section{Introduction} \label{section: introduction}
\IEEEPARstart{D}{riven} by significant advancements in robotics and automation, mobile robots are experiencing widespread adoption in industrial applications, including transportation, logistics, and search-and-rescue operations \cite{rubio2019review}. Among all the essential capabilities, all-time, all-weather autonomous navigation is a fundamental manifestation of robotic autonomy. However, achieving highly resilient localization in low-illumination nocturnal scenarios during navigation typically requires the deployment of high-cost and specialized sensors, such as Light Detection And Ranging (LiDAR) sensors, infrared cameras, high-precision Global Navigation Satellite Systems (GNSS), and Inertial Navigation Systems (INS). In contrast, standard cameras have the advantages of portability, low cost, and informative measurements. However, utilizing standard cameras for robot localization and navigation in nighttime environments seems highly counterintuitive, and notably, the key challenge lies in the nocturnal visual state estimation. 

\begin{figure}[!t] 
    \centerline{\includegraphics[width=0.47\textwidth]{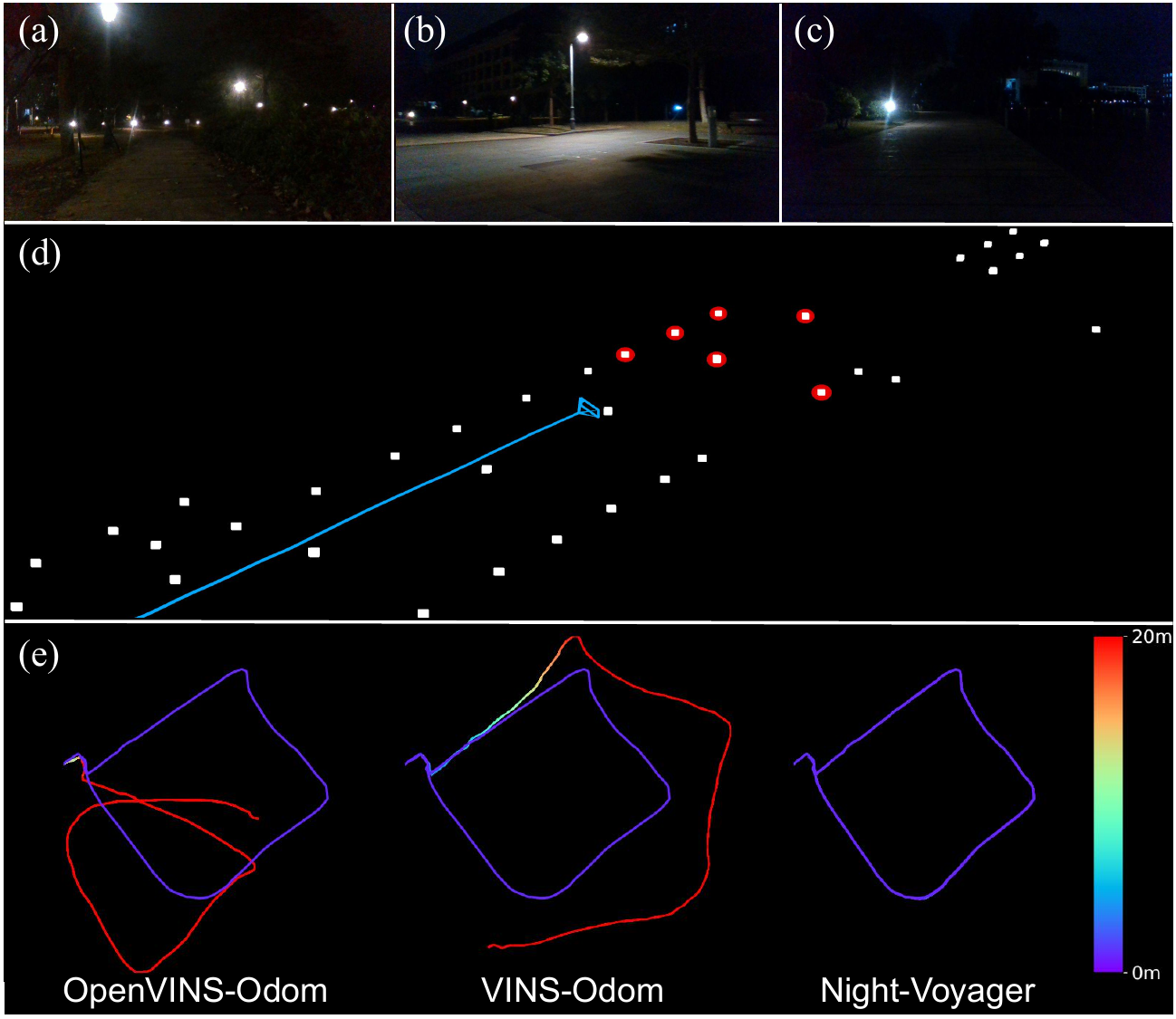}} 
    \caption{Comparison of different vision-aided state estimation methods. (a), (b), and (c) are camera images captured in nocturnal scenes. (d) displays the online localization (the blue curve) of Night-Voyager within the streetlight map (white boxes) and the matches (red spheres) via the proposed data association approach. (e) depicts the trajectories estimated by the odometer-aided OpenVINS \cite{geneva2020openvins} (OpenVINS-Odom), the odometer-aided VINS-Mono \cite{qin2018vins, liu2019visual} (VINS-Odom), and Night-Voyager, respectively. The color bar indicates the trajectory error scale with respect to the ground truth (purple curves).}
    \label{Fig: First Page}
    \vspace{-10pt}
\end{figure}

\subsection{Nocturnal Visual State Estimation} 
\begin{figure}[!t] 
    \centerline{\includegraphics[width=0.47\textwidth]{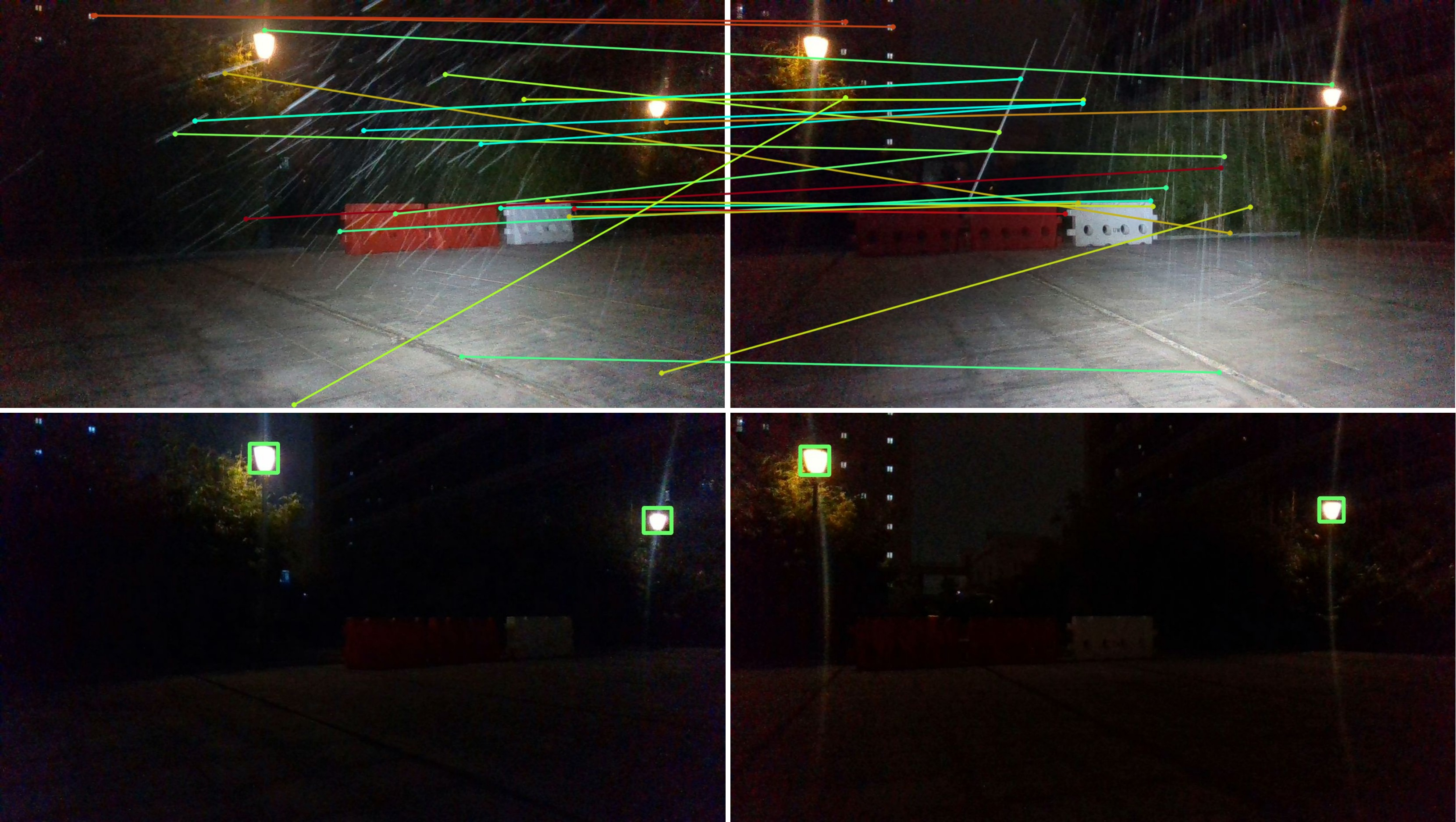}} 
    \caption{Top row: active lighting can further aggravate inconsistent and imbalanced illumination issues while amplifying the backscatter effect caused by sparkling particles, resulting in a significant number of erroneous feature matches (colored dots and lines). Bottom row: object-level detection remains extraordinarily robust to varying lighting conditions. Even in low-light or completely dark nighttime scenarios, streetlights can consistently serve as stable and salient object-level features for detection (green detection boxes).}
    \label{Fig: Particles}
    \vspace{-10pt}
\end{figure}

Therefore, a natural question arises: Is it feasible to achieve accurate and robust state estimation in low-illumination nocturnal environments using low-cost, standard camera-centric sensors with generic parameters, which are primarily designed for well-illuminated conditions? Unfortunately, conventional vision-aided state estimation or Simultaneous Localization And Mapping (SLAM) methods \cite{forster2016svo, engel2017dso, mur2017orb, bloesch2017rovio, qin2018vins, geneva2020openvins, campos2021orb, liu2019visual} are prone to failures, as shown in Fig. \ref{Fig: First Page}. This is attributed to the fact that unfavorable, unstable, and inconsistent illumination conditions could lead to insufficient or erroneous data association. Besides, the limited visual information available in dark scenes can increase pose drifts, potentially resulting in system failures. Intuitively, a practical solution is to install active heading lights on robots. However, as shown in Fig. \ref{Fig: Particles}, the empirical evidence suggests a reality that is contrary to initial expectations. On the one hand, active lighting and dynamic motions can exacerbate the inconsistent, imbalanced, and radical illumination variance, thereby increasing the challenge of data association \cite{kim2020proactive, kim2020dark_synthetic_vision, lin2024breaking}. On the other hand, illuminated and enhanced airborne particles and dust\cite{bohren2008absorption, baron2008adobe}, as well as high-reflectivity objects\cite{kim2020dark_synthetic_vision}, are sparkling in camera images owing to backscattering \cite{grimm1997basic}. This phenomenon could affect the feature tracking, leading to deteriorated and chaotic data association results. To solve these issues from active lighting, sophisticated camera lens design or low-illumination environment-specific parameter tuning in Image Signal Processing (ISP) algorithms or camera control functions is necessary in different environments \cite{blahnik2016reduction_zeiss, zhang2017active, yang2018challenges}. Consequently, although the active lighting solution is direct and aligns with common intuition, it fails to fundamentally address the underlying problem of nocturnal visual state estimation in low-illumination environments. 

\subsection{Bottleneck and Crux} \label{subsection: crux}
The underlying problem of nocturnal visual state estimation stems from two primary bottleneck factors: \textit{insufficiency} and \textit{inconsistency}. First, the insufficiency indicates the lack of prominent visual features in dark nocturnal environments, which prevents the system from leveraging effective and accurate observations to update states in time. Second, even when prominent visual features are present (e.g., from active lighting or locally illuminated areas), these features are often inconsistent, transient, and typically originate from dynamic objects (e.g., illuminated dust particles). To alleviate the \textit{insufficiency} issue in low-illumination environments, researchers explore numerous classical image processing methods \cite{yamasaki2008denighting, dong2010fast, fotiadou2014low} and deep learning-based methods \cite{lore2017llnet, jiang2021enlightengan, ma2022toward, yang2023implicit, wu2023learning, wang2024zero} to enhance low-light images. Nevertheless, the generalization and real-time performance of these methods fail to meet the requirements of robotic applications, rendering them challenging to be deployed on robotic platforms. Crucially, these methods are unable to maintain temporal photometric consistency between images, which consequently impacts the performance of data association and exaggerates the \textit{inconsistency} issue. In parallel, extensive works are proposed to achieve robust and consistent image feature matching \cite{lowe2004SIFT, rosten2006FAST, rublee2011ORB, leutenegger2011brisk, sarlin2019coarse, detone2018superpoint, sarlin2020superglue, sun2021loftr, ma2021image, jiang2021review}. However, these methods are also found to be challenging to apply in low-illumination images and dark environments \cite{song2021dark_matching, he2023darkfeat}. Furthermore, the issues of generalization, parameter tuning, and real-time performance also represent bottlenecks for online robotic applications. 

Significantly, the two key bottleneck factors in nocturnal visual state estimation highlighted above -- namely, \textit{insufficiency} and \textit{inconsistency} -- point to a common crux and a fundamental insight: current conventional visual methods heavily rely on \textit{pixel-level} visual features and \textit{pixel-level} data association approaches. Therefore, relying solely on the \textit{ pixel-level} feature and data association is not an essentially seminal and ultimate solution for nocturnal visual state estimation. As for tackling the issues of \textit{insufficiency} and \textit{inconsistency}, is there a solution that can provide salient visual features in nighttime environments, while ensuring these features remain consistent, stable, and non-pixel-level? This represents an essential problem in nocturnal visual state estimation. Inspired by the stellar-inertial navigation of spacecraft in deep space (e.g., the Voyager Mission) \cite{mckinley1976mariner, jacobson1976navigation, kohlhase1977voyager_mission_description, kuritsky1983inertial} or the human celestial navigation in the wilderness \cite{touche2004wilderness}, which exploits priori ephemeris data and star coordinates to achieve localization, we realize that utilizing the priori environmental map information represents a fundamental and effective avenue for addressing the aforementioned challenges in nocturnal scenes. Fortunately, for low-illumination nocturnal environments, an important discernment is that sufficient static streetlights can provide a stable source of rare object-level perceptive information in most urban scenarios (e.g., nighttime campuses, parks, factories, or streets). Additionally, the light saturation makes streetlights easily detectable from images, crucially at the object level instead of the pixel level. Consequently, these insights motivate us to explore priori lightweight streetlight maps and object-level association methods to achieve globally consistent and drift-bounded nocturnal visual state estimation. 

\begin{figure*}[t]
    \centerline{\includegraphics[width=0.85\textwidth]{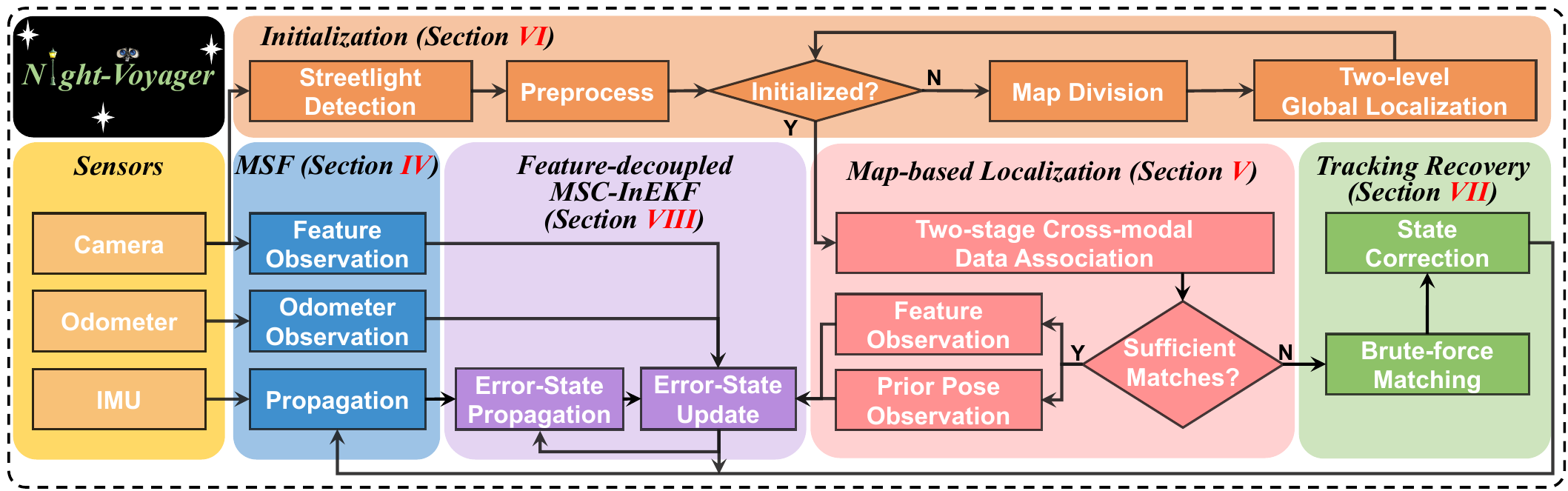}}
    \caption{System overview of the proposed consistent and efficient nocturnal vision-aided state estimation framework, Night-Voyager.}
    \label{Fig: system_overview}
    \vspace{-10pt}
\end{figure*}

\subsection{Proposed Methodology and Framework} 
Despite the benefits of object-level streetlight detection, there are several challenges in leveraging it to achieve generic visual state estimation. First, due to the high similarity between different streetlight detections in images, designing discriminating descriptors for streetlight data association or feature maps is not feasible. Hence, the localization based on conventional visual-feature correspondences or visual maps is unreliable and error-prone. Second, the substantial uncertainties inherent in nighttime visual observations coupled with the considerable errors in initialization compel us to design a robust, consistent, and efficient state estimator in theory. Third, a fast and accurate initialization module, which is essentially dominated by a global localization problem within the prior map, is indispensable for the convergence of online state estimator. In addition, owing to the intrinsic sparsity of streetlights, degenerate and tracking lost cases are inevitable when few streetlights are visible within the camera's Field of View (FoV). Accordingly, pose estimates are susceptible to rapid drifts in the absence of global constraints. 

To address these issues, this work introduces Night-Voyager, a hybrid object-level and pixel-level state estimation framework, achieving robust and versatile nocturnal vision-aided state estimation in object maps. The Night-Voyager framework is presented in Fig. \ref{Fig: system_overview}. For nighttime cases, the reuse of collected LiDAR maps can be combined with object detection methods to generate extremely lightweight streetlight maps. Night-Voyager starts with the multi-sensor fusion module (MSF) (Section \ref{section: multi-sensor fusion module}) and the initialization (Section \ref{section: initialization}) concurrently. The initialization is performed in a divide-and-conquer scheme, thereby dividing the map-based global localization problem into a series of \textit{Perspective-Three-Point} (P3P) problems \cite {ding2023revisiting}. A two-stage cross-modal data association approach further ensures the accuracy and robustness of the detection-map matching process, providing reliable observations for state update (Section \ref{section: map-based localization}). Besides, the MSF module that fuses keypoints with other sensor measurements is independent of other modules. This design enables the system to achieve accurate state estimation in both dark nighttime and well-illuminated scenarios. Thus, Night-Voyager actually constitutes a complete and all-day vision-aided state estimation framework. More importantly, with a novel matrix Lie group formulation, a feature-decoupled Multi-State-Constraint Invariant Extended Kalman Filter (MSC-InEKF) is designed (Section \ref{section: feature-decoupled MSC-InEKF}), endowing Night-Voyager with a consistent and efficient state estimator that attains the significantly accurate localization at night. The designed matrix Lie group-based filter can ensure stability, convergence, and consistency in state estimation, making it particularly robust for nocturnal state estimation, even in scenarios characterized by high observation noise, prolonged absence of valid observations, and considerable initialization errors. When streetlight matches are absent for a while, the tracking recovery module (Section \ref{section: tracking recovery}) is triggered to correct the state estimate. Extensive experiments on simulation, public sequences, and the collected dataset validate the effectiveness and performance of the proposed methodology. Compared with the previous work Night-Rider \cite{gao2024night}, Night-Voyager features novel advancements in: 
\begin{enumerate}
    \item A novel and efficient filter design. In contrast to Night-Rider only utilizing current observations, the multi-state-constraint and the feature-decoupled traits substantially enhance the estimation accuracy and efficiency. 
    
    \item A more robust cross-modal data association approach. The innovative error design improves the ability to identify correct matches and the adaptability to adjust the requirement for finding matches.
    
    \item A versatile and generic framework architecture. Night-Rider requires a manually specified initial pose and operates only at night. In comparison, the fast initialization and MSF enable Night-Voyager to perform autonomous state estimation throughout diverse all-day scenarios. 
    
    \item The usage of pixel-level features and prior poses for accuracy improvement and generality. In contrast, Night-Rider only utilizes streetlight detections for localization. 
    
    \item A precise modeling of streetlight observations. The virtual center of streetlight clusters significantly improves localization accuracy, while Night-Rider exploits the geometric center, leading to inaccurate constraints. 
    
    \item A robust tracking recovery module design. Night-Rider relies on a single image to determine the correct match for pose recovery, whereas Night-Voyager leverages the information accumulated over a short period. 
\end{enumerate}

In summary, the contributions of this work are as follows: 
\begin{enumerate} 
    \item An innovative object map-empowered nocturnal vision-aided state estimation framework is developed to achieve object-level correspondences, which essentially resolve bottlenecks and the crux in nocturnal visual tasks. Notably, this work bridges the critical gap in nocturnal visual state estimation. 
    
    \item A robust two-stage cross-modal data association approach is introduced, which effectively addresses the challenge of cross-modal correspondences between visual measurements and prior maps. 
    
    \item A hierarchical initialization method, efficiently solving the global localization problem within the map, is proposed without the aid of manual intervention or GNSS. 
    
    \item A novel matrix Lie group formulation and a feature-decoupled MSC-InEKF are designed, equipping Night-Voyager with a consistent and efficient state estimator. In addition, multiple state representations are thoroughly compared via an in-depth observability analysis. 

    \item To foster research and field applications, the code and the nocturnal multi-sensor dataset are publicly released. 
\end{enumerate}

Ultimately, the proposed Night-Voyager framework endows robots with extraordinary localization capabilities, enabling reliable autonomous navigation in nighttime urban environments and substantially expanding the applicability of such systems for round-the-clock operations. 


\section{Related Work} \label{section: related work} 
In this section, we discuss the most related research works on priori LiDAR map-based visual localization, along with nocturnal visual state estimation and SLAM. 

\subsection{LiDAR Map-based Visual Localization}
Due to the long-term stability and rich geometric information of LiDAR maps, cross-modal localization attracts substantial research attention. Early research works focus on comparing the similarity between images and projected LiDAR data by utilizing normalized mutual information (NMI) \cite{wolcott2014visual}, normalized information distance (NID) \cite{stewart2012laps}, intensity \cite{pascoe2015direct}, and depth \cite{neubert2017sampling, zuo2019visual}. Some methods \cite{ding2019persistent, caselitz2016monocular, zhang2023cross, zuo2020multimodal} propose to transform the 2D-3D matching problem into a 3D-3D matching problem, thus the structure information can be utilized. Caselitz et al. \cite{caselitz2016monocular} employ a visual odometry algorithm to reconstruct 3D points from visual features. The sparse reconstructed points are matched with the map for tracking. In \cite{zhang2023cross}, semantic consistency between visual features and maps is introduced to improve the accuracy of point-to-plane Iterative Closest Point (ICP) problems. Structural features (e.g., lines) \cite{yu2020monocular, zhou2021visual, leng2024cross} are also utilized for sufficient matches. Recently, deep learning techniques are introduced into cross-modal localization tasks \cite{miao2023poses, shubodh2024lip, zhao2023attention, yin2021i3dloc, cattaneo2019cmrnet, chen2022i2d, wu2024lhmap_icra2024}. While these methods achieve reasonable accuracy, they necessitate dense maps and abundant visual information for cross-modal correspondences, which can be challenging to apply in nighttime scenes. 

To alleviate the huge memory cost of dense maps, common landmarks (e.g., lanes, poles, signboards) are utilized to establish lightweight maps \cite{qin2021light, pauls2020monocular, liao2020coarse, wang2021visual}. In \cite{pauls2020monocular}, the authors leverage semantic segmentation and a distance transformation method for data association. Liao et al. \cite{liao2020coarse} develop a coarse-to-fine localization method, using the coarse pose from wheel odometry to perform the data association of poles and refining the pose from matches. Wang et al. \cite{wang2021visual} propose a new data association method that exploits the local structure, global pattern, and temporal consistency. Compared to the landmarks utilized in the aforementioned works, streetlights are stable, salient, and easily detectable in low-illumination environments, making them suitable for nighttime visual localization. 

\subsection{Nocturnal Visual State Estimation and SLAM}
Conventional visual-inertial odometry (VIO) and vision-aided SLAM methods \cite{forster2016svo, engel2017dso, mur2017orb, bloesch2017rovio, qin2018vins, geneva2020openvins, campos2021orb, liu2019visual, leutenegger2015keyframe, usenko2019visual}, such as VINS-Mono \cite{qin2018vins} and Multi-State-Constraint Extended Kalman Filter (MSCKF) \cite{geneva2020openvins, mourikis2007multi},  typically rely on pixel-level features from images. Unfortunately, these methods struggle in nocturnal environments \cite{kim2020dark_synthetic_vision}, as discussed in Section \ref{subsection: crux}. Thanks to the power of data-driven machine learning techniques, methods for low-light image enhancement \cite{lore2017llnet, jiang2021enlightengan, ma2022toward, yang2023implicit, wu2023learning, wang2024zero, gomez2018learning, jung2020multi} are developed to improve feature extraction. In \cite{gomez2018learning}, a deep neural network is designed to generate more informative images for visual odometry (VO). Jung et al. \cite{jung2020multi} further improve the VO by introducing temporal and spatial constraints into the neural network. However, the issues of generalization and heavy computational burden significantly hinder the application of these methods in robotic tasks. Moreover, some studies focus on designing robust descriptors \cite{alismail2016direct, pascoe2017nid}, explicitly modeling camera parameters using photometric models \cite{engel2017dso, bergmann2017online, bloesch2017rovio, yang2018challenges}, and controlling camera attributes like exposure time and gain \cite{litvinov2005addressing, lu2010camera, zhang2017active, weidner2017underwater, kim2020dark_synthetic_vision, kim2020proactive}. In \cite{kim2020dark_synthetic_vision}, an exposure control scheme that considers inter-frame consistency is proposed, which effectively increases the matched features for visual SLAM. This work is further extended by jointly considering exposure time and gain \cite{kim2020proactive}. Nevertheless, these approaches require an in-depth understanding of the specific sensor and heuristic tuning of parameters, which impedes their generalization to different setups. In addition, these methods still essentially rely on pixel-level correspondences, which cannot resolve the fundamental \textit{insufficiency} and \textit{inconsistency} issues in nocturnal visual state estimation, as discussed in Section \ref{subsection: crux}. Few studies investigate consistent and stable visual features for nocturnal state estimation. Nelson et al. \cite{nelson2015dusk} propose NightNav, a visual localization system that uses a prior map storing streetlight images from different views and positions. However, the assumption that the area of streetlight blobs is inversely proportional to the distance of streetlights from the camera is overly stringent. When occlusion and perspective change cases occur, the distance-area function is unreliable. Besides, the pre-stored images of all streetlights in NightNav increase the storage requirement. Our previous work, Night-Rider \cite{gao2024night}, only uses the object-level visual association from streetlights, demonstrating accurate localization performance within the streetlight map. However, Night-Rider ignores historical information and suffers from degradation and degeneration cases. Additionally, Night-Rider can only be applied in nighttime scenes rather than all-day scenarios. 

In this work, a novel state estimation framework named Night-Voyager is proposed, leveraging hybrid object-level and pixel-level visual correspondences to tackle the fundamental bottlenecks of \textit{insufficiency} and \textit{inconsistency} in nocturnal visual cases, thereby addressing the critical crux inherent in mainstream pixel-level methodologies. 


\section{Preliminary and State Definition} \label{section: preliminary and state definition} 

\subsection{Theoretical Background}
\textbf{Lie Group and Lie Algebra}: Let $\mathcal{G}$ denote a matrix Lie group with its corresponding Lie algebra $\mathfrak{g}$. Elements in the Euclidean space can be mapped to the Lie group by the exponential mapping $\mathrm{Exp}: \mathbb{R}^{dim\mathfrak{g}}\rightarrow \mathcal{G}$ \cite{barrau2016invariant}: 
\begin{equation}
    \label{equation: exponential map}
    \mathbf{X}=\mathrm{Exp}(\bm{\xi})=\exp(\bm{\xi}^{\curlywedge})
\end{equation}
where $dim\mathfrak{g}$ represents the dimension of the Euclidean space of Lie algebra. $\bm{\xi}\in\mathbb{R}^{dim\mathfrak{g}}$ and $\mathbf{X}\in\mathcal{G}$ are the elements in the corresponding Euclidean space and Lie group. $\exp(\cdot)$ is the standard exponential map. $(\cdot)^{\curlywedge}: \mathbb{R}^{dim\mathfrak{g}}\rightarrow\mathfrak{g}$ converts the vector representation to the isomorphic matrix representation of the Lie algebra. We denote $\mathrm{Log}: \mathcal{G}\rightarrow\mathbb{R}^{dim\mathfrak{g}}$ as the inverse operation that maps the Lie group to the Euclidean space. 

The commonly used Lie groups include the special orthogonal groups $SO(3)$, the special Euclidean groups $SE(3)$, and the extended special Euclidean group $SE_{2}(3)$, which is an extension of $SE(3)$, containing a rotation matrix and two vectors. It is possible to extend the Lie groups $SE(3)$ or $SE_{2}(3)$ by associating them with $K$ additional vectors, resulting in the Lie groups $SE_{1+K}(3)$ or $SE_{2+K}(3)$, respectively \cite{hartley2020contact}.

\textbf{Invariant Error}: Suppose $\mathbf{X}\in \mathcal{G}$ and $\widehat{\mathbf{X}}\in \mathcal{G}$ are, respectively, the true and estimated states, the left- and right-invariant errors are defined as \cite{barrau2016invariant}: 
\begin{align}
    \label{equation: left invariant error}
    \bm{\eta}^{l} &= \mathbf{X}^{-1}\widehat{\mathbf{X}} \quad (\text{Left-invariant Error}) \\
    \label{equation: right invariant error}
    \bm{\eta}^{r} &= \widehat{\mathbf{X}}\mathbf{X}^{-1} \quad (\text{Right-invariant Error}).
\end{align}
In this work, we use the right-invariant error representation \cite{hartley2020contact, brossard2018invariant, song2021right} to formulate the state estimation problem.

\subsection{Notation and State Definition} \label{section: Problem Description}
In this work, ${^{G}}(\cdot)$, ${^{L}}(\cdot)$, ${^{C}}(\cdot)$, ${^{O}}(\cdot)$, and ${^{I}}(\cdot)$ respectively represent a 3D vector in the global map, local, camera, odometer, and Inertial Measurement Unit (IMU) frames. Both the map and local frames are fixed, and the $z$ direction of the local frame is aligned with the constant gravity vector $\mathbf{g}$. To distinguish the frames between mapping and online state estimation, we use primed letters to represent the frames in the mapping system, e.g., ${I^{\prime}}$. The IMU frame coincides with the body frame. ${^{L}}\mathbf{T}_{I}\in SE(3)$ is the transformation from the local frame to body frame, including the rotation ${^{L}}\mathbf{R}_{I} \in SO(3)$ and the position ${^{L}}\mathbf{p}_{I} \in \mathbb{R}^{3}$. All velocities are with respect to the local frame. $^{L}\mathbf{v}_{I} \in \mathbb{R}^{3}$ is the velocity of the body frame with respect to the local frame. Time index $k$ and IMU sample index $i$ indicate their corresponding time $t_{k}$ and $t_{i}$. $\widetilde{(\cdot)}$ and $\widehat{(\cdot)}$ represent the measurement and the state estimate, respectively. 

The map-based state estimation problem is decomposed into two key parts: estimating odometry in the local frame and determining the relative transformation from the local frame to the map frame. Thus, the system state $\mathbf{X}_{k}$ at time $t_{k}$ is divided into: the local state $\mathbf{X}_{L_{k}}$ (including the body navigation state $\mathbf{X}_{\bm{I}_{k}}\in SE_{2}(3)$, the IMU bias $\mathbf{B}_{k}\in \mathbb{R }^{6}$, and the historical pose clone $\mathbf{X}_{\bm{C}_{k}}$) and the compound state $\mathbf{X}_{\bm{G}_{k}}\in SE_{1+K}(3)$: 
\begin{align}
    \mathbf{X}_{k} &= (\mathbf{X}_{\bm{L}_{k}},\mathbf{X}_{\bm{G}_{k}}) \notag \\ 
    \mathbf{X}_{\bm{L}_{k}} &= (\mathbf{X}_{\bm{I}_{k}}, \mathbf{B}_{k}, \mathbf{X}_{\bm{C}_{k}}) \notag \\ 
    \mathbf{X}_{\bm{I}_{k}} &= \begin{bmatrix}
        \begin{array}{c|cc}
            {^{L}}\mathbf{R}_{I_{k}} & {^{L}}\mathbf{p}_{I_{k}} & {^{L}}\mathbf{v}_{I_{k}}\\\hline
            \mathbf{0}_{2\times3} & \multicolumn{2}{c}{\mathbf{I}_{2\times2}} 
        \end{array}
    \end{bmatrix}, \mathbf{B}_{k} = \begin{bmatrix} \mathbf{b}_{g_{k}}^{\top} & \mathbf{b}_{a_{k}}^{\top} \end{bmatrix}^{\top} \notag \\ 
    \mathbf{X}_{\bm{C}_{k}} &= (\mathbf{X}_{CP_{k_{k}}},\cdots,\mathbf{X}_{CP_{k-c+1_{k}}}), 
    \mathbf{X}_{CP_{j_{k}}} = {\setlength\arraycolsep{2pt} \begin{bmatrix} {^{L}}\mathbf{R}_{I_{j_{k}}} & {^{L}}\mathbf{p}_{I_{j_{k}}} \\ \mathbf{0}_{1\times 3} & 1 \end{bmatrix}} \notag \\ 
    \mathbf{X}_{\bm{G}_{k}} &= \begin{bmatrix}
    \begin{array}{c|cccc}
        {^{L}}\mathbf{R}_{G_{k}} & {^{L}}\mathbf{p}_{G_{k}} & {^{L}}\mathbf{p}_{f_{1_{k}}} & \cdots & {^{L}}\mathbf{p}_{f_{K_{k}}} \label{equation: all states} \\\hline \mathbf{0}_{(1+K)\times 3} & \multicolumn{4}{c}{\mathbf{I}_{(1+K)\times(1+K)}}
    \end{array}
    \end{bmatrix} 
\end{align}
where ${^{L}}\mathbf{R}_{I_{k}} \in SO(3)$, ${^{L}}\mathbf{p}_{I_{k}} \in \mathbb{R}^{3}$, and ${^{L}}\mathbf{v}_{I_{k}} \in \mathbb{R}^{3}$ represent the rotation, position, and velocity of body frame in the local frame, respectively. The bias state includes the gyroscope bias $\mathbf{b}_{g_{k}}\in \mathbb{R}^{3}$ and the accelerometer bias $\mathbf{b}_{a_{k}}\in \mathbb{R}^{3}$. The state also contains $c$ historical IMU pose clones $\mathbf{X}_{CP_{j_{k}}} \in SE(3), j=k-c+1,\cdots,k$, each consists of a rotation matrix $^{L}\mathbf{R}_{I_{j_{k}}}\in SO(3)$ and a translation vector ${^{L}}\mathbf{p}_{I_{j_{k}}}\in\mathbb{R}^{3}$. The index $j_{k}$ denotes the pose cloned at time $t_{j}$ with the estimate at $t_{k}$. $\mathbf{X}_{\bm{G}_{k}}$ is composed of the rotation ${^{L}}\mathbf{R}_{G_{k}}\in SO(3)$ and position ${^{L}}\mathbf{p}_{G_{k}}\in \mathbb{R}^{3}$ of global map frame with respect to the local frame, as well as the positions of the tracked feature points expressed in the local frame ${^{L}}\mathbf{p}_{f_{l_{k}}}\in\mathbb{R}^{3}, l=1,\cdots,K$. 

The right-invariant error in (\ref{equation: right invariant error}) between the estimated state $\widehat{\mathbf{X}}_{k}$ and the true state $\mathbf{X}_{k}$ is leveraged to derive \cite{brossard2018invariant}: 
\begin{align}
    \label{equation: error state function}
    {^{L}}\mathbf{R}_{I_{k}}&=\mathrm{Exp}(\bm{\xi}_{\bm{R}_{LI_{k}}}){^{L}}\widehat{\mathbf{R}}_{I_{k}}\notag\\
    {^{L}}\mathbf{p}_{I_{k}}&=\mathrm{Exp}(\bm{\xi}_{\bm{R}_{LI_{k}}}){^{L}}\widehat{\mathbf{p}}_{I_{k}}+\mathbf{J}_{l}(\bm{\xi}_{\bm{R}_{LI_{k}}})\bm{\xi}_{\bm{p}_{LI_{k}}}\notag\\
    {^{L}}\mathbf{v}_{I_{k}}&=\mathrm{Exp}(\bm{\xi}_{\bm{R}_{LI_{k}}}){^{L}}\widehat{\mathbf{v}}_{I_{k}}+\mathbf{J}_{l}(\bm{\xi}_{\bm{R}_{LI_{k}}})\bm{\xi}_{\bm{v}_{LI_{k}}}\notag\\
    {^{L}}\mathbf{R}_{I_{j_{k}}}&=\mathrm{Exp}(\bm{\xi}_{\bm{R}_{CP_{j_{k}}}}){^{L}}\widehat{\mathbf{R}}_{I_{j_{k}}}\notag\\
    {^{L}}\mathbf{p}_{I_{j_{k}}}&=\mathrm{Exp}(\bm{\xi}_{\bm{R}_{CP_{j_{k}}}}){^{L}}\widehat{\mathbf{p}}_{I_{j_{k}}}+\mathbf{J}_{l}(\bm{\xi}_{\bm{R}_{CP_{j_{k}}}})\bm{\xi}_{\bm{p}_{CP_{j_{k}}}}\notag\\
    {^{L}}\mathbf{R}_{G_{k}}&=\mathrm{Exp}(\bm{\xi}_{\bm{R}_{LG_{k}}}){^{L}}\widehat{\mathbf{R}}_{G_{k}}\notag\\
    {^{L}}\mathbf{p}_{G_{k}}&=\mathrm{Exp}(\bm{\xi}_{\bm{R}_{LG_{k}}}){^{L}}\widehat{\mathbf{p}}_{G_{k}}+\mathbf{J}_{l}(\bm{\xi}_{\bm{R}_{LG_{k}}})\bm{\xi}_{\bm{p}_{LG_{k}}}\notag\\
    {^{L}}\mathbf{p}_{f_{l_{k}}}&=\mathrm{Exp}(\bm{\xi}_{\bm{R}_{LG_{k}}}){^{L}}\widehat{\mathbf{p}}_{f_{l_{k}}}+\mathbf{J}_{l}(\bm{\xi}_{\bm{R}_{LG_{k}}})\bm{\xi}_{\bm{p}_{Lf_{l_{k}}}}\notag\\
    \mathbf{b}_{g_{k}}&=\widehat{\mathbf{b}}_{g_{k}} + \delta\mathbf{b}_{g_{k}},\quad\mathbf{b}_{a_{k}}=\widehat{\mathbf{b}}_{a_{k}} + \delta\mathbf{b}_{a_{k}}
\end{align}
where $\setlength\arraycolsep{2pt}{\bm{\xi}_{\bm{I}_{k}}=\left[\begin{smallmatrix} \bm{\xi}_{\bm{R}_{LI_{k}}}^{\top} & \bm{\xi}_{\bm{p}_{LI_{k}}}^{\top} & \bm{\xi}_{\bm{v}_{LI_{k}}}^{\top}\end{smallmatrix}\right]^{\top}\in\mathbb{R}^{9}}$ is the body error state. The subscript $(\cdot)_{\bm{R}_{LI_{k}}}$ indicates that the error state $\bm{\xi}_{\bm{R}_{LI_{k}}}$ corresponds to the rotation state ${^{L}}\mathbf{R}_{I_{k}}$, $(\cdot)_{\bm{p}_{LI_{k}}}$ indicates that the error state $\bm{\xi}_{\bm{p}_{LI_{k}}}$ corresponds to the position state ${^{L}}\mathbf{p}_{I_{k}}$, and $(\cdot)_{\bm{v}_{LI_{k}}}$ indicates that the error state $\bm{\xi}_{\bm{v}_{LI_{k}}}$ corresponds to the velocity state ${^{L}}\mathbf{v}_{I_{k}}$. $\bm{\xi}_{\bm{C}_{k}}=\left[\begin{smallmatrix}
    {\bm{\xi}_{CP_{k_{k}}}^{\top}} \cdots {\bm{\xi}_{CP_{k-c+1_{k}}}^{\top}}
\end{smallmatrix}\right]^{\top} \in \mathbb{R}^{6c}$ contains all the error states of pose clones, with each one formulated as $\bm{\xi}_{CP_{j_{k}}}=\left[\begin{smallmatrix}
    \bm{\xi}_{\bm{R}_{CP_{j_{k}}}}^{\top} & \bm{\xi}_{\bm{p}_{CP_{j_{k}}}}^{\top}
\end{smallmatrix}\right]^{\top}\in\mathbb{R}^{6}$. $\bm{\xi}_{\bm{G}_{k}}=\left[\begin{smallmatrix} \bm{\xi}_{\bm{R}_{LG_{k}}}^{\top} & \bm{\xi}_{\bm{p}_{LG_{k}}}^{\top} & \bm{\xi}_{\bm{p}_{Lf_{1_{k}}}}^{\top} & \cdots \bm{\xi}_{\bm{p}_{Lf_{K_{k}}}}^{\top}\end{smallmatrix}\right]^{\top}\in\mathbb{R}^{6+3K}$ is the compound error state, where $\bm{\xi}_{\bm{T}_{LG_{k}}}=\left[\begin{smallmatrix}
    \bm{\xi}_{\mathbf{R}_{LG_{k}}}^{\top} & \bm{\xi}_{\mathbf{p}_{LG_{k}}}^{\top}
\end{smallmatrix}\right]^{\top}\in\mathbb{R}^{6}$ denotes the error state of relative transformation and $\bm{\xi}_{\bm{p}_{Lf_{l_{k}}}}\in\mathbb{R}^{3}$ is the feature error state. $\bm{\zeta}_{\bm{B}_{k}}=\left[\begin{smallmatrix}
    \delta\mathbf{b}_{g_{k}}^{\top} & \delta\mathbf{b}_{a_{k}}^{\top}
\end{smallmatrix}\right]^{\top}\in\mathbb{R}^{6}$ represents the bias error state. $\mathbf{J}_{l}(\cdot)$ denotes the left Jacobian of $SO(3)$\cite{yang2020analytic}. The error state $\bm{\xi}_{k}=\left[\begin{smallmatrix} \bm{\xi}_{\bm{I}_{k}}^{\top} & \bm{\zeta}_{\bm{B}_{k}}^{\top} & \bm{\xi}_{\bm{C}_{k}}^{\top} & \bm{\xi}_{\bm{G}_{k}}^{\top} \end{smallmatrix}\right]^{\top}\in\mathbb{R}^{21+6c+3K}$ is used to formulate the state estimation problem in the following sections. For simplicity, we only use one feature point ${^{L}}\mathbf{p}_{f_{k}}$ to formulate the state in the rest sections. 


\section{Multi-Sensor Fusion Module} \label{section: multi-sensor fusion module}
We introduce the multi-sensor fusion module (MSF), a general module for vision-aided state estimation, fusing multi-sensor measurements in a tightly-coupled scheme. 

\subsection{State Propagation}
\label{subsection: state propagation}
The system dynamics for state propagation can be derived from the IMU kinematic model \cite{potokar2021invariant} as: 
\begin{equation}
    \label{equation: dynamics model}
    \left\{
        \begin{array}{@{\hspace{0em}}r@{\hspace{0.2em}}l}
            {^{L}}\dot{\mathbf{R}}_{I} &\!=\! {^{L}}\mathbf{R}_{I} (\widetilde{\bm{\omega}} \!-\! \mathbf{b}_{g} \!-\! \mathbf{n}_{g})_{\times},\ 
            {^{L}}\dot{\mathbf{p}}_{I} \!=\! {^{L}}\mathbf{v}_{I}\\
            {^{L}}\dot{\mathbf{v}}_{I} &\!=\! {^{L}}\mathbf{R}_{I}(\widetilde{\bm{a}} \!-\! \mathbf{b}_{a} \!-\! \mathbf{n}_{a}) \!+\! \mathbf{g}, \  
            \dot{\mathbf{b}}_{g} \!=\! \mathbf{n}_{bg}, \  \dot{\mathbf{b}}_{a} \!=\! \mathbf{n}_{ba}\\
            {^{L}}\dot{\mathbf{R}}_{I_{j}} &\!=\! \mathbf{0}_{3\times 3},\ {^{L}}\dot{\mathbf{p}}_{I_{j}} \!=\! \mathbf{0}_{3\times 1},\ j\!=\!k,\cdots,k-c+1\\
            {^{L}}\dot{\mathbf{R}}_{G} &\!=\! \mathbf{0}_{3\times 3},\ {^{L}}\dot{\mathbf{p}}_{G} \!=\! \mathbf{0}_{3\times 1},\ {^{L}}\dot{\mathbf{p}}_{f}\!=\!\mathbf{0}_{3\times 1}
        \end{array}
    \right.
\end{equation}
where $\widetilde{\bm{\omega}}$ and $\widetilde{\bm{a}}$ represent IMU readings of angular velocity and specific force, contaminated respectively by Gaussian noises $\mathbf{n}_{g}$ and $\mathbf{n}_{a}$. $(\cdot)_{\times}$ is the 3-dimensional skew-symmetric matrix. $\mathbf{n}_{bg}$ and $\mathbf{n}_{ba}$ are the random walk noises of IMU biases. The discrete system dynamics can be derived by integrating IMU measurements from $t_{i}$ to $t_{i+1}$:
\begin{equation}
\label{equation: discretized dynamics}
\!\!\!\left\{
    \begin{array}{@{\hspace{0em}}r@{\hspace{0.2em}}l}
        {^{L}}\mathbf{R}_{I_{i+1}} &\!=\! {^{L}}\mathbf{R}_{I_{i}}\Delta\mathbf{R}_{i}  \\ 
        {^{L}}\mathbf{p}_{I_{i+1}} &\!=\! {^{L}}\mathbf{p}_{I_{i}} \!+\! {^{L}}\mathbf{v}_{I_{i}} \delta t \!+\! {^{L}}\mathbf{R}_{I_{i}}\Delta \mathbf{p}_{i} \!+\! \frac{1}{2}\mathbf{g}\delta t^{2}  \\ 
        {^{L}}\mathbf{v}_{I_{i+1}} &\!=\! {^{L}}\mathbf{v}_{I_{i}} \!+\! {^{L}}\mathbf{R}_{I_{i}}\Delta \mathbf{v}_{i} \!+\! \mathbf{g} \delta t \\
        \mathbf{b}_{g_{i+1}} &\!=\! \mathbf{b}_{g_{i}} \!+\! \int_{t_{i}}^{t_{i\!+\!1}}\mathbf{n}_{bg}d\tau, \ 
        \mathbf{b}_{a_{i\!+\!1}} \!=\! \mathbf{b}_{a_{i}} \!+\! \int_{t_{i}}^{t_{i\!+\!1}}\mathbf{n}_{ba}d\tau
    \end{array}
\right.\!\!\!\!
\end{equation}
where $\Delta\mathbf{R}_{i}\!=\!\mathrm{Exp}(\int_{t_{i}}^{t_{i+1}}\bm{\omega}_{\tau}d\tau)$, $\Delta \mathbf{v}_{i}\!=\!\int_{t_{i}}^{t_{i+1}}\!{^{I_{i}}}\mathbf{R}_{I_{\tau}}\bm{a}_{\tau}d\tau$, and $\Delta \mathbf{p}_{i}\!=\!\int_{t_{i}}^{t_{i+1}}\int_{t_{i}}^{s}{^{I_{i}}}\mathbf{R}_{I_{\tau}}\bm{a}_{\tau}d\tau ds$. $\delta t=t_{i+1}-t_{i}$ is the time interval. Following \cite{yang2020analytic}, we use the analytical combined IMU integration method to approximate $\Delta\mathbf{R}_{i}, \Delta \mathbf{p}_{i}, \Delta \mathbf{v}_{i}$, and substitute the error-state functions (\ref{equation: error state function}) into (\ref{equation: discretized dynamics}) to derive the linearized error propagation function as: 
\begin{gather}
    \label{equation: linearized function}
    \bm{\xi}_{i+1}=\begin{bmatrix}
        \bm{\xi}_{\bm{I}_{i+1}} \\ \bm{\zeta}_{\bm{B}_{i+1}} \\ \bm{\xi}_{\bm{C}_{i+1}} \\ \bm{\xi}_{\bm{T}_{LG_{i+1}}} \\ \bm{\xi}_{\bm{p}_{Lf_{i+1}}}
    \end{bmatrix}=\bm{\Phi}_{i}^{i+1}\begin{bmatrix}
        \bm{\xi}_{\bm{I}_{i}} \\ \bm{\zeta}_{\bm{B}_{i}} \\ \bm{\xi}_{\bm{C}_{i}} \\ \bm{\xi}_{\bm{T}_{LG_{i}}} \\ \bm{\xi}_{\bm{p}_{Lf_{i}}}
    \end{bmatrix}+\bm{G}_{i}\begin{bmatrix}
        \mathbf{n}_{dg} \\ \mathbf{n}_{da} \\ \mathbf{n}_{dbg} \\ \mathbf{n}_{dba}
    \end{bmatrix}\\
{\setlength\arraycolsep{0.7pt}
    \bm{\Phi}_{i}^{i\!+\!1}\!\!=\!\!\begin{bmatrix}
        \bm{\Phi}_{\bm{I}\bm{I}} & \bm{\Phi}_{\bm{I}\bm{B}} & \mathbf{0}_{9\times 6c} & \mathbf{0}_{9\times 6} & \mathbf{0}_{9\times 3}\\
        \mathbf{0}_{6\times 9} & \mathbf{I}_{6\times 6} & \mathbf{0}_{6\times 6c} & \mathbf{0}_{6\times 9} & \mathbf{0}_{6\times 3}\\
        \mathbf{0}_{6c\times 9} & \mathbf{0}_{6c\times 6} & \mathbf{I}_{6c\times 6c} & \mathbf{0}_{6c\times 9} & \mathbf{0}_{6c\times 3}\\
        \mathbf{0}_{6\times 9} & \mathbf{0}_{6\times 6} & \mathbf{0}_{6\times 6c} & \mathbf{I}_{6\times 6} & \mathbf{0}_{6\times 3} \\
        \mathbf{0}_{3\times 9} & \mathbf{0}_{3\times 6} & \mathbf{0}_{3\times 6c} & \mathbf{0}_{3\times 6} & \mathbf{I}_{3\times 3}
    \end{bmatrix}}
{\setlength\arraycolsep{0.7pt}
    \mathbf{G}_{i}\!\!=\!\!\begin{bmatrix}
        \mathbf{G}_{\bm{I}\bm{I}} & \mathbf{0}_{9\times 6}\\
        \mathbf{0}_{6\times 6} & \mathbf{I}_{6\times 6}\delta t \\
        \mathbf{0}_{6c\times 6} & \mathbf{0}_{6c\times 6} \\
        \mathbf{0}_{6\times 6} & \mathbf{0}_{6\times 6} \\
        \mathbf{0}_{3\times 6} & \mathbf{0}_{3\times 6}
    \end{bmatrix}}\notag
\end{gather}
where $\mathbf{\Phi}_{i}^{i+1}\in\mathbb{R}^{(21+6c+3)\!\times\!(21+6c+3)}$ and $\mathbf{G}_{i}\in\mathbb{R}^{(21+6c+3)\times12}$ are respectively the transition matrix and the noise Jacobian matrix. $\mathbf{n}_{d}=\begin{bmatrix}
    \mathbf{n}_{dg}^{\top} & \mathbf{n}_{da}^{\top} & \mathbf{n}_{dbg}^{\top} & \mathbf{n}_{dba}^{\top}
\end{bmatrix}^{\top}$ is the discretized noise. The matrix block components are formulated as: 
\begin{gather}
\label{equation: components of transition matrix}
    {\setlength\arraycolsep{2pt}
    \mathbf{\Phi}_{\bm{I}\bm{I}}=\begin{bmatrix}
        \mathbf{I}_{3\times 3} & \mathbf{0}_{3\times 3} & \mathbf{0}_{3\times 3}\\
        \frac{1}{2}(\mathbf{g})_{\times}\delta t^{2} & \mathbf{I}_{3\times 3} & \mathbf{I}_{3\times 3}\delta t\\
        (\mathbf{g})_{\times}\delta t & \mathbf{0}_{3\times 3} & \mathbf{I}_{3\times 3}
    \end{bmatrix}} \\
    \bm{\Phi}_{\bm{I}\bm{B}}\!=\!\mathbf{G}_{\bm{I}\bm{I}}\!=\!{\setlength\arraycolsep{2pt}
    \begin{bmatrix}
        -\Delta \mathbf{R}\mathbf{J} & \mathbf{0}_{3\times 3} &  \\
        -({^{L}}\widehat{\mathbf{p}}_{I_{i+1}})_{\times} \Delta \mathbf{R}\mathbf{J} + {^{L}}\widehat{\mathbf{R}}_{I_{i}}\mathbf{\Xi}_{4} & -{^{L}}\widehat{\mathbf{R}}_{I_{i}}\mathbf{\Xi}_{2} \\
        -({^{L}}\widehat{\mathbf{v}}_{I_{i+1}})_{\times} \Delta \mathbf{R}\mathbf{J} + {^{L}}\widehat{\mathbf{R}}_{I_{i}}\mathbf{\Xi}_{3} & -{^{L}}\widehat{\mathbf{R}}_{I_{i}}\mathbf{\Xi}_{1}\notag
    \end{bmatrix}}
\end{gather}
where $\Delta \mathbf{R}\mathbf{J}={^{L}}\widehat{\mathbf{R}}_{I_{i+1}}\mathbf{J}_{r}(\mathrm{Log}(\Delta \widehat{\mathbf{R}}_{i}))\delta t$ with $\mathbf{J}_{r}(\cdot)$ denoting the right Jacobian of $SO(3)$\cite{yang2020analytic}. $\mathbf{\Xi}_{1}, \mathbf{\Xi}_{2}, \mathbf{\Xi}_{3}, \mathbf{\Xi}_{4}$ are the integration components and their formulations can be found in \cite{yang2022decoupled}. More derivation details are provided in the corresponding supplementary material (in the GitHub repository). 

Consequently, the estimated state $\widehat{\mathbf{X}}_{i}$ and covariance $\widehat{\mathbf{P}}_{i}$ are propagated to time $t_{i+1}$ by: 
\begin{equation}
    \label{equation: propagation function}
    \!\!\!\!\!\!\!\!\left\{
       \begin{array}{@{\hspace{0em}}r@{\hspace{0.2em}}l}
          {^{L}}\widehat{\mathbf{R}}_{I_{i\!+\!1}}&\!=\!{^{L}}\widehat{\mathbf{R}}_{I_{i}}\mathrm{Exp}((\widetilde{\bm{\omega}}_{i}\!-\!\widehat{\mathbf{b}}_{\omega_{i}})\delta t)\\
          {^{L}}\widehat{\mathbf{p}}_{I_{i\!+\!1}}&\!=\!{^{L}}\widehat{\mathbf{p}}_{I_{i}}\!+\!{^{L}}\widehat{\mathbf{v}}_{I_{i}}\delta t\!+\!{^{L}}\widehat{\mathbf{R}}_{I_{i\!+\!1}}\mathbf{\Xi}_{2}(\widetilde{\bm{a}}_{i}\!-\!\widehat{\mathbf{b}}_{a_{i}})\!+\!\frac{1}{2}\mathbf{g}\delta t^{2}\\
          {^{L}}\widehat{\mathbf{v}}_{I_{i\!+\!1}} &\!=\! {^{L}}\widehat{\mathbf{v}}_{I_{i}} \!+\! {^{L}}\widehat{\mathbf{R}}_{I_{i}}\mathbf{\Xi}_{1}(\widetilde{\bm{a}}_{i}-\widehat{\mathbf{b}}_{a_{i}}) \!+\! \mathbf{g} \delta t\\
          \widehat{\mathbf{b}}_{g_{i\!+\!1}} &\!=\! \widehat{\mathbf{b}}_{g_{i}},\ \widehat{\mathbf{b}}_{a_{i\!+\!1}} \!=\! \widehat{\mathbf{b}}_{a_{i}}\\
          {^{L}}\widehat{\mathbf{R}}_{I_{j_{i\!+\!1}}} &= {^{L}}\widehat{\mathbf{R}}_{I_{j_{i}}},\ {^{L}}\widehat{\mathbf{p}}_{I_{j_{i\!+\!1}}} \!=\! {^{L}}\widehat{\mathbf{p}}_{I_{j_{i}}},\ j\!=\!k,\cdots,k\!-\!c\!+\!1\\
          {^{L}}\widehat{\mathbf{R}}_{G_{i\!+\!1}} &\!=\! {^{L}}\widehat{\mathbf{R}}_{G_{i}},\ {^{L}}\widehat{\mathbf{p}}_{G_{i\!+\!1}}\!=\! {^{L}}\widehat{\mathbf{p}}_{G_{i}},\ {^{L}}\widehat{\mathbf{p}}_{f_{i\!+\!1}}\!=\!{^{L}}\widehat{\mathbf{p}}_{f_{i}}\\
          \widehat{\mathbf{P}}_{i\!+\!1} &\!=\! \mathbf{\Phi}_{i}^{i\!+\!1} \widehat{\mathbf{P}}_{i}{\mathbf{\Phi}_{i}^{i\!+\!1}}^{\top} \!+\! \mathbf{G}_{i}\mathbf{Q}_{d}\mathbf{G}_{i}^{\top}
       \end{array}
    \right.\!\!\!\!\!\!\!
\end{equation}
where $\mathbf{Q}_{d}$ denotes the covariance of $\mathbf{n}_{d}$. The initial values of $\widehat{\mathbf{X}}_{0}$ and $\widehat{\mathbf{P}}_{0}$ at time $t_{i}=t_{0}$ are set to the most recently updated state and covariance (obtained at the closest odometer or camera measurement time $t_{k}$). The system state will be propagated to $t_{k+1}$ when a new odometer or camera measurement arrives. If the measurement is captured from camera, we augment $\widehat{\mathbf{X}}_{\bm{C}_{k+1}}$ by cloning the body pose and the oldest clone will be marginalized after the state update. 

\subsection{Odometer Observation}
It is common to equip wheeled robots with odometers, which measure wheel rates and provide linear velocity measurements under the non-slip assumption. We assume the velocity measurement ${^{O}}\widetilde{\mathbf{v}}_{O_{k}}$ to be corrupted by the Gaussian white noise $\mathbf{n}_{o_{k}} \sim \mathcal{N}(\mathbf{0}_{3 \times 1}, \mathbf{\Sigma}_{o})$:
\begin{equation}
\label{noisy_velocity} 
    {^{O}}\widetilde{\mathbf{v}}_{O_{k}} = {^{O}}\mathbf{v}_{O_{k}} + {\mathbf{n}_{o_{k}}}.
\end{equation}
${^{O}}\mathbf{v}_{O_{k}}$ can be transformed from the velocity expressed in the body (IMU) frame ${^{I}}\mathbf{v}_{I_{k}}$ through the calibrated extrinsic parameters $({^{O}}\mathbf{R}_{I}, {^{O}}\mathbf{p}_{I})$ between the odometer and body frame:
\begin{equation}
    {^{O}\mathbf{v}_{O_{k}}}={^{O}}\mathbf{R}_{I}{^{I}}\mathbf{v}_{I_{k}}+({^{O}}\mathbf{p}_{I})_{\times}{^{O}}\mathbf{R}_{I}(\widetilde{\bm{\omega}}_{k}-\mathbf{b}_{g_{k}}). 
\end{equation}
In our system, the IMU and odometer are assumed to reside in the same location with only ${^{O}}\mathbf{R}_{I}$ being considered, hence the odometer-based observation model is formulated as: 
\begin{equation} 
\label{equation: odometer-based linearized observation}
    \widetilde{\mathbf{y}}_{o_{k}} = {^{O}}\mathbf{R}_{I}{^{L}}\mathbf{R}_{I_{k}}^{\top}{^{L}} \mathbf{v}_{I_{k}} + \mathbf{n}_{o_{k}}
\end{equation}
where $\widetilde{\mathbf{y}}_{o_{k}} = {^{O}}\widetilde{\mathbf{v}}_{O_{k}}$ is the measured velocity in the odometer frame. With the error state functions in (\ref{equation: error state function}), we can linearize the measurement error with respect to $\bm{\xi}_{k}$ as: 
\begin{equation}
    \mathbf{z}_{o_{k}}=\widetilde{\mathbf{y}}_{o_{k}}-\widehat{\mathbf{y}}_{o_{k}}\approx\mathbf{H}_{o_{k}}\bm{\xi}_{k}+\mathbf{n}_{o_{k}}
\end{equation}
where $\widehat{\mathbf{y}}_{o_{k}}$ is the estimated velocity. The Jacobian $\mathbf{H}_{o_{k}}\triangleq\frac{\partial \mathbf{z}_{o_{k}}}{\partial \bm{\xi}_{k}} \in \mathbb{R}^{{3} {\times} {(21 + 6c + 3K)}}$ can be found in the supplementary material. Using Kalman filtering theory \cite{barrau2016invariant}, the updated states and covariance can be obtained. 

\subsection{Feature Observation}
The pixel-level visual features can provide effective supplementary constraints for nocturnal state estimation, especially when the streetlights are invisible. Additionally, leveraging visual features enables all-day localization. Features that are tracked longer than the size of sliding window are initialized into the active state through triangulation. For each observation of ${^{L}}\mathbf{p}_{f_{k}}$ in pose clones, we have the following observation:
\begin{equation}
    \label{feature-based observation}
    \widetilde{\mathbf{y}}_{f_{j_{k}}}=\bm{h} \left( {^{C}}\mathbf{R}_{I} {^{L}}\mathbf{R}_{I_{j_{k}}}^{\top}(^{L}\mathbf{p}_{f_{k}}-{^{L}}\mathbf{p}_{I_{j_{k}}}) + {^{C}}\mathbf{p}_{I} \right) + \mathbf{n}_{f_{k}}
\end{equation}
where $\widetilde{\mathbf{y}}_{f_{j_{k}}}$ denotes the 2D feature point measurement captured at time $t_{j}$. $\bm{h}(\cdot)$ is the camera projection function. $({^{C}}\mathbf{R}_{I}, {^{C}}\mathbf{p}_{I})$ constitute the extrinsic parameters between camera and IMU. $\mathbf{n}_{f_{k}}\sim\mathcal{N}(\mathbf{0}_{2\times1}, \mathbf{\Sigma}_{f_{k}})$ is the Gaussion noise. Then the linearized measurement error function is derived as:
\begin{equation}
    \label{equation: feature-based linearized error function}
    \mathbf{z}_{f_{j_{k}}}=\widetilde{\mathbf{y}}_{f_{j_{k}}}-\widehat{\mathbf{y}}_{f_{j_{k}}}\approx \mathbf{H}_{f_{j_{k}}}\bm{\xi}_{{k}}+\mathbf{n}_{f_{k}}
\end{equation}
where $\mathbf{H}_{f_{j_{k}}}\triangleq\frac{\partial\mathbf{z}_{f_{j_{k}}}}{\partial\bm{\xi}_{k}} \in \mathbb{R}^{{2} {\times} {(21 + 6c + 3K)}}$ is the observation Jacobian matrix. By stacking all observations from different pose clones, the state and covariance can be updated. Detailed formulations of matrix blocks and corresponding derivations can be found in the supplementary material. 


\section{Map-Based Localization}\label{section: map-based localization}
\begin{figure}[t]
    \centerline{\includegraphics[width=0.47\textwidth]{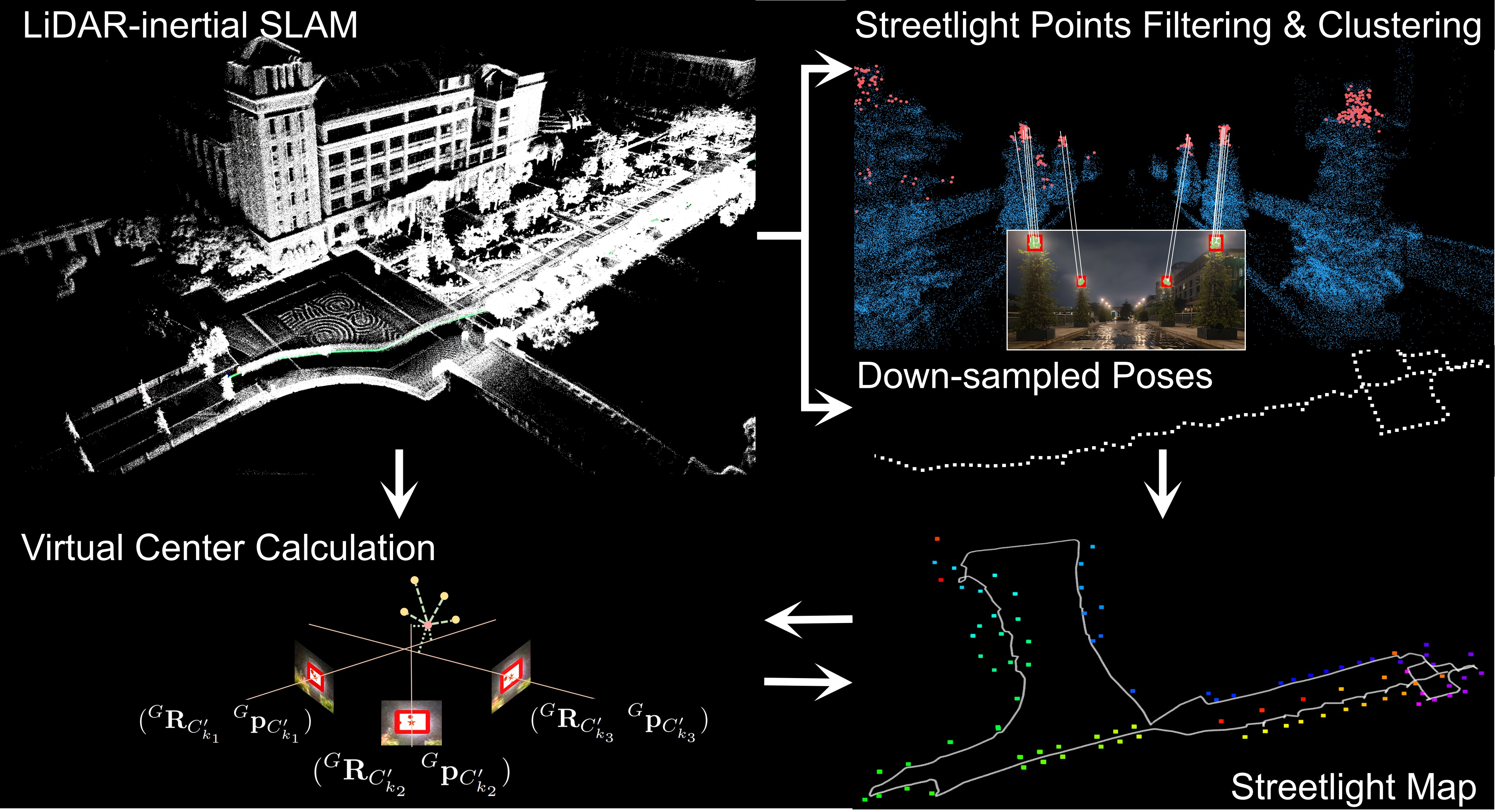}}
    \caption{Generation process of the streetlight map. The points of each LiDAR scan are projected onto the image and the streetlight points are filtered according to streetlight detections in the image. The final streetlight map is constructed by clustering the streetlight points and storing the poses estimated during the mapping process. With the streetlight map, the LiDAR-inertial SLAM is run again to calculate the virtual center of each streetlight cluster.}
    \label{Fig: map creation}
    \vspace{-10pt}
\end{figure}

The essential component for achieving accurate nocturnal visual state estimation is the object-level streetlight map. This section details the generation of the streetlight map and virtual centers, the process of streetlight association and observation, along with the prior pose observation. The map comprises 3D points of streetlights, prior poses generated during the mapping process, and virtual centers of streetlight clusters. The first two can provide prior information, while the virtual centers enable more accurate modeling for visual constraints. 

\subsection{Streetlight Map and  Virtual Center Generation} \label{subsection: Streetlight Map Generation} 
The streetlight mapping process is shown in Fig. \ref{Fig: map creation}. We employ an image-based object detection method \cite{wang2023yolov7} to filter luminous streetlight points from LiDAR measurements to construct the streetlight map. For each LiDAR scan, points whose projections fall into the detection boxes are labeled as ``streetlight''. The original LiDAR map is generated using a modified LiDAR-inertial SLAM system\cite{xu2022fast}, improved with a fast and accurate loop closure detection method\cite{lidar_iris} and a back-end optimization module \cite{dellaert2012factor}. The map points with ``streetlight'' labels \cite{gao2024night} and the prior poses estimated during the mapping process are retained. The final neat 3D streetlight map is displayed in the right bottom of Fig. \ref{Fig: map creation} where the colored blocks denote streetlights and the white dotted line indicates the prior poses. Since the mapping process is offline, one can also fuse additional sensors like GNSS to obtain a more accurate and consistent map. With the streetlight map, we can generate virtual centers for map-based localization. 

\begin{figure}[t] 
    \centering 
    \includegraphics[width=0.45\textwidth]{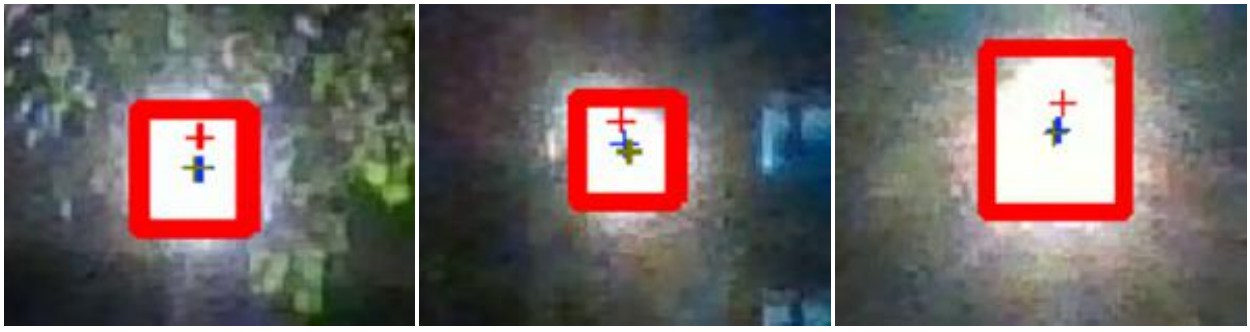} 
    \caption{The projected geometric centers often deviate from streetlight observation centers. The red, blue, and brown crosses are the projected geometric centers, virtual centers, and detection box centers.} 
    \label{Fig: virtual centers}
    \vspace{-10pt}
\end{figure}

It is naive to directly use geometric centers of streetlight clusters and detection box centers to constrain the state. However, as Fig. \ref{Fig: virtual centers} shows, they do not satisfy the reprojection constraint, which inevitably influences the state estimation accuracy. The main reasons are twofold: 1) the centers of light bulbs are not exactly consistent with those of streetlight clusters, and 2) it is difficult to reconstruct the complete shape of streetlights, thus making the centers of streetlight clusters deviate from the true geometric centers. Therefore, we propose the virtual center, which is calculated from both the LiDAR and image data during the mapping process. 

When performing LiDAR-inertial SLAM for the second time, the nearby streetlight clusters are projected onto the image according to the estimated pose. If all the projected points of a streetlight cluster lie within a bounding box, it is considered matching with the streetlight cluster. Therefore, the detection boxes from multiple viewpoints corresponding to a specific streetlight cluster are collected. Suppose there are $Q$ points and $V$ detection boxes from $V$ different views belonging to a streetlight cluster, a distance function of a virtual point ${^{G}}\mathbf{c}_{s}$ is formulated as: 
\begin{align}
\label{Virtual Center}
    \frac{1}{Q}\sum^{Q}_{q=1}\lVert{^{G}}\mathbf{c}_{s}-{^{G}}\widetilde{\mathbf{p}}_{s}^{q}\rVert_{2}^{2}+\frac{\lambda}{V}\sum_{v=1}^{V}\frac{\lVert{^{G}}\widetilde{\mathbf{n}}_{v}-{^{G}}\mathbf{c}_{s} {\times}{^{G}}\widetilde{\mathbf{d}}_{v}\rVert_{2}^{2}}{\lVert{^{G}}\widetilde{\mathbf{d}}_{v}\rVert_{2}^{2}}
\end{align}
where ${^{G}}\widetilde{\mathbf{p}}_{s}^{q}$ is the streetlight point. The first term represents the average distance from the virtual center to all streetlight points, and the second denotes the average distance from the virtual center to the back-projected rays of the detection box centers\cite{zhang2015building}. $\lambda$ is a weight to balance the influence of these two terms. We utilize the Pl\"ucker coordinates ${^{G}}\widetilde{\bm{\mathcal{P}}}_{v}=\left[\begin{smallmatrix} {^{G}}\widetilde{\mathbf{d}}_{v}^{\top} & {^{G}}\widetilde{\mathbf{n}}_{v}^{\top} \end{smallmatrix}\right]^{\top}\in\mathbb{R}^{6}$ \cite{zhang2015building} to represent the back-projection line in the map frame, which is transformed from the camera frame by: 
\begin{align}
    {^{G}}\widetilde{\bm{\mathcal{P}}}_{v}&=\mathbf{Ad}(\begin{bmatrix} {^{G}}\widetilde{\mathbf{R}}_{C^{\prime}_{v}} & {^{G}}\widetilde{\mathbf{p}}_{C^{\prime}_{v}} \\ \mathbf{0}_{1\times 3} & 1 \end{bmatrix})\begin{bmatrix} \mathbf{K}^{-1}\overline{\widetilde{\mathbf{p}}}_{s_{v}} \\ \mathbf{0}_{3\times 1} \end{bmatrix}
\end{align}
where $({^{G}}\widetilde{\mathbf{R}}_{C^{\prime}_{v}}, {^{G}}\widetilde{\mathbf{p}}_{C^{\prime}_{v}})$ is the camera pose from the mapping SLAM system, and $\mathbf{Ad(\cdot)}$ denotes the adjoint matrix \cite{barfoot2024state}. $\overline{\widetilde{\mathbf{p}}}_{s_{v}}$ is the homogeneous coordinate of the detection box center. $\mathbf{K}$ represents the camera intrinsics matrix. By forcing the derivative of (\ref{Virtual Center}) to zero, the virtual center is calculated.

\subsection{Data Association and Streetlight Observation} \label{subsection: streetlight clusters}
Unlike feature points, designing descriptors for cross-modal data association is challenging. Therefore, we propose employing spatial information for data association, which has proven to be sufficiently effective. To maximize the association of image streetlight observations with the streetlight map, we propose a two-stage data association method by leveraging learning-based and binary-based detections.
\subsubsection{Learning-based Data Association} In this stage, 2D streetlight observations are detected using an object detection network, such as \cite{wang2023yolov7}. Suppose that there are $n_{k}$ detected boxes and $m_{k}$ 3D streetlight clusters near the robot's current position. The detected bounding boxes are represented as $\{\widetilde{\bm{\mathcal{B}}}_{d_{k}}=(\widetilde{\mathbf{p}}_{s_{d_{k}}},\widetilde{\bm{l}}_{s_{d_{k}}})\}_{d=1}^{n_{k}}$ where $\widetilde{\mathbf{p}}_{s_{d_{k}}}=\left[\begin{smallmatrix}\widetilde{u}_{s_{d_{k}}}&\widetilde{v}_{s_{d_{k}}}\end{smallmatrix}\right]^{\top}$ is the image coordinate of the box center and $\widetilde{\bm{l}}_{s_{d_{k}}}=\left[\begin{smallmatrix}\widetilde{lu}_{s_{d_{k}}} & \widetilde{lv}_{s_{d_{k}}}\end{smallmatrix}\right]^{\top}$ contains the width and length of box. The streetlight clusters in the map are represented as $\{\widetilde{\bm{\mathcal{L}}}_{g}=({^{G}}\widetilde{\mathbf{c}}_{s_{g}},{^{G}}\widetilde{\mathbf{p}}_{s_{g}}^{1},...,{^{G}}\widetilde{\mathbf{p}}_{s_{g}}^{Q})\}_{g=1}^{m_{k}}$ where ${^{G}}\widetilde{\mathbf{c}}_{s_{g}}$ is the virtual center and ${^{G}}\widetilde{\mathbf{p}}_{s_{g}}^{q}\in \mathbb{R}^{3},q=1,\cdots,Q$ denotes the 3D coordinate of each streetlight point. To handle potential false detections such as car headlights and reflective objects, $(-1)$ is added to the set to indicate the absence of a matching streetlight cluster for any given detection. We score all combinations of streetlight matches and the one with the highest score is determined as the correct combination. The score is calculated based on the following two errors:

\textbf{Reprojection Error.} Given a pose, an incorrect match generally results in a large error between the box center and the projected virtual center of the streetlight cluster. We model and score the residual ${\mathbf{r}}_{p,dg}$ using a Gaussian distribution. Then for the detected box $\widetilde{\bm{\mathcal{B}}}_{d_{k}}$ and the streetlight cluster $\widetilde{\bm{\mathcal{L}}}_{g_{k}}$, the reprojection error-based score $s_{p,dg}$ is calculated as: 
\begin{align}
    \label{equation: reprojection score} 
    s_{p,dg} &= {\mathcal{N}_1({\lVert {{\mathbf{r}}_{p,dg}} \rVert_{2}} | 0, \sigma^{2}_{p,dg})} \\ 
    \label{equation: reprojection error}
    {\mathbf{r}}_{p,dg} &= {\widetilde{\mathbf{p}}_{s_{d_{k}}}} - {\bm{h}({^{C}}\widehat{\mathbf{c}}_{s_{g}})}
\end{align}
where ${\mathcal{N}_1({\lVert {{\mathbf{r}}_{p,dg}} \rVert_{2}} | 0, \sigma^{2}_{p,dg})}$ represents the one-dimensional Gaussian probability density function with mean $0$ and variance $\sigma^{2}_{p,dg}$. ${^{C}}\widehat{\mathbf{c}}_{s_{g}}={^{C}}\mathbf{R}_{I} {^{L}}\widehat{\mathbf{R}}^{^{\top}}_{I_{k}}({^{L}}\widehat{\mathbf{R}}_{G_{k}}{^{G}}\widetilde{\mathbf{c}}_{s_{g}} + {^{L}}\widehat{\mathbf{p}}_{G_{k}} - {^{L}}\widehat{\mathbf{p}}_{I_{k}}) + {^{C}}\mathbf{p}_{I}$ is the projected virtual center in the camera frame. 

\textbf{Angle Error.} We also introduce the score $s_{a,dg}$ based on angle error between the back-projection ray of the center of the box $\widetilde{\bm{\mathcal{B}}}_{d_{k}}$ and the line connecting the virtual center of the streetlight cluster $\widetilde{\bm{\mathcal{L}}}_{g_{k}}$ and the camera optical center: 
\begin{align}
    \label{eequation: angle score}
    s_{a,dg} &= \mathcal{N}_1({\lVert {{\mathbf{r}}_{a,dg}} \rVert_{2}} | 0, \sigma^{2}_{a,dg}) \\ 
    \label{equation: angle error}
    {{\mathbf{r}}_{a,dg}} &= \frac{ {^{C}}\widetilde{\mathbf{p}}_{s_{d_{k}}} {\times} {{^{C}}\widehat{\mathbf{c}}_{s_{g}}} }{\lVert{^{C}}\widetilde{\mathbf{p}}_{s_{d_{k}}}\rVert_{2} \lVert{^{C}}\widehat{\mathbf{c}}_{s_{g}}\rVert_{2}}
\end{align}
where ${^{C}}\widetilde{\mathbf{p}}_{s_{d_{k}}} = \bm{h}^{-1}(\widetilde{\mathbf{p}}_{s_{d_{k}}})$ is the back-projection ray. $\sigma_{p,dg}^{2}$ and $\sigma^{2}_{a,dg}$ are derived in the supplementary material.

The total score of each potential match is calculated by: 
\begin{equation}
\label{equation: total score}
    s_{dg}=\omega s_{p,dg}+(1-\omega)s_{a,dg}
\end{equation}
where $0 \leq \omega \leq 1$ is the preset weight. For the streetlight $\widetilde{\bm{\mathcal{B}}}_{d_{k}}$ without any correspondence, the score is calculated as $s_{d(m_{k}+1)} = 1 - \sum_{g = 1}^{m_{k}} {s_{dg}}$.

Based on (\ref{equation: total score}), each potential match is evaluated and assigned a score. We use $l_{dg}$ to indicate whether the $d$-th streetlight observation matches the $g$-th 3D streetlight cluster. $\widetilde{\bm{\mathcal{D}}}_{k} = \{((\widetilde{\bm{\mathcal{B}}}_{d_{k}}, \widetilde{\bm{\mathcal{L}}}_{g_{k}}) | l_{dg} = 1)\}_{d = 1}^{n_{k}}$ denotes the possible match combinations. The set of clusters is augmented to $m_{k}+n_{k}$, where the extra $n_{k}$ elements represent unmatched streetlight observations, i.e., $\widetilde{\bm{\mathcal{B}}}_{d_{k}}$ is assigned to $(-1)$. In this way, $\widetilde{\bm{\mathcal{D}}}_{k}$ is made injective, meaning that each observation corresponds uniquely to one element in the set of clusters. To obtain the optimal combination $\widetilde{\bm{\mathcal{D}}}^{\star}_{k}$, the problem is formulated as the classical assignment problem:
\begin{align}
\label{equation: assignment problem} 
    & \widetilde{\bm{\mathcal{D}}}^{\star}_{k}\! =\! \mathop{\arg\max} \limits_{\widetilde{\bm{\mathcal{D}}}_{k}}\! \sum_{d = 1}^{n_{k}}\! \sum_{g = 1}^{m_{k} + n_{k}}\! l_{dg} s_{dg} , \  \text{s.t. } \sum_{d = 1}^{n_{k}}\! l_{dg}\! \leq\! 1, \sum_{g = 1}^{m_{k} + n_{k}}\! l_{dg}\! =\! 1 . \notag 
\end{align}
Using Hungarian algorithm\cite{kuhn1955hungarian}, the optimization problem can be efficiently solved ($\sim$0.2 ms per frame). It should be noted that when there are few visible streetlights, the cumulative error of state increases the above two match errors, making it possible to find no matches, i.e., $\widetilde{\bm{\mathcal{D}}}_{k}^{\star}=\{(\widetilde{\bm{\mathcal{B}}}_{d_{k}}, -1)\}^{n_{k}}_{d=1}$. However, since we model the score as a Gaussian distribution whose variance is related to the covariance of state, the increased covariance caused by few streetlights makes the proposed method able to accept matches with larger errors. In other words, our method can adaptively lower or raise the requirement for finding matches, which, to some extent, avoids the problem of tracking failure caused by few observations. 

For each match of $\widetilde{\bm{\mathcal{D}}}^{\star}_{k}$, the reprojection constraint between the virtual center ${^{G}}\widetilde{\mathbf{c}}_{s}$ of streetlight cluster and the box center $\widetilde{\mathbf{p}}_{s}$ of streetlight observation is formulated as: 
\begin{equation}
    \widetilde{\mathbf{y}}_{s_{k}} \! = \! \bm{h} \left( ^{C}\mathbf{R}_{I} {^{L}}\mathbf{R}_{I_{k}}^{\top}({^{L}}\mathbf{R}_{G_{k}}{^{G}}\widetilde{\mathbf{c}}_{s} \! + \! {^{L}}\mathbf{p}_{G_{k}} \! - \! {^{L}}\mathbf{p}_{I_{k}}) + {^{C}}\mathbf{p}_{I} \right) + \mathbf{n}_{s_{k}}
\end{equation}
where $\widetilde{\mathbf{y}}_{s_{k}}=\widetilde{\mathbf{p}}_{s}$ and $\mathbf{n}_{s_{k}}\sim\mathcal{N}(\mathbf{0}_{2\times1},\mathbf{\Sigma}_{s_{k}})$ is the Gaussian measurement noise. The linearized error function is: 
\begin{equation}
    \label{equation: streetlight-based linearized error function}
    \mathbf{z}_{s_{k}}=\widetilde{\mathbf{y}}_{s_{k}}-\widehat{\mathbf{y}}_{s_{k}} \approx\mathbf{H}_{s_{k}}\bm{\xi_{k}}+\mathbf{n}_{s_{k}}
\end{equation}
where the Jacobian $\mathbf{H}_{s_{k}}\triangleq\frac{\partial\mathbf{z}_{s_{k}}}{\partial\bm{\xi}_{k}} \in \mathbb{R}^{{2} {\times} {(21 + 6c + 3K)}}$. Derivation details are provided in the supplementary material. 

\begin{figure}[t] 
    \centering 
    \includegraphics[width=0.47\textwidth]{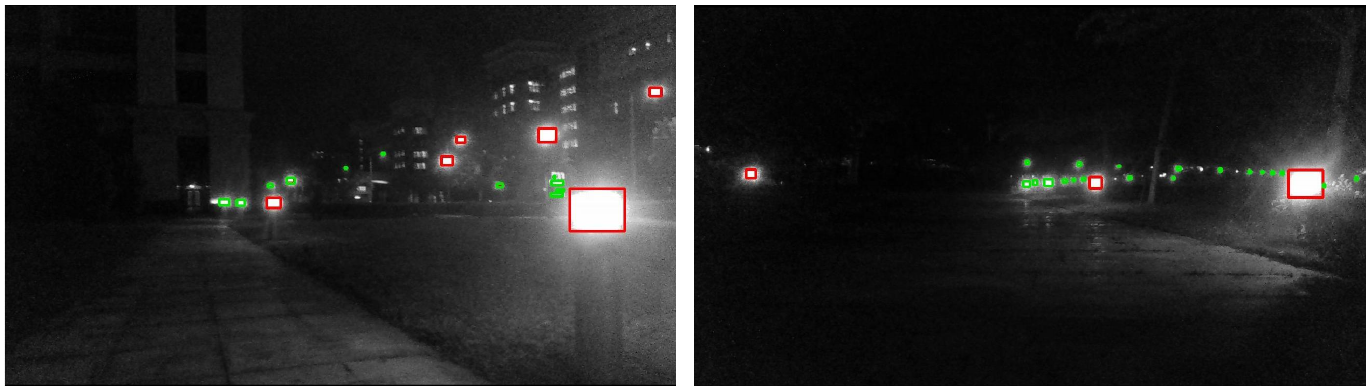} 
    \caption{Examples of match extension. The red boxes represent the detection by the learning-based method and the green boxes are detected by the binary-based method. The images show that the learning-based method cannot detect small streetlights while the binary-based method makes up for this shortcoming.}
    \label{Fig: match_extenstion}
    \vspace{-10pt}
\end{figure}

\subsubsection{Binary-based Data Association}
In the first stage, small streetlights are frequently missed by the object detection network, leading to the underutilization of potential streetlight matches. To remedy this issue, we introduce the binary-based streetlight detection method in the second stage. Fig. \ref{Fig: match_extenstion} shows the examples of match extension. Pixels with intensity above a specific threshold are processed by the contour tracing algorithm \cite{suzuki1985topological} to generate bounding boxes that represent streetlight observations. With the updated pose, all streetlight clusters without 2D matches are projected onto the image. For a single streetlight cluster, since the updated pose is relatively accurate, all boxes containing the projected points are considered as the candidate matches. For each box, we then calculate the ratio of the number of projected points within the box to the total point number within the cluster. The box with the maximum ratio is determined as the match for the cluster. By traversing all streetlight clusters, new matches are identified, enabling an update of the robot state. In addition, historical streetlight matches are preserved for state update. This strategy mitigates the impact of streetlight mismatches within the current frame, thereby enhancing robustness and accuracy. 

\subsection{Prior Pose Observation} \label{section: prior poses}
Since streetlights are commonly installed along both sides of roads, the wheeled robot is required to travel these roads multiple times to collect streetlights during the mapping process. When the robot is in operation, it typically navigates along the nearby roads, resulting in similar trajectories in mapping and localization. This similarity inspires us to preserve the accurately estimated poses during mapping. These prior poses are leveraged in three ways: 1) Provide constraints for localization. 2) Divide the streetlight map into different regions to accelerate initialization. 3) Exclude unreasonable solutions in the initialization and tracking recovery modules. The last two points are elaborated on in Section \ref{section: initialization} and Section \ref{section: tracking recovery} and this section focuses on the first point. 

Using the estimated poses ${^{L}}\widehat{\mathbf{T}}_{I_{k}}$ and ${^{L}}\widehat{\mathbf{T}}_{G_{k}}$, radius search is first conducted to identify the nearby prior poses. The prior poses ${^{G}}\widetilde{\mathbf{T}}_{I^{\prime}}=({^{G}}\widetilde{\mathbf{R}}_{I^{\prime}},{^{G}}\widetilde{\mathbf{p}}_{I^{\prime}})$ within a threshold distance from the current pose, i.e., $\lVert{^{L}}\widehat{\mathbf{R}}_{G_{k}}^{\top}({^{L}}\widehat{\mathbf{p}}_{I_{k}}-{^{L}}\widehat{\mathbf{p}}_{G_{k}})-{^{G}}\widetilde{\mathbf{p}}_{I^{\prime}}\rVert<\sigma$, are selected to impose planar constraints. We assume that the road surface is locally planar, thus the constraint is: 
\begin{equation}
    \label{equation: prior pose constraint}
    \widetilde{\mathbf{y}}_{p_{k}}\!\!=\!\!\begin{bmatrix} ({^{G}}\widetilde{\mathbf{R}}_{I^{\prime}} {^{O}}\mathbf{R}_{I}^{\top} \mathbf{e}_{3})^{\top} ({^{L}}\mathbf{R}_{G_{k}}^{\top}({^{L}}\mathbf{p}_{I_{k}}\!-\!{^{L}}\mathbf{p}_{G_{k}})\!-\!{^{G}}\widetilde{\mathbf{p}}_{I^{\prime}}) \\ ({^{G}}\widetilde{\mathbf{R}}_{I^{\prime}} {^{O}}\mathbf{R}_{I}^{\top} \mathbf{e}_{3})^{\top} ({^{L}}\mathbf{R}_{G_{k}}^{\top}{^{L}}\mathbf{R}_{I_{k}}{^{O}}\mathbf{R}_{I}^{\top}\mathbf{e}_{3}) \end{bmatrix}+\mathbf{n}_{p_{k}}
\end{equation}
where $\widetilde{\mathbf{y}}_{p_{k}}\!=\!{\setlength\arraycolsep{2.5pt}\begin{bmatrix} 0 & 1 \end{bmatrix}^{\top}}$, $\mathbf{e}_{3}\!=\!{\setlength\arraycolsep{2.5pt}\begin{bmatrix}0 & 0 & 1\end{bmatrix}^{\top}}$, and the Gaussian noise $\mathbf{n}_{p_{k}}\!\sim\! \mathcal{N}(\mathbf{0}_{2\times 1}, \mathbf{\Sigma}_{p_{k}})$. 

With (\ref{equation: prior pose constraint}), we can derive the linearized error function as: 
\begin{equation}
    \label{equation: prior pose-based linearized error function}
    \mathbf{z}_{p_{k}}=\widetilde{\mathbf{y}}_{p_{k}}-\widehat{\mathbf{y}}_{p_{k}}\approx\mathbf{H}_{p_{k}}\bm{\xi}_{k}+\mathbf{n}_{p_{k}}
\end{equation}
where the Jacobian matrix $\mathbf{H}_{p_{k}}\triangleq\frac{\partial\mathbf{z}_{p_{k}}}{\partial\bm{\xi}_{k}} \in \mathbb{R}^{{2} {\times} {(21 + 6c + 3K)}}$. More details are provided in the supplementary material. 


\section{Initialization} \label{section: initialization}
\begin{figure}[t] 
    \centering 
    \includegraphics[width=0.47\textwidth]{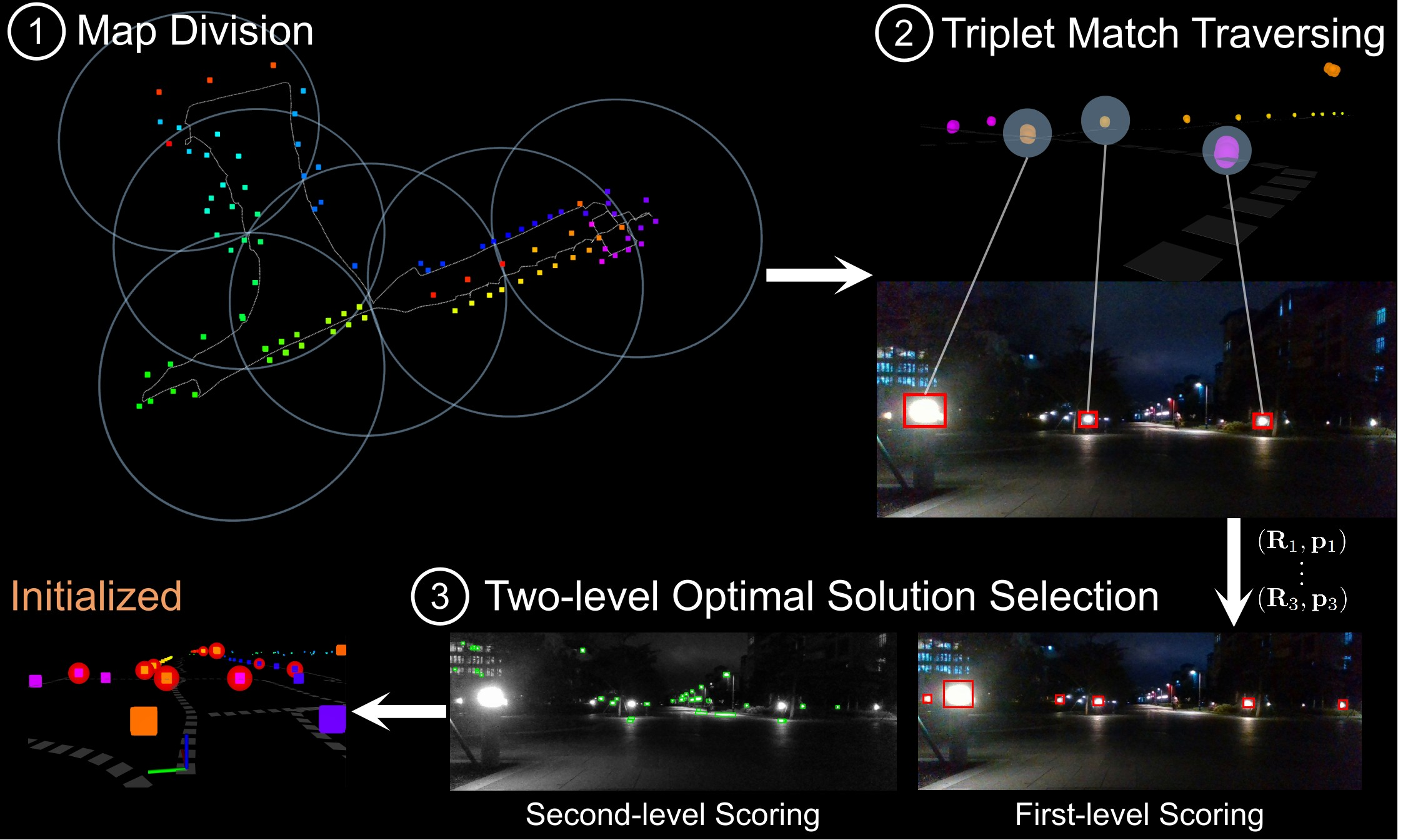} 
    \caption{Process of initialization. The streetlights are first divided into different regions according to the sampled prior poses. Next, all possible triplet matches are fed into the P3P algorithm \cite{ding2023revisiting} concurrently for solving the corresponding poses. The optimal solution is ultimately selected based on the designed two-level scoring strategy.} 
    \label{Fig: pose initialization}
    \vspace{-10pt}
\end{figure}

Accurate initial estimates are essential for estimator convergence. However, in the absence of manual intervention or GNSS, the initialization is restricted to rely solely on visual perception and the streetlight map, making it an extremely challenging task. Hence, we design an initialization module to efficiently estimate the initial relative transformation between the map frame and the local frame at system startup. For robust initialization, at least six streetlight detections are required. As shown in Fig. \ref{Fig: pose initialization}, the initialization is conducted in a divide-and-conquer manner with the following three steps. 

\textbf{Map Division.} Leveraging the sparsity of the streetlight map, we extract all the triplets of streetlights, thus transforming the initialization problem into determining the optimal P3P solution\cite{ding2023revisiting}. However, given $M$ streetlight clusters and $N$ detections, the number of possible triplet matches is $C_{N}^{3}A_{M}^{3}$, making it impractical to traverse them all. Fortunately, the hidden traversable area information in prior poses prompts a reasonable division of streetlight map. We sample the prior poses at intervals of $\rho$ meters, with each pose corresponding to a circular region with radius $\rho$. This sampling reduces the total number of triplet matches to $C_{N}^{3}\sum_{r=1}^{R}A_{M_{r}}^{3}$, where $M_{r}$ is the number of streetlights within the $r$-th region. Additionally, the division strategy enables parallel computation, which further accelerates the iteration through triplet matches. 

\textbf{Triplet Match Traversing.} Each triplet match is processed via the P3P algorithm \cite{ding2023revisiting} to derive a pose solution in $SE(3)$. Unreasonable solutions are discarded if the pose estimation deviates largely from the nearest prior pose in $z$-axis.

\textbf{Optimal Solution Selection.} For the remaining solutions, we design a two-level optimal solution selection method. In the first level, we use the solutions to find the matches of unselected streetlight detections. Given $n_{0}$ unmatched detections and $m_{0}$ unmatched streetlight clusters in a region, each potential match is scored using the sine of angle error in (\ref{equation: angle error}). We use the Hungarian algorithm \cite{kuhn1955hungarian} for the optimization:
\begin{equation}
    \widetilde{\bm{\mathcal{D}}}^{\star}_{0}\! =\! \mathop{\arg\min} \limits_{\widetilde{\bm{\mathcal{D}}}_{0}}\! \sum_{d = 1}^{n_{0}}\! \sum_{g = 1}^{m_{0} + n_{0}}\! l_{dg} r_{a,dg}, \  \text{s.t. } \sum_{d = 1}^{n_{0}}\! l_{dg}\! \leq\! 1, \! \sum_{g = 1}^{m_{0} + n_{0}}\! l_{dg}\! =\! 1. \notag 
\end{equation}
The optimal matches for the detections can be identified. To better distinguish the quality of different solutions, we calculate the total reprojection error in (\ref{equation: reprojection error}) across all matches as the punishment score $s_{1}$. The top-$p$ solutions with the lowest scores in each region are then selected. 

In the second level, we lower the threshold for binary-based streetlight detection. This adjustment increases the number of detections, but also introduces more noise from non-streetlight light sources. Each detection box is associated with the nearest projected streetlight cluster and the distance is used to calculate a reward score according to (\ref{equation: reprojection score}). Let $s_{2}$ denote the summation of all rewards. The total score in the second level for each solution is given by $s=-s_{1}+\gamma s_{2}$, where $\gamma$ is a preset weight. The solution with the highest score $s$ is set to the initial estimate of the relative transformation. 


\section{Tracking Recovery} \label{section: tracking recovery}
In the map regions with sparsely distributed streetlights, the robot may experience extended periods without streetlight detections. The accumulated pose drift over time makes it difficult to identify the correct matches for new streetlight detections, potentially resulting in tracking failures. To improve robustness, a tracking recovery module is proposed \cite{gao2024night}. 

When the robot travels for a distance without matching streetlight detections, the tracking is considered lost, and the tracking recovery module is triggered. The limited number of streetlight detections makes it computationally feasible to employ the brute-force method to enumerate all possible streetlight match combinations. To select the optimal match, we harness the key insight that incorrect match combinations result in increasing deviations of the estimated trajectories from the true values, whereas the correct match combination effectively minimizes pose drifts. Specifically, the state is cloned multiple times, with each copy being updated according to the corresponding match combination. The states with poses that deviate a lot from the nearest prior poses in the $z$-axis are discarded. All the remaining states will be propagated and updated in parallel until the traveled distance exceeds the set threshold and at least two streetlights are detected. During this process, incorrect match combinations exacerbate state errors, whereas correct matches constrain pose drifts, leading to significant differences in reprojection errors. We then re-adopt the two-level scoring strategy in the initialization module to identify the optimal state. The optimal state is regarded as the recovered state, indicating that the state is successfully corrected and ensuring the estimated trajectory continuity. 


\section{Feature-Decoupled MSC-InEKF} \label{section: feature-decoupled MSC-InEKF}
Most InEKF-based VIO systems \cite{brossard2018invariant} \cite{zhang2023toward} \cite{wu2017invariant} maintain the positions of visual features within the body navigation state to form the $SE_{2+K}(3)$ group. However, this feature-coupled approach results in the state transition matrix containing feature-related elements, which increases computational costs during propagation \cite{yang2022decoupled}. Herein, we analyze the prevalent feature-coupled MSC-InEKF, termed MSC-InEKF-FC. Following an in-depth theoretical analysis of multiple solutions, we propose a novel feature-decoupled MSC-InEKF that significantly improves efficiency without sacrificing accuracy. 

\subsection{MSC-InEKF-FC}
For MSC-InEKF-FC, the feature positions are associated with the body navigation state to form the $SE_{2+K}(3)$ group. In the system state, $\mathbf{B}_{k}$ and $\mathbf{X}_{\bm{C}_{k}}$ remain unchanged as in (\ref{equation: all states}), while $\mathbf{X}_{\bm{I}_{k}}$ and $\mathbf{X}_{\bm{G}_{k}}$ are modified as: 
\begin{gather}
    \mathbf{X}_{\bm{I}_{k}}\!=\!{\setlength\arraycolsep{0.4pt}\begin{bmatrix}
        \begin{array}{c|ccc}
            {^{L}}\mathbf{R}_{I_{k}} & {^{L}}\mathbf{p}_{I_{k}} & {^{L}}\mathbf{v}_{I_{k}} & {^{L}}\mathbf{p}_{f_{k}} \\\hline
            \mathbf{0}_{3\times3} & \multicolumn{3}{c}{\mathbf{I}_{3\times3}} 
        \end{array}
    \end{bmatrix}} , \mathbf{X}_{\bm{G}_{k}}\!=\!{\setlength\arraycolsep{0.4pt}\begin{bmatrix}
        {^{L}}\mathbf{R}_{G_{k}} & {^{L}}\mathbf{p}_{G_{k}} \\ \mathbf{0}_{1\times 3} & 1 \\
    \end{bmatrix}}.
\end{gather}
All the error states remain unchanged except for the feature error state, which is reformulated as: 
\begin{equation}
    \label{equation: feature error state FC}
    {^{L}}\mathbf{p}_{f_{k}}=\mathrm{Exp}(\bm{\xi}_{\bm{R}_{LI_{k}}}){^{L}}\widehat{\mathbf{p}}_{I_{k}}+\mathbf{J}_{l}(\bm{\xi}_{\bm{R}_{LI_{k}}})\bm{\xi}_{\bm{p}_{Lf_{k}}}.
\end{equation}
Following similar procedures in Section \ref{subsection: state propagation}, the state transition matrix and the noise Jacobian matrix are formulated as: 
\begin{gather}
{\setlength\arraycolsep{0.88pt}
    \bm{\Phi}_{i}^{i\!+\!1} \!\! = \!\!\! \begin{bmatrix}
        \bm{\Phi}_{\bm{I}\bm{I}} & \bm{\Phi}_{\bm{I}\bm{B}} & \mathbf{0}_{9\times 6c} & \mathbf{0}_{9\times 6} & \mathbf{0}_{9\times 3}\\
        \mathbf{0}_{6\times 9} & \mathbf{I}_{6\times 6} & \mathbf{0}_{6\times 6c} & \mathbf{0}_{6\times 6} & \mathbf{0}_{6\times 3}\\
        \mathbf{0}_{6c\times 9} & \mathbf{0}_{6c\times 6} & \mathbf{I}_{6c\times 6c} & \mathbf{0}_{6c\times 6} & \mathbf{0}_{6c\times 3}\\
        \mathbf{0}_{6\times 9} & \mathbf{0}_{6\times 6} & \mathbf{0}_{6\times 6c} & \mathbf{I}_{6\times 6} & \mathbf{0}_{6\times 3} \\
        \mathbf{0}_{3\times 9} & \mathbf{\Phi}_{\bm{f}\bm{B}} & \mathbf{0}_{3\times 6c} & \mathbf{0}_{3\times 6} & \mathbf{I}_{3\times 3} 
    \end{bmatrix}} \! , \! 
{\setlength\arraycolsep{0.88pt}
    \mathbf{G}_{i} \!\! = \!\!\! \begin{bmatrix}
        \mathbf{G}_{\bm{I}\bm{I}} & \mathbf{0}_{9\times 6}\\
        \mathbf{0}_{6\times 6} & \mathbf{I}_{6\times 6}\delta t \\
        \mathbf{0}_{6c\times 6} & \mathbf{0}_{6c\times 6} \\
        \mathbf{0}_{6\times 6} & \mathbf{0}_{6\times 6} \\
        \mathbf{G}_{\bm{f}\bm{I}} & \mathbf{0}_{3\times 6}
    \end{bmatrix}} \notag 
\end{gather}
where the block components $\mathbf{\Phi}_{\bm{I}\bm{I}}$, $\mathbf{\Phi}_{\bm{I}\bm{B}}$, and $\mathbf{G}_{\bm{I}\bm{I}}$ have the same formulations with (\ref{equation: components of transition matrix}). $\mathbf{\Phi}_{\bm{f}\bm{B}}$ and $\mathbf{G}_{\bm{f}\bm{I}}$ are derived as: 
\begin{equation}
    \mathbf{\Phi}_{\bm{f}\bm{B}}=\mathbf{G}_{\bm{f}\bm{I}}=\begin{bmatrix}
        -({^{L}}\widehat{\mathbf{p}}_{f_{i+1}})_{\times} \Delta \mathbf{R}\mathbf{J} & \mathbf{0}_{3\times 3}
    \end{bmatrix}. 
\end{equation}
Notably, $\mathbf{\Phi}_{\bm{fB}}$ causes an extra calculation in covariance propagation. Given $K$ features, the covariance propagation requires a computation of $O(K^{2})$ flops. Compared to MSCKF which only requires a level of about $O(K)$ \cite{yang2022decoupled}, the MSC-InEKF-FC process slows considerably when the number of features is large. The underlying reason is that the features are associated with the body navigation state, thus the state transition matrix is correlated with the features through (\ref{equation: feature error state FC}).

\subsection{MSC-InEKF-FDN}
A naive feature-decoupled method (termed MSC-InEKF-FDN) represents the features in $\mathbb{R}^{3}$ space, without maintaining them in any Lie group. In this way, only the relative transformation in $\mathbf{X}_{\bm{G}_{k}}$ is changed as following: 
\begin{gather}\label{equation: all states in MSC-InEKF-FDN}
\mathbf{X}_{\bm{G}_{k}} = \begin{bmatrix}
    {^{L}}\mathbf{R}_{G_{k}} & {^{L}}\mathbf{p}_{G_{k}} \\    \mathbf{0}_{1\times 3} & 1
\end{bmatrix}.
\end{gather}
The error state of the feature is formulated as: 
\begin{equation}
    {^{L}}\mathbf{p}_{f_{k}}={^{L}}\widehat{\mathbf{p}}_{f_{k}}+{\bm{\xi}_{\mathbf{p}_{Lf_{k}}}}.
\end{equation}
The state transition matrix and the noise Jacobian matrix share the same formulation as (\ref{equation: linearized function}). Although MSC-InEKF-FDN decouples the feature position states, the observability analysis reveals that it suffers from the inconsistency issue, introducing spurious information to state estimation. We first present the inherent unobservable dimensions of the system. The local frame aligns with gravity. When applying a 4-DoF transformation (three for translation and one for rotation around the gravity) to the local frame, the relative pose from frame $I$ to $G$ is not affected. Therefore, the observations provided by prior poses and streetlight clusters remain unchanged. Similar to visual-inertial SLAM systems, the observations of features and velocity are also not affected \cite{zhang2023toward}. In summary, the dimension of the unobservable subspace is 4. 

The observability matrix of the system is: 
\begin{equation}
    \bm{\mathcal{O}}\!=\!{\setlength\arraycolsep{0.6pt}\begin{bmatrix}
        \bm{\mathcal{O}}_{0}^{\top} & \bm{\mathcal{O}}_{1}^{\top} & \cdots & \bm{\mathcal{O}}_{k}^{\top}
    \end{bmatrix}\!^{\top}}\!=\!{\setlength\arraycolsep{0.6pt}\begin{bmatrix}
        \bm{\mathbf{H}}_{0}^{\top} & \mathbf{\Phi}^{1^{\top}}_{0}\!\bm{\mathbf{H}}_{1}^{\top} & \cdots & \mathbf{\Phi}^{k^{\top}}_{0}\!\bm{\mathbf{H}}_{k}^{\top}
    \end{bmatrix}\!^{\top}}
\end{equation}
where $\mathbf{H}_{k}=\left[\begin{smallmatrix} \mathbf{H}_{o_{k}}^{\top} & \mathbf{H}_{f_{k}}^{\top} & \mathbf{H}_{s_{k}}^{\top} & \mathbf{H}_{p_{k}}^{\top} \end{smallmatrix}\right]^{\top}$. For simplicity, we do not include the biases in the state vector as they are proved to be observable \cite{li2013high} for general motions. In addition, both MSC-InEKF and InEKF are derived based on the same propagation model and linearized error models with only the observation Jacobians calculated using different estimates \cite{li2013high}. Therefore, it is equivalent to the InEKF observability analysis, as long as we adjust the linearization points. The final considered state for MSC-InEKF-FDN is given by $\mathbf{X}_{k}=(\mathbf{X}_{\bm{I}_{k}},\mathbf{X}_{\bm{G}_{k}},{^{L}}\mathbf{p}_{f_{k}})$. Then, the right null space $\bm{\mathcal{N}}_{N}$ is: 
\begin{equation}
    \bm{\mathcal{N}}_{N}=\setlength\arraycolsep{2.2pt}{\begin{bmatrix}
        \mathbf{g}^{\top} & \mathbf{0}_{1\times3} & \mathbf{0}_{1\times3} & \mathbf{0}_{1\times3} & \mathbf{0}_{1\times3} & \mathbf{g}^{\top}({^{L}}\mathbf{\widehat{p}}_{f_{k}})_{\times} \\
        \mathbf{0}_{3\times3} & \mathbf{I}_{3\times3} & \mathbf{0}_{3\times3} & \mathbf{0}_{3\times3} & \mathbf{I}_{3\times3} & \mathbf{I}_{3\times3}
    \end{bmatrix}^{\top}}. 
\end{equation}
It shows that the unobservable subspace is affected by the feature estimate ${^{L}}\widehat{\mathbf{p}}_{f_{k}}$, which causes an inconsistency problem.

\begin{figure}[t] 
    \centering 
    \includegraphics[width=0.45\textwidth]{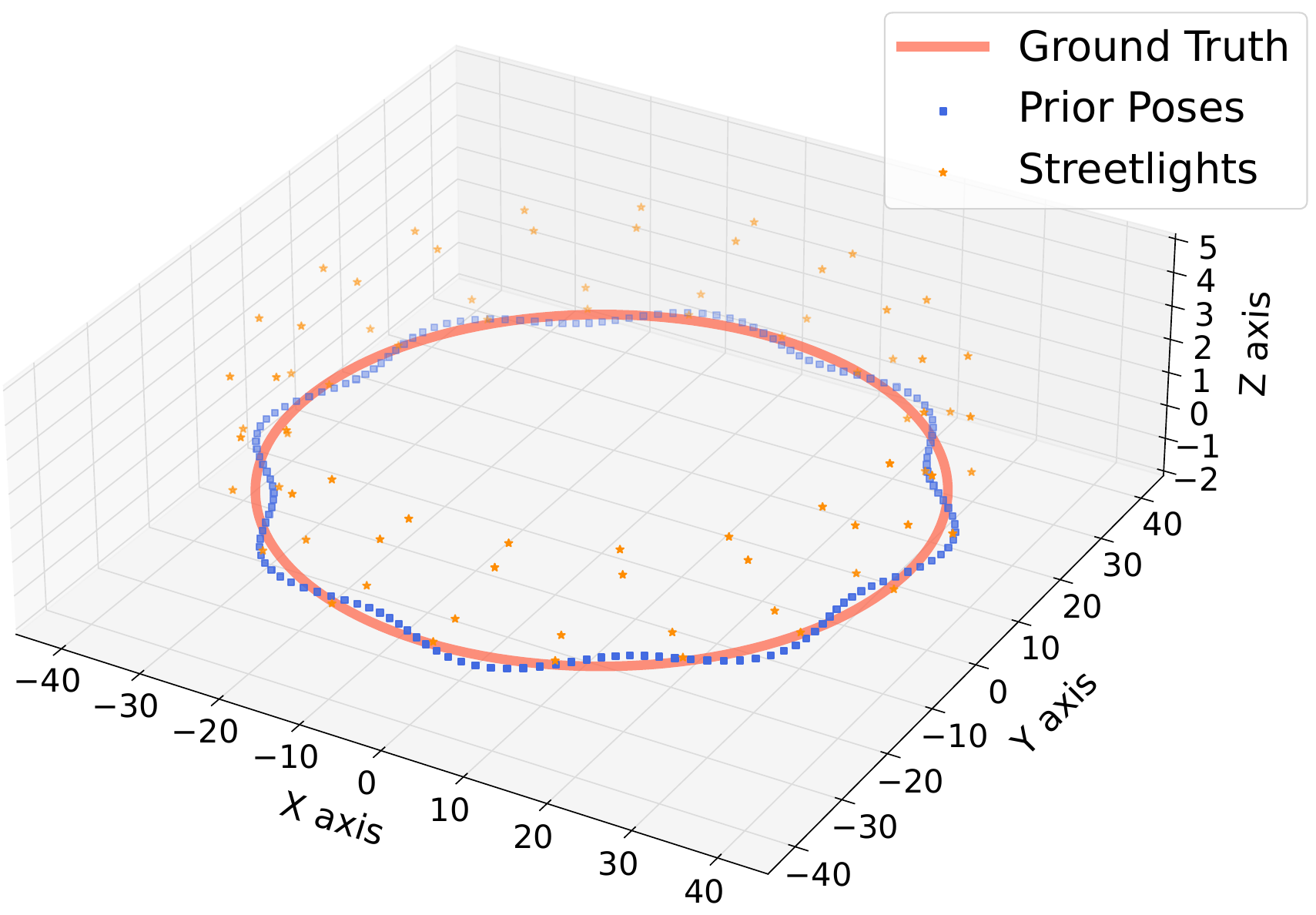} 
    \caption{The data used in the simulation. The pink trajectory is the ground-truth trajectory. The blue squares are prior poses generated in the mapping process and the orange stars represent the streetlights.} 
    \label{Fig: simulation scene}
    \vspace{-10pt}
\end{figure}

\subsection{MSC-InEKF-FDR} \label{section: MSC-InEKF-FDR}
Another feature-decoupled method associates the feature with the relative transformation in $\mathbf{X}_{\bm{G}_{k}}$ to form $SE_{1+K}(3)$. Notably, this method is natural, since both the feature position state and the relative transformation state exhibit identical dynamics, characterized by stable states with zero velocities. We term it as MSC-InEKF-FDR. The state definition, propagation, and update are provided in the previous sections. The right null space of the observability matrix is spanned by: 
\begin{equation}
    \label{right null space of MSC-InEKF-FDR1}
    \bm{\mathcal{N}}_{R_{1}}=\setlength\arraycolsep{1.5pt}{\begin{bmatrix}
        \mathbf{g}^{\top} & \mathbf{0}_{1\times3} & \mathbf{0}_{1\times3} & \mathbf{g}^{\top} & \mathbf{0}_{1\times3} & \mathbf{0}_{1\times3} \\
        \mathbf{0}_{3\times3} & \mathbf{I}_{3\times3} & \mathbf{0}_{3\times3} & \mathbf{0}_{3\times3} & \mathbf{I}_{3\times3} & \mathbf{I}_{3\times3}
    \end{bmatrix}^{\top}}. 
\end{equation}

Unlike MSC-InEKF-FDN, the unobservable subspace of MSC-InEKF-FDR is not affected by the state estimate, thus ensuring the consistency of the system. However, when the map-based information (i.e., streetlights and prior poses) is not available, the unobservable subspace of the system is: 
\begin{equation}
\label{equation: observability without streetlights}
\bm{\mathcal{N}}_{R_{2}}=\setlength\arraycolsep{3.5pt}{\begin{bmatrix}
        \mathbf{g}^{\top} & \mathbf{0}_{1\times3} & \mathbf{0}_{1\times3} & \mathbf{g}^{\top} & \mathbf{0}_{1\times3} & \mathbf{0}_{1\times3} \\
        \mathbf{0}_{3\times3} & \mathbf{I}_{3\times3} & \mathbf{0}_{3\times3} & \mathbf{0}_{3\times3} & \mathbf{0}_{3\times3} & \mathbf{I}_{3\times3} \\
        \mathbf{0}_{3\times3} & \mathbf{0}_{3\times3} & \mathbf{0}_{3\times3} & {^{L}}\widehat{\mathbf{p}}_{f_{k}} & \mathbf{0}_{3\times3} & \mathbf{0}_{3\times3} \\
        \mathbf{0}_{3\times3} & \mathbf{0}_{3\times3} & \mathbf{0}_{3\times3} & \mathbf{0}_{3\times3} & \mathbf{I}_{3\times3} & \mathbf{0}_{3\times3}
    \end{bmatrix}^{\top}}. 
\end{equation}
The lack of streetlight observations and prior poses results in insufficient constraints on the relative transformation, thus the dimension of unobservable subspace should be 10. However, the existence of ${^{L}}\widehat{\mathbf{p}}_{f_{k}}$ leads to inconsistency.

\subsection{MSC-InEKF-FDRC} \label{section: MSC-InEKF-FDRC}
Based on MSC-InEKF-FDR, we propose MSC-InEKF-FDRC that changes the feature association to the closest cloned pose $\mathbf{X}_{CP_{k_{k}}}$ when no streetlight match is obtained. Therefore, the definitions of $\mathbf{X}_{CP_{k_{k}}}$ and $\mathbf{X}_{\bm{G}_{k}}$ are updated as: 
\begin{gather}\label{equation: all states in MSC-InEKF-FDRC}
    \mathbf{X}_{CP_{k_{k}}}\!=\!{\setlength\arraycolsep{0.8pt}\begin{bmatrix}
        \begin{array}{c|cc}
        {^{L}}\mathbf{R}_{I_{k_{k}}} & {^{L}}\mathbf{p}_{I_{k_{k}}} &
        {^{L}}\mathbf{p}_{f_{k}} \\\hline
        \mathbf{0}_{2\times3} & \multicolumn{2}{c}{\mathbf{I}_{2\times2}}
        \end{array}
    \end{bmatrix}, \mathbf{X}_{\bm{G}_{k}}\!=\!{\setlength\arraycolsep{0.8pt}\begin{bmatrix}
    {^{L}}\mathbf{R}_{G_{k}} & {^{L}}\mathbf{p}_{G_{k}} \\    \mathbf{0}_{1\times 3} & 1 \\
\end{bmatrix}}}. \notag 
\end{gather}
The formulation of other states is the same as (\ref{equation: all states}). The error state of the feature is then changed to: 
\begin{equation}
    {^{L}}\mathbf{p}_{f_{k}}=\mathrm{Exp}(\bm{\xi}_{\bm{R}_{CP_{k_{k}}}}){^{L}}\widehat{\mathbf{p}}_{f_{k}}+\mathbf{J}_{l}(\bm{\xi}_{\bm{R}_{CP_{k_{k}}}}){\bm{\xi}_{\bm{p}_{Lf_{k}}}}. 
\end{equation}
The change in the representation of the feature error state results in the modification of the covariance. By denoting the state before the change with subscript 0 and the state after the change with subscript 1, we have: 
\begin{gather}
    {^{L}}\mathbf{p}_{f}=\mathrm{Exp}(\bm{\xi}_{\bm{R}_{0}}){^{L}}\widehat{\mathbf{p}}_{f_{0}}+\mathbf{J}_{l}(\bm{\xi}_{\bm{R}_{0}})\bm{\xi}_{\bm{p}_{Lf_{0}}}\notag\\
    {^{L}}\mathbf{p}_{f}=\mathrm{Exp}(\bm{\xi}_{\bm{R}_{1}}){^{L}}\widehat{\mathbf{p}}_{f_{1}}+{\mathbf{J}_{l}(\bm{\xi}_{\bm{R}_{1}})\bm{\xi}_{\bm{p}_{Lf_{1}}}}. 
\end{gather}
The feature state estimate remains unchanged, i.e., ${^{L}}\widehat{\mathbf{p}}_{f}\triangleq{^{L}}\widehat{\mathbf{p}}_{f_{0}}={^{L}}\widehat{\mathbf{p}}_{f_{1}}$, hence the new feature error state is: 
\begin{equation}
    \label{equation: new feature error state}
    \bm{\xi}_{\bm{p}_{Lf_{1}}}\approx-({^{L}}\widehat{\mathbf{p}}_{f})_{\times}\bm{\xi}_{\bm{R}_{0}}+({^{L}}\widehat{\mathbf{p}}_{f})_{\times}\bm{\xi}_{\bm{R}_{1}}+{\bm{\xi}_{\bm{p}_{Lf_{0}}}}. 
\end{equation}
The new covariance is then updated with (\ref{equation: new feature error state}). For the new association way, the right null space is formulated as: 
\begin{equation}
    \bm{\mathcal{N}}_{C}=\begin{bmatrix}
        \mathbf{g}^{\top} & \mathbf{0}_{1\times3} & \mathbf{0}_{1\times3} & \mathbf{0}_{1\times3} & \mathbf{0}_{1\times3} & \mathbf{0}_{1\times3} \\
        \mathbf{0}_{3\times3} & \mathbf{I}_{3\times3} & \mathbf{0}_{3\times3} & \mathbf{0}_{3\times3} & \mathbf{0}_{3\times3} & \mathbf{I}_{3\times3} \\
        \mathbf{0}_{3\times3} & \mathbf{0}_{3\times3} & \mathbf{0}_{3\times3} & \mathbf{I}_{3\times3} & \mathbf{0}_{3\times3} & \mathbf{0}_{3\times3} \\
        \mathbf{0}_{3\times3} & \mathbf{0}_{3\times3} & \mathbf{0}_{3\times3} & \mathbf{0}_{3\times3} & \mathbf{I}_{3\times3} & \mathbf{0}_{3\times3}
    \end{bmatrix}^{\top}. 
\end{equation}
Once the pose clone is marginalized, the features must change the associated pose clone. According to (\ref{equation: new feature error state}), the association change also involves a covariance update. If the sliding window size is small, the frequent covariance update can lead to an increased computation cost. To mitigate this issue, the features are re-associated with the relative transformation in $\mathbf{X}_{\bm{G}_{k}}$ when streetlight matches are available, as described in Section \ref{section: MSC-InEKF-FDR}. This strategy improves efficiency and avoids the inconsistency problem in (\ref{equation: observability without streetlights}). Further details of these filter representation variants are in the supplementary material. 

\begin{table}[!t]
\renewcommand{\arraystretch}{1.5}
\caption{The Noise Standard Deviation in the Simulation}
\centering
\begin{tabular*}{\hsize}{@{}@{\extracolsep{\fill}}ccc@{}}
\toprule
Type & Measurement               & Noise Standard Deviation \\
\midrule
\multirow{4}{*}{IMU}    & Angular Velocity         & 0.001\ rad/s            \\
                        & Specific Force      & 0.02\ m/$\text{s}^{\text{2}}$           \\
                        & Gyroscope Bias           & 0.001\ rad/$\text{s}^{\text{2}}$        \\
                        & Accelerometer Bias       & 0.001\ m/$\text{s}^{\text{3}}$          \\
\midrule
Odometer                & Velocity                 & 0.01\ m/s               \\
\midrule
\multirow{2}{*}{Camera} & Image Features           & 1.0\ pixel                \\
                        & Streetlight Observations & 1.0\ pixel                \\
\midrule
Prior Pose & Rotation/Position         & 0.02\ rad/0.02\ m              \\
\midrule
Initial Relative Pose & Rotation/Position & 0.04\ rad/0.1\ m           \\
\bottomrule
\end{tabular*}
\label{Table: noise sd}
\vspace{-10pt}
\end{table}


\section{Experimental Result} \label{section: experimental result} 
We conduct comprehensive experiments on the simulation, public dataset sequences, and the real-world collected dataset to thoroughly evaluate the proposed framework. 

\subsection{Evaluation Metric}
We utilize the Absolute Trajectory Error (ATE) to measure the localization accuracy, the Relative Trajectory Error (RPE) to reflect the pose drift, and the Normalized Estimation Error Squared (NEES) to evaluate the consistency \cite{geneva2020openvins}. For NEES, the value should be close to 1.0 if the system is consistent. 

\subsection{Simulation Experiment}

\begin{table}[!t]
\renewcommand{\arraystretch}{1.5}
\caption{ATE Results for Translation (m)/Rotation ($^{\circ}$) of Different Algorithms in the Simulation Experiment}
\centering
\begin{tabular*}{\hsize}{@{}@{\extracolsep{\fill}}cccc}
\toprule
Algorithm/Result  & Loc-traj & Rel-trans & Map-traj\\ \midrule
MSCKF          & 0.44/0.36 & \textbf{0.02}/0.07 & 0.44/0.40 \\ 
MSC-InEKF-FC   & \textbf{0.25}/\textbf{0.15} & \textbf{0.02}/\textbf{0.05} & \textbf{0.26}/\textbf{0.17} \\ 
MSC-InEKF-FDN  & 0.45/0.40 & \textbf{0.02}/\textbf{0.05} & 0.45/0.38 \\ 
MSC-InEKF-FDR  & 0.46/0.35 & \textbf{0.02}/\textbf{0.05} & 0.46/0.38 \\ 
MSC-InEKF-FDRC & \textbf{0.25}/\textbf{0.15} & \textbf{0.02}/\textbf{0.05} & \textbf{0.26}/\textbf{0.17} \\
\bottomrule
\end{tabular*}
\label{Table: simulation ate}
\vspace{-10pt}
\end{table}

\begin{figure}[!t] 
    \centering 
    \includegraphics[width=0.47\textwidth]{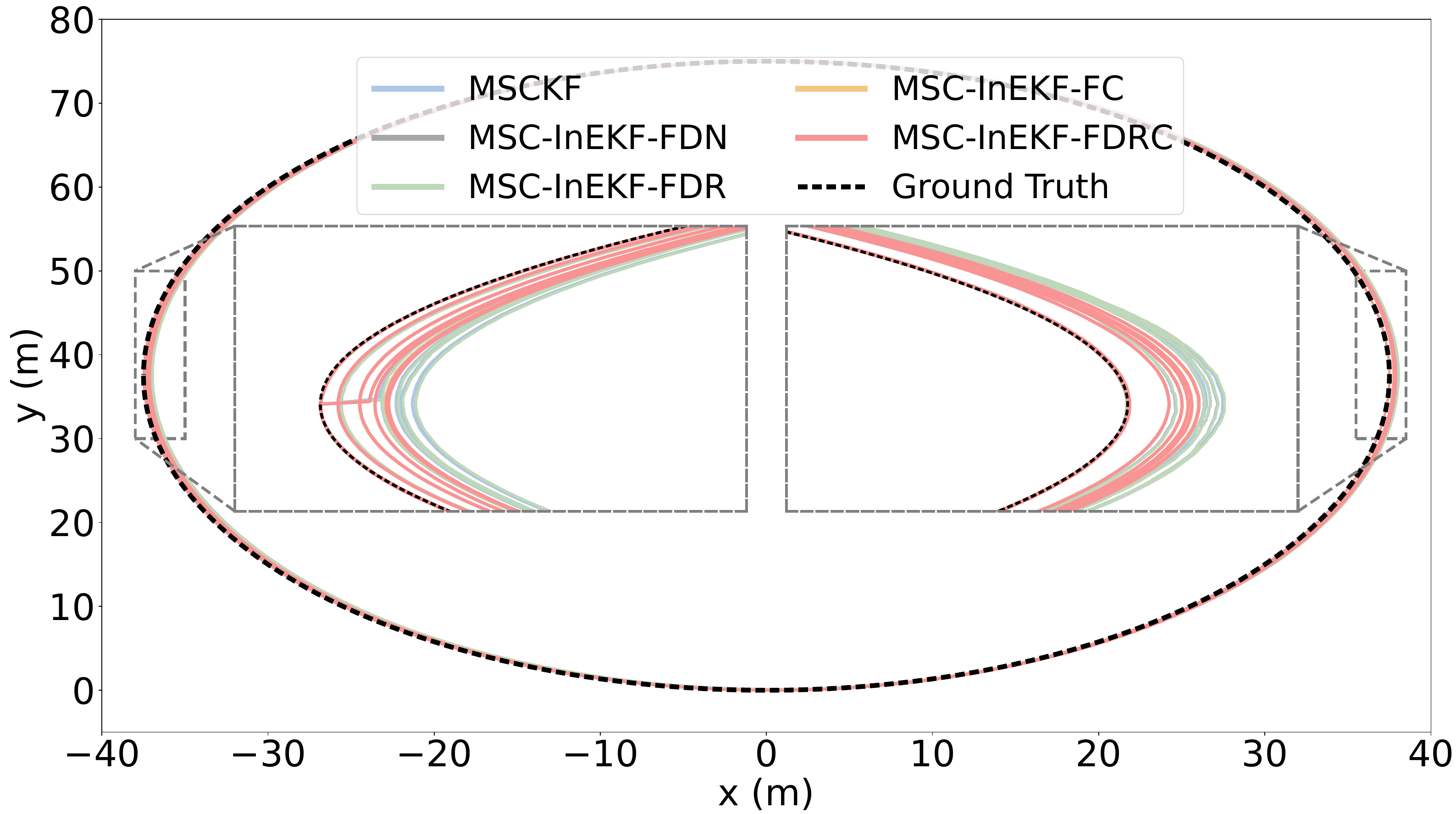} 
    \caption{Comparison of trajectories estimated by different algorithms.} 
    \label{Fig: simulation trajectories}
    \vspace{-10pt}
\end{figure}

We perform simulation experiments to compare the proposed MSC-InEKF-FDRC with MSCKF, MSC-InEKF-FDC, MSC-InEKF-FDN, and MSC-InEKF-FDR in terms of accuracy, consistency, and efficiency. In the simulation, the shape of the ground truth trajectory in localization is a circle (the pink trajectory in Fig. \ref{Fig: simulation scene}). The ground truth is then fed into the simulator of OpenVINS \cite{geneva2020openvins} to simulate the IMU, odometer, and feature point measurements. We also generate a mapping trajectory that is shaped like a wave swinging around the circle to simulate the prior poses (blue squares in Fig. \ref{Fig: simulation scene}). Points selected in the 3D space at a given interval are simulated as streetlights (orange stars in Fig. \ref{Fig: simulation scene}) so that the robot can observe 2 to 8 streetlights in each camera frame. The IMU, odometer, and visual data are generated, respectively, at 200, 10, and 25 Hz. All measurements are corrupted by additive Gaussian noises, as specified in Table \ref{Table: noise sd}. The initial relative pose is given by adding Gaussian noises (Table \ref{Table: noise sd}) to the true value. The robot runs ten times around the pink trajectory, and the total traveling distance is around 2513 m. In the ten loops, the streetlights and prior poses are only available in the first two and the last two loops, so the map information is absent for a long term (during the middle six loops). 

\textbf{Evaluation of Accuracy.} We provide three different result types for comparison, namely the robot trajectory in the local reference frame (Loc-traj), the relative transformation between the local and the map frames (Rel-trans), and the trajectory in the map frame (Map-traj). Table \ref{Table: simulation ate} contains the ATE results for rotation and translation between different algorithms. As expected, MSC-InEKF-FDRC performs better than MSCKF because the combination of InEKF and MSCKF solves the inconsistency problem. Although MSC-InEKF-FDN and MSC-InEKF-FDR also combine InEKF with MSCKF, the inconsistency in feature association methods makes their performance not as accurate as MSC-InEKF-FDRC, and even worse than MSCKF in the translation part. It is not surprising that MSC-InEKF-FC achieves comparable accuracy with MSC-InEKF-FDRC as both their feature association ways preserve the correct observability. Moreover, it can be observed that all filters achieve competitive accuracy in Rel-trans. The reason is that the map-based information (i.e., streetlights and prior poses) effectively bounds the drift of relative transformation while features and odometer measurements only constrain the local state. When map-based information is absent (i.e. during the middle 6 loops), the local trajectories drift, while the relative transformation is minimally affected. Fig. \ref{Fig: simulation trajectories} visually compares the trajectories generated by different algorithms. The zoom-in areas show that the estimated trajectory of MSC-InEKF-FDRC (the pink one) is the closest to the ground truth.

\begin{table}[!t]
\renewcommand{\arraystretch}{1.5}
\caption{NEES Results for Translation/Rotation of Different Algorithms in the Simulation Experiment}
\centering
\begin{tabular*}{\hsize}{@{}@{\extracolsep{\fill}}cccc}
\toprule
Algorithm/Result  & Loc-traj & Rel-trans & Map-traj \\ \midrule
MSCKF          & 10.14/13.31 & 0.93/11.35 & 9.73/14.43 \\ 
MSC-InEKF-FC   & \textbf{1.29}/\textbf{1.92} & \textbf{0.94}/\textbf{1.34} & 0.58/\textbf{1.48} \\ 
MSC-InEKF-FDN  & 7.63/16.15 & \textbf{0.94}/6.62 & 7.50/12.89 \\ 
MSC-InEKF-FDR  & 7.67/2.47 & \textbf{0.94}/\textbf{1.34} & 5.28/2.05 \\ 
MSC-InEKF-FDRC & \textbf{1.29}/\textbf{1.92} & \textbf{0.94}/\textbf{1.34} & \textbf{0.59}/\textbf{1.48} \\
\bottomrule
\end{tabular*}
\label{Table: simulation nees}
\vspace{-10pt}
\end{table}

\begin{figure*}[!t]
    \centering 
    \includegraphics[width=0.85\textwidth]{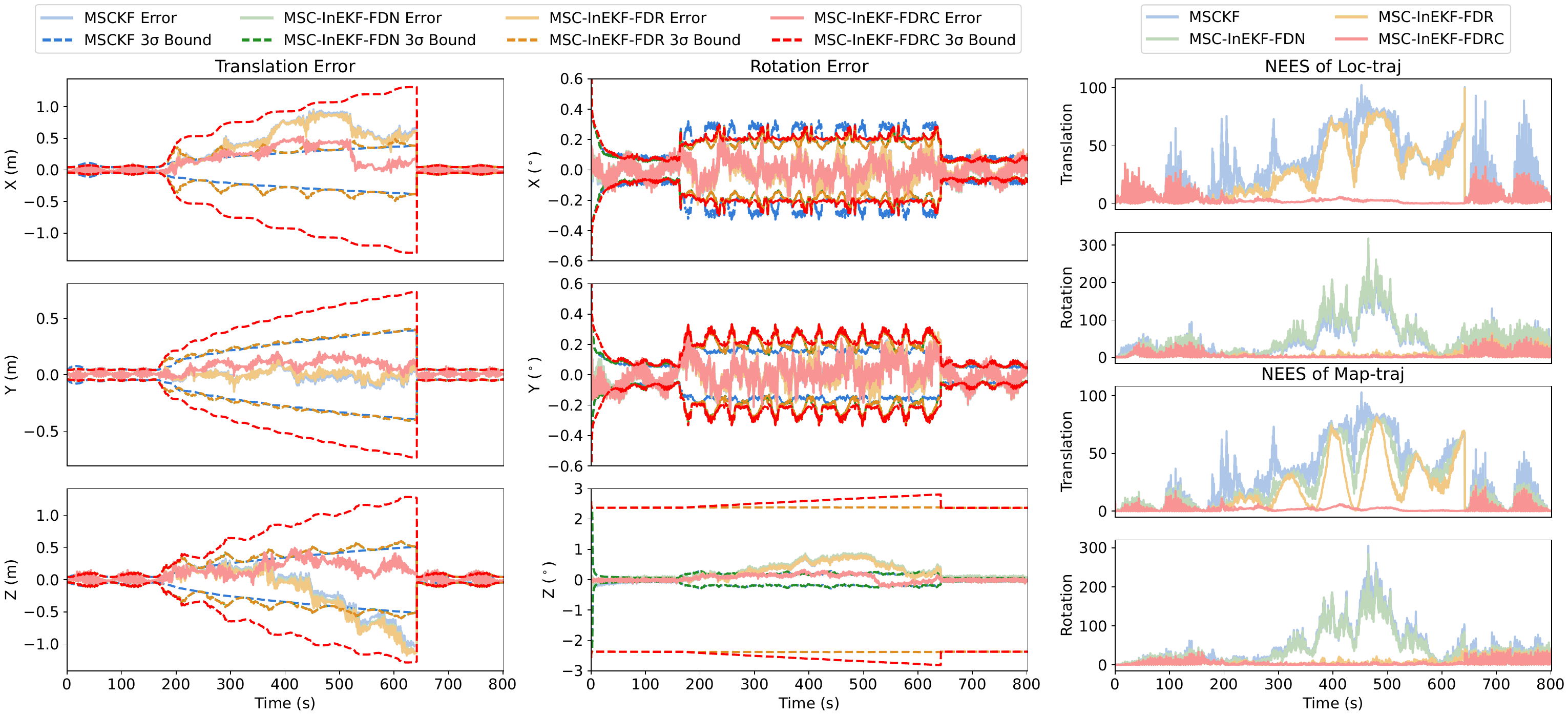} 
    \caption{Comparison of the estimation consistency. The left two columns show the local trajectory errors and the corresponding $\pm$3$\sigma$ bounds of different algorithms. The right column shows the NEES of local and map trajectories of different algorithms.} 
    \label{Fig: simulation error and nees}
    \vspace{-10pt}
\end{figure*}

\textbf{Evaluation of Consistency.} Table \ref{Table: simulation nees} includes the NEES results between different algorithms. Note that although the covariance of the pose in the map frame is not directly estimated, we do have the covariance of both ${^{L}}\mathbf{T}_{G}$ and ${^{L}}\mathbf{T}_{I}$. Therefore, we can derive it using the relationship ${^{G}}\mathbf{T}_{I}={^{L}}\mathbf{T}_{G}^{-1}{^{L}}\mathbf{T}_{I}$. Inflated NEES values are observed in MSCKF, MSC-InEKF-FDN, and MSC-InEKF-FDR, due to the underestimated state covariance caused by inconsistency. In contrast, MSC-InEKF-FC and MSC-InEKF-FDRC keep the correct dimension of the unobservable subspace, thus their NEES values are around 1, indicating consistent estimators. The qualitative results in Fig. \ref{Fig: simulation error and nees} display the local-pose error with 3$\sigma$ bounds and the NEES variation of Loc-traj and Map-traj over time for different algorithms. The plots reveal that the estimation errors of MSC-InEKF-FDRC are well characterized in the uncertainty bounds. The NEES values for all filters initially remain at a low level owing to the map-based constraints. Subsequently, when the map information is absent, the MSC-InEKF-FDRC curve exhibits minor fluctuations, whereas the other curves increase rapidly, indicating the consistency of MSC-InEKF-FDRC. 

\begin{figure}[!t]
    \centering 
    \includegraphics[width=0.47\textwidth]{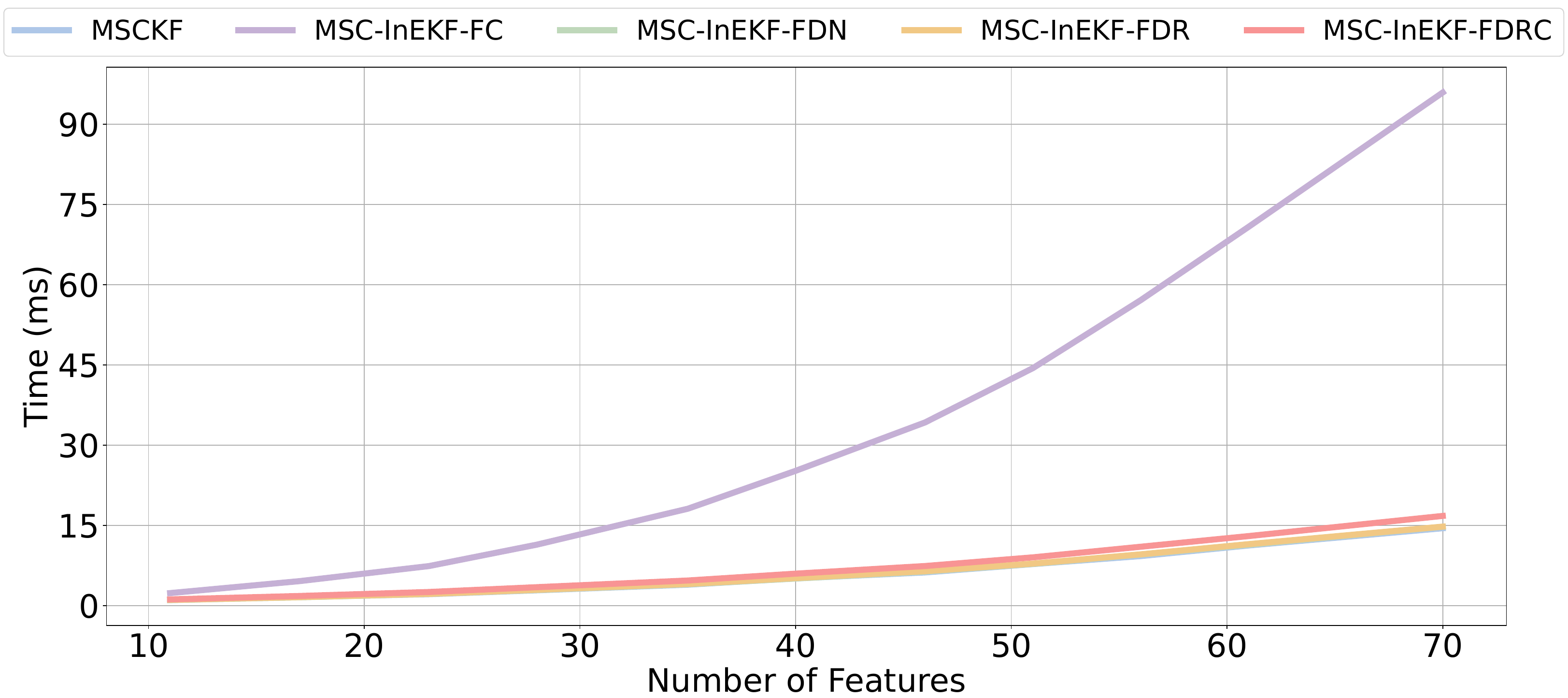} 
    \caption{Average runtime per step over the different number of feature points in the simulation.} 
    \label{Fig: simulation runtime}
    \vspace{-10pt}
\end{figure}

\textbf{Evaluation of Efficiency.} The average runtime per step over the number of feature points is shown in Fig. \ref{Fig: simulation runtime}, validating that MSC-InEKF-FC is time-consuming compared to other filters. For example, when the number of features increases to 50, the time consumption of MSC-InEKF-FC is approximately three times that of the others. Thanks to the decoupled feature association, the proposed MSC-InEKF-FDRC is much more time-saving and has efficiency similar to that of MSCKF. 

Overall, the proposed MSC-InEKF-FDRC is an accurate, efficient, and consistent filter and representation design, playing a pivotal role within the nocturnal state estimation framework.

\subsection{Evaluation on the Public Dataset}
\begin{figure}[!t]
    \centering 
    \includegraphics[width=0.47\textwidth]{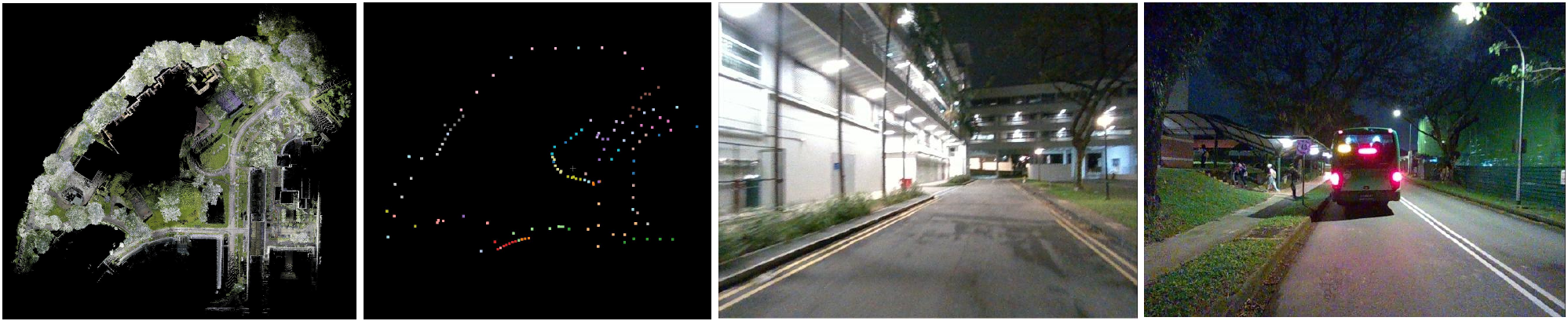} 
    \caption{The prior map, the constructed streetlight map, and sampled images from the MCD dataset \cite{nguyen2024mcd}.} 
    \label{Fig: MCD datasets}
    \vspace{-10pt}
\end{figure}

\textbf{Setup.} There is a paucity of publicly available multi-sensor datasets for urban environments at night. Accordingly, we use the nighttime sequences from the public MCD dataset \cite{nguyen2024mcd} to evaluate different algorithms. The MCD dataset consists of 18 sequences collected from 3 different campuses, including 3 sequences (\textit{ntu\_night\_04}, \textit{ntu\_night\_08}, and \textit{ntu\_night\_13}) gathered by an all-terrain vehicle on campus NTU during the night. The character of the 3 sequences lies in the relatively rich illumination (Fig. \ref{Fig: MCD datasets}). We leverage the images in these 3 sequences and the provided accurate survey-grade prior map to construct a streetlight map. The mapping process is similar to the description in Section \ref{subsection: Streetlight Map Generation}. The final streetlight map of the NTU campus is shown in the second image in Fig. \ref{Fig: MCD datasets}. Since the odometer is not provided, we add noise to the interpolated velocity \cite{geneva2020openvins} to simulate the measurements. 

\begin{figure}[!t]
    \centering 
    \includegraphics[width=0.47\textwidth]{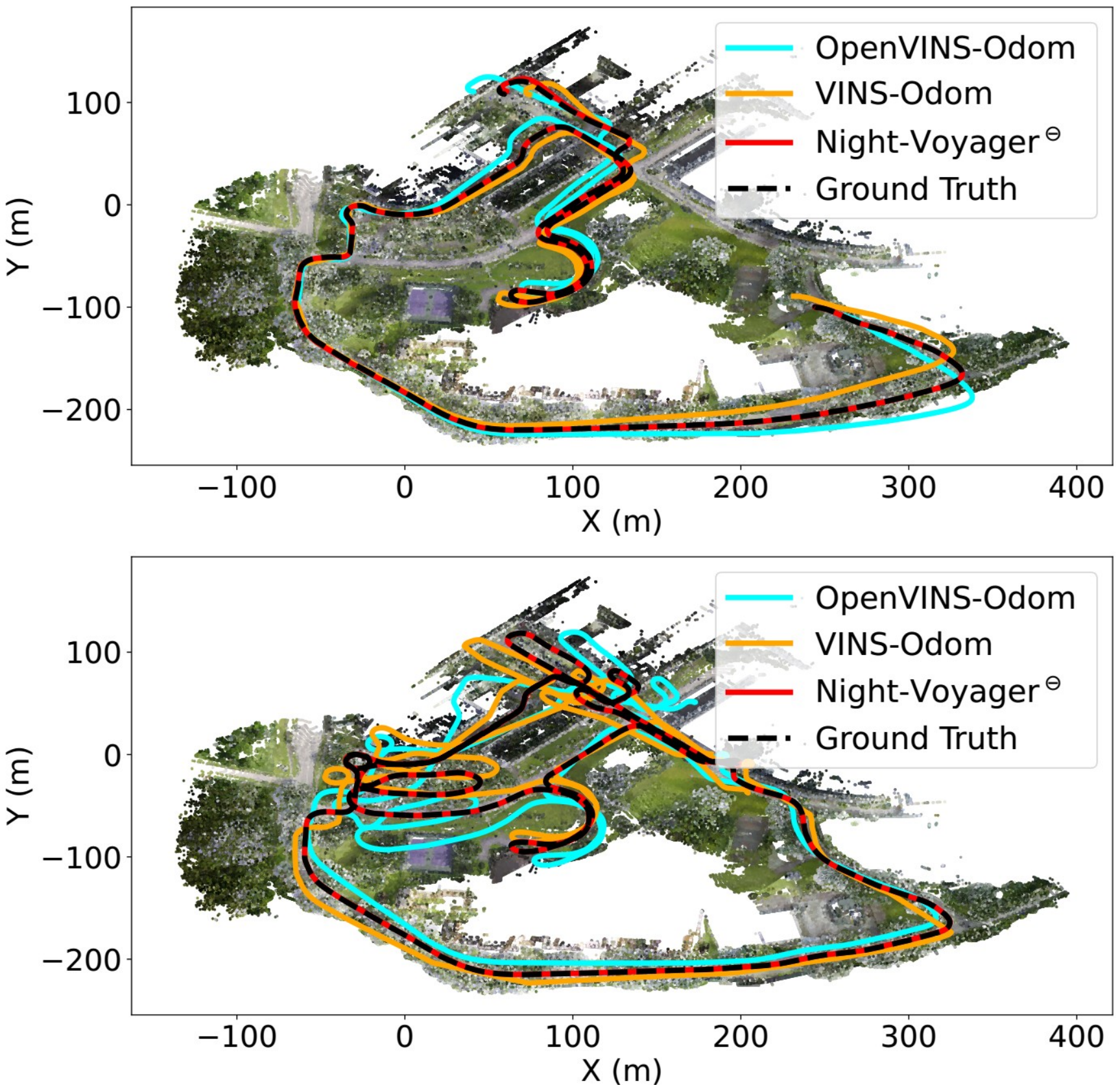} 
    \caption{Estimated trajectories of different algorithms on the prior map. Top: \textit{ntu\_night\_04}. Bottom: \textit{ntu\_night\_08}.} 
    \label{Fig: MCD Trajectories}
    \vspace{-10pt}
\end{figure}

\textbf{Result and Analysis.} For each sequence, we benchmark the proposed system without prior pose observations (Night-Voyager$^{\ominus}$), and compare the estimated trajectory in the map frame with those of state-of-the-art (SOTA) vision-aided state estimation methods augmented with odometer observations, including the odometer-aided VINS-Mono \cite{qin2018vins, liu2019visual} (VINS-Odom) and the odometer-aided OpenVINS \cite{geneva2020openvins} (OpenVINS-Odom). We also compare several popular visual localization algorithms which leverage different map types, including methods based on feature point maps (Hloc \cite{sarlin2019coarse}), 3D line maps (VL-Line \cite{yu2020monocular}), and point cloud maps (CMRNet \cite{cattaneo2019cmrnet} and LHMap \cite{wu2024lhmap_icra2024}). For Hloc, SuperPoint \cite{detone2018superpoint} and SuperGlue \cite{sarlin2020superglue} are utilized for feature map reconstruction and visual localization. Due to insufficient feature points at night, constructing a feature point map is challenging. Therefore, we use daytime sequences to build the map. In VL-Line, VINS-Mono \cite{qin2018vins} provides prior poses to identify line matches. For a fair comparison, we replace it with VINS-Odom. CMRNet and LHMap refine coarse poses with neural networks using images and projected point clouds as input. MSF is used to generate the coarse poses and integrate the refined poses. We select the \textit{ntu\_night\_08} sequence for training the networks introduced in the aforementioned methods. Table \ref{Table: ntu_comparison} reports the ATE results of different methods. Even without the assistance of prior pose observations, Night-Voyager$^{\ominus}$ still significantly outperforms other algorithms in all sequences. The estimated trajectories are presented in Fig. \ref{Fig: MCD Trajectories}. Thanks to the streetlight map, the trajectories estimated by Night-Voyager$^{\ominus}$ align well with the ground truth. 
Nevertheless, although well-lit sequences promote the feature association, the estimated trajectories of VINS-Odom and OpenVINS-Odom inevitably suffer from pose drifts, indicating the critical role of object-level correspondences. Besides, 3D line maps have limited effectiveness in visual localization because a structured scene is required for VL-Line. However, the textural lines of traffic lanes and floor tile gaps, along with spurious line detections caused by uneven lighting, influence the line association. Essentially, the poor performance of VL-Line stems from the \textit{insufficiency} and \textit{inconsistency} issues of pixel-level features. Due to the limited information at night, the improvement of CMRNet and LHMap is much inferior to that of Night-Voyager, which demonstrates the substantial benefits brought by the object-level map and data association. 

\setlength{\tabcolsep}{5.5pt}
\begin{table}[!t]
\centering
\renewcommand{\arraystretch}{1.5}
\caption{ATE Results (m/$^{\circ}$) on the MCD Dataset}
\begin{tabular}{cccc}
\toprule
Method/Sequence               & \begin{tabular}[c]{@{}c@{}}\textit{ntu\_night\_04}\\(1459 m)\end{tabular} & \begin{tabular}[c]{@{}c@{}}\textit{ntu\_night\_08}\\(2421 m)\end{tabular} & \begin{tabular}[c]{@{}c@{}}\textit{ntu\_night\_13}\\(1231 m)\end{tabular} \\\hline
OpenVINS-Odom & 14.30/7.72 & 20.03/11.71 & 14.64/7.07 \\ 
VINS-Odom      & 11.15/5.66 & 18.43/7.97 & 16.74/7.21  \\ 
Hloc & --- & --- & --- \\ 
VL-Line & 10.34/4.73 & 21.44/5.85 & 11.27/13.28 \\ 
CMRNet & 26.67/11.21 & ${\mathcal{\#}}$ & 20.25/9.07 \\ 
LHMap & 21.55/9.83 & ${\mathcal{\#}}$ & 16.24/7.99 \\ 
MSCKF$^{\ominus}$          & 0.81/1.20 & 40.87/12.39 & 6.28/4.48     \\ 
MSC-InEKF-FC$^{\ominus}$   & 0.81/1.29 & \textbf{0.25}/\textbf{0.80} & 0.27/0.73\\ 
MSC-InEKF-FDN$^{\ominus}$  & 0.86/1.29 & 34.28/11.38 & 27.85/13.21\\ 
MSC-InEKF-FDR$^{\ominus}$  & 0.93/1.40 & \ding{56} & \textbf{0.24}/\textbf{0.71}\\ 
Night-Voyager$^{\ominus}$ w/o FP & 0.86/\textbf{1.15} & 1.03/1.41 & 0.57/0.98\\ 
Night-Voyager$^{\ominus}$  & \textbf{0.79}/1.21 & \textbf{0.25}/0.82 & 0.27/0.73 \\\bottomrule
\end{tabular}
\label{Table: ntu_comparison}
\captionsetup{justification=raggedright,singlelinecheck=false}
\caption*{--- denotes that Hloc fails to construct the feature map. \\ ${\mathcal{\#}}$ denotes the training sequence for CMRNet and LHMap. \\ \ding{56} indicates the algorithm fails on the corresponding sequence. \\ The optimal results are highlighted in \textbf{bold}.}
\vspace{-10pt}
\end{table}

Ablation experiments are also performed. By replacing MSC-InEKF-FDRC with different filters, we obtain 4 versions of Night-Voyager$^{\ominus}$, namely MSCKF$^{\ominus}$, MSC-InEKF-FC$^{\ominus}$, MSC-InEKF-FDN$^{\ominus}$, and MSC-InEKF-FDR$^{\ominus}$. Consistent with the simulation results, the two Night-Voyager$^{\ominus}$ variants built on MSC-InEKF-FC and MSC-InEKF-FDRC demonstrate the best results. In comparison, MSCKF$^{\ominus}$, MSC-InEKF-FDN$^{\ominus}$, and MSC-InEKF-FDR$^{\ominus}$ exhibit inferior performance. In \textit{ntu\_night\_08}, there is a long distance with multiple false streetlight detections. Since the streetlight matching depends on the estimated pose, pose errors caused by inconsistency impact the results of data association, further exacerbating the errors. Consequently, MSCKF$^{\ominus}$, MSC-InEKF-FDN$^{\ominus}$, and MSC-InEKF-FDR$^{\ominus}$ fail to estimate the trajectory. 

\begin{figure}[!t]
    \centering 
    \includegraphics[width=0.47\textwidth]{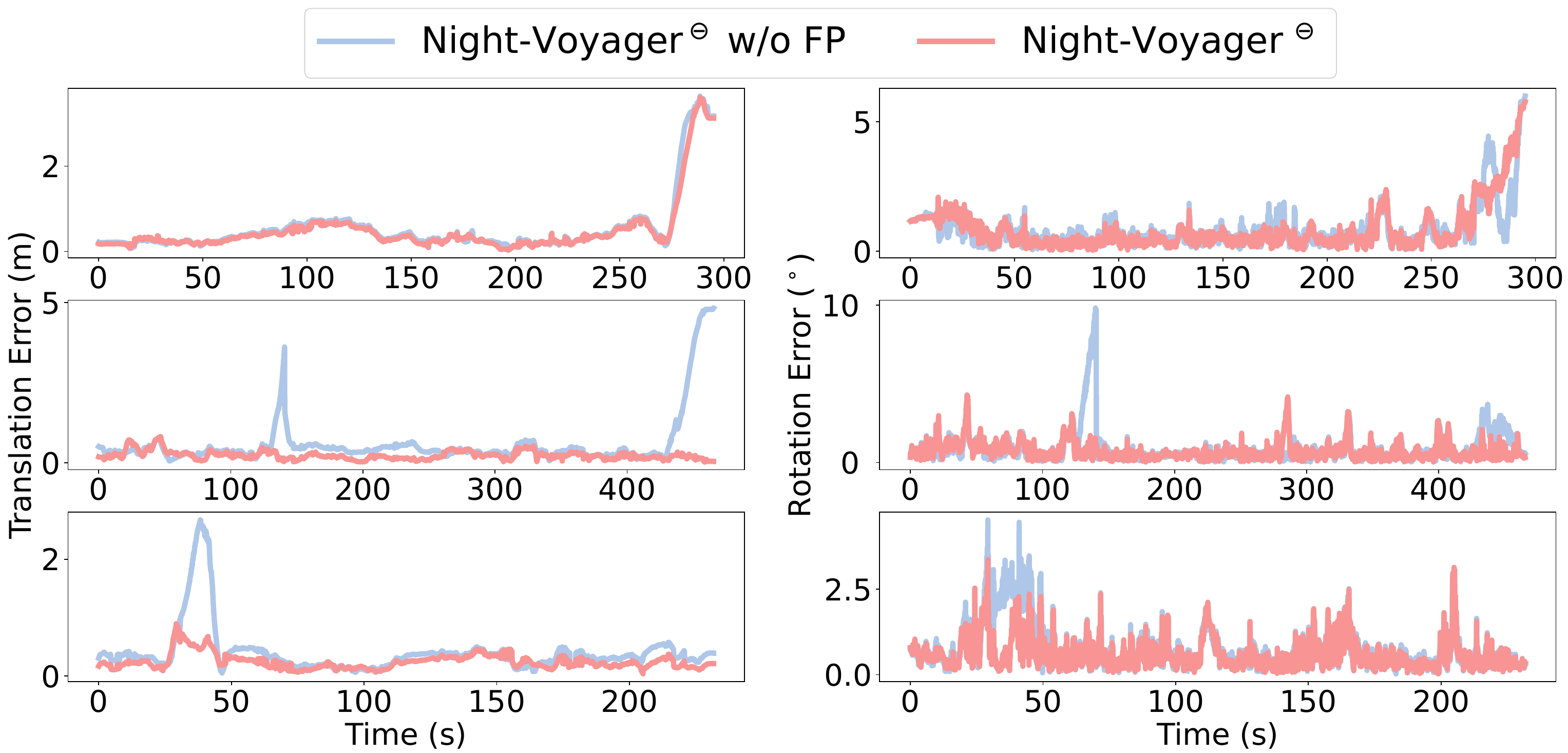} 
    \caption{Comparison of translation and rotation errors between the Night-Voyager$^{\ominus}$ with and without feature points. Top: \textit{ntu\_night\_04}. Middle: \textit{ntu\_night\_08}. Bottom: \textit{ntu\_night\_13}.} 
    \label{Fig: MCD Ablation} 
    \vspace{-10pt}
\end{figure}

Furthermore, we analyze the role of feature points in nighttime localization. We record the ATE result of Night-Voyager$^{\ominus}$ without feature points (Night-Voyager$^{\ominus}$ w/o FP) in Table \ref{Table: ntu_comparison}. The accuracy decreases without feature points, especially in \textit{ntu\_night\_08}. Fig. \ref{Fig: MCD Ablation} further presents the trajectory errors of both Night-Voyager$^{\ominus}$ and Night-Voyager$^{\ominus}$ w/o FP. Compared to Night-Voyager$^{\ominus}$, Night-Voyager$^{\ominus}$ w/o FP has higher estimation errors, indicating the effectiveness of leveraging hybrid object-level and pixel-level features. 

\subsection{Evaluation on the Collected Dataset}
\begin{figure}[!t]
    \centering 
    \includegraphics[width=0.47\textwidth]{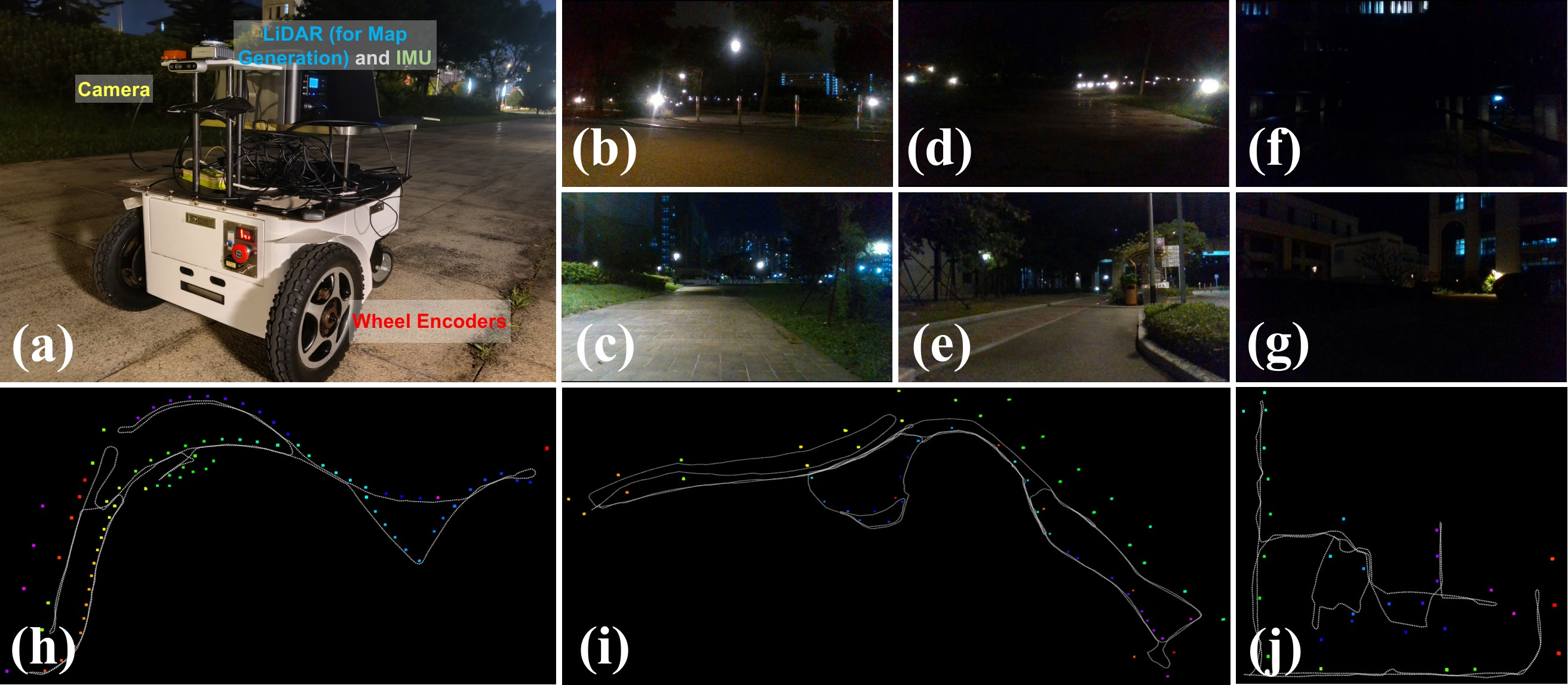} 
    \caption{The robot for data collection and the sequence examples. (a) displays the sensors on the robot. (b), (c), (d), (e), (f), and (g) are sampled images from the collected dataset. (h), (i), and (j) are the constructed streetlight maps of \textit{Scene\_02}, \textit{Scene\_03}, and \textit{Scene\_04}.} 
    \label{Fig: real scenes}
    \vspace{-10pt}
\end{figure}

\textbf{Setup.} In addition to evaluations on public datasets, further validation of the proposed framework is required through its deployment on real robots across diverse nocturnal urban scenarios. We utilize the wheeled robot equipped with multiple sensors for data collection. As Fig. \ref{Fig: real scenes} shows, the sensors consist of a front-facing RGB camera (30 Hz 1280$\times$720 images), an odometer (10 Hz), and a Livox 360$^{\circ}$ FoV LiDAR (10 Hz) integrating an ICM40609 IMU (200 Hz). Only IMU, odometer, and monocular camera measurements are used for nocturnal state estimation. The camera's intrinsics and all extrinsics among sensors are pre-calibrated, and the trajectory ground truth is provided by RTK GNSS sensors. To comprehensively evaluate the proposed framework, we collect nighttime data from ten different scenes. Each scenario in the collected dataset comprises two sets: one for constructing the streetlight map and the other for validating nocturnal state estimation algorithms. The trajectories of these two sets are different and have distinct starting points. The dataset focuses on areas such as campuses, alleys, streets, and gardens, and exhibits poorer lighting conditions compared to the MCD dataset. Table \ref{Table: sequences} summarizes the characters of each sequence, including the distance, the scene type, and the distribution density of streetlights (a mixed distribution indicates the existence of both sparse and dense streetlight areas). Fig. \ref{Fig: real scenes} shows the example images and streetlight maps. 

We utilize the YOLOv7 series \cite{wang2023yolov7} as the streetlight detector due to its high accuracy and real-time performance. About 3000 images labeled from different nighttime scenes are used to train the detection network. A desktop computer with an Intel i9-10900X CPU and an NVIDIA GeForce RTX 3090 GPU is used for network training, while all evaluation experiments are performed on an Intel NUC with an i7-1165G7 CPU. For all data-driven methods (i.e., Hloc \cite{sarlin2019coarse}, VL-Line \cite{yu2020monocular}, CMRNet \cite{cattaneo2019cmrnet}, and LHMap \cite{wu2024lhmap_icra2024}), except for Hloc, which is trained using additional mapping sequences collected during the daytime, the networks are trained with nighttime mapping sequences. These four methods are trained and evaluated on the aforementioned desktop computer. 

\begin{figure}[!t] 
    \centerline{\includegraphics[width=0.47\textwidth]{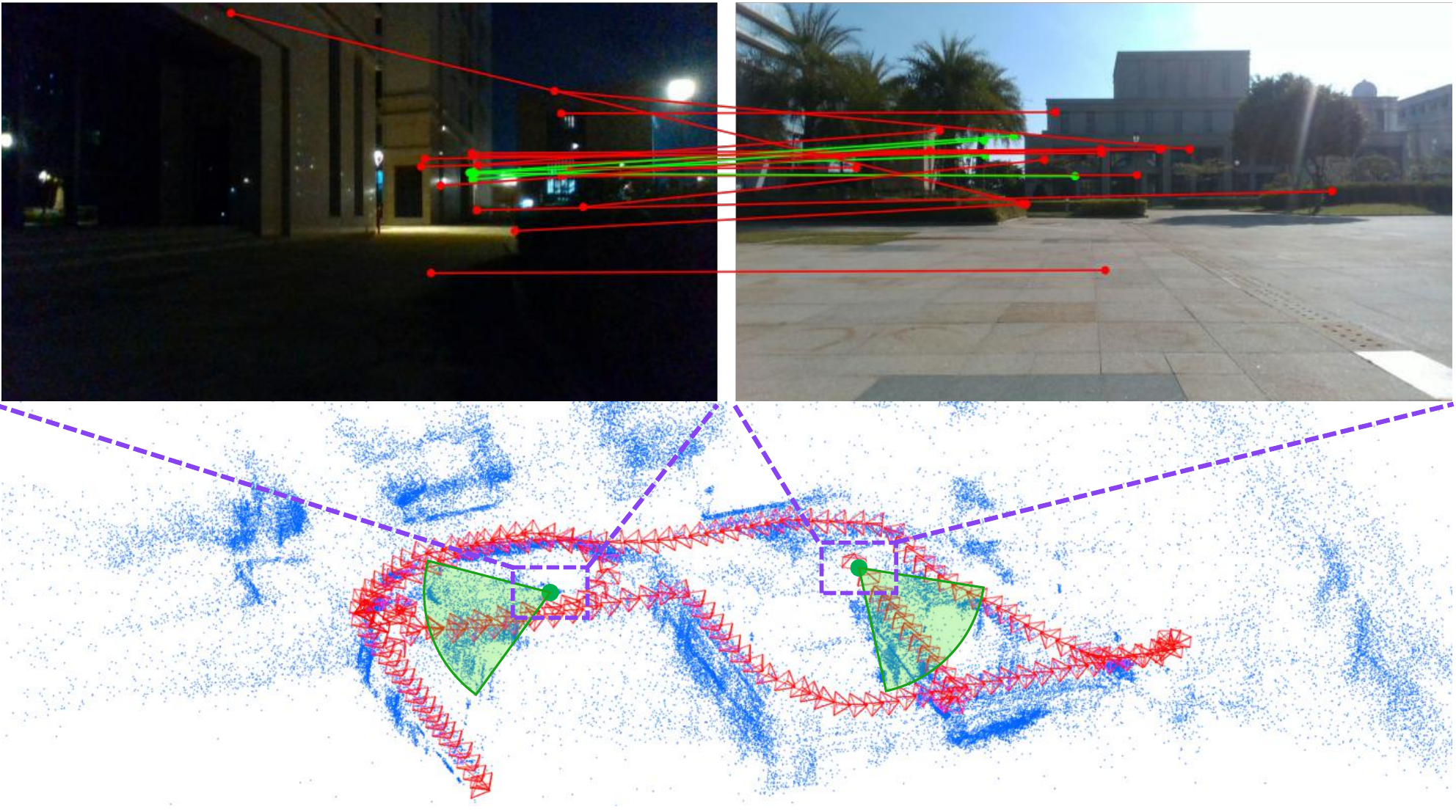}} 
    \caption{Hloc failure case. Top left: captured image during nocturnal localization. Top right: queried image from the mapping sequence. Bottom: constructed feature map and mapping trajectory.}
    \label{Fig: Hloc_failure}
    \vspace{-10pt}
\end{figure}

\begin{figure*}[!t]
    \centering 
    \includegraphics[width=0.85\textwidth]{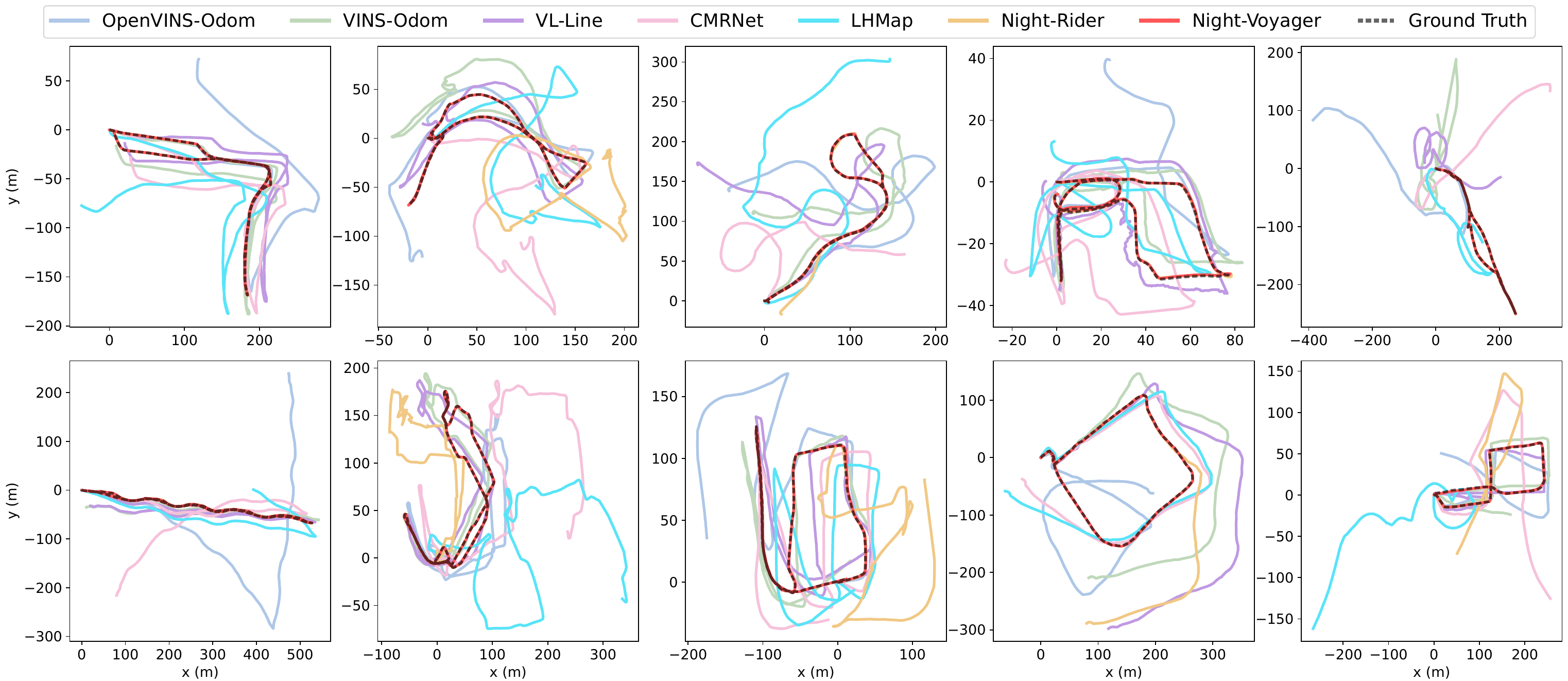} 
    \caption{Comparison of the estimated trajectories of different algorithms on the collected dataset. (a) \textit{Scene\_01}. (b) \textit{Scene\_02}. (c) \textit{Scene\_03}. (d) \textit{Scene\_04}. (e) \textit{Scene\_05}. (f) \textit{Scene\_06}. (g) \textit{Scene\_07}. (h) \textit{Scene\_08}. (i) \textit{Scene\_09}. (j) \textit{Scene\_10}.} 
    \label{Fig: trajectories}
    \vspace{-10pt}
\end{figure*}

\textbf{Evaluation and Analysis.} In the collected dataset, we compare Night-Voyager with the aforementioned methods and the previous work, Night-Rider \cite{gao2024night}. The quantitative results and estimated trajectories are shown in Table \ref{Table: collected dataset comparison} and Fig. \ref{Fig: trajectories}. Contrary to the results observed in the brighter scenarios of the MCD dataset, both OpenVINS-Odom and VINS-Odom perform significantly worse in the darker scenarios of the collected dataset, due to the limited tracked features under unfavorable lighting conditions. As shown in Fig. \ref{Fig: Hloc_failure}, the performance of Hloc is significantly affected by the severe illumination differences between daytime and nighttime images. VL-Line, CMRNet, and LHMap also perform poorly in dark environments. Hloc even fails across all sequences, poor lighting conditions pose challenges in identifying reliable matches between images and feature maps. In contrast, Night-Voyager has a much smaller trajectory error due to the streetlight map. Night-Rider shows significant performance variations across different sequences. Two reasons lead to its poor robustness: 1) Night-Rider leverages InEKF as the estimator. Compared to MSC-InEKF-FDRC, which utilizes historical information in the sliding window, InEKF emphasizes the current observation, making Night-Rider prone to failures when incorrect matches are determined. 2) Night-Rider is highly dependent on streetlights, causing poor robustness when no streetlight is visible. In Night-Voyager, all feature points, prior poses, and virtual centers serve as system observations. Therefore, regardless of the density of the streetlight distribution, Night-Voyager achieves remarkably accurate and robust localization in all scenes ($<$0.2\% relative error of the total trajectory). 

\begin{table}[!t]
\setlength{\tabcolsep}{2.8pt}
\caption{Details of the Collected Dataset}
\centering
\renewcommand{\arraystretch}{1.5}
\begin{tabular}{cccccccc}
\toprule
\multirow{2}{*}{Seq.} & \multirow{2}{*}{Dist. (m)} & \multirow{2}{*}{Scene} & \multirow{2}{*}{Distrib.} & \multicolumn{4}{c}{Map Storage (MB)}\\\cline{5-8}
&&&& St. & Line & PCD & Feat. \\\hline
\textit{Scene\_01}   & 724        & Avenue \& Garden          & Sparse          & 3.0        & 0.4 & 81.3 & 648.6  \\ 
\textit{Scene\_02}   & 613        & Lane        & Dense          & 2.9     & 0.1 & 57.4 & 318.1 \\ 
\textit{Scene\_03}   & 635        & Lane \& Pavement           & Sparse       & 6.0       & 0.2 & 61.4 & 534.9   \\ 
\textit{Scene\_04}   & 305        & Garden            & Sparse       & 3.3      & 0.3 & 58.4 & 577.4    \\ 
\textit{Scene\_05}   & 777        & Seaside Road            & Dense        & 3.8       & 0.1 & 54.1 & 283.4   \\ 
\textit{Scene\_06}   & 1071       & Seaside Road            & Dense        & 3.5       & 0.1 & 54.1 & 361.2   \\ 
\textit{Scene\_07}   & 892        & Avenue \& Garden           & Mixed          & 5.2     & 0.5 & 75.6 & 446.2   \\ 
\textit{Scene\_08}   & 719        & Alley \& Campus           & Mixed        & 4.5       & 0.2 & 71.6 & 621.4   \\ 
\textit{Scene\_09}   & 842        & Campus           & Mixed         & 6.1     & 0.6  & 86.1 & 511.7  \\ 
\textit{Scene\_10}   & 601        & Bridge \& Avenue           & Mixed          & 2.9    & 0.6 & 112.0 & 446.2   \\
\bottomrule
\end{tabular}
\caption*{Seq., Dist., and Distrib. are short for sequence, distance, and streetlight distribution. St., Line, PCD, and Feat. denote the streetlight map, the line map, the point-cloud map, and the feature-point map.}
\label{Table: sequences}
\vspace{-10pt}
\end{table}

\setlength{\tabcolsep}{4.0pt}
\begin{table*}[!t]
\caption{ATE results (m/$^{\circ}$) of Different Algorithms on the Collected Dataset}
\centering
\renewcommand{\arraystretch}{1.5}
\begin{tabular}{ccccccccccc}
\toprule
Method/Sequence & \textit{Scene\_01} & \textit{Scene\_02} & \textit{Scene\_03} & \textit{Scene\_04} & \textit{Scene\_05} & \textit{Scene\_06} & \textit{Scene\_07} & \textit{Scene\_08} & \textit{Scene\_09} & \textit{Scene\_10} \\\hline
 OpenVINS-Odom & \ding{56} & 39.44/56.31 & \ding{56} & 18.10/17.58 & \ding{56} & \ding{56} & 41.90/23.29 & \ding{56} & \ding{56} & 26.13/15.36 \\ 
 VINS-Odom & 11.38/4.05 & 39.45/23.25 & \ding{56} & 7.17/13.23 & \ding{56} & 13.86/11.34 & 27.74/11.47 & 18.78/11.31 & \ding{56} & \ding{56} \\ 
 Hloc & \ding{56} & \ding{56} & \ding{56} & \ding{56} & \ding{56} & \ding{56} & \ding{56} & \ding{56} & \ding{56} & \ding{56}\\ 
 VL-Line & 7.27/3.56 & 17.02/18.94 & \ding{56} & 5.05/4.63 & \ding{56} & 9.33/5.03 & 21.86/19.16 & 18.80/24.65 & \ding{56} & \ding{56}\\ 
 CMRNet & 25.58/9.29 & 41.04/37.11 & \ding{56} & 10.02/17.29 & \ding{56} & \ding{56} & \ding{56} & 28.56/16.22 & 20.61/11.53 &\ding{56}\\ 
 LHMap & 20.11/15.12 & 40.92/36.50 & \ding{56} & 8.56/17.92 & \ding{56} & 20.54/13.93 & \ding{56} & 22.65/19.03 & 30.52/17.14 & \ding{56} \\ 
 Night-Rider & 0.88/3.71 & \ding{56} & 4.26/5.42 & 0.36/1.86 & 0.33/1.67 & 0.38/\textbf{0.56} & 18.85/20.03 & \ding{56} & \ding{56} & 27.98/24.38 \\ 
 MSF & 11.84/9.70 & 46.56/58.41 & \ding{56} & 8.21/28.11 & \ding{56} & \ding{56} & \ding{56} & \ding{56} & \ding{56} & 34.37/21.26 \\ 
 \begin{tabular}[c]{@{}c@{}}Night-Voyager\\w/o FP\end{tabular} & 0.20/\textbf{0.57} & 0.39/0.87 & 0.54/\textbf{0.93} & 0.27/0.88 & \textbf{0.29}/\textbf{0.77} & 0.38/\textbf{0.56} & 0.49/\textbf{1.12} & \textbf{0.41}/\textbf{0.84} & 0.55/0.83 & \textbf{0.45}/\textbf{0.96}\\ 
 \begin{tabular}[c]{@{}c@{}}Night-Voyager\\w/o PP\end{tabular} & 0.20/0.60 & \textbf{0.31}/\textbf{0.83} & 0.52/0.94 & 0.31/\textbf{0.86} & 0.30/0.78 & \textbf{0.32}/0.64 & 0.49/1.13 & 0.43/\textbf{0.84} & 0.58/0.86 & \textbf{0.45}/\textbf{0.96}\\ 
 \begin{tabular}[c]{@{}c@{}}Night-Voyager\\w/o VC\end{tabular} & 0.20/0.60 & 0.57/1.00 & 0.54/0.99 & 0.29/1.06 & 0.34/0.94 & 0.37/0.64 & 0.48/1.15 & 0.46/0.96 & 0.61/0.93 & 0.46/0.97 \\ 
 Night-Voyager & \textbf{0.19}/\textbf{0.57} & 0.38/0.87 & \textbf{0.49}/\textbf{0.93} & \textbf{0.26}/0.90 & \textbf{0.29}/0.78 & 0.38/\textbf{0.56} & \textbf{0.44}/\textbf{1.12} & \textbf{0.41}/\textbf{0.84} & \textbf{0.54}/\textbf{0.80} & \textbf{0.45}/\textbf{0.96} \\
\bottomrule
\end{tabular}
\captionsetup{justification=raggedright,singlelinecheck=false}
\caption*{\ding{56} indicates the algorithm fails to run the corresponding sequence. The optimal results are highlighted in \textbf{bold}.}
\label{Table: collected dataset comparison}
\vspace{-10pt}
\end{table*}

We then conduct an ablation study to assess the impact of feature points, prior poses, and virtual centers. Four variants of the proposed algorithm are evaluated, including 1) MSF: Night-Voyager only fusing with the measurements of IMU, odometer, and feature points; 2) Night-Voyager w/o FP: Night-Voyager without utilizing feature points; 3) Night-Voyager w/o PP: Night-Voyager without utilizing prior poses; 4) Night-Voyager w/o VC: Night-Voyager utilizing the geometric centers of streetlight clusters instead of the virtual centers. In Table \ref{Table: collected dataset comparison}, the results of MSF and Night-Voyager verify the importance of the prior streetlight map. In addition, the proposed data association algorithm determines the correct streetlight matches, thus the streetlight information is fully leveraged for nocturnal state estimation. When feature points are not utilized, there is almost no loss of accuracy in most sequences. This phenomenon is natural since it is difficult to perform feature tracking in scenes under poor lighting conditions. In most sequences, prior poses improve the performance of Night-Voyager, indicating that they can provide versatile information for estimation. We also observe that Night-Voyager achieves improvements in almost all sequences compared with Night-Voyager w/o VC. This is because virtual centers mitigate the inconsistency between the light bulb center and the geometric center of the streetlight cluster, providing more accurate and physically plausible constraints for state update. 

\begin{figure}[!t]
    \centering 
    \includegraphics[width=0.47\textwidth]{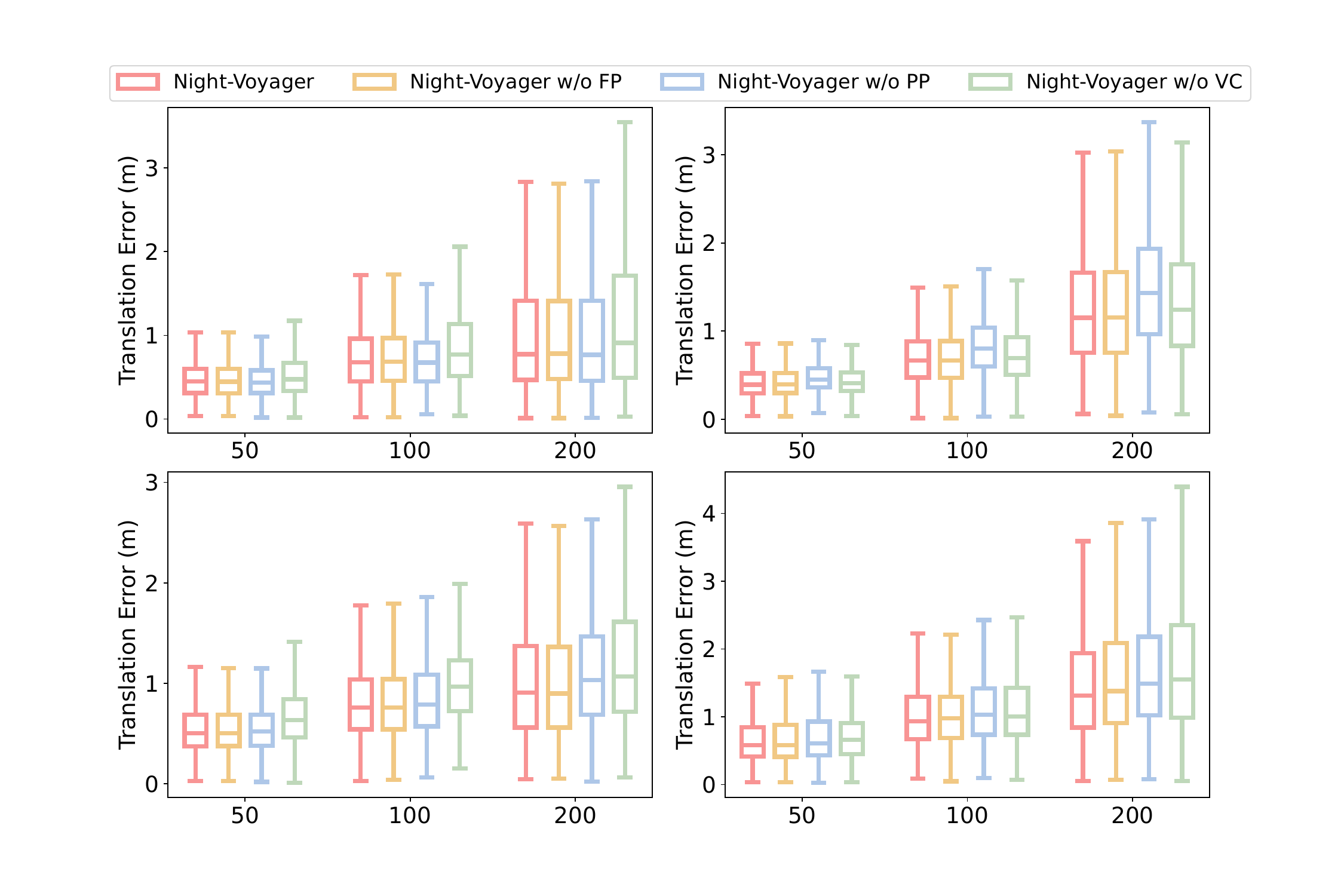} 
    \caption{Comparison of RPE results on different sequences. (a) \textit{Scene\_02}. (b) \textit{Scene\_06}. (c) \textit{Scene\_08}. (d) \textit{Scene\_09}.} 
    \label{Fig: relative pose error}
    \vspace{-10pt}
\end{figure}

In Fig. \ref{Fig: relative pose error}, we give the RPE results of Night-Voyager w/o FP, Night-Voyager w/o PP, Night-Voyager w/o VC, and Night-Voyager to demonstrate the effect of feature points, prior poses, and virtual centers. Three different segment lengths (50 m, 100 m, and 200 m) are selected for statistical RPE analysis. In Fig. \ref{Fig: relative pose error}, the translation errors increase with the segment length, indicating the pose drift in nocturnal localization. By comparing the translation errors among Night-Voyager, Night-Voyager w/o PP, and Night-Voyager w/o VC, we can see that the drift in most sequences is alleviated, which further proves the effectiveness of prior poses and virtual centers. 

\subsection{Evaluation of Initialization}
From the mapping set of various scenes, we select images with at least six streetlight detection boxes and feed them into the initialization module to estimate the corresponding relative transformation. We gather the following statistics to evaluate the initialization module: 1) \textit{Success Rate (SR)}: The ratio of the number of successful initializations to the total number of input images for each sequence. Initialization is considered successful when the Absolute Pose Error (APE) between the estimated relative pose and the true value is less than a set threshold (0.5 m for translation and 3$^{\circ}$ for rotation). 2) \textit{Average Running Time (ART)}: The average running time of the initialization in different sequences. 3) \textit{Average APE (AAPE)}: Average APE between the estimated relative pose and the true value of the successful initializations. The results are presented in Table \ref{Table: initialization cases}. 

\setlength{\tabcolsep}{5.5pt}
\begin{table*}[!t]
\centering
\renewcommand{\arraystretch}{1.5}
\caption{SR (\%), ART (s), and AAPE (m/$^{\circ}$) of the Initialization Module on Different Sequences}
\begin{tabular}{ccccccccccccc}
\toprule
Method & Metric & \textit{Scene\_01} & \textit{Scene\_02} & \textit{Scene\_03} & \textit{Scene\_04} & \textit{Scene\_05} & \textit{Scene\_06} & \textit{Scene\_07} & \textit{Scene\_08} & \textit{Scene\_09} & \textit{Scene\_10} \\\hline
\multirow{3}{*}{\begin{tabular}[c]{@{}c@{}}w/o\\ Coarse Position\end{tabular}} & SR       & 81.3 & 79.2 & 100.0 & 100.0 & 58.9 & 60.9 & 75.0 & 100.0 & 100.0 & 81.6 \\ 
& ART      & 2.47 & 5.38 & 0.76 & 0.69 & 4.82 & 1.76 & 6.49 & 5.72 & 2.11 & 6.11 \\ 
& AAPE     & 0.25/0.40 & 0.18/0.51 & 0.32/1.66 & 0.31/0.28 & 0.24/0.74 & 0.21/0.63 & 0.25/0.34 & 0.35/0.74 & 0.34/0.45 & 0.25/0.57 \\\hline
\multirow{3}{*}{\begin{tabular}[c]{@{}c@{}}w/\\ Coarse Position\\(10 m)\end{tabular}} & SR       & 88.9 & 94.2 & 100.0 & 100.0 & 85.7 & 87.8 & 81.8 & 100.0 & 100.0 & 90.0\\ 
& ART      & 0.27 & 0.50 & 0.09 & 0.11 & 0.51 & 0.18 & 0.35 & 0.47 & 0.13 & 0.41\\ 
& AAPE     & 0.25/0.48 & 0.20/0.64 & 0.32/1.66 & 0.31/0.28 & 0.27/0.70 & 0.22/0.57 & 0.23/0.31 & 0.35/0.74 & 0.34/0.45 & 0.25/0.55\\
\bottomrule
\end{tabular}
\label{Table: initialization cases}
\vspace{-10pt}
\end{table*}

\begin{figure}[!t]
    \centering 
    \includegraphics[width=0.47\textwidth]{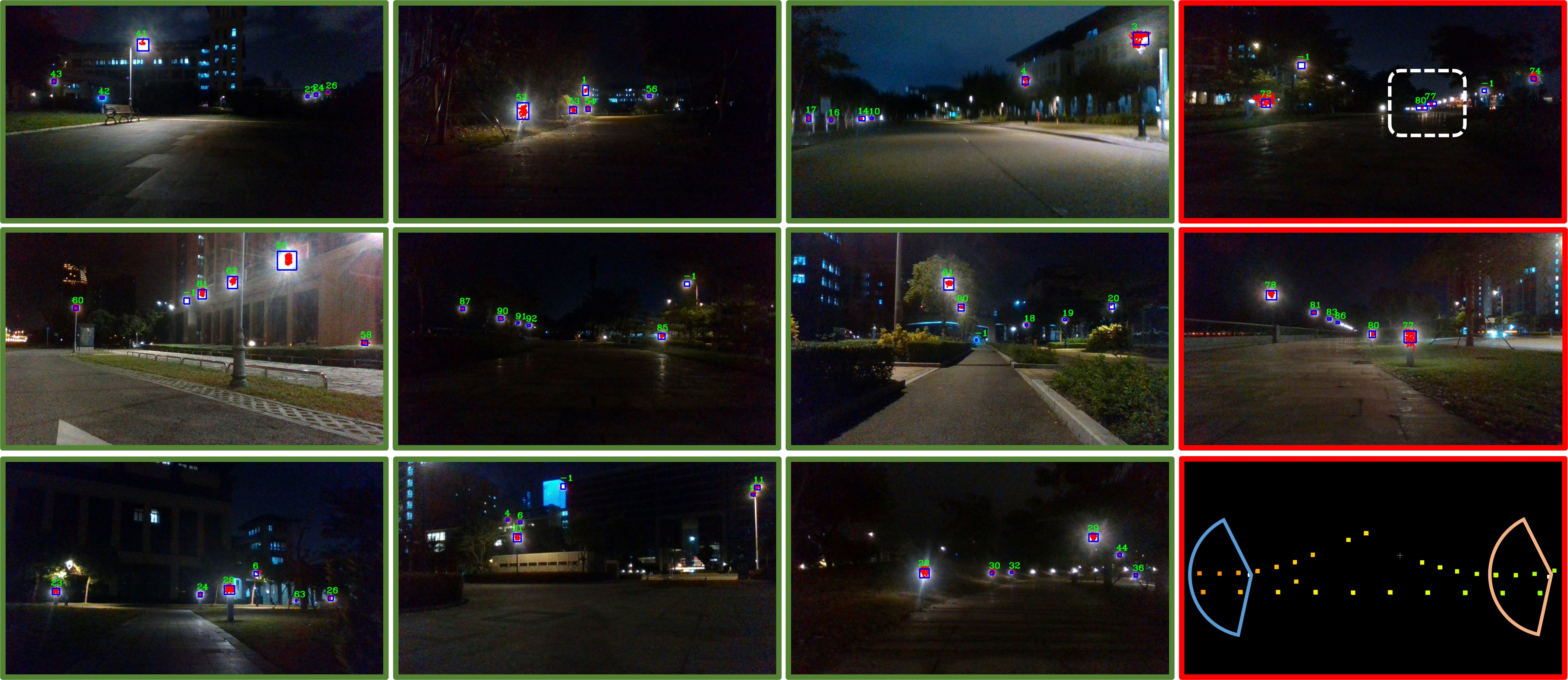} 
    \caption{Examples of successful initialization (green outlines) and failed initialization (red outlines). The red dots in the images represent the projected streetlight points. The blue boxes are streetlight detections. The green numbers above detections indicate the indices of matched streetlight clusters ($-$1: no matched streetlight).} 
    \label{Fig: intialization cases}
    \vspace{-10pt}
\end{figure}

In general, the proposed initialization module exhibits high success rates, especially in scenarios with sparsely distributed streetlights. Fig. \ref{Fig: intialization cases} shows some successful cases (figures with green outlines). We can see that the streetlight points (red dots) projected from the estimated relative poses all fall in the streetlight detection boxes (blue boxes), which proves the high accuracy of the relative pose estimation. We also observe that the initialization module is robust to incorrect streetlight detections (i.e., false detection or streetlights not constructed in the map). All incorrect detections are identified as unmatched streetlights (marked as $-$1). In sequences with a dense distribution of streetlights like \textit{Scene\_02}, \textit{Scene\_05}, and \textit{Scene\_06}, the success rate obviously declines. The fourth column of Fig. \ref{Fig: intialization cases} depicts two common failure cases to illustrate the reasons: 1) Overlapped streetlight detections. For densely distributed streetlights, the detection boxes are prone to overlap, resulting in a significant deviation of the box centers to the true coordinates. Therefore, the affected score in initialization leads to the selection of incorrect triplet matches. 2) Repetitive streetlight distribution patterns. In scenarios with a dense streetlight distribution, there is a higher likelihood of areas with similar streetlight distribution patterns, thus confusing the initialization module. The sector-shaped areas on the bottom right of Fig. \ref{Fig: intialization cases} illustrate this situation. Erroneous initialization can lead to localization failures. To minimize the likelihood of this situation, we add an interface to the initialization module to receive a coarse initial position. When users provide this value, the module excludes solutions that are more than $10$ m from the given position. Significant improvements can be seen in Table \ref{Table: initialization cases}. 
The running time results are shown in Table \ref{Table: initialization cases}. By utilizing prior poses for streetlight map division, the initialization module demonstrates commendable time efficiency, allowing the robot to quickly determine its pose within the map and execute tasks efficiently. 

\subsection{Evaluation of Tracking Recovery}
\begin{figure}[!t]
    \centering 
    \includegraphics[width=0.47\textwidth]{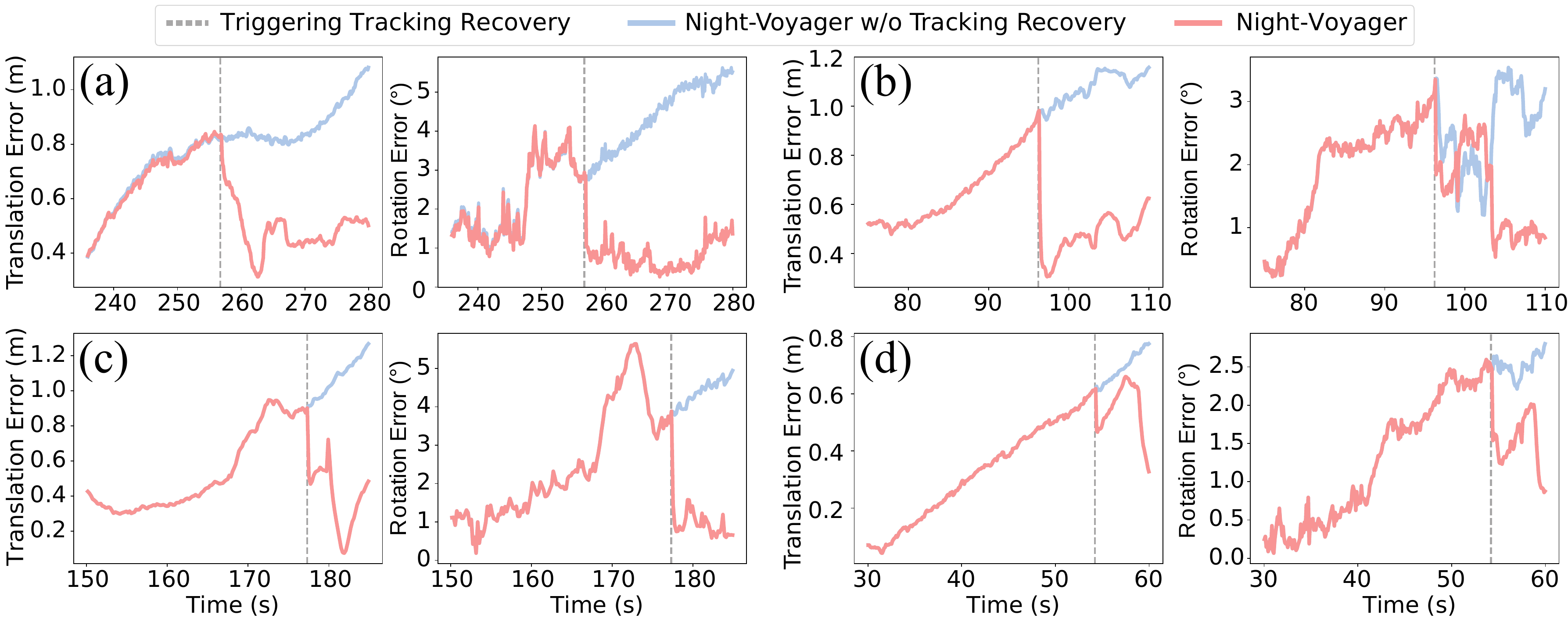} 
    \caption{Comparison of translation and rotation errors with and without the tracking recovery module on different sequences. (a) \textit{Scene\_02}. (b) \textit{Scene\_05}. (c) \textit{Scene\_07}. (d) \textit{Scene\_08}.} 
    \label{Fig: Tracking Recovery Experiment}
    \vspace{-10pt}
\end{figure}

To test the tracking recovery module, we randomly select segments (lasting 30 to 60 seconds) from the collected sequences, during which we turn off the data association module to test whether the tracking recovery module can correct pose drifts. The segments are chosen at moments when few streetlights are visible, effectively imitating scenarios of tracking loss. Table \ref{Table: Tracking Recovery Verification} presents the ATE results of Night-Voyager with and without the tracking recovery module. The results indicate that Night-Voyager without tracking recovery fails to correct the robot pose in most sequences. The failure significantly impacts localization, as large drifts lead to erroneous streetlight matches, thus further exacerbating errors. In contrast, the tracking recovery module enables rapid pose correction through the brute-force matching method and the scoring strategy, thereby mitigating potential accuracy loss. Fig. \ref{Fig: Tracking Recovery Experiment} visually displays the changes in translation and rotation errors between the two variants. When tracking recovery is triggered (gray dashed line), the pose errors decrease immediately. However, without tracking recovery, errors continue to increase, eventually leading to localization failures. 

\setlength{\tabcolsep}{7.0pt}
\begin{table*}[!t]
\centering
\renewcommand{\arraystretch}{1.5}
\caption{ATE Results (m/$^{\circ}$) with and without the Tracking Recovery Module on Different Sequences}
\begin{tabular}{ccccccccccc}
\toprule
Method & \textit{Scene\_01} & \textit{Scene\_02} & \textit{Scene\_03} & \textit{Scene\_04} & \textit{Scene\_05} & \textit{Scene\_06} & \textit{Scene\_07} & \textit{Scene\_08} & \textit{Scene\_09} & \textit{Scene\_10} \\\hline 
\begin{tabular}[c]{@{}c@{}}Night-Voyager\\ w/o TR\end{tabular} & \textbf{0.20}/\textbf{0.56} & \ding{56} & \ding{56} & 15.50/24.74 & \ding{56} & \ding{56} & \ding{56} & \ding{56} & \ding{56} & \ding{56} \\ 
Night-Voyager & \textbf{0.20}/\textbf{0.56} & \textbf{0.39}/\textbf{0.88} & \textbf{0.61}/\textbf{1.06} & \textbf{0.32}/\textbf{1.08} & \textbf{0.30}/\textbf{0.89} & \textbf{0.39}/\textbf{0.57} & \textbf{0.51}/\textbf{1.20} & \textbf{0.50}/\textbf{1.16} & \textbf{0.55}/\textbf{1.05} & \textbf{0.45}/\textbf{0.96} \\
\bottomrule
\end{tabular}
\captionsetup{justification=raggedright,singlelinecheck=false}
\caption*{\ding{56} indicates the algorithm fails on the corresponding sequence. The optimal results are highlighted in \textbf{bold}.}
\label{Table: Tracking Recovery Verification}
\vspace{-10pt}
\end{table*}

\subsection{Running Time and Map Storage}
\begin{figure}[!t]
    \centering 
    \includegraphics[width=0.47\textwidth]{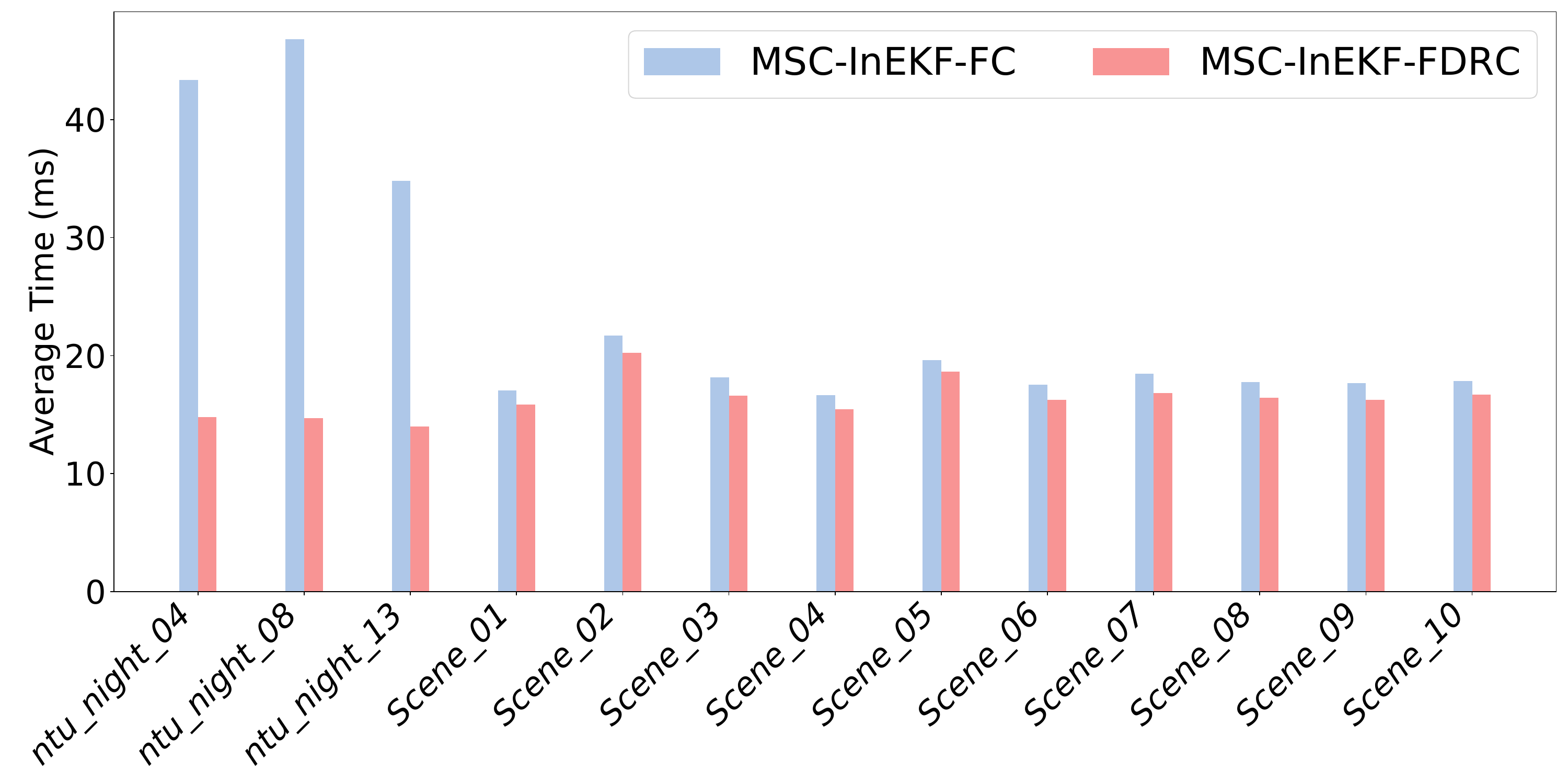} 
    \caption{Time consumption of MSC-InEKF-FC and MSC-InEKF-FDRC on both the MCD and the collected datasets.} 
    \label{Fig: Running Time}
    \vspace{-10pt}
\end{figure}

Fig. \ref{Fig: Running Time} shows the time consumption of Night-Voyager on both the MCD and the collected datasets. Meanwhile, we compare the computation cost of the Night-Voyager variant which exploits MSC-InEKF-FC. For a fair comparison of estimators, the time spent on streetlight detection is excluded from the results. Compared to MSC-InEKF-FC, MSC-InEKF-FDRC (the default estimator of Night-Voyager, see Section \ref{section: MSC-InEKF-FDRC}) is much more efficient in well-lit scenes, such as MCD sequences, confirming the efficiency of MSC-InEKF-FDRC. In the collected dataset, the runtime of MSC-InEKF-FC is slightly higher than that of MSC-InEKF-FDRC. The adverse lighting condition reduces the number of feature points, and thus the computation cost of MSC-InEKF-FC remains low. The map storage is also compared in Table \ref{Table: sequences}. The constructed streetlight maps are highly space-efficient and require significantly less storage than point-cloud maps or feature-point maps. In summary, Night-Voyager delivers accurate and real-time localization results at night with minimal storage requirements, independent of varying lighting conditions. 

\section{Conclusion}
In this work, we propose Night-Voyager, a consistent and efficient framework that harnesses rare object-level information from streetlight maps to fundamentally resolve the \textit{insufficiency} and \textit{inconsistency} bottlenecks inherent in nocturnal visual tasks. A fast global initialization is designed to precisely boot the system. Leveraging cross-modal correspondences, streetlight observations can provide accurate state updates. Additionally, pixel-level visual features endow Night-Voyager with the capability of all-day localization. With the feature-decoupled MSC-InEKF, Night-Voyager achieves consistent and efficient state estimation. Comprehensive simulation and real-world experiments demonstrate the efficacy, robustness, efficiency, and high accuracy of Night-Voyager. 

This work reveals that conventional visual methods are prone to failure in low-light cases, even with active lighting or image enhancement. The genesis is the reliance on pixel-level correspondences. Unlike pixel-level features, object-level features are immune to inconsistency and transience encountered in tracking, leading to substantial accuracy improvements and a fundamental solution for nocturnal visual state estimation. While primarily designed for urban environments, Night-Voyager can be effectively deployed in other dark scenarios featuring stable light sources (e.g., factories, mines, parking lots, wharves, and tunnels). Significantly, this work reinforces a seminal insight that remains equally essential for both model-based and data-driven methodologies: Prior is all you need -- whether from sensors, physical models, or learned models. 

In future work, we aim to develop an all-day navigation framework, leveraging adaptive object-level perception approaches. Another promising direction is to design generalizable methods that integrate semantic information or object tracking to incorporate temporal constraints without relying on prior maps. Finally, an all-day visual place recognition model is also essential for robust long-term navigation. 


\bibliographystyle{IEEEtran} 
\bibliography{references}

\clearpage 


\section*{Supplementary material}

# Supplementary Material for
# Night-Voyager: Consistent and Efficient Nocturnal Vision-Aided State Estimation in Object Maps

This document provides the supplementary material for the manuscript: "Night-Voyager: Consistent and Efficient Nocturnal Vision-Aided State Estimation in Object Maps".

## APPENDIX A
### DERIVATION OF STATE PROPAGATION FUNCTION

In this section, we derive the IMU propagation function for MSC-InEKF-FDRC. For simplicity, we only investigate the states $\mathbf{X}_{I_i}$ and $\mathbf{B}_i$ since other states are not correlated with the covariance propagation. We first rewrite the dynamic propagation model in **Eq. (7)** of **Section IV-A. State Propagation** in the manuscript as:

$$\mathbf{X}_{I_{i+1}} = \mathbf{\Gamma}_i \mathbf{\Psi}(\mathbf{X}_{I_i}) \mathbf{\Upsilon}_i \tag{1}$$

where $\mathbf{\Gamma}_i$, $\mathbf{\Psi}(\mathbf{X}_{I_i})$, and $\mathbf{\Upsilon}_i$ are formulated as:

$$\mathbf{\Gamma}_i = \begin{bmatrix} \mathbf{I}_{3\times3} & \frac{1}{2}\mathbf{g}\delta t^2 & \delta t \\ \mathbf{0}_{1\times3} & 1 & 0 \\ \mathbf{0}_{1\times3} & 0 & 1 \end{bmatrix}, \mathbf{\Upsilon}_i = \begin{bmatrix} \Delta\mathbf{R}_i & \Delta\mathbf{p}_i & \Delta\mathbf{v}_i \\ \mathbf{0}_{1\times3} & 1 & 0 \\ \mathbf{0}_{1\times3} & 0 & 1 \end{bmatrix}$$

$$\mathbf{\Psi}(\mathbf{X}_{I_i}) = \begin{bmatrix} {}^L\mathbf{R}_{I_i} & {}^L\mathbf{p}_{I_i} + {}^L\mathbf{v}_{I_i}\delta t & {}^L\mathbf{v}_{I_i} \\ \mathbf{0}_{1\times3} & 1 & 0 \\ \mathbf{0}_{1\times3} & 0 & 1 \end{bmatrix}. \tag{2}$$

For $\mathbf{\Psi}(\mathbf{X}_{I_i})$, it has the following properties:

$$\mathbf{\Psi}(\mathbf{X}_1\mathbf{X}_2) = \mathbf{\Psi}(\mathbf{X}_1)\mathbf{\Psi}(\mathbf{X}_2) \tag{3}$$

$$\mathbf{\Psi}(\mathrm{Exp}(\boldsymbol{\xi})) \approx \mathrm{Exp}(\mathbf{F}\boldsymbol{\xi}), \quad \mathbf{F} = \begin{bmatrix} \mathbf{I}_{3\times3} & \mathbf{0}_{3\times3} & \mathbf{0}_{3\times3} \\ \mathbf{0}_{3\times3} & \mathbf{I}_{3\times3} & \delta t\mathbf{I}_{3\times3} \\ \mathbf{0}_{3\times3} & \mathbf{0}_{3\times3} & \mathbf{I}_{3\times3} \end{bmatrix}. \tag{4}$$

Using **Eq. (3)** of **Section III-A. Theoretical Background** in the manuscript, (1) can be formulated as:

$$\mathrm{Exp}(\boldsymbol{\xi}_{I_{i+1}})\widehat{\mathbf{X}}_{I_{i+1}} = \mathbf{\Gamma}_i \mathbf{\Psi}(\mathrm{Exp}(\boldsymbol{\xi}_{I_i})\widehat{\mathbf{X}}_{I_i})\mathbf{\Upsilon}_i$$
$$= \mathbf{\Gamma}_i \mathbf{\Psi}(\mathrm{Exp}(\boldsymbol{\xi}_{I_i}))\mathbf{\Psi}(\widehat{\mathbf{X}}_{I_i})\mathbf{\Upsilon}_i \approx \mathbf{\Gamma}_i\mathrm{Exp}(\mathbf{F}\boldsymbol{\xi}_{I_i})\mathbf{\Psi}(\widehat{\mathbf{X}}_{I_i})\mathbf{\Upsilon}_i$$
$$= \mathrm{Exp}(\mathbf{Ad}(\mathbf{\Gamma}_i)\mathbf{F}\boldsymbol{\xi}_{I_i})\mathbf{\Gamma}_i\mathbf{\Psi}(\widehat{\mathbf{X}}_{I_i})\mathbf{\Upsilon}_i$$
$$= \mathrm{Exp}(\mathbf{Ad}(\mathbf{\Gamma}_i)\mathbf{F}\boldsymbol{\xi}_{I_i})\widehat{\mathbf{X}}_{I_{i+1}}. \tag{5}$$

Therefore, we can derive the formulation of $\mathbf{\Phi}_{II}$ as:

$$\mathbf{\Phi}_{II} = \mathbf{Ad}(\mathbf{\Gamma}_i)\mathbf{F} = \begin{bmatrix} \mathbf{I}_{3\times3} & \mathbf{0}_{3\times3} & \mathbf{0}_{3\times3} \\ \frac{1}{2}(\mathbf{g})_\times \delta t^2 & \mathbf{I}_{3\times3} & \mathbf{I}_{3\times3}\delta t \\ (\mathbf{g})_\times \delta t & \mathbf{0}_{3\times3} & \mathbf{I}_{3\times3} \end{bmatrix}. \tag{6}$$

To derive the relationship between $\boldsymbol{\xi}_{I_{i+1}}$ and $\boldsymbol{\zeta}_{I_i}$, we first rewrite $\mathbf{\Upsilon}_i$ as:

$$\mathbf{\Upsilon}_i = \widehat{\mathbf{\Upsilon}}_i \widetilde{\mathbf{\Upsilon}}_i = \begin{bmatrix} \Delta\widehat{\mathbf{R}}_i & \Delta\widehat{\mathbf{p}}_i & \Delta\widehat{\mathbf{v}}_i \\ \mathbf{0}_{1\times3} & 1 & 0 \\ \mathbf{0}_{1\times3} & 0 & 1 \end{bmatrix} \begin{bmatrix} \Delta\widetilde{\mathbf{R}}_i & \Delta\widehat{\mathbf{R}}_i^\top \Delta\widetilde{\mathbf{p}}_i & \Delta\widehat{\mathbf{R}}_i^\top \Delta\widetilde{\mathbf{v}}_i \\ \mathbf{0}_{1\times3} & 1 & 0 \\ \mathbf{0}_{1\times3} & 0 & 1 \end{bmatrix}$$

$$= \widehat{\mathbf{\Upsilon}}_i \mathrm{Exp}\left(\begin{bmatrix} \delta\Delta\boldsymbol{\theta}_i \\ \mathbf{J}_l^{-1}(\delta\Delta\boldsymbol{\theta}_i)\Delta\widehat{\mathbf{R}}_i^\top\Delta\widetilde{\mathbf{p}}_i \\ \mathbf{J}_l^{-1}(\delta\Delta\boldsymbol{\theta}_i)\Delta\widehat{\mathbf{R}}_i^\top\Delta\widetilde{\mathbf{v}}_i \end{bmatrix}\right) \approx \widehat{\mathbf{\Upsilon}}_i \mathrm{Exp}\left(\begin{bmatrix} \delta\Delta\boldsymbol{\theta}_i \\ \Delta\widehat{\mathbf{R}}_i^\top\Delta\widetilde{\mathbf{p}}_i \\ \Delta\widehat{\mathbf{R}}_i^\top\Delta\widetilde{\mathbf{v}}_i \end{bmatrix}\right) \tag{7}$$

where $\Delta\widetilde{\mathbf{R}}_i$, $\Delta\widetilde{\mathbf{p}}_i$ and $\Delta\widetilde{\mathbf{v}}_i$ are respectively formulated as:

$$\Delta\widetilde{\mathbf{R}}_i = \mathrm{Exp}(-\mathbf{J}_r(\widetilde{\boldsymbol{\omega}}_i\delta t)(\delta\mathbf{b}_{g_i} + \mathbf{n}_{dg})\delta t) = \mathrm{Exp}(\delta\Delta\boldsymbol{\theta}_i)$$
$$\Delta\widetilde{\mathbf{p}}_i = \mathbf{\Xi}_4(\delta\mathbf{b}_{g_i} + \mathbf{n}_{dg}) - \mathbf{\Xi}_2(\delta\mathbf{b}_{a_i} + \mathbf{n}_{da}) \tag{8}$$
$$\Delta\widetilde{\mathbf{v}}_i = \mathbf{\Xi}_3(\delta\mathbf{b}_{g_i} + \mathbf{n}_{dg}) - \mathbf{\Xi}_1(\delta\mathbf{b}_{a_i} + \mathbf{n}_{da})$$

where $\mathbf{n}_{dg} \sim \mathcal{N}(\mathbf{0}_{3\times1}, \frac{\sigma_g^2}{\delta t}\mathbf{I}_{3\times3})$ and $\mathbf{n}_{da} \sim \mathcal{N}(\mathbf{0}_{3\times1}, \frac{\sigma_a^2}{\delta t}\mathbf{I}_{3\times3})$ are the discretized noises. The measurements $\widetilde{\boldsymbol{\omega}}_i$ and $\widetilde{\mathbf{a}}_i$ are assumed to be constant during propagation, thus we have $\Delta\widehat{\boldsymbol{\theta}}_i = \widetilde{\boldsymbol{\omega}}_i\delta t = \mathrm{Log}(\Delta\widehat{\mathbf{R}}_i)$. Detailed derivation can be referred to [1], [2]. By substituting (7) into (1), we can derive:

$$\mathrm{Exp}(\boldsymbol{\xi}_{I_{i+1}})\widehat{\mathbf{X}}_{i+1} = \mathbf{\Gamma}_i\mathbf{\Psi}(\mathrm{Exp}(\mathbf{X}_i)\widehat{\mathbf{\Upsilon}}_i\widetilde{\mathbf{\Upsilon}}_i$$

$$\Rightarrow \mathrm{Exp}(\boldsymbol{\xi}_{I_{i+1}}) = \widehat{\mathbf{X}}_{i+1}\widetilde{\mathbf{\Upsilon}}_i\widehat{\mathbf{X}}_{i+1}^{-1} = \mathrm{Exp}(\mathbf{Ad}_{\widehat{\mathbf{X}}_{i+1}}\begin{bmatrix} \delta\Delta\boldsymbol{\theta}_i \\ \Delta\widehat{\mathbf{R}}_i^\top\Delta\widetilde{\mathbf{p}}_i \\ \Delta\widehat{\mathbf{R}}_i^\top\Delta\widetilde{\mathbf{v}}_i \end{bmatrix})$$

$$= \mathrm{Exp}(\mathbf{Ad}_{\widehat{\mathbf{X}}_{i+1}}\begin{bmatrix} \mathbf{I}_{3\times3} & \mathbf{0}_{3\times3} & \mathbf{0}_{3\times3} \\ \mathbf{0}_{3\times3} & \Delta\widehat{\mathbf{R}}_i^\top & \mathbf{0}_{3\times3} \\ \mathbf{0}_{3\times3} & \mathbf{0}_{3\times3} & \Delta\widehat{\mathbf{R}}_i^\top \end{bmatrix}\begin{bmatrix} \delta\Delta\boldsymbol{\theta}_i \\ \Delta\widetilde{\mathbf{p}}_i \\ \Delta\widetilde{\mathbf{v}}_i \end{bmatrix}) \tag{9}$$

$$= \mathrm{Exp}(\mathbf{Ad}_{\widehat{\mathbf{X}}_{i+1}}\begin{bmatrix} \mathbf{I}_{3\times3} & \mathbf{0}_{3\times3} & \mathbf{0}_{3\times3} \\ \mathbf{0}_{3\times3} & \Delta\widehat{\mathbf{R}}_i^\top & \mathbf{0}_{3\times3} \\ \mathbf{0}_{3\times3} & \mathbf{0}_{3\times3} & \Delta\widehat{\mathbf{R}}_i^\top \end{bmatrix}\begin{bmatrix} -\mathbf{J}_r(\Delta\widehat{\boldsymbol{\theta}}_i)\delta t & \mathbf{0}_{3\times3} \\ \mathbf{\Xi}_4 & -\mathbf{\Xi}_2 \\ \mathbf{\Xi}_3 & -\mathbf{\Xi}_1 \end{bmatrix}\begin{bmatrix} \delta\mathbf{b}_{g_i} + \mathbf{n}_{dg} \\ \delta\mathbf{b}_{a_i} + \mathbf{n}_{da} \end{bmatrix}).$$

Then we can derive $\mathbf{\Phi}_{IB}$ and $\mathbf{G}_{II}$ as:

$$\mathbf{\Phi}_{IB} = \mathbf{G}_{II} = \mathbf{Ad}_{\widehat{\mathbf{X}}_{i+1}}\begin{bmatrix} \mathbf{I}_{3\times3} & \mathbf{0}_{3\times3} & \mathbf{0}_{3\times3} \\ \mathbf{0}_{3\times3} & \Delta\widehat{\mathbf{R}}_i^\top & \mathbf{0}_{3\times3} \\ \mathbf{0}_{3\times3} & \mathbf{0}_{3\times3} & \Delta\widehat{\mathbf{R}}_i^\top \end{bmatrix}\begin{bmatrix} -\mathbf{J}_r(\Delta\widehat{\boldsymbol{\theta}}_i) & \mathbf{0}_{3\times3} \\ \mathbf{\Xi}_4 & -\mathbf{\Xi}_2 \\ \mathbf{\Xi}_3 & -\mathbf{\Xi}_1 \end{bmatrix}$$

$$= \begin{bmatrix} -{}^L\widehat{\mathbf{R}}_{I_{i+1}}\mathbf{J}_r(\Delta\widehat{\boldsymbol{\theta}}_i)\delta t & \mathbf{0}_{3\times3} \\ -({}^L\widehat{\mathbf{p}}_{I_{i+1}})_\times {}^L\widehat{\mathbf{R}}_{I_{i+1}}\mathbf{J}_r(\Delta\widehat{\boldsymbol{\theta}}_i)\delta t + \widehat{\mathbf{R}}_{I_i}\mathbf{\Xi}_4 & -\widehat{\mathbf{R}}_{I_i}\mathbf{\Xi}_2 \\ -({}^L\widehat{\mathbf{v}}_{I_{i+1}})_\times {}^L\widehat{\mathbf{R}}_{I_{i+1}}\mathbf{J}_r(\Delta\widehat{\boldsymbol{\theta}}_i)\delta t + \widehat{\mathbf{R}}_{I_i}\mathbf{\Xi}_3 & -\widehat{\mathbf{R}}_{I_i}\mathbf{\Xi}_1 \end{bmatrix}. \tag{10}$$

On the other hand, the biases can be discretized as:

$$\mathbf{b}_{g_{i+1}} = \mathbf{b}_{g_i} + \mathbf{n}_{dbg}\delta t \tag{11}$$
$$\mathbf{b}_{a_{i+1}} = \mathbf{b}_{a_i} + \mathbf{n}_{dba}\delta t \tag{12}$$

where $\mathbf{n}_{dbg} \sim \mathcal{N}(\mathbf{0}_{3\times1}, \frac{\sigma_{bg}^2}{\delta t}\mathbf{I}_{3\times3})$ and $\mathbf{n}_{dba} \sim \mathcal{N}(\mathbf{0}_{3\times1}, \frac{\sigma_{ba}^2}{\delta t}\mathbf{I}_{3\times3})$ are the discretized noises. We can then derive:

$$\delta\mathbf{b}_{g_{i+1}} = \delta\mathbf{b}_{g_i} + \mathbf{n}_{dbg}\delta t \tag{13}$$
$$\delta\mathbf{b}_{a_{i+1}} = \delta\mathbf{b}_{a_i} + \mathbf{n}_{dba}\delta t. \tag{14}$$

Therefore, $\Phi_i^{i+1}$ and $\mathbf{G}_i$ are formulated as:

$$\Phi_i^{i+1} = \begin{bmatrix} \Phi_{II} & \Phi_{IB} \\ \mathbf{0}_{6\times 9} & \mathbf{I}_{6\times 6} \end{bmatrix}, \quad \mathbf{G}_i = \begin{bmatrix} \mathbf{G}_{II} & \mathbf{0}_{9\times 6} \\ \mathbf{0}_{6\times 6} & \mathbf{I}_{6\times 6}\delta t \end{bmatrix}. \quad (15)$$

## APPENDIX B
## DERIVATION OF OBSERVATION JACOBIAN MATRICES

### A. Odometer Observation

By replacing the state in error function **Eq. (14)** of **Section IV-B. Odometer Observation** with **Eq. (5)** of **Section III-B. Notation and State Definition** in the manuscript, the linearized observation error function is derived as:

$$\begin{aligned}
\mathbf{z}_{o_k} &= \widetilde{\mathbf{y}}_{o_k} - \widehat{\mathbf{y}}_{o_k} \\
&= {}^O\mathbf{R}_I(\text{Exp}(\boldsymbol{\xi}_{\boldsymbol{R}_{LI_k}})^L\widehat{\mathbf{R}}_{I_k})^\top \left[ \text{Exp}(\boldsymbol{\xi}_{\boldsymbol{R}_{LI_k}})^L\widehat{\mathbf{v}}_{I_k} \right. \\
&\quad \left. + \mathbf{J}_l(\boldsymbol{\xi}_{\boldsymbol{R}_{LI_k}})\boldsymbol{\xi}_{\boldsymbol{v}_{LI_k}} \right] - {}^O\mathbf{R}_I{}^L\widehat{\mathbf{R}}_{I_k}^\top {}^L\widehat{\mathbf{v}}_{I_k} + \mathbf{n}_{o_k} \\
&\approx {}^O\mathbf{R}_I{}^L\widehat{\mathbf{R}}_{I_k}^\top \text{Exp}(\boldsymbol{\xi}_{\boldsymbol{R}_{LI_k}})^\top \left[ \text{Exp}(\boldsymbol{\xi}_{\boldsymbol{R}_{LI_k}})^L\widehat{\mathbf{v}}_{I_k} + \boldsymbol{\xi}_{\boldsymbol{v}_{LI_k}} \right] \\
&\quad - {}^O\mathbf{R}_I{}^L\widehat{\mathbf{R}}_{I_k}^\top {}^L\widehat{\mathbf{v}}_{I_k} + \mathbf{n}_{o_k} \\
&\approx {}^O\mathbf{R}_I{}^L\widehat{\mathbf{R}}_{I_k}^\top \text{Exp}(\boldsymbol{\xi}_{\boldsymbol{R}_{LI_k}})^\top \boldsymbol{\xi}_{\boldsymbol{v}_{LI_k}} + \mathbf{n}_{o_k} \\
&\approx {}^O\mathbf{R}_I{}^L\widehat{\mathbf{R}}_{I_k}^\top \boldsymbol{\xi}_{\boldsymbol{v}_{LI_k}} + \mathbf{n}_{o_k}. \quad (16)
\end{aligned}$$

Then the Jacobian matrix $\mathbf{H}_{o_k}$ of $\mathbf{z}_{o_k}$ with respect to $\boldsymbol{\xi}_k$ is formulated as:

$$\begin{aligned}
\mathbf{H}_{o_k} &= \begin{bmatrix} \frac{\partial \mathbf{z}_{o_k}}{\partial \boldsymbol{\xi}_{R_{LI_k}}} & \frac{\partial \mathbf{z}_{o_k}}{\partial \boldsymbol{\xi}_{p_{LI_k}}} & \frac{\partial \mathbf{z}_{o_k}}{\partial \boldsymbol{\xi}_{v_{LI_k}}} & \frac{\partial \mathbf{z}_{o_k}}{\partial \boldsymbol{\xi}_{B_k}} & \frac{\partial \mathbf{z}_{o_k}}{\partial \boldsymbol{\xi}_{C_k}} & \frac{\partial \mathbf{z}_{o_k}}{\partial \boldsymbol{\xi}_{G_k}} \end{bmatrix} \quad (17) \\
&= \begin{bmatrix} \mathbf{0}_{3\times 3} & \mathbf{0}_{3\times 3} & {}^O\mathbf{R}_I{}^L\widehat{\mathbf{R}}_{I_k}^\top & \mathbf{0}_{3\times 6} & \mathbf{0}_{3\times 6c} & \mathbf{0}_{3\times 9} \end{bmatrix}. \quad (18)
\end{aligned}$$

### B. Feature Observation

By replacing the state in error function **Eq. (16)** of **Section IV-C. Feature Observation** with **Eq. (5)** of **Section III-B. Notation and State Definition** in the manuscript, the linearized observation error function is derived as:

$$\begin{aligned}
\mathbf{z}_{f_{j_k}} &= \widetilde{\mathbf{y}}_{f_{j_k}} - \widehat{\mathbf{y}}_{f_{j_k}} \\
&= h\left[{}^C\mathbf{R}_I{}^L\mathbf{R}_{I_{j_k}}^\top({}^L\mathbf{p}_{f_k} - {}^L\mathbf{p}_{I_{j_k}}) + {}^C\mathbf{p}_I\right] - \widehat{\mathbf{y}}_{f_{j_k}} + \mathbf{n}_{f_k} \\
&\approx \mathcal{H}_{j_k}\left[{}^C\mathbf{R}_I{}^L\mathbf{R}_{I_{j_k}}^\top({}^L\mathbf{p}_{f_k} - {}^L\mathbf{p}_{I_{j_k}}) + {}^C\mathbf{p}_I \right. \\
&\quad \left. - {}^C\mathbf{R}_I{}^L\widehat{\mathbf{R}}_{I_{j_k}}^\top({}^L\widehat{\mathbf{p}}_{f_k} - {}^L\widehat{\mathbf{p}}_{I_{j_k}}) - {}^C\mathbf{p}_I\right] + \mathbf{n}_{f_k} \\
&= \mathcal{H}_{j_k}\left[{}^C\mathbf{R}_I(\text{Exp}(\boldsymbol{\xi}_{\boldsymbol{R}_{LI_k}})^L\widehat{\mathbf{R}}_{I_{j_k}})^\top \left[\text{Exp}(\boldsymbol{\xi}_{\boldsymbol{R}_{LG_k}})^L\widehat{\mathbf{p}}_{f_k} + \right. \right. \\
&\quad \left. \left. \mathbf{J}_l(\boldsymbol{\xi}_{\boldsymbol{R}_{LG_k}})\boldsymbol{\xi}_{\boldsymbol{p}_{Lf_k}} - \text{Exp}(\boldsymbol{\xi}_{\boldsymbol{R}_{LI_k}})^L\widehat{\mathbf{p}}_{I_{j_k}} - \mathbf{J}_l(\boldsymbol{\xi}_{\boldsymbol{R}_{LI_k}})\boldsymbol{\xi}_{\boldsymbol{p}_{LI_k}}\right] \right. \\
&\quad \left. - {}^C\mathbf{R}_I{}^L\widehat{\mathbf{R}}_{I_{j_k}}^\top({}^L\widehat{\mathbf{p}}_{f_k} - {}^L\widehat{\mathbf{p}}_{I_{j_k}})\right] + \mathbf{n}_{f_k} \\
&\approx \mathcal{H}_{j_k}\left[{}^C\mathbf{R}_I{}^L\widehat{\mathbf{R}}_{I_{j_k}}^\top(\mathbf{I}_{3\times 3} + (\boldsymbol{\xi}_{\boldsymbol{R}_{LI_k}})\times)\left[(\mathbf{I}_{3\times 3} \right. \right. \\
&\quad \left. \left. -(\boldsymbol{\xi}_{\boldsymbol{R}_{LG_k}})\times)^L\widehat{\mathbf{p}}_{f_k} + \boldsymbol{\xi}_{\boldsymbol{p}_{Lf_k}} - (\mathbf{I}_{3\times 3} + (\boldsymbol{\xi}_{\boldsymbol{R}_{LI_k}})\times)^L\widehat{\mathbf{p}}_{I_{j_k}} \right. \right. \\
&\quad \left. \left. -\boldsymbol{\xi}_{\boldsymbol{p}_{LI_k}}\right] - {}^C\mathbf{R}_I{}^L\widehat{\mathbf{R}}_{I_{j_k}}^\top({}^L\widehat{\mathbf{p}}_{f_k} - {}^L\widehat{\mathbf{p}}_{I_{j_k}})\right] + \mathbf{n}_{f_k} \\
&\approx \mathcal{H}_{j_k}{}^C\mathbf{R}_I{}^L\widehat{\mathbf{R}}_{I_{j_k}}^\top \left[({}^L\widehat{\mathbf{p}}_{f_k})\times\boldsymbol{\xi}_{\boldsymbol{R}_{LI_k}} - ({}^L\widehat{\mathbf{p}}_{f_k})\times\boldsymbol{\xi}_{\boldsymbol{R}_{LG_k}} + \boldsymbol{\xi}_{\boldsymbol{p}_{LI_k}} \right. \\
&\quad \left. -\boldsymbol{\xi}_{\boldsymbol{p}_{Lf_k}}\right] + \mathbf{n}_{f_k}. \quad (19)
\end{aligned}$$

The Jacobian matrices $\mathbf{H}_{f_k}$ of $\mathbf{z}_{f_{j_k}}$ with respect to $\boldsymbol{\xi}_k$ and $\boldsymbol{\xi}_{\boldsymbol{p}_{Lf}}$ are formulated as:

$$\begin{aligned}
\mathbf{H}_{f_{j_k}} &= \begin{bmatrix} \frac{\partial \mathbf{z}_{f_{j_k}}}{\partial \boldsymbol{\xi}_{I_k}} & \frac{\partial \mathbf{z}_{f_{j_k}}}{\partial \boldsymbol{\zeta}_{B_k}} & \frac{\partial \mathbf{z}_{f_{j_k}}}{\partial \boldsymbol{\xi}_{C_k}} & \frac{\partial \mathbf{z}_{f_{j_k}}}{\partial \boldsymbol{\xi}_{T_{LG_k}}} & \frac{\partial \mathbf{z}_{f_{j_k}}}{\partial \boldsymbol{\xi}_{\boldsymbol{p}_{Lf_k}}} \end{bmatrix} \\
&= \begin{bmatrix} \mathbf{0}_{2\times 9} & \mathbf{0}_{2\times 6} & \mathbf{H}_{f_{j_k}}^{\mathbf{X}_C} & \mathbf{H}_{f_{j_k}}^{\mathbf{X}_G} & \mathbf{H}_{f_{j_k}}^{\boldsymbol{p}_{Lf}} \end{bmatrix} \quad (20)
\end{aligned}$$

$$\mathbf{H}_{f_{j_k}}^{\mathbf{X}_C} = \mathcal{H}_{j_k}\begin{bmatrix} \mathbf{0}_{3\times 6} & \cdots & {}^C\mathbf{R}_I{}^L\widehat{\mathbf{R}}_{I_{j_k}}({}^L\widehat{\mathbf{p}}_f)\times & -{}^C\mathbf{R}_I{}^L\widehat{\mathbf{R}}_{I_{j_k}} & \cdots & \mathbf{0}_{3\times 6} \end{bmatrix}$$

$$\mathbf{H}_{f_{j_k}}^{\mathbf{X}_G} = \mathcal{H}_{j_k}\begin{bmatrix} -{}^C\mathbf{R}_I{}^L\widehat{\mathbf{R}}_{I_{j_k}}({}^L\widehat{\mathbf{p}}_f)\times & \mathbf{0}_{3\times 3} \end{bmatrix}, \quad \mathbf{H}_{f_{j_k}}^{\boldsymbol{p}_{Lf}} = \mathcal{H}_{j_k}{}^C\mathbf{R}_I{}^L\widehat{\mathbf{R}}_{I_{j_k}}. \quad (21)$$

### C. Streetlight Observation

By replacing the state in error function **Eq. (25)** of **Section V-B. Data Association and Streetlight Observation** with **Eq. (5)** of **Section III-B. Notation and State Definition** in the manuscript, the observation error function is linearized as:

$$\begin{aligned}
\mathbf{z}_{s_k} &= \widetilde{\mathbf{y}}_{s_k} - \widehat{\mathbf{y}}_{s_k} \\
&= h\left[{}^C\mathbf{R}_I{}^L\mathbf{R}_{I_k}^\top({}^L\mathbf{R}_{G_k}{}^G\widetilde{\mathbf{c}}_s + {}^L\mathbf{p}_{G_k} - {}^L\mathbf{p}_{I_k}) + {}^C\mathbf{p}_I\right] - \widehat{\mathbf{y}}_{s_k} + \mathbf{n}_{s_k} \\
&\approx \mathcal{H}_{k_k}\left[{}^C\mathbf{R}_I{}^L\mathbf{R}_{I_k}^\top({}^L\mathbf{R}_{G_k}{}^G\widetilde{\mathbf{c}}_s + {}^L\mathbf{p}_{G_k} - {}^L\mathbf{p}_{I_k}) + {}^C\mathbf{p}_I \right. \\
&\quad \left. -{}^C\mathbf{R}_I{}^L\widehat{\mathbf{R}}_{I_k}^\top({}^L\widehat{\mathbf{R}}_{G_k}{}^G\widetilde{\mathbf{c}}_s + {}^L\widehat{\mathbf{p}}_{G_k} - {}^L\widehat{\mathbf{p}}_{I_k}) - {}^C\mathbf{p}_I\right] + \mathbf{n}_{s_k} \\
&= \mathcal{H}_{k_k}\left[{}^C\mathbf{R}_I(\text{Exp}(\boldsymbol{\xi}_{\boldsymbol{R}_{LI_k}})^L\widehat{\mathbf{R}}_{I_k})^\top \left[\text{Exp}(\boldsymbol{\xi}_{\boldsymbol{R}_{LG_k}})^L\widehat{\mathbf{R}}_{G_k}{}^G\widetilde{\mathbf{c}}_s + \right. \right. \\
&\quad \left. \left. \text{Exp}(\boldsymbol{\xi}_{\boldsymbol{R}_{LG_k}})^L\widehat{\mathbf{p}}_{G_k} + \mathbf{J}_l(\boldsymbol{\xi}_{\boldsymbol{R}_{LG_k}})\boldsymbol{\xi}_{\boldsymbol{p}_{LG_k}} - \text{Exp}(\boldsymbol{\xi}_{\boldsymbol{R}_{LI_k}})^L\widehat{\mathbf{p}}_{I_k} \right. \right. \\
&\quad \left. \left. -\mathbf{J}_l(\boldsymbol{\xi}_{\boldsymbol{R}_{LI_k}})\boldsymbol{\xi}_{\boldsymbol{p}_{LI_k}}\right] - {}^C\mathbf{R}_I{}^L\widehat{\mathbf{R}}_{I_k}^\top({}^L\widehat{\mathbf{R}}_{G_k}{}^G\widetilde{\mathbf{c}}_s + {}^L\widehat{\mathbf{p}}_{G_k} \right. \\
&\quad \left. -{}^L\widehat{\mathbf{p}}_{I_k})\right] + \mathbf{n}_{s_k} \\
&\approx \mathcal{H}_{k_k}\left[{}^C\mathbf{R}_I{}^L\widehat{\mathbf{R}}_{I_k}^\top(\mathbf{I}_{3\times 3} - (\boldsymbol{\xi}_{\boldsymbol{R}_{LI_k}})\times)\left[(\mathbf{I}_{3\times 3} + (\boldsymbol{\xi}_{\boldsymbol{R}_{LG_k}})\times)^L\widehat{\mathbf{R}}_{G_k}{}^G\widetilde{\mathbf{c}}_s \right. \right. \\
&\quad \left. \left. +(\mathbf{I}_{3\times 3} + (\boldsymbol{\xi}_{\boldsymbol{R}_{LG_k}})\times)^L\widehat{\mathbf{p}}_{G_k} + \boldsymbol{\xi}_{\boldsymbol{p}_{LG_k}} - (\mathbf{I}_{3\times 3} + (\boldsymbol{\xi}_{\boldsymbol{R}_{LI_k}})\times)^L\widehat{\mathbf{p}}_{I_k}\right) \right. \\
&\quad \left. -\boldsymbol{\xi}_{\boldsymbol{p}_{LI_k}}\right] - {}^C\mathbf{R}_I{}^L\widehat{\mathbf{R}}_{I_k}^\top({}^L\widehat{\mathbf{R}}_{G_k}{}^G\widetilde{\mathbf{c}}_s + {}^L\widehat{\mathbf{p}}_{G_k} - {}^L\widehat{\mathbf{p}}_{I_k})\right] + \mathbf{n}_{s_k} \\
&\approx \mathcal{H}_{k_k}{}^C\mathbf{R}_I{}^L\widehat{\mathbf{R}}_{I_k}^\top \left[({}^L\widehat{\mathbf{R}}_{G_k}{}^G\widetilde{\mathbf{c}}_s + {}^L\widehat{\mathbf{p}}_{G_k})\times(\boldsymbol{\xi}_{\boldsymbol{R}_{LI_k}} - \boldsymbol{\xi}_{\boldsymbol{R}_{LG_k}}) \right. \\
&\quad \left. +\boldsymbol{\xi}_{\boldsymbol{p}_{LG_k}} - \boldsymbol{\xi}_{\boldsymbol{p}_{LI_k}}\right] + \mathbf{n}_{s_k}. \quad (22)
\end{aligned}$$

Then the Jacobian matrix $\mathbf{H}_{s_k}$ of $\mathbf{z}_{s_k}$ with respect to $\boldsymbol{\xi}_k$ is formulated as:

$$\begin{aligned}
\mathbf{H}_{s_k} &= \begin{bmatrix} \frac{\partial \mathbf{z}_{p_k}}{\partial \boldsymbol{\xi}_{I_k}} & \frac{\partial \mathbf{z}_{p_k}}{\partial \boldsymbol{\zeta}_{B_k}} & \frac{\partial \mathbf{z}_{p_k}}{\partial \boldsymbol{\xi}_{C_k}} & \frac{\partial \mathbf{z}_{p_k}}{\partial \boldsymbol{\xi}_{T_{LG_k}}} & \frac{\partial \mathbf{z}_{p_k}}{\partial \boldsymbol{\xi}_{\boldsymbol{p}_{Lf_k}}} \end{bmatrix} \\
&= \begin{bmatrix} \mathbf{H}_{s_k}^{\mathbf{X}_I} & \mathbf{0}_{2\times 6} & \mathbf{0}_{2\times 6c} & \mathbf{H}_{s_k}^{\mathbf{X}_{T_{LG}}} & \mathbf{0}_{2\times 3} \end{bmatrix} \quad (23)
\end{aligned}$$

$$\mathbf{H}_{s_k}^{\mathbf{X}_I} = \mathcal{H}_{k_k}\begin{bmatrix} {}^C\mathbf{R}_I{}^L\widehat{\mathbf{R}}_{I_k}^\top({}^L\widehat{\mathbf{c}}_s)\times & -{}^C\mathbf{R}_I{}^L\widehat{\mathbf{R}}_{I_k}^\top & \mathbf{0}_{3\times 3} \end{bmatrix}$$

$$\mathbf{H}_{s_k}^{\mathbf{X}_{T_{LG}}} = \mathcal{H}_{k_k}\begin{bmatrix} -{}^C\mathbf{R}_I{}^L\widehat{\mathbf{R}}_{I_k}^\top({}^L\widehat{\mathbf{c}}_s)\times & {}^C\mathbf{R}_I{}^L\widehat{\mathbf{R}}_{I_k}^\top \end{bmatrix}. \quad (24)$$

### D. Prior Pose Observation

By replacing the state in error function **Eq. (27)** of **Section V-C. Prior Pose Observation** with **Eq. (5)** of **Section III-B. Notation and State Definition** in the manuscript, the linearized observation error function is derived as:

$$\begin{aligned}
\mathbf{z}_{p_k} &= \widetilde{\mathbf{y}}_{p_k} - \widehat{\mathbf{y}}_{p_k} \quad (25) \\
&= \begin{bmatrix} \mathbf{e}_3^\top{}^O\mathbf{R}_I{}^G\widetilde{\mathbf{R}}_{I'}^\top({}^L\mathbf{R}_{G_k}^\top({}^L\mathbf{p}_{I_k} - {}^L\mathbf{p}_{G_k}) - {}^G\widetilde{\mathbf{p}}_{I'}) \\ \mathbf{e}_3^\top{}^O\mathbf{R}_I{}^G\widetilde{\mathbf{R}}_{I'}^\top{}^L\mathbf{R}_{G_k}^\top{}^L\mathbf{R}_{I_k}{}^O\mathbf{R}_I^\top\mathbf{e}_3 \end{bmatrix}
\end{aligned}$$

$$-\begin{bmatrix} \mathbf{e}_3^\top {}^O\mathbf{R}_I {}^G\widetilde{\mathbf{R}}_{I'}^\top ({}^L\widehat{\mathbf{R}}_{G_k}^\top ({}^L\widehat{\mathbf{p}}_{I_k} - {}^L\widehat{\mathbf{p}}_{G_k}) - {}^G\widetilde{\mathbf{p}}_{I'}) \\ \mathbf{e}_3^\top {}^O\mathbf{R}_I {}^G\widetilde{\mathbf{R}}_{I'}^\top {}^L\widehat{\mathbf{R}}_{G_k}^\top {}^L\widehat{\mathbf{R}}_{I_k} {}^O\mathbf{R}_I^\top \mathbf{e}_3 \end{bmatrix} + \mathbf{n}_{p_k}.$$

For the first element of $\mathbf{z}_{p_k}$, we have:

$$\begin{aligned} z_{p_k,1} &= \mathbf{e}_3^\top {}^O\mathbf{R}_I {}^G\widetilde{\mathbf{R}}_{I'}^\top ({}^L\mathbf{R}_{G_k}^\top ({}^L\mathbf{p}_{I_k} - {}^L\mathbf{p}_{G_k}) - {}^G\widetilde{\mathbf{p}}_{I'}) - \\ &\quad - \mathbf{e}_3^\top {}^O\mathbf{R}_I {}^G\widetilde{\mathbf{R}}_{I'}^\top ({}^L\widehat{\mathbf{R}}_{G_k}^\top ({}^L\widehat{\mathbf{p}}_{I_k} - {}^L\widehat{\mathbf{p}}_{G_k}) - {}^G\widetilde{\mathbf{p}}_{I'}) \\ &= \mathbf{e}_3^\top {}^O\mathbf{R}_I {}^G\widetilde{\mathbf{R}}_{I'}^\top \Big[ (\mathrm{Exp}(\boldsymbol{\xi}_{R_{LI_k}}){}^L\widehat{\mathbf{R}}_{G_k})^\top \big[ (\mathrm{Exp}(\boldsymbol{\xi}_{R_{LI_k}}){}^L\widehat{\mathbf{p}}_{I_k} \\ &\quad + \mathbf{J}_l(\boldsymbol{\xi}_{R_{LI_k}})\boldsymbol{\xi}_{p_{LI_k}} - \mathrm{Exp}(\boldsymbol{\xi}_{R_{LG_k}}){}^L\widehat{\mathbf{p}}_{G_k} - \mathbf{J}_l(\boldsymbol{\xi}_{R_{LG_k}})\boldsymbol{\xi}_{p_{LG_k}} \big] \\ &\quad - {}^G\widetilde{\mathbf{p}}_{I'} - {}^L\widehat{\mathbf{R}}_{G_k}^\top ({}^L\widehat{\mathbf{p}}_{I_k} - {}^L\widehat{\mathbf{p}}_{G_k}) + {}^G\widetilde{\mathbf{p}}_{I'} \Big] + n_{p_k,1} \\ &\approx \mathbf{e}_3^\top {}^O\mathbf{R}_I {}^G\widetilde{\mathbf{R}}_{I'}^\top \Big[ {}^L\widehat{\mathbf{R}}_{G_k}^\top (\mathbf{I}_{3\times 3} - (\boldsymbol{\xi}_{R_{LG_k}})_\times) \big[ (\mathbf{I}_{3\times 3} \\ &\quad + (\boldsymbol{\xi}_{R_{LI_k}})_\times){}^L\widehat{\mathbf{p}}_{I_k} + \boldsymbol{\xi}_{p_{LI_k}} - (\mathbf{I}_{3\times 3} + (\boldsymbol{\xi}_{R_{LG_k}})_\times){}^L\widehat{\mathbf{p}}_{G_k} \\ &\quad - \boldsymbol{\xi}_{p_{LG_k}} \big] - {}^L\widehat{\mathbf{R}}_{G_k}^\top ({}^L\widehat{\mathbf{p}}_{I_k} - {}^L\widehat{\mathbf{p}}_{G_k}) \Big] + n_{p_k,1} \\ &\approx \mathbf{e}_3^\top {}^O\mathbf{R}_I {}^G\widetilde{\mathbf{R}}_{I'}^\top {}^L\widehat{\mathbf{R}}_{G_k}^\top \Big[ ({}^L\widehat{\mathbf{p}}_{I_k})_\times \boldsymbol{\xi}_{R_{LG_k}} - ({}^L\widehat{\mathbf{p}}_{I_k})_\times \boldsymbol{\xi}_{R_{LI_k}} \\ &\quad + \boldsymbol{\xi}_{p_{LI_k}} - \boldsymbol{\xi}_{p_{LG_k}} \Big] + n_{p_k,1}. \end{aligned} \quad (26)$$

For the second element of $\mathbf{z}_{p_k}$, we have:

$$\begin{aligned} z_{p_k,2} &= \mathbf{e}_3^\top {}^O\mathbf{R}_I {}^G\widetilde{\mathbf{R}}_{I'}^\top {}^L\mathbf{R}_{G_k}^\top {}^L\mathbf{R}_{I_k} {}^O\mathbf{R}_I^\top \mathbf{e}_3 \\ &\quad - \mathbf{e}_3^\top {}^O\mathbf{R}_I {}^G\widetilde{\mathbf{R}}_{I'}^\top {}^L\widehat{\mathbf{R}}_{G_k}^\top {}^L\widehat{\mathbf{R}}_{I_k} {}^O\mathbf{R}_I^\top \mathbf{e}_3 + n_{p_k,2} \\ &= \mathbf{e}_3^\top {}^O\mathbf{R}_I {}^G\widetilde{\mathbf{R}}_{I'}^\top (\mathrm{Exp}(\boldsymbol{\xi}_{R_{LG_k}}){}^L\widehat{\mathbf{R}}_{G_k})^\top (\mathrm{Exp}(\boldsymbol{\xi}_{R_{LI_k}}){}^L\widehat{\mathbf{R}}_{I_k}) \\ &\quad {}^O\mathbf{R}_I^\top \mathbf{e}_3 - \mathbf{e}_3^\top {}^O\mathbf{R}_I {}^G\widetilde{\mathbf{R}}_{I'}^\top {}^L\widehat{\mathbf{R}}_{G_k}^\top {}^L\widehat{\mathbf{R}}_{I_k} {}^O\mathbf{R}_I^\top \mathbf{e}_3 + n_{p_k,2} \\ &\approx \mathbf{e}_3^\top {}^O\mathbf{R}_I {}^G\widetilde{\mathbf{R}}_{I'}^\top {}^L\widehat{\mathbf{R}}_{G_k}^\top (\mathbf{I}_{3\times 3} - (\boldsymbol{\xi}_{R_{LG_k}})_\times)(\mathbf{I}_{3\times 3} + (\boldsymbol{\xi}_{R_{LI_k}})_\times) \\ &\quad {}^L\widehat{\mathbf{R}}_{I_k} {}^O\mathbf{R}_I^\top \mathbf{e}_3 - \mathbf{e}_3^\top {}^O\mathbf{R}_I {}^G\widetilde{\mathbf{R}}_{I'}^\top {}^L\widehat{\mathbf{R}}_{G_k}^\top {}^L\widehat{\mathbf{R}}_{I_k} {}^O\mathbf{R}_I^\top \mathbf{e}_3 + n_{p_k,2} \\ &\approx \mathbf{e}_3^\top {}^O\mathbf{R}_I {}^G\widetilde{\mathbf{R}}_{I'}^\top {}^L\widehat{\mathbf{R}}_{G_k}^\top ((\boldsymbol{\xi}_{R_{LI_k}})_\times - (\boldsymbol{\xi}_{R_{LG_k}})_\times){}^L\widehat{\mathbf{R}}_{I_k} {}^O\mathbf{R}_I^\top \mathbf{e}_3 \\ &\quad + n_{p_k,2} \\ &= \mathbf{e}_3^\top {}^O\mathbf{R}_I {}^G\widetilde{\mathbf{R}}_{I'}^\top {}^L\widehat{\mathbf{R}}_{G_k}^\top ({}^L\widehat{\mathbf{R}}_{I_k} {}^O\mathbf{R}_I^\top \mathbf{e}_3)_\times \boldsymbol{\xi}_{R_{LG_k}} - \mathbf{e}_3^\top {}^O\mathbf{R}_I {}^G\widetilde{\mathbf{R}}_{I'}^\top \\ &\quad {}^L\widehat{\mathbf{R}}_{G_k}^\top ({}^L\widehat{\mathbf{R}}_{I_k} {}^O\mathbf{R}_I^\top \mathbf{e}_3)_\times \boldsymbol{\xi}_{R_{LI_k}} + n_{p_k,2}. \end{aligned} \quad (27)$$

Then the Jacobian matrix $\mathbf{H}_{p_k}$ of $\mathbf{z}_{p_k}$ with respect to $\boldsymbol{\xi}_k$ is formulated as:

$$\begin{aligned} \mathbf{H}_{p_k} &= \begin{bmatrix} \frac{\partial \mathbf{z}_{p_k}}{\partial \boldsymbol{\xi}_{I_k}} & \frac{\partial \mathbf{z}_{p_k}}{\partial \boldsymbol{\zeta}_{B_k}} & \frac{\partial \mathbf{z}_{p_k}}{\partial \boldsymbol{\xi}_{C_k}} & \frac{\partial \mathbf{z}_{p_k}}{\partial \boldsymbol{\xi}_{T_{LG}}} & \frac{\partial \mathbf{z}_{p_k}}{\partial \boldsymbol{\xi}_{P_{L f_k}}} \end{bmatrix} \\ &= \begin{bmatrix} \mathbf{H}_{p_k}^{\mathbf{X}_I} & \mathbf{0}_{2\times 6} & \mathbf{0}_{2\times 6c} & \mathbf{H}_{p_k}^{\mathbf{X}_{T_{LG}}} & \mathbf{0}_{2\times 3} \end{bmatrix} \end{aligned} \quad (28)$$

$$\mathbf{H}_{p_k}^{\mathbf{X}_I} = \mathbf{e}_3^\top {}^I\mathbf{R}_O^\top {}^G\widetilde{\mathbf{R}}_{I'}^\top \begin{bmatrix} -{}^L\widehat{\mathbf{R}}_{G_k}^\top ({}^L\widehat{\mathbf{p}}_{I_k})_\times & {}^L\widehat{\mathbf{R}}_{G_k} & \mathbf{0}_{1\times 3} \\ -{}^L\widehat{\mathbf{R}}_{G_k}^\top ({}^L\widehat{\mathbf{R}}_{I_k} {}^O\mathbf{R}_I^\top \mathbf{e}_3)_\times & \mathbf{0}_{1\times 3} & \mathbf{0}_{1\times 3} \end{bmatrix}$$

$$\mathbf{H}_{p_k}^{\mathbf{X}_{T_{LG}}} = \mathbf{e}_3^\top {}^I\mathbf{R}_O^\top {}^G\widetilde{\mathbf{R}}_{I'}^\top \begin{bmatrix} {}^L\widehat{\mathbf{R}}_{G_k} ({}^L\widehat{\mathbf{p}}_{I_k})_\times & -{}^L\widehat{\mathbf{R}}_{G_k} \\ {}^L\widehat{\mathbf{R}}_{G_k}^\top ({}^L\widehat{\mathbf{R}}_{I_k} {}^O\mathbf{R}_I^\top \mathbf{e}_3)_\times & \mathbf{0}_{1\times 3} \end{bmatrix}. \quad (29)$$

## APPENDIX C
### DERIVATION OF VARIANCE $\sigma^2_{p,dg}$ AND $\sigma^2_{a,dg}$

To derive the formulation of variance in a more compact form, we use the transformation matrix form to represent the pose. Hence the reprojection error in **Eq. (20)** of **Section V-B. Data Association and Streetlight Observation** in the manuscript can be rewritten as

$$\mathbf{r}_{p,dg} = \widetilde{\mathbf{p}}_{s_{d_k}} - \boldsymbol{h}({}^C\mathbf{T}_I {}^L\widehat{\mathbf{T}}_{I_k}^{-1} {}^L\widehat{\mathbf{T}}_{G_k} {}^G\bar{\widetilde{\mathbf{c}}}_{s_g}) \quad (30)$$

where ${}^G\bar{\widetilde{\mathbf{c}}}_{s_g}$ is the homogeneous coordinate of ${}^G\widetilde{\mathbf{c}}_{s_g}$. The error states of ${}^L\widehat{\mathbf{T}}_{I_k}$ and ${}^L\widehat{\mathbf{T}}_{G_k}$ are respectively denoted as $\boldsymbol{\xi}_{T_{LI_k}} = \begin{bmatrix} \boldsymbol{\xi}_{R_{LI_k}}^\top & \boldsymbol{\xi}_{p_{LI_k}}^\top \end{bmatrix}^\top$ and $\boldsymbol{\xi}_{T_{LG_k}} = \begin{bmatrix} \boldsymbol{\xi}_{R_{LG_k}}^\top & \boldsymbol{\xi}_{p_{LG_k}}^\top \end{bmatrix}^\top$. We denote $r_{p,dg} = \|\mathbf{r}_{p,dg}\|_2$. The Jacobian matrices of $\delta r_{p,dg}$ with respect to $\boldsymbol{\xi}_{T_{LI_k}}$ and $\boldsymbol{\xi}_{T_{LG_k}}$ are derived as:

$$\frac{\partial \delta r_{p,dg}}{\partial \boldsymbol{\xi}_{T_{LI_k}}} = -\frac{(\widetilde{\mathbf{p}}_{s_{d_k}} - \boldsymbol{h}({}^C\mathbf{T}_I {}^L\widehat{\mathbf{T}}_{I_k}^{-1} {}^L\widehat{\mathbf{T}}_{G_k} {}^G\bar{\widetilde{\mathbf{c}}}_{s_g}))^\top \boldsymbol{\mathcal{H}}_k \boldsymbol{\Pi}_1 \frac{\partial \delta^C\bar{\widetilde{\mathbf{c}}}_{s_g}}{\partial \boldsymbol{\xi}_{T_{LI_k}}}}{r_{p,dg}}$$

$$\frac{\partial \delta^C\widehat{\mathbf{c}}_{s_g}}{\partial \boldsymbol{\xi}_{T_{LI_k}}} = \lim_{\boldsymbol{\xi}_{T_{LI_k}} \to 0} \frac{({}^C\mathbf{T}_I \mathrm{Exp}(-\boldsymbol{\xi}_{T_{LI_k}}){}^L\widehat{\mathbf{T}}_{I_k}^{-1} - {}^C\mathbf{T}_I {}^L\widehat{\mathbf{T}}_{I_k}^{-1}){}^L\widehat{\mathbf{T}}_{G_k}{}^G\bar{\widetilde{\mathbf{c}}}_{s_g}}{\boldsymbol{\xi}_{T_{LI_k}}}$$

$$= \lim_{\boldsymbol{\xi}_{T_{LI_k}} \to 0} \frac{-{}^C\mathbf{T}_I {}^L\widehat{\mathbf{T}}_{I_k}^{-1}{}^L\widehat{\mathbf{T}}_{G_k}(\mathrm{Ad}({}^L\widehat{\mathbf{T}}_{G_k}^{-1}{}^L\widehat{\mathbf{T}}_{I_k})\boldsymbol{\xi}_{T_{LI_k}})^\wedge {}^G\bar{\widetilde{\mathbf{c}}}_{s_g}}{\boldsymbol{\xi}_{T_{LI_k}}}$$

$$= -{}^C\mathbf{T}_I {}^L\widehat{\mathbf{T}}_{I_k}^{-1}{}^L\widehat{\mathbf{T}}_{g_k}{}^G\bar{\widetilde{\mathbf{c}}}_{s_g}^\odot \mathrm{Ad}({}^L\widehat{\mathbf{T}}_{G_k}^{-1}{}^L\widehat{\mathbf{T}}_{I_k})$$

$$\frac{\partial \delta r_{p,dg}}{\partial \boldsymbol{\xi}_{T_{LG_k}}} = -\frac{(\widetilde{\mathbf{p}}_{s_{d_k}} - \boldsymbol{h}({}^C\mathbf{T}_I {}^L\widehat{\mathbf{T}}_{I_k}^{-1} {}^L\widehat{\mathbf{T}}_{G_k} {}^G\bar{\widetilde{\mathbf{c}}}_{s_g}))\boldsymbol{\mathcal{H}}_k \boldsymbol{\Pi}_1 \frac{\partial \delta^C\bar{\widetilde{\mathbf{c}}}_{s_g}}{\partial \boldsymbol{\xi}_{T_{LG_k}}}}{r_{p,dg}}$$

$$\frac{\partial \delta^C\bar{\widetilde{\mathbf{c}}}_{s_g}}{\partial \boldsymbol{\xi}_{T_{LG_k}}} = \lim_{\boldsymbol{\xi}_{T_{LG_k}} \to 0} \frac{{}^C\mathbf{T}_I {}^L\widehat{\mathbf{T}}_{I_k}^{-1}(\mathrm{Exp}(\boldsymbol{\xi}_{T_{LG_k}}){}^L\widehat{\mathbf{T}}_{G_k}{}^G\bar{\widetilde{\mathbf{c}}}_{s_j} - {}^L\widehat{\mathbf{T}}_{G_k}{}^G\bar{\widetilde{\mathbf{c}}}_{s_j})}{\boldsymbol{\xi}_{T_{LG_k}}}$$

$$= \lim_{\boldsymbol{\xi}_{T_{LG_k}} \to 0} \frac{{}^C\mathbf{T}_I {}^L\widehat{\mathbf{T}}_{I_k}^{-1}{}^L\widehat{\mathbf{T}}_{G_k}(\mathrm{Ad}({}^L\widehat{\mathbf{T}}_{G_k}^{-1})\boldsymbol{\xi}_{T_{LG_k}})^\wedge {}^G\bar{\widetilde{\mathbf{c}}}_{s_j}}{\boldsymbol{\xi}_{T_{LG_k}}}$$

$$= {}^C\mathbf{T}_I {}^L\widehat{\mathbf{T}}_{I_k}^{-1}{}^L\widehat{\mathbf{T}}_{G_k}{}^G\bar{\widetilde{\mathbf{c}}}_{s_g}^\odot \mathrm{Ad}({}^L\widehat{\mathbf{T}}_{G_k}^{-1}) \quad (31)$$

where ${}^G\bar{\widetilde{\mathbf{c}}}_{s_g}^\odot = \begin{bmatrix} \mathbf{I}_{3\times 3} & -({}^G\widetilde{\mathbf{c}}_{s_g})_\times \\ \mathbf{0}_{1\times 3} & \mathbf{0}_{1\times 3} \end{bmatrix}$ and $\boldsymbol{\Pi}_1 = \begin{bmatrix} \mathbf{I}_{3\times 3} & \mathbf{0}_{3\times 1} \end{bmatrix}$. $\mathbf{P}_{LG_k}^{LI_k}$ denotes the covariance matrix block related to $\boldsymbol{\xi}_{T_{LI_k}}$ and $\boldsymbol{\xi}_{T_{LG_k}}$. $\mathrm{Ad}(\cdot)$ represents the adjacent opertor which can be formulated as the adjoint matrix for $SE(3)$: $\mathrm{Ad}(\mathbf{T}) = \begin{bmatrix} \mathbf{R} & \mathbf{0}_{3\times 3} \\ (\mathbf{p})_\times \mathbf{R} & \mathbf{R} \end{bmatrix}$. Therefore, the variance is as:

$$\sigma^2_{p,dg} = \begin{bmatrix} \frac{\partial \delta r_{p,dg}}{\partial \boldsymbol{\xi}_{T_{LI_k}}} & \frac{\partial \delta r_{p,dg}}{\partial \boldsymbol{\xi}_{T_{LG_k}}} \end{bmatrix} \mathbf{P}_{LG_k}^{LI_k} \begin{bmatrix} \frac{\partial \delta r_{p,dg}}{\partial \boldsymbol{\xi}_{T_{LI_k}}^\top} & \frac{\partial \delta r_{p,dg}}{\partial \boldsymbol{\xi}_{T_{LG_k}}^\top} \end{bmatrix}^\top + \boldsymbol{\Sigma}_{s_d} \quad (32)$$

where $\boldsymbol{\Sigma}_{s_d}$ is the covariance corresponding to the box center.

Similar to $\sigma^2_{p,dg}$, $\sigma^2_{a,dg}$ is also related to ${}^L\widehat{\mathbf{T}}_{I_k}$ and ${}^L\widehat{\mathbf{T}}_{G_k}$. We denote $r_{a,dg} = \|\mathbf{r}_{a,dg}\|_2$ and the Jacobian matrices of $\delta r_{a,dg}$ with respect to $\boldsymbol{\xi}_{T_{LI_k}}$ and $\boldsymbol{\xi}_{T_{LG_k}}$ are:

$$\frac{\partial \delta r_{a,dg}}{\partial \boldsymbol{\xi}_{T_{LI_k}}} = \frac{({}^C\widetilde{\mathbf{p}}_{s_{d_k}} \times {}^C\widehat{\mathbf{c}}_{s_g})^\top}{\|{}^C\widetilde{\mathbf{p}}_{s_{d_k}} \times {}^C\widehat{\mathbf{c}}_{s_g}\|_2}({}^C\widetilde{\mathbf{p}}_{s_{d_k}})_\times (\frac{\mathbf{I}_{3\times 3}}{\|{}^C\widehat{\mathbf{c}}_{s_g}\|_2} \quad (33)$$

$$- \frac{{}^C\widehat{\mathbf{c}}_{s_g}{}^C\widehat{\mathbf{c}}_{s_g}^\top}{\|{}^C\widehat{\mathbf{c}}_{s_g}\|_2^3})\boldsymbol{\Pi}_1 \frac{\partial \delta^C\bar{\widetilde{\mathbf{c}}}_{s_g}}{\partial \boldsymbol{\xi}_{T_{LI_k}}}$$

$$\frac{\partial \delta r_{a,dg}}{\partial \boldsymbol{\xi}_{T_{LG_k}}} = \frac{({}^C\widetilde{\mathbf{p}}_{s_{d_k}} \times {}^C\widehat{\mathbf{c}}_{s_g})^\top}{\|{}^C\widetilde{\mathbf{p}}_{s_{d_k}} \times {}^C\widehat{\mathbf{c}}_{s_g}\|_2}({}^C\widetilde{\mathbf{p}}_{s_{d_k}})_\times (\frac{\mathbf{I}_{3\times 3}}{\|{}^C\widehat{\mathbf{c}}_{s_g}\|_2} \quad (34)$$

$$- \frac{{}^C\widehat{\mathbf{c}}_{s_g}{}^C\widehat{\mathbf{c}}_{s_g}^\top}{\|{}^C\widehat{\mathbf{c}}_{s_g}\|_2^3})\boldsymbol{\Pi}_1 \frac{\partial \delta^C\bar{\widetilde{\mathbf{c}}}_{s_g}}{\partial \boldsymbol{\xi}_{T_{LG_k}}}.$$

The Jacobian matrix of $\delta r_{a,dg}$ with respect to $\delta^C\mathbf{p}_{s_{d_k}}$ is derived as:

$$\frac{\partial \delta r_{a,dg}}{\partial \delta^C\mathbf{p}_{s_{d_k}}} = \frac{({}^C\widetilde{\mathbf{p}}_{s_{d_k}} \times {}^C\widehat{\mathbf{c}}_{s_g})^\top}{\|{}^C\widetilde{\mathbf{p}}_{s_{d_k}} \times {}^C\widehat{\mathbf{c}}_{s_g}\|_2}({}^C\widehat{\mathbf{c}}_{s_g})_\times (\frac{\mathbf{I}_{3\times 3}}{\|{}^C\widetilde{\mathbf{p}}_{s_{d_k}}\|_2}$$

$$-\frac{{}^C\widetilde{\mathbf{p}}_{s_{d_k}}{}^C\widetilde{\mathbf{p}}_{s_{d_k}}^\top}{\|{}^C\widetilde{\mathbf{p}}_{s_{d_k}}\|_2^3}). \tag{35}$$

The final variance is propagated by:

$$\sigma_{a,dg}^2 = \begin{bmatrix}\frac{\partial \delta r_{a,dg}}{\partial \boldsymbol{\xi}_{\boldsymbol{T}_{LI_k}}} & \frac{\partial \delta r_{a,dg}}{\partial \boldsymbol{\xi}_{\boldsymbol{T}_{LG_k}}}\end{bmatrix} \mathbf{P}_{LG_k}^{LI_k} \begin{bmatrix}\frac{\partial \delta r_{a,dg}}{\partial \boldsymbol{\xi}_{\boldsymbol{T}_{LI_k}}^\top} & \frac{\partial \delta r_{a,dg}}{\partial \boldsymbol{\xi}_{\boldsymbol{T}_{LG_k}}^\top}\end{bmatrix}^\top$$
$$+ \frac{\partial \delta r_{a,dg}}{\partial \delta^C \mathbf{p}_{s_{d_k}}} \mathbf{\Pi}_2 \mathbf{\Sigma}_{s_d} \mathbf{\Pi}_2^\top \frac{\partial \delta r_{a,dg}}{\partial \delta^C \mathbf{p}_{s_{d_k}}}^\top \tag{36}$$

where $\mathbf{\Pi}_2 = \begin{bmatrix}\mathbf{I}_{2\times 2} & \mathbf{0}_{2\times 1}\end{bmatrix}^\top$.

# APPENDIX D
# MSC-INEKF-FC

## A. State Representation

For MSC-InEKF-FC, the features are associated with IMU navigation state to form $SE_{2+K}(3)$ group. The system state is defined as:

$$\mathbf{X}_k = (\mathbf{X}_{\boldsymbol{I}_k}, \mathbf{B}_k, \mathbf{X}_{\boldsymbol{C}_k}, \mathbf{X}_{\boldsymbol{G}_k})$$
$$\mathbf{X}_{\boldsymbol{I}_k} = \begin{bmatrix}{}^L\mathbf{R}_{I_k} & {}^L\mathbf{p}_{I_k} & {}^L\mathbf{v}_{I_k} & {}^L\mathbf{p}_{f_k} \\ \mathbf{0}_{3\times 3} & & \mathbf{I}_{3\times 3}\end{bmatrix}$$
$$\mathbf{B}_k = \begin{bmatrix}\mathbf{b}_{g_k}^\top & \mathbf{b}_{a_k}^\top\end{bmatrix}^\top$$
$$\mathbf{X}_{\boldsymbol{C}_k} = (\mathbf{X}_{CP_{k_k}}, \cdots, \mathbf{X}_{CP_{k-c+1_k}}), \quad \mathbf{X}_{CP_{j_k}} = \begin{bmatrix}{}^L\mathbf{R}_{I_{j_k}} & {}^L\mathbf{p}_{I_{j_k}} \\ \mathbf{0}_{1\times 3} & 1\end{bmatrix}$$
$$\mathbf{X}_{\boldsymbol{G}_k} = \begin{bmatrix}{}^L\mathbf{R}_{G_k} & {}^L\mathbf{p}_{G_k} \\ \mathbf{0}_{1\times 3} & 1\end{bmatrix}. \tag{37}$$

We keep the error states in the same formula with that in MSC-InEKF-FDRC, i.e., $\boldsymbol{\xi}_k = \begin{bmatrix}\boldsymbol{\xi}_{I_k}^\top & \boldsymbol{\zeta}_{B_k}^\top & \boldsymbol{\xi}_{C_k}^\top & \boldsymbol{\xi}_{T_{LG_k}}^\top & \boldsymbol{\xi}_{p_{Lf_k}}^\top\end{bmatrix}^\top$. The feature error state $\boldsymbol{\xi}_{p_{Lf_k}}$ is formulated as:

$${}^L\mathbf{p}_{f_k} = \mathrm{Exp}(\boldsymbol{\xi}_{R_{LI_k}}){}^L\widehat{\mathbf{p}}_{f_k} + \mathbf{J}_l(\boldsymbol{\xi}_{R_{LI_k}})\boldsymbol{\xi}_{p_{Lf_k}}. \tag{38}$$

Since $\mathbf{X}_{CP_{k_k}}$ is the clone of current pose, (38) is equivalent to:

$${}^L\mathbf{p}_{f_k} = \mathrm{Exp}(\boldsymbol{\xi}_{R_{CP_{k_k}}}){}^L\widehat{\mathbf{p}}_{f_k} + \mathbf{J}_l(\boldsymbol{\xi}_{p_{CP_{k_k}}})\boldsymbol{\xi}_{p_{Lf_k}}. \tag{39}$$

## B. State Propagation

We also only investigate the states $\mathbf{X}_{\boldsymbol{I}_i}$ and $\mathbf{B}_i$ for simplicity. Since the feature point is associated with propagation, the dynamic propagation model in **Eq. (7)** of **Section IV-A. State Propagation** in the manuscript is rewritten as:

$$\mathbf{X}_{\boldsymbol{I}_{i+1}} = \boldsymbol{\Gamma}_i \boldsymbol{\Psi}(\mathbf{X}_{\boldsymbol{I}_i}) \boldsymbol{\Upsilon}_i \tag{40}$$

where $\boldsymbol{\Gamma}_i$, $\boldsymbol{\Psi}(\mathbf{X}_{\boldsymbol{I}_i})$, and $\boldsymbol{\Upsilon}_i$ are formulated as:

$$\boldsymbol{\Gamma}_i = \begin{bmatrix}\mathbf{I}_{3\times 3} & \frac{1}{2}\mathbf{g}\delta t^2 & \delta t & \mathbf{0}_{3\times 1} \\ \mathbf{0}_{3\times 3} & & \mathbf{I}_{3\times 3}\end{bmatrix}, \boldsymbol{\Upsilon}_i = \begin{bmatrix}\Delta\mathbf{R}_i & \Delta\mathbf{p}_i & \Delta\mathbf{v}_i & \mathbf{0}_{3\times 1} \\ \mathbf{0}_{3\times 3} & & \mathbf{I}_{3\times 3}\end{bmatrix}$$
$$\boldsymbol{\Psi}(\mathbf{X}_{\boldsymbol{I}_i}) = \begin{bmatrix}{}^L\mathbf{R}_{I_i} & {}^L\mathbf{p}_{I_i} + {}^L\mathbf{v}_{I_i}\delta t & {}^L\mathbf{v}_{I_i} & {}^G\mathbf{p}_f \\ \mathbf{0}_{3\times 3} & & \mathbf{I}_{3\times 3}\end{bmatrix}. \tag{41}$$

Using **Eq. (3)** of **Section III-A. Theoretical Background** in the manuscript, (40) can be formulated as:

$$\mathrm{Exp}(\boldsymbol{\xi}_{\boldsymbol{I}_{i+1}})\widehat{\mathbf{X}}_{\boldsymbol{I}_{i+1}} = \boldsymbol{\Gamma}_i\boldsymbol{\Psi}(\mathrm{Exp}(\boldsymbol{\xi}_{\boldsymbol{I}_i})\widehat{\mathbf{X}}_{\boldsymbol{I}_i})\boldsymbol{\Upsilon}_i$$
$$= \boldsymbol{\Gamma}_i\boldsymbol{\Psi}(\mathrm{Exp}(\boldsymbol{\xi}_{\boldsymbol{I}_i}))\boldsymbol{\Psi}(\widehat{\mathbf{X}}_{\boldsymbol{I}_i})\boldsymbol{\Upsilon}_i \approx \boldsymbol{\Gamma}_i\mathrm{Exp}(\mathbf{F}\boldsymbol{\xi}_{\boldsymbol{I}_i})\boldsymbol{\Psi}(\widehat{\mathbf{X}}_{\boldsymbol{I}_i})\boldsymbol{\Upsilon}_i$$
$$= \mathrm{Exp}(\mathbf{Ad}(\boldsymbol{\Gamma}_i)\mathbf{F}\boldsymbol{\xi}_{\boldsymbol{I}_i})\boldsymbol{\Gamma}_i\boldsymbol{\Psi}(\widehat{\mathbf{X}}_{\boldsymbol{I}_i})\boldsymbol{\Upsilon}_i$$
$$= \mathrm{Exp}(\mathbf{Ad}(\boldsymbol{\Gamma}_i)\mathbf{F}\boldsymbol{\xi}_{\boldsymbol{I}_i})\widehat{\mathbf{X}}_{\boldsymbol{I}_{i+1}} \tag{42}$$

where the matrix $\mathbf{F}$ is formulated as:

$$\mathbf{F} = \begin{bmatrix}\mathbf{I}_{3\times 3} & \mathbf{0}_{3\times 3} & \mathbf{0}_{3\times 3} & \mathbf{0}_{3\times 3} \\ \mathbf{0}_{3\times 3} & \mathbf{I}_{3\times 3} & \delta t\mathbf{I}_{3\times 3} & \mathbf{0}_{3\times 3} \\ \mathbf{0}_{3\times 3} & \mathbf{0}_{3\times 3} & \mathbf{I}_{3\times 3} & \mathbf{0}_{3\times 3} \\ \mathbf{0}_{3\times 3} & \mathbf{0}_{3\times 3} & \mathbf{0}_{3\times 3} & \mathbf{I}_{3\times 3}\end{bmatrix}. \tag{43}$$

By expanding $\mathbf{Ad}(\boldsymbol{\Gamma}_i)\mathbf{F}$, we can obtain:

$$\mathbf{Ad}(\boldsymbol{\Gamma}_i)\mathbf{F} = \begin{bmatrix}\mathbf{I}_{3\times 3} & \mathbf{0}_{3\times 3} & \mathbf{0}_{3\times 3} & \mathbf{0}_{3\times 3} \\ \frac{1}{2}(\mathbf{g})_\times \delta t^2 & \mathbf{I}_{3\times 3} & \mathbf{I}_{3\times 3}\delta t & \mathbf{0}_{3\times 3} \\ (\mathbf{g})_\times \delta t & \mathbf{0}_{3\times 3} & \mathbf{I}_{3\times 3} & \mathbf{0}_{3\times 3} \\ \mathbf{0}_{3\times 3} & \mathbf{0}_{3\times 3} & \mathbf{0}_{3\times 3} & \mathbf{I}_{3\times 3}\end{bmatrix} = \begin{bmatrix}\boldsymbol{\Phi}_{II} & \mathbf{0}_{9\times 3} \\ \mathbf{0}_{3\times 9} & \mathbf{I}_{3\times 3}\end{bmatrix}. \tag{44}$$

To derive the relationship between $\boldsymbol{\xi}_{\boldsymbol{I}_{i+1}}$ and $\boldsymbol{\zeta}_{\boldsymbol{I}_i}$, we first rewrite $\boldsymbol{\Upsilon}_i$ as:

$$\boldsymbol{\Upsilon}_i = \widehat{\boldsymbol{\Upsilon}}_i\widetilde{\boldsymbol{\Upsilon}}_i$$
$$= \begin{bmatrix}\Delta\widehat{\mathbf{R}}_i & \Delta\widehat{\mathbf{p}}_i\ \Delta\widehat{\mathbf{v}}_i\ \mathbf{0}_{3\times 1} \\ \mathbf{0}_{3\times 3} & \mathbf{I}_{3\times 3}\end{bmatrix}\begin{bmatrix}\Delta\widetilde{\mathbf{R}}_i & \Delta\widehat{\mathbf{R}}_i^\top\Delta\widetilde{\mathbf{p}}_i\ \Delta\widehat{\mathbf{R}}_i^\top\Delta\widetilde{\mathbf{v}}_i\ \mathbf{0}_{3\times 1} \\ \mathbf{0}_{3\times 3} & \mathbf{I}_{3\times 3}\end{bmatrix}$$
$$= \widehat{\boldsymbol{\Upsilon}}_i\mathrm{Exp}\left(\begin{bmatrix}\delta\Delta\boldsymbol{\theta}_i \\ \mathbf{J}_l^{-1}(\delta\Delta\boldsymbol{\theta}_i)\Delta\widehat{\mathbf{R}}_i^\top\Delta\widetilde{\mathbf{p}}_i \\ \mathbf{J}_l^{-1}(\delta\Delta\boldsymbol{\theta}_i)\Delta\widehat{\mathbf{R}}_i^\top\Delta\widetilde{\mathbf{v}}_i \\ \mathbf{0}_{3\times 1}\end{bmatrix}\right)$$
$$\approx \widehat{\boldsymbol{\Upsilon}}_i\mathrm{Exp}\left(\begin{bmatrix}\delta\Delta\boldsymbol{\theta}_i \\ \Delta\widehat{\mathbf{R}}_i^\top\Delta\widetilde{\mathbf{p}}_i \\ \Delta\widehat{\mathbf{R}}_i^\top\Delta\widetilde{\mathbf{v}}_i \\ \mathbf{0}_{3\times 1}\end{bmatrix}\right). \tag{45}$$

Then similar to the procedure in (9), we can obtain:

$$\mathrm{Exp}(\boldsymbol{\xi}_{\boldsymbol{I}_{i+1}})\widehat{\mathbf{X}}_{i+1} = \boldsymbol{\Gamma}_i\boldsymbol{\Psi}(\mathrm{Exp}(\mathbf{X}_i))\widehat{\boldsymbol{\Upsilon}}_i\widetilde{\boldsymbol{\Upsilon}}_i$$
$$\Rightarrow \mathrm{Exp}(\boldsymbol{\xi}_{\boldsymbol{I}_{i+1}}) = \widehat{\mathbf{X}}_{i+1}\widetilde{\boldsymbol{\Upsilon}}_i\widehat{\mathbf{X}}_{i+1}^{-1}$$
$$= \mathrm{Exp}(\mathbf{Ad}_{\widehat{\mathbf{X}}_{i+1}}\begin{bmatrix}\delta\Delta\boldsymbol{\theta}_i \\ \Delta\widehat{\mathbf{R}}_i^\top\Delta\widetilde{\mathbf{p}}_i \\ \Delta\widehat{\mathbf{R}}_i^\top\Delta\widetilde{\mathbf{v}}_i \\ \mathbf{0}_{3\times 1}\end{bmatrix})$$
$$= \mathrm{Exp}\left(\mathbf{Ad}_{\widehat{\mathbf{X}}_{i+1}}\begin{bmatrix}\mathbf{I}_{3\times 3} & \mathbf{0}_{3\times 3} & \mathbf{0}_{3\times 3} \\ \mathbf{0}_{3\times 3} & \Delta\widehat{\mathbf{R}}_i^\top & \mathbf{0}_{3\times 3} \\ \mathbf{0}_{3\times 3} & \mathbf{0}_{3\times 3} & \Delta\widehat{\mathbf{R}}_i^\top \\ \mathbf{0}_{3\times 3} & \mathbf{0}_{3\times 3} & \mathbf{0}_{3\times 3}\end{bmatrix}\begin{bmatrix}\delta\Delta\boldsymbol{\theta}_i \\ \Delta\widetilde{\mathbf{p}}_i \\ \Delta\widetilde{\mathbf{v}}_i\end{bmatrix}\right). \tag{46}$$

By expanding the expression in the exponential function, the following equation is obtained:

$$\mathbf{Ad}_{\widehat{\mathbf{X}}_{i+1}}\begin{bmatrix}\mathbf{I}_{3\times 3} & \mathbf{0}_{3\times 3} & \mathbf{0}_{3\times 3} \\ \mathbf{0}_{3\times 3} & \Delta\widehat{\mathbf{R}}_i^\top & \mathbf{0}_{3\times 3} \\ \mathbf{0}_{3\times 3} & \mathbf{0}_{3\times 3} & \Delta\widehat{\mathbf{R}}_i^\top \\ \mathbf{0}_{3\times 3} & \mathbf{0}_{3\times 3} & \mathbf{0}_{3\times 3}\end{bmatrix}\begin{bmatrix}\delta\Delta\boldsymbol{\theta}_i \\ \Delta\widetilde{\mathbf{p}}_i \\ \Delta\widetilde{\mathbf{v}}_i\end{bmatrix}$$
$$= \mathbf{Ad}_{\widehat{\mathbf{X}}_{i+1}}\begin{bmatrix}\mathbf{I}_{3\times 3} & \mathbf{0}_{3\times 3} & \mathbf{0}_{3\times 3} \\ \mathbf{0}_{3\times 3} & \Delta\widehat{\mathbf{R}}_i^\top & \mathbf{0}_{3\times 3} \\ \mathbf{0}_{3\times 3} & \mathbf{0}_{3\times 3} & \Delta\widehat{\mathbf{R}}_i^\top \\ \mathbf{0}_{3\times 3} & \mathbf{0}_{3\times 3} & \mathbf{0}_{3\times 3}\end{bmatrix}\begin{bmatrix}-\mathbf{J}_r(\Delta\widehat{\boldsymbol{\theta}}_i) & \mathbf{0}_{3\times 3} \\ \boldsymbol{\Xi}_4 & -\boldsymbol{\Xi}_2 \\ \boldsymbol{\Xi}_3 & -\boldsymbol{\Xi}_1\end{bmatrix}$$

$$
\begin{aligned}
&= \begin{bmatrix}
{}^L\widehat{\mathbf{R}}_{I_{i+1}} & \mathbf{0}_{3\times 3} & \mathbf{0}_{3\times 3} & \mathbf{0}_{3\times 3} \\
({}^L\widehat{\mathbf{p}}_{I_{i+1}})_\times {}^L\widehat{\mathbf{R}}_{I_{i+1}} & {}^L\widehat{\mathbf{R}}_{I_{i+1}} & \mathbf{0}_{3\times 3} & \mathbf{0}_{3\times 3} \\
({}^L\widehat{\mathbf{v}}_{I_{i+1}})_\times {}^L\widehat{\mathbf{R}}_{I_{i+1}} & \mathbf{0}_{3\times 3} & {}^L\widehat{\mathbf{R}}_{I_{i+1}} & \mathbf{0}_{3\times 3} \\
({}^L\widehat{\mathbf{p}}_{f_{i+1}})_\times {}^L\widehat{\mathbf{R}}_{I_{i+1}} & \mathbf{0}_{3\times 3} & \mathbf{0}_{3\times 3} & {}^L\widehat{\mathbf{R}}_{I_{i+1}}
\end{bmatrix} \\
&\quad \begin{bmatrix}
\mathbf{I}_{3\times 3} & \mathbf{0}_{3\times 3} & \mathbf{0}_{3\times 3} \\
\mathbf{0}_{3\times 3} & \Delta\widehat{\mathbf{R}}_i^\top & \mathbf{0}_{3\times 3} \\
\mathbf{0}_{3\times 3} & \mathbf{0}_{3\times 3} & \Delta\widehat{\mathbf{R}}_i^\top \\
\mathbf{0}_{3\times 3} & \mathbf{0}_{3\times 3} & \mathbf{0}_{3\times 3}
\end{bmatrix}
\begin{bmatrix}
-\mathbf{J}_r(\Delta\boldsymbol{\theta}_i) & \mathbf{0}_{3\times 3} \\
\Xi_4 & -\Xi_2 \\
\Xi_3 & -\Xi_1
\end{bmatrix} \\
&= \begin{bmatrix}
-{}^L\widehat{\mathbf{R}}_{I_{i+1}}\mathbf{J}_r(\Delta\boldsymbol{\theta}_i)\delta t & \mathbf{0}_{3\times 3} \\
-({}^L\widehat{\mathbf{p}}_{I_{i+1}})_\times {}^L\widehat{\mathbf{R}}_{I_{i+1}}\mathbf{J}_r(\Delta\boldsymbol{\theta}_i)\delta t + \widehat{\mathbf{R}}_{I_i}\Xi_4 & -\widehat{\mathbf{R}}_{I_i}\Xi_2 \\
-({}^L\widehat{\mathbf{v}}_{I_{i+1}})_\times {}^L\widehat{\mathbf{R}}_{I_{i+1}}\mathbf{J}_r(\Delta\boldsymbol{\theta}_i)\delta t + \widehat{\mathbf{R}}_{I_i}\Xi_3 & -\widehat{\mathbf{R}}_{I_i}\Xi_1 \\
-({}^L\widehat{\mathbf{p}}_{f_{i+1}})_\times {}^L\widehat{\mathbf{R}}_{I_{i+1}}\mathbf{J}_r(\Delta\boldsymbol{\theta}_i)\delta t & \mathbf{0}_{3\times 3}
\end{bmatrix} \\
&= \begin{bmatrix} \boldsymbol{\Phi}_{IB} \\ \boldsymbol{\Phi}_{fB} \end{bmatrix}. \tag{47}
\end{aligned}
$$

We can find that the propagation of feature point error state is associated with the biases. Organizing the above equations, the error state propagation function of MSC-InEKF-FC is formulated as:

$$
\boldsymbol{\xi}_{i+1} = \begin{bmatrix} \boldsymbol{\xi}_{I_{i+1}} \\ \boldsymbol{\xi}_{B_{i+1}} \\ \boldsymbol{\xi}_{\boldsymbol{p}_{Lf_{i+1}}} \end{bmatrix} = \boldsymbol{\Phi}_i^{i+1} \begin{bmatrix} \boldsymbol{\xi}_{I_i} \\ \boldsymbol{\xi}_{B_i} \\ \boldsymbol{\xi}_{\boldsymbol{p}_{Lf_i}} \end{bmatrix} + \mathbf{G}_i \begin{bmatrix} \mathbf{n}_{dg} \\ \mathbf{n}_{da} \\ \mathbf{n}_{dbg} \\ \mathbf{n}_{dba} \end{bmatrix} \tag{48}
$$

$$
= \begin{bmatrix} \boldsymbol{\Phi}_{II} & \boldsymbol{\Phi}_{IB} & \mathbf{0}_{9\times 3} \\ \mathbf{0}_{6\times 9} & \mathbf{I}_{6\times 6} & \mathbf{0}_{3\times 3} \\ \mathbf{0}_{3\times 9} & \boldsymbol{\Phi}_{fB} & \mathbf{I}_{3\times 3} \end{bmatrix} \begin{bmatrix} \boldsymbol{\xi}_{I_i} \\ \boldsymbol{\xi}_{B_i} \\ \boldsymbol{\xi}_{\boldsymbol{p}_{Lf_i}} \end{bmatrix} + \begin{bmatrix} \mathbf{G}_{II} & \mathbf{0}_{9\times 6} \\ \mathbf{0}_{6\times 9} & \mathbf{I}_{6\times 6}\delta t \\ \mathbf{G}_{fI} & \mathbf{0}_{3\times 6} \end{bmatrix} \begin{bmatrix} \mathbf{n}_{dg} \\ \mathbf{n}_{da} \\ \mathbf{n}_{dbg} \\ \mathbf{n}_{dba} \end{bmatrix}.
$$

To analyze the computation cost, we partition $\boldsymbol{\Phi}_i^{i+1}$ and $\mathbf{G}_i$ into several matrix blocks:

$$
\boldsymbol{\Phi}_i^{i+1} \triangleq \begin{bmatrix} \boldsymbol{\Phi}_{\bar{f}\bar{f}} & \mathbf{0}_{15\times 3} \\ \boldsymbol{\Phi}_{f\bar{f}} & \mathbf{I}_{3\times 3} \end{bmatrix} \quad \mathbf{G}_i \triangleq \begin{bmatrix} \mathbf{G}_{\bar{f}\bar{f}} \\ \mathbf{G}_{f\bar{f}} \end{bmatrix} \tag{49}
$$

where $\boldsymbol{\Phi}_{\bar{f}\bar{f}} \in \mathbb{R}^{15\times 15}$ and $\mathbf{G}_{\bar{f}\bar{f}} \in \mathbb{R}^{15\times 15}$ represent the blocks that are independent of the feature, and $\boldsymbol{\Phi}_{f\bar{f}} \in \mathbb{R}^{3\times 15}, \mathbf{G}_{f\bar{f}} \in \mathbb{R}^{3\times 15}$ represent the blocks that are associated with the feature. Then the state covariance matrix is:

$$
\begin{aligned}
\widehat{\mathbf{P}}_{i+1} &= \boldsymbol{\Phi}_i^{i+1} \begin{bmatrix} \widehat{\mathbf{P}}_{\bar{f}\bar{f}} & \widehat{\mathbf{P}}_{\bar{f}f} \\ \widehat{\mathbf{P}}_{f\bar{f}} & \widehat{\mathbf{P}}_{ff} \end{bmatrix} \boldsymbol{\Phi}_i^{i+1\top} + \mathbf{G}_i \mathbf{Q}_d \mathbf{G}_i^\top \\
&= \begin{bmatrix} \boldsymbol{\Phi}_{\bar{f}\bar{f}}\widehat{\mathbf{P}}_{\bar{f}\bar{f}}\boldsymbol{\Phi}_{\bar{f}\bar{f}}^\top & \boldsymbol{\Phi}_{\bar{f}\bar{f}}\widehat{\mathbf{P}}_{\bar{f}\bar{f}}\boldsymbol{\Phi}_{f\bar{f}}^\top + \boldsymbol{\Phi}_{\bar{f}\bar{f}}\widehat{\mathbf{P}}_{\bar{f}f} \\ \boldsymbol{\Phi}_{f\bar{f}}\widehat{\mathbf{P}}_{\bar{f}\bar{f}}\boldsymbol{\Phi}_{\bar{f}\bar{f}}^\top + \widehat{\mathbf{P}}_{f\bar{f}}\boldsymbol{\Phi}_{\bar{f}\bar{f}}^\top & \boldsymbol{\Phi}_{f\bar{f}}\widehat{\mathbf{P}}_{\bar{f}\bar{f}}\boldsymbol{\Phi}_{f\bar{f}}^\top + \boldsymbol{\Phi}_{f\bar{f}}\widehat{\mathbf{P}}_{\bar{f}f} \\ & + \widehat{\mathbf{P}}_{f\bar{f}}\boldsymbol{\Phi}_{f\bar{f}}^\top + \widehat{\mathbf{P}}_{ff} \end{bmatrix} \\
&\quad + \begin{bmatrix} \mathbf{G}_{\bar{f}\bar{f}}\mathbf{Q}_d\mathbf{G}_{\bar{f}\bar{f}}^\top & \mathbf{G}_{\bar{f}\bar{f}}\mathbf{Q}_d\mathbf{G}_{f\bar{f}}^\top \\ \mathbf{G}_{f\bar{f}}\mathbf{Q}_d\mathbf{G}_{\bar{f}\bar{f}}^\top & \mathbf{G}_{f\bar{f}}\mathbf{Q}_d\mathbf{G}_{f\bar{f}}^\top \end{bmatrix}
\end{aligned} \tag{50}
$$

where $\widehat{\mathbf{P}}_{\bar{f}\bar{f}} \in \mathbb{R}^{(21+6c)\times(21+6c)}, \widehat{\mathbf{P}}_{\bar{f}f} \in \mathbb{R}^{(21+6c)\times 3}, \widehat{\mathbf{P}}_{f\bar{f}} \in \mathbb{R}^{3\times(21+6c)}, \widehat{\mathbf{P}}_{ff} \in \mathbb{R}^{3\times 3}$ are the block components of covariance matrix $\widehat{\mathbf{P}}_i$. It can be seen that since $\boldsymbol{\Phi}_{f\bar{f}}$ and $\mathbf{G}_{f\bar{f}}$ are not zero matrices, the calculation of $\boldsymbol{\Phi}_{f\bar{f}}\widehat{\mathbf{P}}_{\bar{f}\bar{f}}\boldsymbol{\Phi}_{f\bar{f}}^\top$, $\widehat{\mathbf{P}}_{f\bar{f}}\boldsymbol{\Phi}_{\bar{f}\bar{f}}^\top$, $\boldsymbol{\Phi}_{f\bar{f}}\widehat{\mathbf{P}}_{\bar{f}f}\boldsymbol{\Phi}_{f\bar{f}}^\top$, $\mathbf{G}_{\bar{f}\bar{f}}\mathbf{Q}_d\mathbf{G}_{f\bar{f}}^\top$, $\mathbf{G}_{f\bar{f}}\mathbf{Q}_d\mathbf{G}_{\bar{f}\bar{f}}^\top$, and $\mathbf{G}_{f\bar{f}}\mathbf{Q}_d\mathbf{G}_{f\bar{f}}^\top$ is required. If there are $K$ features, computing these terms needs a computation of $O(K^2)$ flops.

## C. State Update

Since only the feature points are associated with the IMU navigation state instead of relative transformation, the feature-based state update is different from (19) while the derivation of other state update equations keeps the same. According to the frame where the feature points are observed, there are two cases to be considered.

1) *Feature points are observed by current clone pose.* By replacing the state in error function **Eq. (16)** of **Section IV-C. Feature Observation** in the manuscript with (38), the linearized observation error function is derived as:

$$
\begin{aligned}
\mathbf{z}_{f_{k_k}} &= \widetilde{\mathbf{y}}_{f_{k_k}} - \widehat{\mathbf{y}}_{f_{k_k}} \\
&= h\left[{}^C\mathbf{R}_I{}^L\mathbf{R}_{I_{k_k}}^\top({}^L\mathbf{p}_{f_k} - {}^L\mathbf{p}_{I_{k_k}}) + {}^C\mathbf{p}_I\right] - \widehat{\mathbf{y}}_{f_{k_k}} + \mathbf{n}_{f_k} \\
&\approx \mathcal{H}_{k_k}\left[{}^C\mathbf{R}_I{}^L\mathbf{R}_{I_{k_k}}^\top({}^L\mathbf{p}_{f_k} - {}^L\mathbf{p}_{I_{k_k}}) + {}^C\mathbf{p}_I\right. \\
&\quad \left. - {}^C\mathbf{R}_I{}^L\widehat{\mathbf{R}}_{I_{k_k}}^\top({}^L\widehat{\mathbf{p}}_{f_k} - {}^L\widehat{\mathbf{p}}_{I_{k_k}}) - {}^C\mathbf{p}_I\right] + \mathbf{n}_{f_k} \\
&= \mathcal{H}_{k_k}\left[{}^C\mathbf{R}_I(\text{Exp}(\boldsymbol{\xi}_{\boldsymbol{R}_{CP_{k_k}}}){}^L\widehat{\mathbf{R}}_{I_{k_k}})^\top\left[\text{Exp}(\boldsymbol{\xi}_{\boldsymbol{R}_{CP_{k_k}}}){}^L\widehat{\mathbf{p}}_{f_k} + \right.\right. \\
&\quad \left.\left. \mathbf{J}_l(\boldsymbol{\xi}_{\boldsymbol{R}_{CP_{k_k}}})\boldsymbol{\xi}_{\boldsymbol{p}_{Lf_k}} - \text{Exp}(\boldsymbol{\xi}_{\boldsymbol{R}_{CP_{k_k}}}){}^L\widehat{\mathbf{p}}_{I_{k_k}} - \mathbf{J}_l(\boldsymbol{\xi}_{\boldsymbol{R}_{CP_{k_k}}})\boldsymbol{\xi}_{\boldsymbol{p}_{CP_{k_k}}}\right]\right. \\
&\quad \left. - {}^C\mathbf{R}_I{}^L\widehat{\mathbf{R}}_{I_{k_k}}^\top({}^L\widehat{\mathbf{p}}_{f_k} - {}^L\widehat{\mathbf{p}}_{I_{k_k}})\right] + \mathbf{n}_{f_k} \\
&\approx \mathcal{H}_{k_k}\left[{}^C\mathbf{R}_I{}^L\widehat{\mathbf{R}}_{I_{k_k}}^\top(\mathbf{I}_{3\times 3} - (\boldsymbol{\xi}_{\boldsymbol{R}_{CP_{k_k}}})_\times)\left[(\mathbf{I}_{3\times 3}\right.\right. \\
&\quad \left.\left. + (\boldsymbol{\xi}_{\boldsymbol{R}_{CP_{k_k}}})_\times){}^L\widehat{\mathbf{p}}_{f_k} + \boldsymbol{\xi}_{\boldsymbol{p}_{Lf_k}} - (\mathbf{I}_{3\times 3} + (\boldsymbol{\xi}_{\boldsymbol{R}_{CP_{k_k}}})_\times){}^L\widehat{\mathbf{p}}_{I_{k_k}}\right.\right. \\
&\quad \left.\left. - \boldsymbol{\xi}_{\boldsymbol{p}_{CP_{k_k}}}\right] - {}^C\mathbf{R}_I{}^L\widehat{\mathbf{R}}_{I_{k_k}}^\top({}^L\widehat{\mathbf{p}}_{f_k} - {}^L\widehat{\mathbf{p}}_{I_{k_k}})\right] + \mathbf{n}_{f_k} \\
&\approx \mathcal{H}_{k_k}{}^C\mathbf{R}_I{}^L\widehat{\mathbf{R}}_{I_{k_k}}^\top(\boldsymbol{\xi}_{\boldsymbol{p}_{Lf_k}} - \boldsymbol{\xi}_{\boldsymbol{p}_{CP_{k_k}}}) + \mathbf{n}_{f_k}.
\end{aligned} \tag{51}
$$

2) *Feature points are observed by historical clone pose.* By replacing the state in error function **Eq. (16)** of **Section IV-C. Feature Observation** in the manuscript with (38), the linearized observation error function is derived as:

$$
\begin{aligned}
\mathbf{z}_{f_{j_k}} &= \widetilde{\mathbf{y}}_{f_{j_k}} - \widehat{\mathbf{y}}_{f_{j_k}} \\
&= h\left[{}^C\mathbf{R}_I{}^L\mathbf{R}_{I_{j_k}}^\top({}^L\mathbf{p}_{f_k} - {}^L\mathbf{p}_{I_{j_k}}) + {}^C\mathbf{p}_I\right] - \widehat{\mathbf{y}}_{f_{j_k}} + \mathbf{n}_{f_k} \\
&\approx \mathcal{H}_{j_k}\left[{}^C\mathbf{R}_I{}^L\mathbf{R}_{I_{j_k}}^\top({}^L\mathbf{p}_{f_k} - {}^L\mathbf{p}_{I_{j_k}}) + {}^C\mathbf{p}_I\right. \\
&\quad \left. - {}^C\mathbf{R}_I{}^L\widehat{\mathbf{R}}_{I_{j_k}}^\top({}^L\widehat{\mathbf{p}}_{f_k} - {}^L\widehat{\mathbf{p}}_{I_{j_k}}) - {}^C\mathbf{p}_I\right] + \mathbf{n}_{f_k} \\
&= \mathcal{H}_{j_k}\left[{}^C\mathbf{R}_I(\text{Exp}(\boldsymbol{\xi}_{\boldsymbol{R}_{CP_{j_k}}}){}^L\widehat{\mathbf{R}}_{I_{j_k}})^\top\left[\text{Exp}(\boldsymbol{\xi}_{\boldsymbol{R}_{CP_{j_k}}}){}^L\widehat{\mathbf{p}}_{f_k} + \right.\right. \\
&\quad \left.\left. \mathbf{J}_l(\boldsymbol{\xi}_{\boldsymbol{R}_{CP_{j_k}}})\boldsymbol{\xi}_{\boldsymbol{p}_{Lf_k}} - \text{Exp}(\boldsymbol{\xi}_{\boldsymbol{R}_{CP_{j_k}}}){}^L\widehat{\mathbf{p}}_{I_{j_k}} - \mathbf{J}_l(\boldsymbol{\xi}_{\boldsymbol{R}_{CP_{j_k}}})\boldsymbol{\xi}_{\boldsymbol{p}_{CP_{j_k}}}\right]\right. \\
&\quad \left. - {}^C\mathbf{R}_I{}^L\widehat{\mathbf{R}}_{I_{j_k}}^\top({}^L\widehat{\mathbf{p}}_{f_k} - {}^L\widehat{\mathbf{p}}_{I_{j_k}})\right] + \mathbf{n}_{f_k} \\
&\approx \mathcal{H}_{j_k}\left[{}^C\mathbf{R}_I{}^L\widehat{\mathbf{R}}_{I_{j_k}}^\top(\mathbf{I}_{3\times 3} - (\boldsymbol{\xi}_{\boldsymbol{R}_{CP_{j_k}}})_\times)\left[(\mathbf{I}_{3\times 3}\right.\right. \\
&\quad \left.\left. + (\boldsymbol{\xi}_{\boldsymbol{R}_{CP_{j_k}}})_\times){}^L\widehat{\mathbf{p}}_{f_k} + \boldsymbol{\xi}_{\boldsymbol{p}_{Lf_k}} - (\mathbf{I}_{3\times 3} + (\boldsymbol{\xi}_{\boldsymbol{R}_{CP_{j_k}}})_\times){}^L\widehat{\mathbf{p}}_{I_{j_k}}\right.\right. \\
&\quad \left.\left. - \boldsymbol{\xi}_{\boldsymbol{p}_{CP_{j_k}}}\right] - {}^C\mathbf{R}_I{}^L\widehat{\mathbf{R}}_{I_{j_k}}^\top({}^L\widehat{\mathbf{p}}_{f_k} - {}^L\widehat{\mathbf{p}}_{I_{j_k}})\right] + \mathbf{n}_{f_k} \\
&\approx \mathcal{H}_{j_k}{}^C\mathbf{R}_I{}^L\widehat{\mathbf{R}}_{I_{j_k}}^\top\left[({}^L\widehat{\mathbf{p}}_{f_k})_\times \boldsymbol{\xi}_{\boldsymbol{R}_{CP_{j_k}}} - ({}^L\widehat{\mathbf{p}}_{f_k})_\times \boldsymbol{\xi}_{\boldsymbol{R}_{CP_{j_k}}}\right. \\
&\quad \left. + \boldsymbol{\xi}_{\boldsymbol{p}_{CP_{j_k}}} - \boldsymbol{\xi}_{\boldsymbol{p}_{Lf_k}}\right] + \mathbf{n}_{f_k}.
\end{aligned} \tag{52}
$$

## D. Observability Analysis of MSCKF

Before deriving the observability analysis of MSC-InEKF-FC, we first introduce the observability of MSCKF. For simplicity, we do not include the biases in the state vector as they have been proved observable [3] for general motion. In addition, according to [3], both MSCKF and EKF are derived based on the same propagation model and the linearized observation error models with only the observation Jacobians are calculated using different estimates. Therefore, it is equivalent to analyze the observability of EKF, as long as we adjust the linearization points. The final considered state for MSCKF is given by $\mathbf{X}_k = (\mathbf{X}_{I_k}, \mathbf{X}_{G_k}, {}^L\mathbf{p}_{f_k})$. The observability matrix is defined as:

$$\mathcal{O} = \begin{bmatrix} \mathcal{O}_0^\top & \mathcal{O}_1^\top & \cdots & \mathcal{O}_k^\top \end{bmatrix}^\top = \begin{bmatrix} \mathbf{H}_0^\top & \boldsymbol{\Phi}_0^{1\top}\mathbf{H}_1^\top & \cdots & \boldsymbol{\Phi}_0^{k\top}\mathbf{H}_k^\top \end{bmatrix}^\top \tag{53}$$

where $\mathbf{H}_k = \begin{bmatrix} \mathbf{H}_{o_k}^\top & \mathbf{H}_{f_k}^\top & \mathbf{H}_{s_k}^\top & \mathbf{H}_{p_k}^\top \end{bmatrix}^\top \in \mathbb{R}^{9 \times 18}$. For MSCKF, the state transition matrix $\boldsymbol{\Phi}_{k-1}^k$ is formulated as:

$$\boldsymbol{\Phi}_{k-1}^k = \begin{bmatrix} \Delta\widehat{\mathbf{R}}_{k-1}^k & \mathbf{0}_{3\times3} & \mathbf{0}_{3\times3} & \mathbf{0}_{3\times3} & \mathbf{0}_{3\times3} & \mathbf{0}_{3\times3} \\ -(\Delta\widehat{\mathbf{p}}_{k-1}^k)_\times {}^L\widehat{\mathbf{R}}_{I_{k-1}} & \mathbf{I}_{3\times3} & \mathbf{I}_{3\times3}\Delta t & \mathbf{0}_{3\times3} & \mathbf{0}_{3\times3} & \mathbf{0}_{3\times3} \\ -(\Delta\widehat{\mathbf{v}}_{k-1}^k)_\times {}^L\widehat{\mathbf{R}}_{I_{k-1}} & \mathbf{0}_{3\times3} & \mathbf{I}_{3\times3} & \mathbf{0}_{3\times3} & \mathbf{0}_{3\times3} & \mathbf{0}_{3\times3} \\ \mathbf{0}_{3\times3} & \mathbf{0}_{3\times3} & \mathbf{0}_{3\times3} & \mathbf{I}_{3\times3} & \mathbf{0}_{3\times3} & \mathbf{0}_{3\times3} \\ \mathbf{0}_{3\times3} & \mathbf{0}_{3\times3} & \mathbf{0}_{3\times3} & \mathbf{0}_{3\times3} & \mathbf{I}_{3\times3} & \mathbf{0}_{3\times3} \\ \mathbf{0}_{3\times3} & \mathbf{0}_{3\times3} & \mathbf{0}_{3\times3} & \mathbf{0}_{3\times3} & \mathbf{0}_{3\times3} & \mathbf{I}_{3\times3} \end{bmatrix} \tag{54}$$

where $\Delta\widehat{\mathbf{R}}_{k-1}^k = {}^L\widehat{\mathbf{R}}_{I_k}^\top {}^L\widehat{\mathbf{R}}_{I_{k-1}}$, $\Delta\widehat{\mathbf{p}}_{k-1}^k = {}^L\widehat{\mathbf{p}}_{I_k} - {}^L\widehat{\mathbf{p}}_{I_{k-1}} - {}^L\widehat{\mathbf{v}}_{I_{k-1}}\Delta t_k + \frac{1}{2}\mathbf{g}\Delta t_k^2$, and $\Delta\widehat{\mathbf{v}}_{k-1}^k = {}^L\widehat{\mathbf{v}}_{I_k} - {}^L\widehat{\mathbf{v}}_{I_{k-1}} + \mathbf{g}\Delta t_k$. The transition matrix $\boldsymbol{\Phi}_0^k$ from time $t_0$ to $t_k$ is derived as:

$$\boldsymbol{\Phi}_0^k = \begin{bmatrix} \Delta\widehat{\mathbf{R}}_0^k & \mathbf{0}_{3\times3} & \mathbf{0}_{3\times3} & \mathbf{0}_{3\times3} & \mathbf{0}_{3\times3} & \mathbf{0}_{3\times3} \\ -(\Delta\widehat{\mathbf{p}}_0^k)_\times {}^L\widehat{\mathbf{R}}_{I_0} & \mathbf{I}_{3\times3} & \mathbf{I}_{3\times3}\Delta t_k & \mathbf{0}_{3\times3} & \mathbf{0}_{3\times3} & \mathbf{0}_{3\times3} \\ -(\Delta\widehat{\mathbf{v}}_0^k)_\times {}^L\widehat{\mathbf{R}}_{I_0} & \mathbf{0}_{3\times3} & \mathbf{I}_{3\times3} & \mathbf{0}_{3\times3} & \mathbf{0}_{3\times3} & \mathbf{0}_{3\times3} \\ \mathbf{0}_{3\times3} & \mathbf{0}_{3\times3} & \mathbf{0}_{3\times3} & \mathbf{I}_{3\times3} & \mathbf{0}_{3\times3} & \mathbf{0}_{3\times3} \\ \mathbf{0}_{3\times3} & \mathbf{0}_{3\times3} & \mathbf{0}_{3\times3} & \mathbf{0}_{3\times3} & \mathbf{I}_{3\times3} & \mathbf{0}_{3\times3} \\ \mathbf{0}_{3\times3} & \mathbf{0}_{3\times3} & \mathbf{0}_{3\times3} & \mathbf{0}_{3\times3} & \mathbf{0}_{3\times3} & \mathbf{I}_{3\times3} \end{bmatrix} \tag{55}$$

where $\Delta\widehat{\mathbf{R}}_0^k = {}^L\widehat{\mathbf{R}}_{I_k}^\top {}^L\widehat{\mathbf{R}}_{I_0}$, $\Delta\widehat{\mathbf{p}}_0^k = {}^L\widehat{\mathbf{p}}_{I_k} - {}^L\widehat{\mathbf{p}}_{I_0} - {}^L\widehat{\mathbf{v}}_{I_0}\Delta t_k + \frac{1}{2}\mathbf{g}\Delta t_k^2$, $\Delta\widehat{\mathbf{v}}_0^k = {}^L\widehat{\mathbf{v}}_{I_k} - {}^L\widehat{\mathbf{v}}_{I_0} + \mathbf{g}\Delta t_k$, and $\Delta t_k = k\Delta t$. The odometry observation Jacobian matrix $\mathbf{H}_{o_k}$ is derived as:

$$\mathbf{H}_{o_k} = \begin{bmatrix} \frac{\partial \mathbf{z}_{o_k}}{\partial \boldsymbol{\xi}_{R_{LI_k}}} & \frac{\partial \mathbf{z}_{o_k}}{\partial \boldsymbol{\xi}_{P_{LI_k}}} & \frac{\partial \mathbf{z}_{o_k}}{\partial \boldsymbol{\xi}_{v_{LI_k}}} & \frac{\partial \mathbf{z}_{o_k}}{\partial \boldsymbol{\xi}_{T_{LG_k}}} & \frac{\partial \mathbf{z}_{o_k}}{\partial \boldsymbol{\xi}_{P_{Lf_k}}} \end{bmatrix}$$
$$= {}^O\mathbf{R}_I \begin{bmatrix} ({}^L\widehat{\mathbf{R}}_{I_k}^\top {}^L\widehat{\mathbf{v}}_{I_k})_\times & \mathbf{0}_{3\times3} & {}^L\widehat{\mathbf{R}}_{I_k} & \mathbf{0}_{3\times 6} & \mathbf{0}_{3\times3} \end{bmatrix}. \tag{56}$$

The feature point observation Jacobian matrix $\mathbf{H}_{f_k}$ is:

$$\mathbf{H}_{f_k} = \begin{bmatrix} \frac{\partial \mathbf{z}_{f_k}}{\partial \boldsymbol{\xi}_{I_k}} & \frac{\partial \mathbf{z}_{f_k}}{\partial \boldsymbol{\xi}_{T_{LG_k}}} & \frac{\partial \mathbf{z}_{f_k}}{\partial \boldsymbol{\xi}_{P_{Lf_k}}} \end{bmatrix} = \begin{bmatrix} \mathbf{H}_{f_k}^{\mathbf{X}_I} & \mathbf{0}_{2\times 6} & \mathbf{H}_{f_k}^{p_{Lf}} \end{bmatrix}$$
$$\mathbf{H}_{f_k}^{\mathbf{X}_I} = \mathcal{H}_k {}^C\mathbf{R}_I \begin{bmatrix} ({}^I\widehat{\mathbf{p}}_{f_k})_\times & -({}^L\widehat{\mathbf{R}}_{I_k})_\times \end{bmatrix}$$
$$\mathbf{H}_{f_k}^{p_{Lf}} = \mathcal{H}_{j_k} {}^C\mathbf{R}_I {}^L\widehat{\mathbf{R}}_{I_{j_k}} ({}^L\widehat{\mathbf{R}}_{I_k})_\times. \tag{57}$$

The Jacobian matrix of streetlight matches $\mathbf{H}_{s_k}$ is derived as:

$$\mathbf{H}_{s_k} = \begin{bmatrix} \frac{\partial \mathbf{z}_{s_k}}{\partial \boldsymbol{\xi}_{I_k}} & \frac{\partial \mathbf{z}_{s_k}}{\partial \boldsymbol{\xi}_{T_{LG_k}}} & \frac{\partial \mathbf{z}_{s_k}}{\partial \boldsymbol{\xi}_{P_{Lf_k}}} \end{bmatrix} = \begin{bmatrix} \mathbf{H}_{s_k}^{\mathbf{X}_I} & \mathbf{H}_{s_k}^{\mathbf{X}_G} & \mathbf{0}_{2\times 3} \end{bmatrix}$$
$$\mathbf{H}_{s_k}^{\mathbf{X}_I} = \mathcal{H}_k {}^C\mathbf{R}_I \begin{bmatrix} ({}^I\widehat{\mathbf{c}}_{s_k})_\times & -{}^L\widehat{\mathbf{R}}_{I_k} & \mathbf{0}_{3\times 3} \end{bmatrix}$$
$$\mathbf{H}_{s_k}^{\mathbf{X}_G} = \begin{bmatrix} {}^L\widehat{\mathbf{R}}_{I_k}^\top {}^L\widehat{\mathbf{R}}_{G_k} ({}^G\widetilde{\mathbf{c}}_s)_\times & {}^L\widehat{\mathbf{R}}_{I_k}^\top \end{bmatrix} \tag{58}$$

where ${}^I\widehat{\mathbf{c}}_{s_k} = {}^L\widehat{\mathbf{R}}_{I_k}^\top ({}^L\widehat{\mathbf{R}}_{G_k} {}^G\widetilde{\mathbf{c}}_s + {}^L\widehat{\mathbf{p}}_{G_k} - {}^L\widehat{\mathbf{p}}_{I_k})$ is the virtual center of streetlight cluster represented in the body frame. The Jacobian matrix of prior poses $\mathbf{H}_{p_k}$ is formulated as:

$$\mathbf{H}_{p_k} = \begin{bmatrix} \frac{\partial \mathbf{z}_{p_k}}{\partial \boldsymbol{\xi}_{I_k}} & \frac{\partial \mathbf{z}_{p_k}}{\partial \boldsymbol{\xi}_{T_{LG_k}}} & \frac{\partial \mathbf{z}_{p_k}}{\partial \boldsymbol{\xi}_{P_{Lf_k}}} \end{bmatrix} = \begin{bmatrix} \mathbf{H}_{p_k}^{\mathbf{X}_I} & \mathbf{H}_{p_k}^{\mathbf{X}_G} & \mathbf{0}_{2\times 3} \end{bmatrix}$$
$$\mathbf{H}_{p_k}^{\mathbf{X}_I} = \mathbf{e}_3^\top {}^O\mathbf{R}_I {}^G\widetilde{\mathbf{R}}_{I'}^\top \begin{bmatrix} \mathbf{0}_{3\times 3} & -{}^L\widehat{\mathbf{R}}_{G_k}^\top & \mathbf{0}_{3\times 3} \\ -{}^L\widehat{\mathbf{R}}_{G_k}^\top {}^L\widehat{\mathbf{R}}_{I_k} ({}^O\mathbf{R}_I^\top \mathbf{e}_3)_\times & \mathbf{0}_{3\times 3} & \mathbf{0}_{3\times 3} \end{bmatrix}$$
$$\mathbf{H}_{p_k}^{\mathbf{X}_G} = \mathbf{e}_3^\top {}^O\mathbf{R}_I {}^G\widetilde{\mathbf{R}}_{I'}^\top \begin{bmatrix} {}^L\widehat{\mathbf{R}}_{G_k}^\top ({}^L\widehat{\mathbf{p}}_{I_k} - {}^L\widehat{\mathbf{p}}_{G_K}))_\times & -{}^L\widehat{\mathbf{R}}_{G_k}^\top \\ ({}^L\widehat{\mathbf{R}}_{G_k}^\top {}^L\widehat{\mathbf{R}}_{I_k} {}^O\mathbf{R}_I^\top \mathbf{e}_3)_\times & \mathbf{0}_{3\times 3} \end{bmatrix}. \tag{59}$$

The $k$-th block row of $\mathcal{O}$ is formulated as:

$$\mathcal{O}_k = \mathbf{H}_k \boldsymbol{\Phi}_0^k = \begin{bmatrix} \mathbf{O}_{11} & \mathbf{0}_{3\times 3} & \mathbf{O}_{13} & \mathbf{0}_{3\times 3} & \mathbf{0}_{3\times 3} & \mathbf{0}_{3\times 3} \\ \mathbf{O}_{21} & \mathbf{O}_{22} & \mathbf{O}_{23} & \mathbf{0}_{3\times 3} & \mathbf{0}_{3\times 3} & \mathbf{O}_{26} \\ \mathbf{O}_{31} & \mathbf{O}_{32} & \mathbf{O}_{33} & \mathbf{O}_{34} & \mathbf{O}_{35} & \mathbf{0}_{3\times 3} \\ \mathbf{O}_{41} & \mathbf{O}_{42} & \mathbf{O}_{43} & \mathbf{O}_{44} & \mathbf{O}_{45} & \mathbf{0}_{3\times 3} \\ \mathbf{O}_{51} & \mathbf{0}_{3\times 3} & \mathbf{0}_{3\times 3} & \mathbf{O}_{54} & \mathbf{0}_{3\times 3} & \mathbf{0}_{3\times 3} \end{bmatrix} \tag{60}$$

where the block matrices are respectively formulated as:

$$\mathbf{O}_{11} = {}^O\mathbf{R}_I {}^L\widehat{\mathbf{R}}_{I_k}^\top ({}^L\widehat{\mathbf{v}}_{I_k} - \Delta\widehat{\mathbf{v}}_0^k)_\times {}^L\widehat{\mathbf{R}}_{I_0}$$
$$\mathbf{O}_{13} = {}^O\mathbf{R}_I {}^L\widehat{\mathbf{R}}_{I_k}^\top \Delta t_k$$
$$\mathbf{O}_{21} = \mathcal{H}_k {}^C\mathbf{R}_I (({}^I\widehat{\mathbf{p}}_{f_k})_\times \Delta\widehat{\mathbf{R}}_0^k + {}^L\widehat{\mathbf{R}}_{I_k}^\top (\Delta\widehat{\mathbf{p}}_0^k)_\times {}^L\widehat{\mathbf{R}}_{I_0})$$
$$\mathbf{O}_{22} = -\mathcal{H}_k {}^C\mathbf{R}_I {}^L\widehat{\mathbf{R}}_{I_k}^\top$$
$$\mathbf{O}_{23} = -\mathcal{H}_k {}^C\mathbf{R}_I {}^L\widehat{\mathbf{R}}_{I_k}^\top \Delta t_k$$
$$\mathbf{O}_{26} = \mathcal{H}_k {}^C\mathbf{R}_I {}^L\widehat{\mathbf{R}}_{I_k}^\top$$
$$\mathbf{O}_{31} = \mathcal{H}_k {}^C\mathbf{R}_I (({}^I\widehat{\mathbf{c}}_{s_k})_\times \Delta\widehat{\mathbf{R}}_0^k + {}^L\widehat{\mathbf{R}}_{I_k}^\top (\Delta\widehat{\mathbf{p}}_0^k)_\times {}^L\widehat{\mathbf{R}}_{I_0})$$
$$\mathbf{O}_{32} = -\mathcal{H}_k {}^C\mathbf{R}_I {}^L\widehat{\mathbf{R}}_{I_k}^\top$$
$$\mathbf{O}_{33} = -\mathcal{H}_k {}^C\mathbf{R}_I {}^L\widehat{\mathbf{R}}_{I_k}^\top \Delta t_k$$
$$\mathbf{O}_{34} = -\mathcal{H}_k {}^C\mathbf{R}_I {}^L\widehat{\mathbf{R}}_{I_k}^\top {}^L\widehat{\mathbf{R}}_{G_k} ({}^G\widetilde{\mathbf{c}}_s)_\times$$
$$\mathbf{O}_{35} = \mathcal{H}_k {}^C\mathbf{R}_I {}^L\widehat{\mathbf{R}}_{I_k}^\top$$
$$\mathbf{O}_{41} = \mathbf{e}_3^\top {}^O\mathbf{R}_I {}^G\widetilde{\mathbf{R}}_{I'}^\top {}^L\widehat{\mathbf{R}}_{G_k} (\Delta\widehat{\mathbf{p}}_0^k)_\times {}^L\widehat{\mathbf{R}}_{I_0}$$
$$\mathbf{O}_{42} = -\mathbf{e}_3^\top {}^O\mathbf{R}_I {}^G\widetilde{\mathbf{R}}_{I'}^\top {}^L\widehat{\mathbf{R}}_{G_k}$$
$$\mathbf{O}_{43} = -\mathbf{e}_3^\top {}^O\mathbf{R}_I {}^G\widetilde{\mathbf{R}}_{I'}^\top {}^L\widehat{\mathbf{R}}_{G_k} \Delta t_k$$
$$\mathbf{O}_{44} = -\mathbf{e}_3^\top {}^O\mathbf{R}_I {}^G\widetilde{\mathbf{R}}_{I'}^\top ({}^L\widehat{\mathbf{R}}_{G_k}^\top ({}^L\widehat{\mathbf{p}}_{I_k} - {}^L\widehat{\mathbf{p}}_{G_K}))_\times$$
$$\mathbf{O}_{45} = \mathbf{e}_3^\top {}^O\mathbf{R}_I {}^G\widetilde{\mathbf{R}}_{I'}^\top {}^L\widehat{\mathbf{R}}_{G_k}$$
$$\mathbf{O}_{51} = -\mathbf{e}_3^\top {}^O\mathbf{R}_I {}^G\widetilde{\mathbf{R}}_{I'}^\top {}^L\widehat{\mathbf{R}}_{G_k} {}^L\widehat{\mathbf{R}}_{I_k} ({}^O\mathbf{R}_I^\top \mathbf{e}_3)_\times$$
$$\mathbf{O}_{54} = \mathbf{e}_3^\top {}^O\mathbf{R}_I {}^G\widetilde{\mathbf{R}}_{I'}^\top ({}^L\widehat{\mathbf{R}}_{G_k}^\top {}^L\widehat{\mathbf{R}}_{I_k} {}^O\mathbf{R}_I^\top \mathbf{e}_3)_\times. \tag{61}$$

Then the right nullspace of $\mathcal{O}_k$ spans as:

$$\mathcal{N}_k = \text{span}\left(\begin{bmatrix} \mathbf{0}_{3\times 3} & {}^L\widehat{\mathbf{R}}_{I_0}^\top \mathbf{g} \\ \mathbf{I}_{3\times 3} & -({}^L\widehat{\mathbf{p}}_{I_0})_\times \mathbf{g} \\ \mathbf{0}_{3\times 3} & -({}^L\widehat{\mathbf{v}}_{I_0})_\times \mathbf{g} \\ \mathbf{0}_{3\times 3} & {}^L\widehat{\mathbf{R}}_{G_k}^\top \mathbf{g} \\ \mathbf{I}_{3\times 3} & -({}^L\widehat{\mathbf{p}}_{G_k})_\times \mathbf{g} \\ \mathbf{I}_{3\times 3} & -({}^L\widehat{\mathbf{p}}_{f_k})_\times \mathbf{g} \end{bmatrix}\right). \tag{62}$$

Ideally, if all Jacobians are calculated using the true states [3], $\mathcal{N}_k$ is constant regardless of the value of $k$. Hence the observability matrix has a null space of dimension four, indicating the system's observability is not disrupted after

linearization. However, the Jacobians are always evaluated using current state estimates. Consequently, $\mathcal{N}_k$ varies with $k$ due to the different values of $^L\widehat{\mathbf{R}}_{G_k}$, $^L\widehat{\mathbf{p}}_{G_k}$, $^L\widehat{\mathbf{p}}_{f_k}$, leading to the null space of $\mathcal{O}$ collapsing to three dimensions. In other words, one dimension of the unobservable space becomes observable, leading to the inconsistency problem.

### E. Observability Analysis of MSC-InEKF-FC

To take the $k$-th element of $\mathcal{O}$ for analysis, $\boldsymbol{\Phi}_1^k$ is formulated as:

$$\boldsymbol{\Phi}_1^k = \left(\begin{bmatrix} \boldsymbol{\Phi}_{II} & \mathbf{0}_{9\times 9} \\ \mathbf{0}_{9\times 9} & \mathbf{I}_{9\times 9} \end{bmatrix}\right)^k \quad (63)$$

$$= \begin{bmatrix} \mathbf{I}_{3\times 3} & \mathbf{0}_{3\times 3} & \mathbf{0}_{3\times 3} & \mathbf{0}_{3\times 3} & \mathbf{0}_{3\times 3} & \mathbf{0}_{3\times 3} \\ \frac{1}{2}(\mathbf{g})_\times \Delta t_k^2 & \mathbf{I}_{3\times 3} & \mathbf{I}_{3\times 3}\Delta t_k & \mathbf{0}_{3\times 3} & \mathbf{0}_{3\times 3} & \mathbf{0}_{3\times 3} \\ (\mathbf{g})_\times \Delta t_k & \mathbf{0}_{3\times 3} & \mathbf{I}_{3\times 3} & \mathbf{0}_{3\times 3} & \mathbf{0}_{3\times 3} & \mathbf{0}_{3\times 3} \\ \mathbf{0}_{3\times 3} & \mathbf{0}_{3\times 3} & \mathbf{0}_{3\times 3} & \mathbf{I}_{3\times 3} & \mathbf{0}_{3\times 3} & \mathbf{0}_{3\times 3} \\ \mathbf{0}_{3\times 3} & \mathbf{0}_{3\times 3} & \mathbf{0}_{3\times 3} & \mathbf{0}_{3\times 3} & \mathbf{I}_{3\times 3} & \mathbf{0}_{3\times 3} \\ \mathbf{0}_{3\times 3} & \mathbf{0}_{3\times 3} & \mathbf{0}_{3\times 3} & \mathbf{0}_{3\times 3} & \mathbf{0}_{3\times 3} & \mathbf{I}_{3\times 3} \end{bmatrix}$$

where $\Delta t_k = k\Delta t$ and $\Delta t$ is the time interval between two consecutive time steps. We then obtain:

$$\mathbf{H}_k \boldsymbol{\Phi}_1^k = \begin{bmatrix} \mathbf{0}_{3\times 3} & \mathbf{0}_{3\times 3} & \mathbf{H}_{13} & \mathbf{0}_{3\times 3} & \mathbf{0}_{3\times 3} & \mathbf{0}_{3\times 3} \\ \mathbf{0}_{2\times 3} & \mathbf{H}_{22} & \mathbf{0}_{2\times 3} & \mathbf{0}_{2\times 3} & \mathbf{0}_{2\times 3} & \mathbf{H}_{26} \\ \mathbf{H}_{31} & \mathbf{H}_{32} & \mathbf{0}_{2\times 3} & \mathbf{H}_{34} & \mathbf{H}_{35} & \mathbf{0}_{2\times 3} \\ \mathbf{H}_{41} & \mathbf{H}_{42} & \mathbf{0}_{2\times 3} & \mathbf{H}_{44} & \mathbf{H}_{45} & \mathbf{0}_{2\times 3} \end{bmatrix}$$

$$\begin{bmatrix} \mathbf{I}_{3\times 3} & \mathbf{0}_{3\times 3} & \mathbf{0}_{3\times 3} & \mathbf{0}_{3\times 3} & \mathbf{0}_{3\times 3} & \mathbf{0}_{3\times 3} \\ \frac{1}{2}(\mathbf{g})_\times \Delta t_k^2 & \mathbf{I}_{3\times 3} & \mathbf{I}_{3\times 3}\Delta t_k & \mathbf{0}_{3\times 3} & \mathbf{0}_{3\times 3} & \mathbf{0}_{3\times 3} \\ (\mathbf{g})_\times \Delta t_k & \mathbf{0}_{3\times 3} & \mathbf{I}_{3\times 3} & \mathbf{0}_{3\times 3} & \mathbf{0}_{3\times 3} & \mathbf{0}_{3\times 3} \\ \mathbf{0}_{3\times 3} & \mathbf{0}_{3\times 3} & \mathbf{0}_{3\times 3} & \mathbf{I}_{3\times 3} & \mathbf{0}_{3\times 3} & \mathbf{0}_{3\times 3} \\ \mathbf{0}_{3\times 3} & \mathbf{0}_{3\times 3} & \mathbf{0}_{3\times 3} & \mathbf{0}_{3\times 3} & \mathbf{I}_{3\times 3} & \mathbf{0}_{3\times 3} \\ \mathbf{0}_{3\times 3} & \mathbf{0}_{3\times 3} & \mathbf{0}_{3\times 3} & \mathbf{0}_{3\times 3} & \mathbf{0}_{3\times 3} & \mathbf{I}_{3\times 3} \end{bmatrix}$$

$$= \begin{bmatrix} \mathbf{M}_{11} & \mathbf{0}_{3\times 3} & \mathbf{M}_{13} & \mathbf{0}_{3\times 3} & \mathbf{0}_{3\times 3} & \mathbf{0}_{3\times 3} \\ \mathbf{M}_{21} & \mathbf{M}_{22} & \mathbf{M}_{23} & \mathbf{0}_{2\times 3} & \mathbf{0}_{2\times 3} & \mathbf{M}_{26} \\ \mathbf{M}_{31} & \mathbf{M}_{32} & \mathbf{M}_{33} & \mathbf{M}_{34} & \mathbf{M}_{35} & \mathbf{0}_{2\times 3} \\ \mathbf{M}_{41} & \mathbf{M}_{42} & \mathbf{M}_{43} & \mathbf{M}_{44} & \mathbf{M}_{45} & \mathbf{0}_{2\times 3} \end{bmatrix} \quad (64)$$

where the matrix elements are respectively formulated as:

$$\mathbf{H}_{13} = {}^O\mathbf{R}_I {}^L\widehat{\mathbf{R}}_{I_k}^\top$$
$$\mathbf{H}_{22} = -\mathbf{H}_{26} = -\boldsymbol{\mathcal{H}}_k {}^C\mathbf{R}_I {}^L\widehat{\mathbf{R}}_{I_k}^\top$$
$$\mathbf{H}_{31} = -\mathbf{H}_{34} = \boldsymbol{\mathcal{H}}_k {}^C\mathbf{R}_I {}^L\widehat{\mathbf{R}}_{I_k}^\top ({}^L\widehat{\mathbf{c}}_s)_\times$$
$$\mathbf{H}_{32} = -\mathbf{H}_{35} = -\boldsymbol{\mathcal{H}}_k {}^C\mathbf{R}_I {}^L\widehat{\mathbf{R}}_{I_k}^\top$$
$$\mathbf{H}_{41} = -\mathbf{H}_{44} = \begin{bmatrix} -\mathbf{e}_3^\top {}^I\mathbf{R}_O^\top {}^G\widetilde{\mathbf{R}}_{I'}^\top {}^L\widehat{\mathbf{R}}_{G_k} ({}^L\widehat{\mathbf{p}}_{I_k})_\times \\ -\mathbf{e}_3^\top {}^I\mathbf{R}_O^\top {}^G\widetilde{\mathbf{R}}_{I'}^\top {}^L\widehat{\mathbf{R}}_{G_k}^\top ({}^L\widehat{\mathbf{R}}_{I_k} {}^O\mathbf{R}_I^\top \mathbf{e}_3)_\times \end{bmatrix}$$
$$\mathbf{H}_{42} = -\mathbf{H}_{45} = \begin{bmatrix} \mathbf{e}_3^\top {}^I\mathbf{R}_O^\top {}^G\widetilde{\mathbf{R}}_{I'}^\top {}^L\widehat{\mathbf{R}}_{G_k} \\ \mathbf{0}_{3\times 3} \end{bmatrix}$$
$$\mathbf{M}_{11} = {}^O\mathbf{R}_I {}^L\widehat{\mathbf{R}}_{I_k}^\top (\mathbf{g})_\times \Delta t_k \quad \mathbf{M}_{13} = {}^O\mathbf{R}_I {}^L\widehat{\mathbf{R}}_{I_k}^\top$$
$$\mathbf{M}_{21} = -\frac{1}{2}\boldsymbol{\mathcal{H}}_k {}^C\mathbf{R}_I {}^L\widehat{\mathbf{R}}_{I_k}^\top (\mathbf{g})_\times \Delta t_k^2 \quad \mathbf{M}_{22} = -\boldsymbol{\mathcal{H}}_k {}^C\mathbf{R}_I {}^L\widehat{\mathbf{R}}_{I_k}^\top$$
$$\mathbf{M}_{23} = -\boldsymbol{\mathcal{H}}_k {}^C\mathbf{R}_I {}^L\widehat{\mathbf{R}}_{I_k}^\top \Delta t_k \quad \mathbf{M}_{26} = \boldsymbol{\mathcal{H}}_k {}^C\mathbf{R}_I {}^L\widehat{\mathbf{R}}_{I_k}^\top$$
$$\mathbf{M}_{31} = \boldsymbol{\mathcal{H}}_k {}^C\mathbf{R}_I {}^L\widehat{\mathbf{R}}_{I_k}^\top ({}^L\widehat{\mathbf{c}}_s)_\times - \frac{1}{2}\boldsymbol{\mathcal{H}}_k {}^C\mathbf{R}_I {}^L\widehat{\mathbf{R}}_{I_k}^\top (\mathbf{g})_\times \Delta t_k^2$$
$$\mathbf{M}_{32} = -\boldsymbol{\mathcal{H}}_k {}^C\mathbf{R}_I {}^L\widehat{\mathbf{R}}_{I_k}^\top \quad \mathbf{M}_{33} = -\boldsymbol{\mathcal{H}}_k {}^C\mathbf{R}_I {}^L\widehat{\mathbf{R}}_{I_k}^\top \Delta t_k$$
$$\mathbf{M}_{34} = -\boldsymbol{\mathcal{H}}_k {}^C\mathbf{R}_I {}^L\widehat{\mathbf{R}}_{I_k}^\top ({}^L\widehat{\mathbf{c}}_s)_\times \quad \mathbf{M}_{35} = \boldsymbol{\mathcal{H}}_k {}^C\mathbf{R}_I {}^L\widehat{\mathbf{R}}_{I_k}^\top$$

$$\mathbf{M}_{41} = \mathbf{e}_3^\top {}^I\mathbf{R}_O^\top {}^G\widetilde{\mathbf{R}}_{I'}^\top \begin{bmatrix} -{}^L\widehat{\mathbf{R}}_{G_k}({}^L\widehat{\mathbf{p}}_{I_k})_\times + \frac{1}{2}{}^L\widehat{\mathbf{R}}_{G_k}(\mathbf{g})_\times \Delta t_k^2 \\ -{}^L\widehat{\mathbf{R}}_{G_k}^\top ({}^L\widehat{\mathbf{R}}_{I_k} {}^O\mathbf{R}_I^\top \mathbf{e}_3)_\times \end{bmatrix}$$

$$\mathbf{M}_{42} = \begin{bmatrix} \mathbf{e}_3^\top {}^I\mathbf{R}_O^\top {}^G\widetilde{\mathbf{R}}_{I'}^\top {}^L\widehat{\mathbf{R}}_{G_k} \\ \mathbf{0}_{3\times 3} \end{bmatrix} \quad \mathbf{M}_{43} = \begin{bmatrix} \mathbf{e}_3^\top {}^I\mathbf{R}_O^\top {}^G\widetilde{\mathbf{R}}_{I'}^\top {}^L\widehat{\mathbf{R}}_{G_k}\Delta t_k \\ \mathbf{0}_{3\times 3} \end{bmatrix}$$

$$\mathbf{M}_{44} = \mathbf{e}_3^\top {}^I\mathbf{R}_O^\top {}^G\widetilde{\mathbf{R}}_{I'}^\top \begin{bmatrix} {}^L\widehat{\mathbf{R}}_{G_k}({}^L\widehat{\mathbf{p}}_{I_k})_\times \\ {}^L\widehat{\mathbf{R}}_{G_k}^\top ({}^L\widehat{\mathbf{R}}_{I_k} {}^O\mathbf{R}_I^\top \mathbf{e}_3)_\times \end{bmatrix}$$

$$\mathbf{M}_{45} = \begin{bmatrix} -\mathbf{e}_3^\top {}^I\mathbf{R}_O^\top {}^G\widetilde{\mathbf{R}}_{I'}^\top {}^L\widehat{\mathbf{R}}_{G_k} \\ \mathbf{0}_{3\times 3} \end{bmatrix}. \quad (65)$$

The right nullspace $\mathcal{N}_{FC}$ of the observability matrix can be deduced by spanning the following unobservable subspace:

$$\mathcal{N}_{FC} = span\left(\begin{bmatrix} \mathbf{g}^\top & \mathbf{0}_{1\times 3} & \mathbf{0}_{1\times 3} & \mathbf{g}^\top & \mathbf{0}_{1\times 3} & \mathbf{0}_{1\times 3} \\ \mathbf{0}_{3\times 3} & \mathbf{I}_{3\times 3} & \mathbf{0}_{3\times 3} & \mathbf{0}_{3\times 3} & \mathbf{I}_{3\times 3} & \mathbf{I}_{3\times 3} \end{bmatrix}\right)^\top. \quad (66)$$

We can see that MSC-InEKF-FC maintains the correct dimension of the nullspace without introducing spurious gain.

## APPENDIX E
## MSC-InEKF-FDN

### A. State Representation

For MSC-InEKF-FDN, the features are not maintained in any Lie group. The system state is defined as:

$$\mathbf{X}_k = (\mathbf{X}_{I_k}, \mathbf{B}_k, \mathbf{X}_{C_k}, \mathbf{X}_{G_k}, {}^L\mathbf{p}_{f_k})$$
$$\mathbf{X}_{I_k} = \begin{bmatrix} {}^L\mathbf{R}_{I_k} & {}^L\mathbf{p}_{I_k} & {}^L\mathbf{v}_{I_k} \\ \mathbf{0}_{2\times 3} & \mathbf{I}_{2\times 2} \end{bmatrix}$$
$$\mathbf{B}_k = \begin{bmatrix} \mathbf{b}_{g_k}^\top & \mathbf{b}_{a_k}^\top \end{bmatrix}^\top$$
$$\mathbf{X}_{C_k} = (\mathbf{X}_{CP_{k_k}}, \cdots, \mathbf{X}_{CP_{k-c+1_k}}), \quad \mathbf{X}_{CP_{j_k}} = \begin{bmatrix} {}^L\mathbf{R}_{I_{j_k}} & {}^L\mathbf{p}_{I_{j_k}} \\ \mathbf{0}_{1\times 3} & 1 \end{bmatrix}$$
$$\mathbf{X}_{G_k} = \begin{bmatrix} {}^L\mathbf{R}_{G_k} & {}^L\mathbf{p}_{G_k} \\ \mathbf{0}_{1\times 3} & 1 \end{bmatrix}. \quad (67)$$

We keep the error states in the same formula with that in MSC-InEKF-FDRC, i.e., $\boldsymbol{\xi}_k = \begin{bmatrix} \boldsymbol{\xi}_{I_k}^\top & \boldsymbol{\zeta}_{B_k}^\top & \boldsymbol{\xi}_{C_k}^\top & \boldsymbol{\xi}_{T_{LG_k}}^\top & \boldsymbol{\xi}_{\boldsymbol{p}_{Lf_k}}^\top \end{bmatrix}^\top$. The feature is decoupled from the Lie group and we have:

$$^L\mathbf{p}_{f_k} = \widehat{\mathbf{p}}_{f_k} + \boldsymbol{\xi}_{\boldsymbol{p}_{Lf_k}}. \quad (68)$$

### B. State Propagation

The derivation of state propagation is same as that in Section. A.

### C. State Update

Since only the feature points are not maintained in Lie group, the feature-based state update is different from (19) while the derivation of other state update equations keeps the same. By replacing the state in error function **Eq. (16)** of **Section IV-C. Feature Observation** in the manuscript with (68), the linearized observation error function is derived as:

$$\mathbf{z}_{f_{j_k}} = \widetilde{\mathbf{y}}_{f_{j_k}} - \widehat{\mathbf{y}}_{f_{j_k}}$$
$$= h\left[{}^C\mathbf{R}_I {}^L\mathbf{R}_{I_{j_k}}^\top ({}^L\mathbf{p}_{f_k} - {}^L\mathbf{p}_{I_{j_k}}) + {}^C\mathbf{p}_I\right] - \widehat{\mathbf{y}}_{f_{j_k}} + \mathbf{n}_{f_k}$$
$$\approx \boldsymbol{\mathcal{H}}_{j_k}\left[{}^C\mathbf{R}_I {}^L\mathbf{R}_{I_{j_k}}^\top ({}^L\mathbf{p}_{f_k} - {}^L\mathbf{p}_{I_{j_k}}) + {}^C\mathbf{p}_I\right.$$

$$\begin{aligned}
&-{}^C\mathbf{R}_I{}^L\widehat{\mathbf{R}}_{I_{j_k}}^\top({}^L\widehat{\mathbf{p}}_{f_k} - {}^L\widehat{\mathbf{p}}_{I_{j_k}}) - {}^C\mathbf{p}_I \Big] + \mathbf{n}_{f_k} \\
&= \boldsymbol{\mathcal{H}}_{j_k} \Big[{}^C\mathbf{R}_I (\text{Exp}(\boldsymbol{\xi}_{\boldsymbol{R}_{LI_k}}){}^L\widehat{\mathbf{R}}_{I_{j_k}})^\top \big[{}^L\widehat{\mathbf{p}}_{f_k} + \boldsymbol{\xi}_{\boldsymbol{p}_{Lf}} \\
&\quad - \text{Exp}(\boldsymbol{\xi}_{\boldsymbol{R}_{LI_k}}){}^L\widehat{\mathbf{p}}_{I_{j_k}} - \mathbf{J}_l(\boldsymbol{\xi}_{\boldsymbol{R}_{LI_k}})\boldsymbol{\xi}_{\boldsymbol{p}_{LI_k}}\big] - {}^C\mathbf{R}_I{}^L\widehat{\mathbf{R}}_{I_{j_k}}^\top \\
&\quad ({}^L\widehat{\mathbf{p}}_{f_k} - {}^L\widehat{\mathbf{p}}_{I_{j_k}})\Big] + \mathbf{n}_{f_k} \\
&\approx \boldsymbol{\mathcal{H}}_{j_k} \Big[{}^C\mathbf{R}_I{}^L\widehat{\mathbf{R}}_{I_{j_k}}^\top (\mathbf{I}_{3\times 3} - (\boldsymbol{\xi}_{\boldsymbol{R}_{LI_k}})_\times) \big[{}^L\widehat{\mathbf{p}}_{f_k} + \boldsymbol{\xi}_{\boldsymbol{p}_{Lf_k}} \\
&\quad - (\mathbf{I}_{3\times 3} + (\boldsymbol{\xi}_{\boldsymbol{R}_{LI_k}})_\times){}^L\widehat{\mathbf{p}}_{I_{j_k}} - \boldsymbol{\xi}_{\boldsymbol{p}_{LI_k}}\big] - {}^C\mathbf{R}_I{}^L\widehat{\mathbf{R}}_{I_{j_k}}^\top \\
&\quad ({}^L\widehat{\mathbf{p}}_{f_k} - {}^L\widehat{\mathbf{p}}_{I_{j_k}})\Big] + \mathbf{n}_{f_k} \\
&\approx \boldsymbol{\mathcal{H}}_{j_k}{}^C\mathbf{R}_I{}^L\widehat{\mathbf{R}}_{I_{j_k}}^\top \Big[({}^L\widehat{\mathbf{p}}_{f_k})_\times \boldsymbol{\xi}_{\boldsymbol{R}_{LI_k}} - \boldsymbol{\xi}_{\boldsymbol{p}_{LI_k}} + \boldsymbol{\xi}_{\boldsymbol{p}_{Lf_k}}\Big] + \mathbf{n}_{f_k}.
\end{aligned} \tag{69}$$

### D. Observability Analysis

In MSC-InEKF-FDN, the formulation of $\Phi_1^k$ is same as (63). Therefore, the $k$-th element of the observability matrix is:

$$\mathbf{H}_k\mathbf{\Phi}_1^k = \begin{bmatrix} \mathbf{0}_{3\times3} & \mathbf{0}_{3\times3} & \mathbf{H}_{13} & \mathbf{0}_{3\times3} & \mathbf{0}_{3\times3} & \mathbf{0}_{3\times3} \\ \mathbf{H}_{21} & \mathbf{H}_{22} & \mathbf{0}_{2\times3} & \mathbf{0}_{2\times3} & \mathbf{0}_{2\times3} & \mathbf{H}_{26} \\ \mathbf{H}_{31} & \mathbf{H}_{32} & \mathbf{0}_{2\times3} & \mathbf{H}_{34} & \mathbf{H}_{35} & \mathbf{0}_{2\times3} \\ \mathbf{H}_{41} & \mathbf{H}_{42} & \mathbf{0}_{2\times3} & \mathbf{H}_{44} & \mathbf{H}_{45} & \mathbf{0}_{2\times3} \end{bmatrix}$$

$$\begin{bmatrix} \mathbf{I}_{3\times3} & \mathbf{0}_{3\times3} & \mathbf{0}_{3\times3} & \mathbf{0}_{3\times3} & \mathbf{0}_{3\times3} & \mathbf{0}_{3\times3} \\ \frac{1}{2}(\mathbf{g})_\times\Delta t_k^2 & \mathbf{I}_{3\times3} & \mathbf{I}_{3\times3}\Delta t_k & \mathbf{0}_{3\times3} & \mathbf{0}_{3\times3} & \mathbf{0}_{3\times3} \\ (\mathbf{g})_\times\Delta t_k & \mathbf{0}_{3\times3} & \mathbf{I}_{3\times3} & \mathbf{0}_{3\times3} & \mathbf{0}_{3\times3} & \mathbf{0}_{3\times3} \\ \mathbf{0}_{3\times3} & \mathbf{0}_{3\times3} & \mathbf{0}_{3\times3} & \mathbf{I}_{3\times3} & \mathbf{0}_{3\times3} & \mathbf{0}_{3\times3} \\ \mathbf{0}_{3\times3} & \mathbf{0}_{3\times3} & \mathbf{0}_{3\times3} & \mathbf{0}_{3\times3} & \mathbf{I}_{3\times3} & \mathbf{0}_{3\times3} \\ \mathbf{0}_{3\times3} & \mathbf{0}_{3\times3} & \mathbf{0}_{3\times3} & \mathbf{0}_{3\times3} & \mathbf{0}_{3\times3} & \mathbf{I}_{3\times3} \end{bmatrix}$$

$$= \begin{bmatrix} \mathbf{M}_{11} & \mathbf{0}_{3\times3} & \mathbf{M}_{13} & \mathbf{0}_{3\times3} & \mathbf{0}_{3\times3} & \mathbf{0}_{3\times3} \\ \mathbf{M}_{21} & \mathbf{M}_{22} & \mathbf{M}_{23} & \mathbf{0}_{2\times3} & \mathbf{0}_{2\times3} & \mathbf{M}_{26} \\ \mathbf{M}_{31} & \mathbf{M}_{32} & \mathbf{M}_{33} & \mathbf{M}_{34} & \mathbf{M}_{35} & \mathbf{0}_{2\times3} \\ \mathbf{M}_{41} & \mathbf{M}_{42} & \mathbf{M}_{43} & \mathbf{M}_{44} & \mathbf{M}_{45} & \mathbf{0}_{2\times3} \end{bmatrix} \tag{70}$$

where $\mathbf{H}_{21} = \boldsymbol{\mathcal{H}}_k{}^C\mathbf{R}_I{}^L\widehat{\mathbf{R}}_{I_k}^\top({}^L\widehat{\mathbf{p}}_{f_k})_\times$, $\mathbf{M}_{21} = -\frac{1}{2}\boldsymbol{\mathcal{H}}_k{}^C\mathbf{R}_I{}^L\widehat{\mathbf{R}}_{I_k}^\top(\mathbf{g})_\times\Delta t_k^2 + \boldsymbol{\mathcal{H}}_k{}^C\mathbf{R}_I{}^L\widehat{\mathbf{R}}_{I_k}^\top({}^L\widehat{\mathbf{p}}_{f_k})_\times$ and other elements keep the same formulation as (65). The right nullspace $\mathcal{N}_N$ of the observability matrix can be deduced by spanning the following unobservable subspace:

$$\mathcal{N}_N = \begin{bmatrix} \mathbf{g}^\top & \mathbf{0}_{1\times3} & \mathbf{0}_{1\times3} & \mathbf{0}_{1\times3} & \mathbf{0}_{1\times3} & \mathbf{g}^\top({}^L\widehat{\mathbf{p}}_{f_k})_\times \\ \mathbf{0}_{3\times3} & \mathbf{I}_{3\times3} & \mathbf{0}_{3\times3} & \mathbf{0}_{3\times3} & \mathbf{I}_{3\times3} & \mathbf{I}_{3\times3} \end{bmatrix}^\top. \tag{71}$$

## APPENDIX F
## MSC-InEKF-FDR

### A. State Representation

The state definition in MSC-InEKF-FDR is formulated as:

$$\mathbf{X}_k = (\mathbf{X}_{\mathbf{L}_k}, \mathbf{X}_{\mathbf{G}_k})$$

$$\mathbf{X}_{\mathbf{L}_k} = (\mathbf{X}_{\mathbf{I}_k}, \mathbf{B}_k, \mathbf{X}_{\mathbf{C}_k})$$

$$\mathbf{X}_{\mathbf{I}_k} = \begin{bmatrix} {}^L\mathbf{R}_{I_k} & {}^L\mathbf{p}_{I_k} & {}^L\mathbf{v}_{I_k} \\ \hline \mathbf{0}_{2\times3} & \mathbf{I}_{2\times2} \end{bmatrix} \qquad \mathbf{B}_k = \begin{bmatrix} \mathbf{b}_{g_k}^\top & \mathbf{b}_{a_k}^\top \end{bmatrix}^\top$$

$$\mathbf{X}_{\mathbf{G}_k} = \begin{bmatrix} {}^L\mathbf{R}_{G_k} & {}^L\mathbf{p}_{G_k} & {}^L\mathbf{p}_{f_k} \\ \hline \mathbf{0}_{2\times3} & \mathbf{I}_{2\times2} \end{bmatrix}$$

$$\mathbf{X}_{\mathbf{C}_k} = (\mathbf{X}_{CP_{k_k}}, \cdots, \mathbf{X}_{CP_{k-c+1_k}}), \ \mathbf{X}_{CP_{j_k}} = \begin{bmatrix} {}^L\mathbf{R}_{I_{j_k}} & {}^L\mathbf{p}_{I_{j_k}} \\ \mathbf{0}_{1\times3} & 1 \end{bmatrix}.$$

The error state keeps the same formulation as **Eq. (5)** of **Section III-B. Preliminary and State Definition** in the manuscript.

### B. State Propagation

The state propagation is same as that in Section A.

### C. State Update

All the derivation of state update equations keep the same as that in Section B.

### D. Observability Analysis

In MSC-InEKF-FDN, the formulation of $\Phi_1^k$ is same as (63). Therefore, the $k$-th element of the observability matrix is:

$$\mathbf{H}_k\mathbf{\Phi}_1^k = \begin{bmatrix} \mathbf{0}_{3\times3} & \mathbf{0}_{3\times3} & \mathbf{H}_{13} & \mathbf{0}_{3\times3} & \mathbf{0}_{3\times3} & \mathbf{0}_{3\times3} \\ \mathbf{H}_{21} & \mathbf{H}_{22} & \mathbf{0}_{2\times3} & \mathbf{H}_{24} & \mathbf{0}_{2\times3} & \mathbf{H}_{26} \\ \mathbf{H}_{31} & \mathbf{H}_{32} & \mathbf{0}_{2\times3} & \mathbf{H}_{34} & \mathbf{H}_{35} & \mathbf{0}_{2\times3} \\ \mathbf{H}_{41} & \mathbf{H}_{42} & \mathbf{0}_{2\times3} & \mathbf{H}_{44} & \mathbf{H}_{45} & \mathbf{0}_{2\times3} \end{bmatrix}$$

$$\begin{bmatrix} \mathbf{I}_{3\times3} & \mathbf{0}_{3\times3} & \mathbf{0}_{3\times3} & \mathbf{0}_{3\times3} & \mathbf{0}_{3\times3} & \mathbf{0}_{3\times3} \\ \frac{1}{2}(\mathbf{g})_\times\Delta t_k^2 & \mathbf{I}_{3\times3} & \mathbf{I}_{3\times3}\Delta t_k & \mathbf{0}_{3\times3} & \mathbf{0}_{3\times3} & \mathbf{0}_{3\times3} \\ (\mathbf{g})_\times\Delta t_k & \mathbf{0}_{3\times3} & \mathbf{I}_{3\times3} & \mathbf{0}_{3\times3} & \mathbf{0}_{3\times3} & \mathbf{0}_{3\times3} \\ \mathbf{0}_{3\times3} & \mathbf{0}_{3\times3} & \mathbf{0}_{3\times3} & \mathbf{I}_{3\times3} & \mathbf{0}_{3\times3} & \mathbf{0}_{3\times3} \\ \mathbf{0}_{3\times3} & \mathbf{0}_{3\times3} & \mathbf{0}_{3\times3} & \mathbf{0}_{3\times3} & \mathbf{I}_{3\times3} & \mathbf{0}_{3\times3} \\ \mathbf{0}_{3\times3} & \mathbf{0}_{3\times3} & \mathbf{0}_{3\times3} & \mathbf{0}_{3\times3} & \mathbf{0}_{3\times3} & \mathbf{I}_{3\times3} \end{bmatrix}$$

$$= \begin{bmatrix} \mathbf{M}_{11} & \mathbf{0}_{3\times3} & \mathbf{M}_{13} & \mathbf{0}_{3\times3} & \mathbf{0}_{3\times3} & \mathbf{0}_{3\times3} \\ \mathbf{M}_{21} & \mathbf{M}_{22} & \mathbf{M}_{23} & \mathbf{M}_{24} & \mathbf{0}_{2\times3} & \mathbf{M}_{26} \\ \mathbf{M}_{31} & \mathbf{M}_{32} & \mathbf{M}_{33} & \mathbf{M}_{34} & \mathbf{M}_{35} & \mathbf{0}_{2\times3} \\ \mathbf{M}_{41} & \mathbf{M}_{42} & \mathbf{M}_{43} & \mathbf{M}_{44} & \mathbf{M}_{45} & \mathbf{0}_{2\times3} \end{bmatrix} \tag{72}$$

where $\mathbf{H}_{21} = -\mathbf{H}_{24} = \boldsymbol{\mathcal{H}}_k{}^C\mathbf{R}_I{}^L\widehat{\mathbf{R}}_{I_k}^\top({}^L\widehat{\mathbf{p}}_{f_k})_\times$, $\mathbf{M}_{21} = -\frac{1}{2}\boldsymbol{\mathcal{H}}_k{}^C\mathbf{R}_I{}^L\widehat{\mathbf{R}}_{I_k}^\top(\mathbf{g})_\times\Delta t_k^2 + \boldsymbol{\mathcal{H}}_k{}^C\mathbf{R}_I{}^L\widehat{\mathbf{R}}_{I_k}^\top({}^L\widehat{\mathbf{p}}_{f_k})_\times$, $\mathbf{M}_{24} = -\boldsymbol{\mathcal{H}}_k{}^C\mathbf{R}_I{}^L\widehat{\mathbf{R}}_{I_k}^\top({}^L\widehat{\mathbf{p}}_{f_k})_\times$ and other elements keep the same formulation as (65). The right nullspace $\mathcal{N}_{R_1}$ of the observability matrix can be deduced by spanning the following unobservable subspace:

$$\mathcal{N}_{R_1} = span\left(\begin{bmatrix} \mathbf{g}^\top & \mathbf{0}_{1\times3} & \mathbf{0}_{1\times3} & \mathbf{g}^\top & \mathbf{0}_{1\times3} & \mathbf{0}_{1\times3} \\ \mathbf{0}_{3\times3} & \mathbf{I}_{3\times3} & \mathbf{0}_{3\times3} & \mathbf{0}_{3\times3} & \mathbf{I}_{3\times3} & \mathbf{I}_{3\times3} \end{bmatrix}^\top\right). \tag{73}$$

We can see that when the streetlight-based observations are available, MSC-InEKF-FDN maintains the correct dimension of the nullspace without introducing spurious gain. When the streetlight-based observations and prior pose-based observations are not available, we have:

$$\mathbf{H}_k\mathbf{\Phi}_1^k = \begin{bmatrix} \mathbf{0}_{3\times3} & \mathbf{0}_{3\times3} & \mathbf{H}_{13} & \mathbf{0}_{3\times3} & \mathbf{0}_{3\times3} & \mathbf{0}_{3\times3} \\ \mathbf{H}_{21} & \mathbf{H}_{22} & \mathbf{0}_{2\times3} & \mathbf{H}_{24} & \mathbf{0}_{2\times3} & \mathbf{H}_{26} \end{bmatrix}$$

$$\begin{bmatrix} \mathbf{I}_{3\times3} & \mathbf{0}_{3\times3} & \mathbf{0}_{3\times3} & \mathbf{0}_{3\times3} & \mathbf{0}_{3\times3} & \mathbf{0}_{3\times3} \\ \frac{1}{2}(\mathbf{g})_\times\Delta t_k^2 & \mathbf{I}_{3\times3} & \mathbf{I}_{3\times3}\Delta t_k & \mathbf{0}_{3\times3} & \mathbf{0}_{3\times3} & \mathbf{0}_{3\times3} \\ (\mathbf{g})_\times\Delta t_k & \mathbf{0}_{3\times3} & \mathbf{I}_{3\times3} & \mathbf{0}_{3\times3} & \mathbf{0}_{3\times3} & \mathbf{0}_{3\times3} \\ \mathbf{0}_{3\times3} & \mathbf{0}_{3\times3} & \mathbf{0}_{3\times3} & \mathbf{I}_{3\times3} & \mathbf{0}_{3\times3} & \mathbf{0}_{3\times3} \\ \mathbf{0}_{3\times3} & \mathbf{0}_{3\times3} & \mathbf{0}_{3\times3} & \mathbf{0}_{3\times3} & \mathbf{I}_{3\times3} & \mathbf{0}_{3\times3} \\ \mathbf{0}_{3\times3} & \mathbf{0}_{3\times3} & \mathbf{0}_{3\times3} & \mathbf{0}_{3\times3} & \mathbf{0}_{3\times3} & \mathbf{I}_{3\times3} \end{bmatrix}$$

$$= \begin{bmatrix} \mathbf{M}_{11} & \mathbf{0}_{3\times3} & \mathbf{M}_{13} & \mathbf{0}_{3\times3} & \mathbf{0}_{3\times3} & \mathbf{0}_{3\times3} \\ \mathbf{M}_{21} & \mathbf{M}_{22} & \mathbf{M}_{23} & \mathbf{M}_{24} & \mathbf{0}_{2\times3} & \mathbf{M}_{26} \end{bmatrix}. \tag{74}$$

All the elements keep the same formulation as (72). The right nullspace $\mathcal{N}_{R_2}$ of the observability matrix can be deduced by spanning the following unobservable subspace:

$$\mathcal{N}_{R_2} = \begin{bmatrix} \mathbf{g}^\top & \mathbf{0}_{1\times 3} & \mathbf{0}_{1\times 3} & \mathbf{g}^\top & \mathbf{0}_{1\times 3} & \mathbf{0}_{1\times 3} \\ \mathbf{0}_{3\times 3} & \mathbf{I}_{3\times 3} & \mathbf{0}_{3\times 3} & \mathbf{0}_{3\times 3} & \mathbf{0}_{3\times 3} & \mathbf{I}_{3\times 3} \\ \mathbf{0}_{3\times 3} & \mathbf{0}_{3\times 3} & \mathbf{0}_{3\times 3} & {}^L\widehat{\mathbf{p}}_{f_k} & \mathbf{0}_{3\times 3} & \mathbf{0}_{3\times 3} \\ \mathbf{0}_{3\times 3} & \mathbf{0}_{3\times 3} & \mathbf{0}_{3\times 3} & \mathbf{0}_{3\times 3} & \mathbf{I}_{3\times 3} & \mathbf{0}_{3\times 3} \end{bmatrix}^\top. \quad (75)$$

The lack of streetlight observations and prior poses results in the insufficient constraints on the relative pose state, thus the dimension of unobservable subspace should be 10. However, the feature estimate ${}^L\widehat{\mathbf{p}}_{L_{f_k}}$ in $\mathcal{N}_{R_2}$ reduces the dimension to 7 and leads to the inconsistency problem.

## APPENDIX G
## MSC-InEKF-FDRC

### A. State Representation

The state definition in MSC-InEKF-FDRC is formulated as:

$$\mathbf{X}_k = (\mathbf{X}_{L_k}, \mathbf{X}_{G_k})$$
$$\mathbf{X}_{L_k} = (\mathbf{X}_{I_k}, \mathbf{B}_k, \mathbf{X}_{C_k})$$
$$\mathbf{X}_{I_k} = \begin{bmatrix} {}^L\mathbf{R}_{I_k} & {}^L\mathbf{p}_{I_k} & {}^L\mathbf{v}_{I_k} \\ \mathbf{0}_{2\times 3} & \mathbf{I}_{2\times 2} \end{bmatrix} \quad \mathbf{B}_k = \begin{bmatrix} \mathbf{b}_{g_k}^\top & \mathbf{b}_{a_k}^\top \end{bmatrix}^\top$$
$$\mathbf{X}_{C_k} = (\mathbf{X}_{CP_{k_k}}, \cdots, \mathbf{X}_{CP_{k-c+1_k}}), \mathbf{X}_{CP_{j_k}} = \begin{bmatrix} {}^L\mathbf{R}_{I_{j_k}} & {}^L\mathbf{p}_{I_{j_k}} \\ \mathbf{0}_{1\times 3} & 1 \end{bmatrix}$$
$$\mathbf{X}_{G_k} = \begin{bmatrix} {}^L\mathbf{R}_{G_k} & {}^L\mathbf{p}_{G_k} & {}^L\mathbf{p}_{f_k} \\ \mathbf{0}_{2\times 3} & \mathbf{I}_{2\times 2} \end{bmatrix}. \quad (76)$$

The error state keeps the same formulation as **Eq. (5)** of **Section III-B. Preliminary and State Definition**. When the feature is associated with the pose clone, the state definition becomes:

$$\mathbf{X}_k = (\mathbf{X}_{I_k}, \mathbf{B}_k, \mathbf{X}_{C_k}, \mathbf{X}_{G_k})$$
$$\mathbf{X}_{I_k} = \begin{bmatrix} {}^L\mathbf{R}_{I_k} & {}^L\mathbf{p}_{I_k} & {}^L\mathbf{v}_{I_k} \\ \mathbf{0}_{2\times 3} & \mathbf{I}_{2\times 2} \end{bmatrix} \quad \mathbf{B}_k = \begin{bmatrix} \mathbf{b}_{g_k}^\top & \mathbf{b}_{a_k}^\top \end{bmatrix}^\top$$
$$\mathbf{X}_{C_k} = (\mathbf{X}_{CP_{k_k}}, \cdots, \mathbf{X}_{CP_{k-c+1_k}}), \mathbf{X}_{CP_{k_k}} = \begin{bmatrix} {}^L\mathbf{R}_{I_{k_k}} & {}^L\mathbf{p}_{I_{k_k}} & {}^L\mathbf{p}_{f_k} \\ \mathbf{0}_{2\times 3} & \mathbf{I}_{2\times 2} \end{bmatrix}$$
$$\mathbf{X}_{G_k} = \begin{bmatrix} {}^L\mathbf{R}_{G_k} & {}^L\mathbf{p}_{G_k} \\ \mathbf{0}_{1\times 3} & 1 \end{bmatrix}. \quad (77)$$

And the feature error state is formulated as:

$${}^L\mathbf{p}_{f_k} = \mathrm{Exp}(\boldsymbol{\xi}_{\boldsymbol{R}_{CP_{k_k}}}){}^L\widehat{\mathbf{p}}_{f_k} + \mathbf{J}_l(\boldsymbol{\xi}_{\boldsymbol{p}_{CP_{k_k}}})\boldsymbol{\xi}_{\boldsymbol{p}_{L_{f_k}}}. \quad (78)$$

### B. State Propagation

The derivation of state propagation is same as that in Section. A regardless of whether the feature point is associated with the relative transformation or the pose clone.

### C. State Update

When the feature point is associated with the relative transformation, all the state update equations are the same as that in MSC-InEKF-FDR. When the feature point is associated with the pose clone, all the state update equations are the same as that in MSC-InEKF-FC.

### D. Observability Analysis

When the feature point is associated with the relative transformation, the unobservable space is same as in (73). When the feature point is associated with the pose clone, the unobservable space is same as in (66). In this way, the dimension of the unobservable subspace is well reserved and the computation cost is significantly reduced compared with MSC-InEKF-FC.

## APPENDIX H
## EVALUATION OF THE EFFECT OF OUTLIER LIGHT SOURCES

In our system, it is not necessary to distinguish between streetlights and other light sources. Several proposed methods can accurately and efficiently detect and exclude unstable or non-map elements of light sources, thereby mitigating their negative impact on state estimation:

**1.** As illustrated in **Section V-B. Data Association and Streetlight Observation**, the set of the streetlight clusters is expanded by $(-1)$ to indicate the absence of a matching streetlight cluster for any given detection. In other words, the set is represented as $\{\widetilde{\mathcal{L}}_1 = ({}^G\widetilde{\mathbf{c}}_{s_1}, {}^G\widetilde{\mathbf{p}}_{s_1}^1, ..., {}^G\widetilde{\mathbf{p}}_{s_1}^Q), \cdots, \widetilde{\mathcal{L}}_{m_k} = ({}^G\widetilde{\mathbf{c}}_{s_{m_k}}, {}^G\widetilde{\mathbf{p}}_{s_{m_k}}^1, ..., {}^G\widetilde{\mathbf{p}}_{s_{m_k}}^Q), (-1)\}$. The matching score of each detection with $(-1)$ is calculated by $1-\sum_{g=1}^{m_k} s_{dg}$ where $s_{dg}$ is the matching score of the detection with other streetlight clusters, which is computed using **Eq. (23)** in **Section V-B. Data Association and Streetlight Observation** in the manuscript. In this way, the non-streetlight sources such as car headlights and reflective objects can be detected during the matching process and be assigned to $(-1)$, indicating there is no correspondence of the false detections.

**2.** Another way of reducing the effect of false detections is the Chi-square test. When the system state and the observation follow the Gaussian distributions, the square of the Mahalanobis distance from the observation to the corresponding prediction should satisfy a Chi-squared distribution [4]. Therefore, before utilizing the identified streetlight matches for state update, we perform the Chi-squared test to remove the incorrect matches.

To demonstrate that our system is not affected by the outlier light sources, we select the scenarios containing outlier light sources. In addition, for adequate verification, we collect extra sequences where two high-powered flashlights held by a walking person are utilized to simulate the outlier light sources. The results are displayed in Fig. 1 and Fig. 2. It can be observed that although the outlier light sources are falsely detected, the matches of detected streetlights are still correctly identified and no correspondence of false detections is determined.

## APPENDIX I
## COMPARISON OF DIFFERENT MAP-BASED VISUAL LOCALIZATION METHODS

In this experiment, we compare Night-Voyager with several popular visual localization algorithms which leverage different map types, including methods based on feature point maps (Hloc [5]), 3D line maps (VL-Line [6]), and point cloud maps

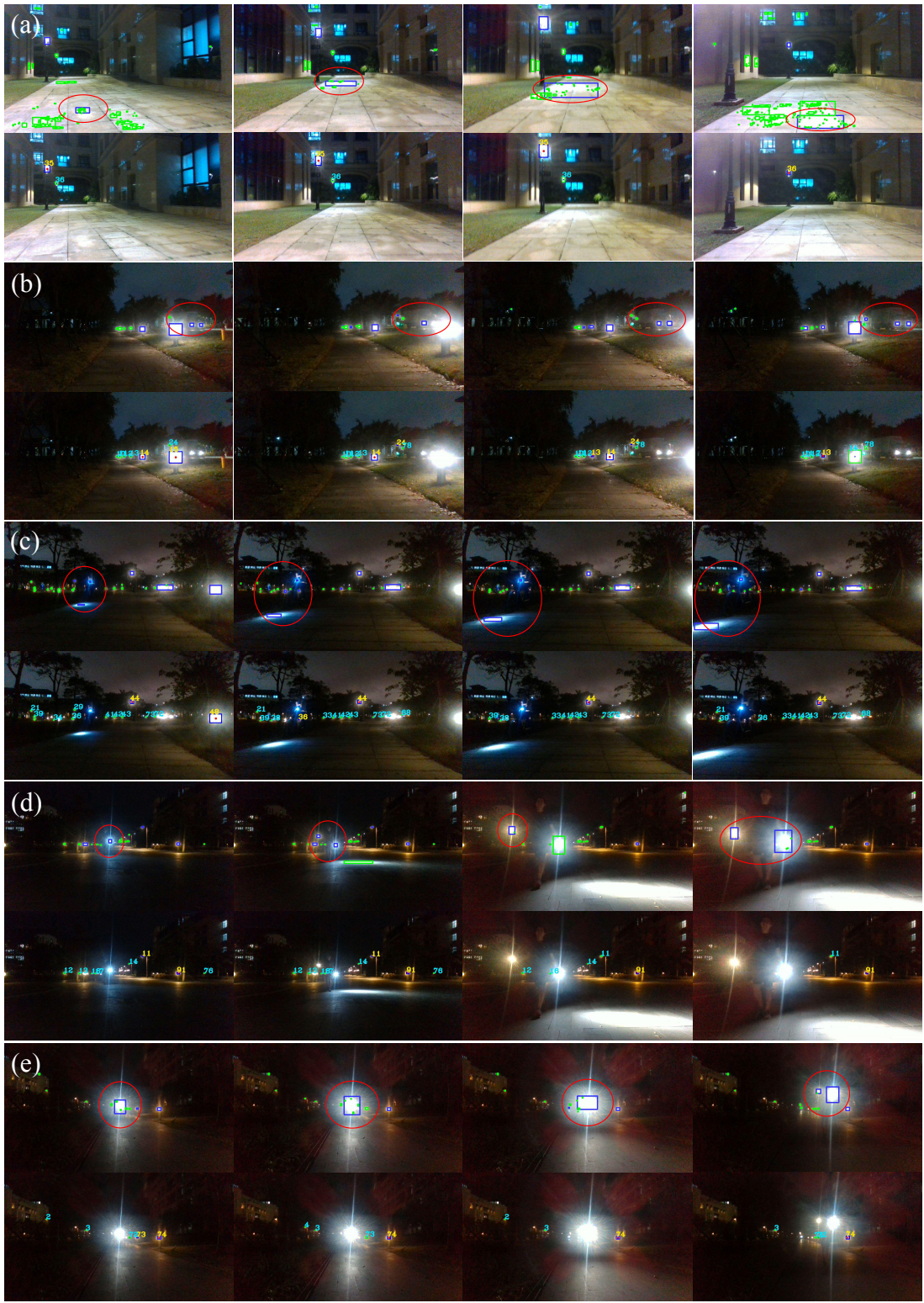

Fig. 1: The matching results using the proposed two strategies. The top figures of each scene are original streetlight detections and the bottom ones represent the final matches. Blue bounding boxes are detected using the learning-based method while the green ones are detected using the binary-based method. The numbers indicate the indices of matched streetlight clusters. For each scenario, we can see that the outliers (red circles) are falsely detected (top figures) while they can be accurately identified and exluded (bottom figures). (a) *Scene_02*. The reflective ground is falsely detected. (b) *Scene_03*. The truck lights are falsely detected. (c) *Scene_04*. A flashlight held by a walking person and the illuminated ground are falsely detected. (d) *Newly Collected Sequence 1*. The flashlights held by a walking person are falsely detected. (e) *Newly Collected Sequence 2*. The flashlights held by a walking person are falsely detected.

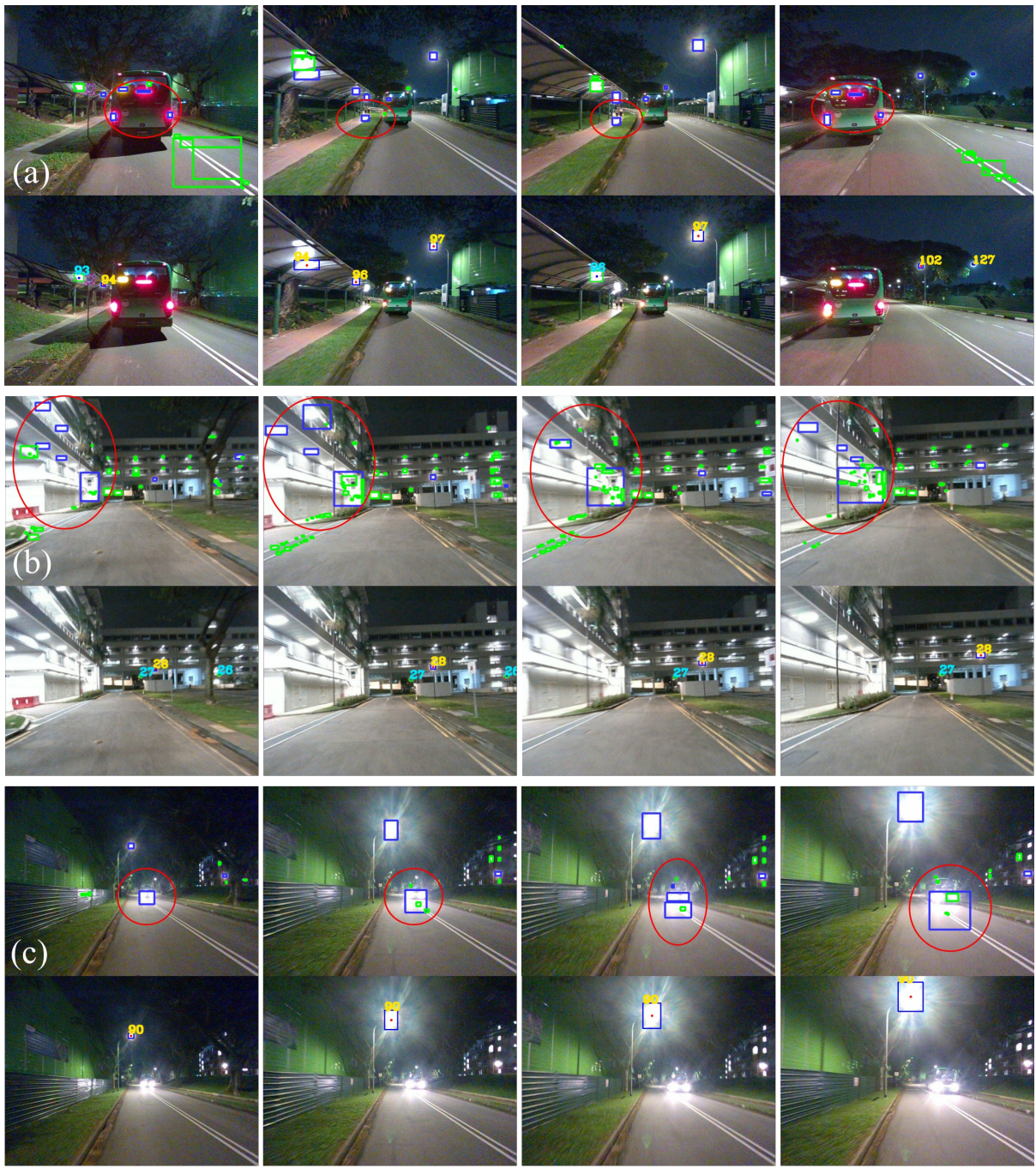

Fig. 2: The matching results in MCD dataset with outlier detections. The top figures of each scene are original streetlight detections and the bottom ones represent the final matches. Blue bounding boxes are detected using the learning-based method while the green ones are detected using the binary-based method. The numbers indicate the indices of matched streetlight clusters. For each scenario, we can see that the outliers (red circles) are falsely detected (top figures) while they can be accurately identified and exluded (bottom figures). (a) *ntu_night_04*. The tail lights of a bus and the reflective ground are falsely detected. (b) *ntu_night_08*. The lights emitted from the buildings are falsely detected. (c) *ntu_night_13*. The head lights of a car and the illuminated ground are falsely detected.

(CMRNet [7] and LHMap [8]). Details of these methods are listed as follows:

**1.** Hloc [5]. With the development of deep learning, many designed networks [9]–[13] are verified to extract more reliable features and descriptors than manually designed features. Hloc further integrates these feature extractors and matchers into a framework, achieving coarse-to-fine localization in the feature maps constructed by the Structure-from-Motion (SfM) method, COLMAP [14], [15]. The popular SuperPoint [10] and SuperGlue [9] are utilized for feature map reconstruction and visual localization. The final constructed feature maps are

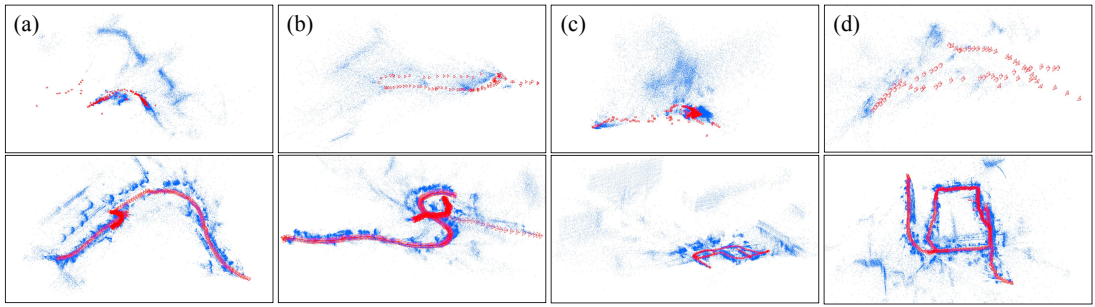

Fig. 3: The constructed feature maps using the nighttime sequences (top row) and the corresponding daytime sequences (bottom row). The sequences in the same column are collected in the identical scene. The 3D features are represented by the blue points and the trajectories are in red. (a) *Scene_02*. (b) *Scene_03*. (c) *Scene_04*. (d) *Scene_08*.

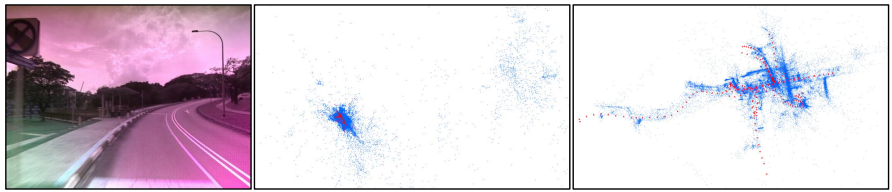

Fig. 4: Failure cases of feature map construction using the daytime or nighttime sequences from MCD dataset. From the left to right: the daytime image with purple blooming, incorrect feature map construction using the daytime sequence, incorrect feature map construction using the nighttime sequence.

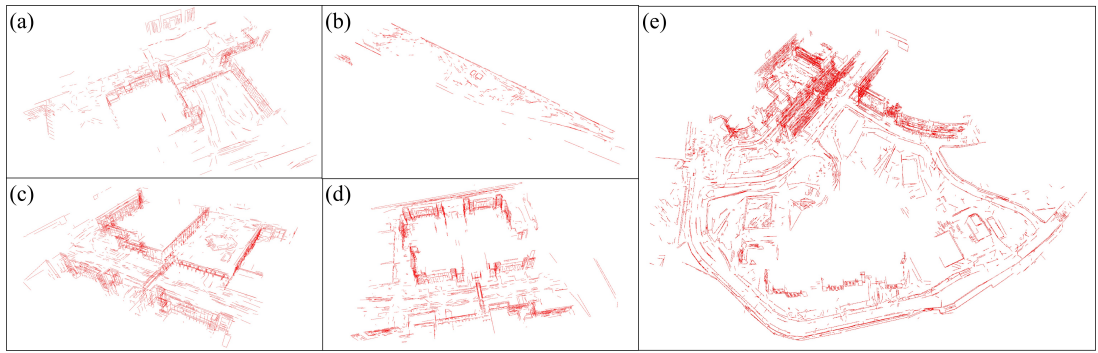

Fig. 5: The constructed line maps for VL-Line using the nighttime sequences. (a) *Scene_01*. (b) *Scene_05*. (c) *Scene_07*. (d) *Scene_09*. (e) Line map for all three sequences of MCD dataset.

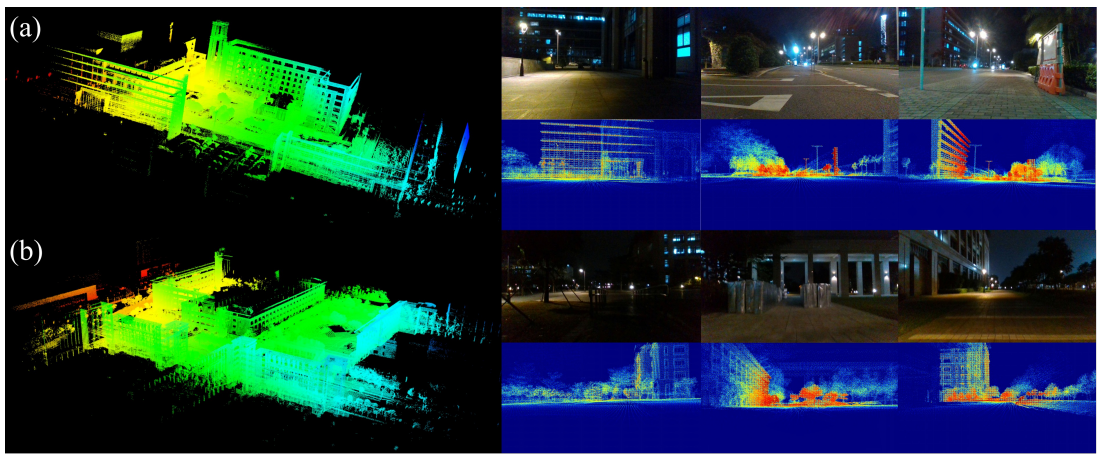

Fig. 6: The constructed point cloud maps using the nighttime sequences. The right figures display the captured images and the corresponding depth images generated by the projection of LiDAR point cloud maps. (a) *Scene_04*. (b) *Scene_07*.

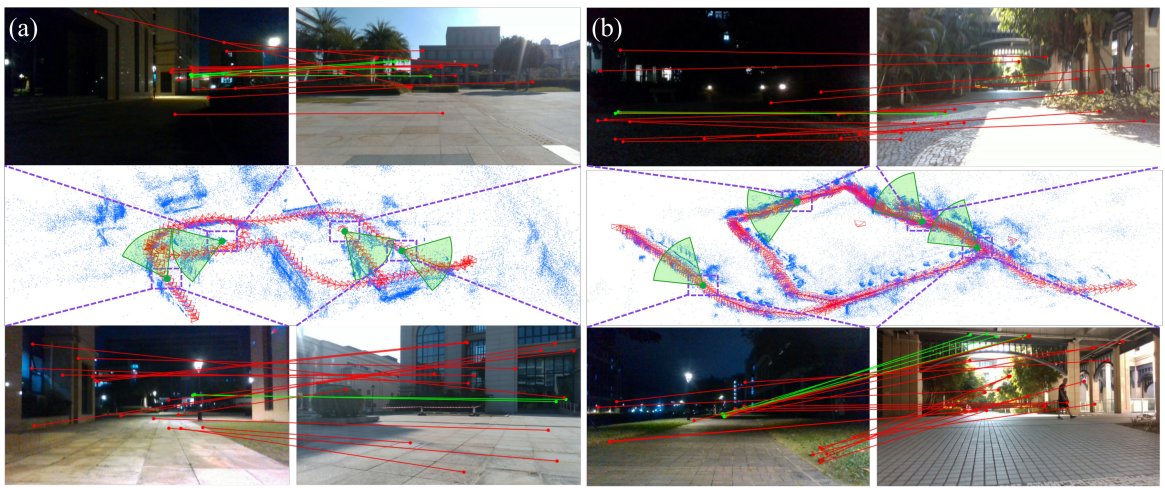

Fig. 7: Hloc failure cases. The middle images show the constructed feature maps and mapping trajectories. In each sequence, the left figures are the captured images during localization and the right ones are the queried images from the mapping sequence. It can be found that there are significantly large localization errors between the localization images and the queried images. (a) *Scene_04*. (b) *Scene_08*.

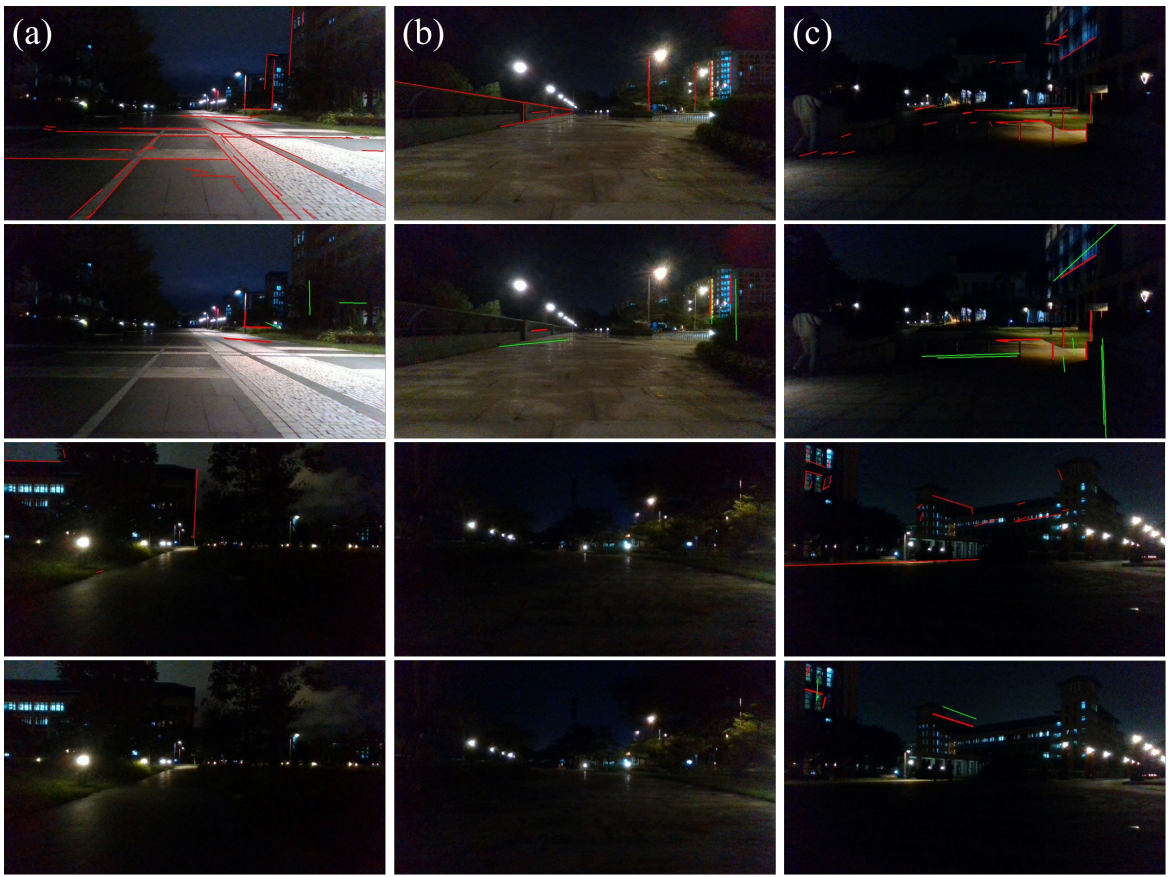

Fig. 8: VL-Line failure cases. The red lines in the images denote the detected 2D lines and the green ones represent the matched projected lines. Despite VL-Line is based on structural line features, it cannot fundamentally address the *insufficiency* issue inherent in nocturnal visual problems. (a) *Scene_01*. (b) *Scene_06*. (c) *Scene_10*.

presented in Fig. 3. Note that the constructed feature maps using the data from nighttime sequences are rambling due to the insufficient and inconsistent feature points. Therefore, we additionally collect daytime sequences for constructing consistent feature maps. As for the public MCD dataset, the constructed feature map using the daytime sequence is not satisfactory as shown in Fig. 4 because the images are affected by a purple cast. Furthermore, the feature map constructed

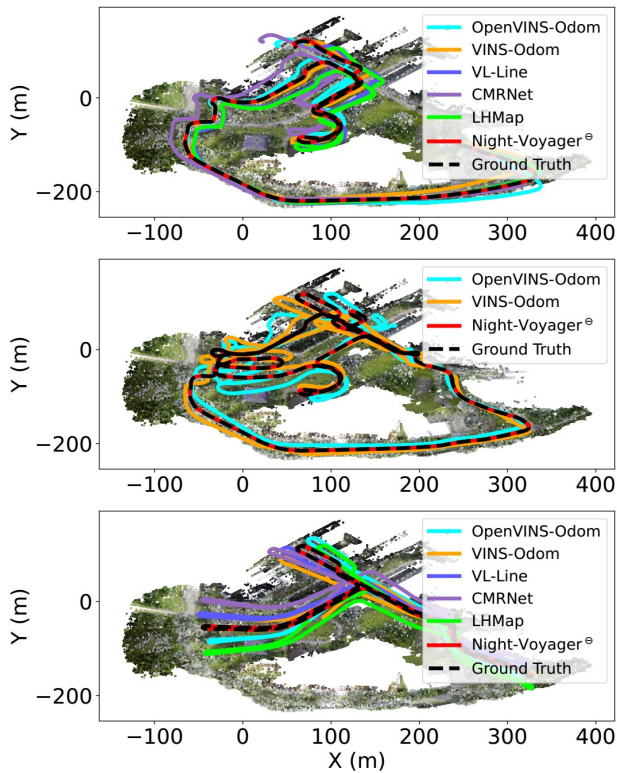

Fig. 9: Estimated trajectories of different algorithms on the prior map. Top: ntu_night_ 04. Middle: ntu_night_08. Bottom: ntu_night_13.

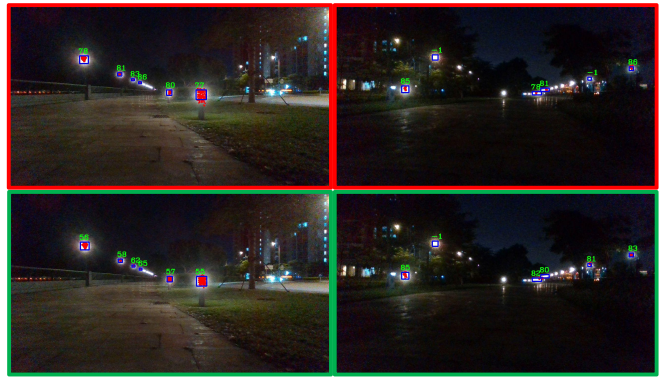

Fig. 10: Initialization cases. The images with red bounds indicate the wrong initialization. The images with green bounds represent the correct initialization after using the provided coarse position. The green numbers represent the indices of the matched streetlight clusters in the map. −1 represents no matches are identified for the detection. It is seen that the incorrect initialization caused by overlapped streetlight detections and repetitive streetlight distribution patterns (as mentioned in **Section IX-E. Evaluation of Initialization** in the manuscript) is corrected thanks to the coarse position.

from the nighttime sequence is also fragmented and chaotic due to the drastically changed illumination. Furthermore, we also try to feed the system even with the ground-truth mapping trajectory for map construction, but the performance still remains significantly poor. Consequently, the comparison for Hloc is conducted only on the collected dataset, not on the MCD dataset.

**2.** VL-Line [6]. This algorithm leverages both geometric point and line features to achieve accurate visual localization in line maps. Compared to VINS-Mono, strengthened with a loop closure detection module, it still demonstrates superior performance in the EuRoC dataset [16]. After constructing the Light Detection And Ranging (LiDAR) point cloud map, VL-Line utilizes a 3D line detection method [17] to extract the structure lines. During localization, AFM [18] is utilized for online 2D line detection. With the prior poses provided by VINS-Mono [19], line matches are identified for pose refinement. We replace the VINS-Mono with the version fused with the odometer, i.e., VINS-Odom [20], for a fair comparison. We employ the LiDAR data in the collected mapping sequences to construct the line maps. The survey-grade prior maps in MCD dataset are directly used for 3D line map construction. The final line maps are shown in Fig. 5.

**3.** CMRNet [21]. CMRNet is a popular state-of-the-art (SOTA) learning-based visual localization method that employs a point cloud map as prior. Taking a color image and a sparse depth image obtained by projecting the LiDAR map according to the coarse pose as input, the designed networks predict a refined pose for localization. For a fair comparison, we design a system which utilizes MSC-InEKF-FDRC as the estimator to fuse IMU and odometer measurements, feature points, and CMRNet pose prediction to achieve continuous and robust state estimation. We present the LiDAR point cloud maps and the projected depth images used for CMRNet in Fig. 6.

**4.** LHMap [8]. Similar to CMRNet, LHMap also takes a pair of color and depth images as the input of the neural network and predicts a refined pose. Based on optical flow estimation and spatial attention, it achieves SOTA performance in both KITTI [22] and Argoverse [23] datasets. To facilitate a fair comparison, we also design a system utilizing MSC-InEKF-FDRC to fuse the LHMap predicted poses with IMU, odometer measurements, and feature points.

To test their performance on MCD dataset, we utilize the data of *ntu_night_08* to finetune the networks (except for Hloc due to the feature map construction failure). When testing the algorithms on the collected datasets, we finetune the networks with the mapping sequences. The final localization results on the two datasets are reported in **Table IV** and **Table VI** in the manuscript, respectively. We can see that Night-Voyager significantly outperforms all the map-based localization methods. As shown in Fig. 7, the performance of Hloc is significantly affected by the severe illumination differences between daytime and nighttime images. 3D line maps have limited effectiveness in visual localization because a structured scene is required for VL-Line. However, as shown in Fig. 8, the textural lines of traffic lanes and floor tile gaps, along with spurious line detections caused by uneven lighting, influence the line association. Essentially, the poor performance of Hloc and VL-Line stems from the *insufficiency*

TABLE I: SR (%), ART (s), and AAPE (m/°) of the initialization module on different sequences.

| | | Scene_01 | Scene_02 | Scene_03 | Scene_04 | Scene_05 | Scene_06 | Scene_07 | Scene_08 | Scene_09 | Scene_10 |
|---|---|---|---|---|---|---|---|---|---|---|---|
| w/o Coarse Position | SR | 81.3 | 79.2 | 100.0 | 100.0 | 58.9 | 60.9 | 75.0 | 100.0 | 100.0 | 81.6 |
| | ART | 2.47 | 5.38 | 0.76 | 0.69 | 4.82 | 1.76 | 6.49 | 5.72 | 2.11 | 6.11 |
| | AAPE | 0.25/0.40 | 0.18/0.51 | 0.32/1.66 | 0.31/0.28 | 0.24/0.74 | 0.21/0.63 | 0.25/0.34 | 0.35/0.74 | 0.34/0.45 | 0.25/0.57 |
| w/ Coarse Position (10 m) | SR | 88.9 | 94.2 | 100.0 | 100.0 | 85.7 | 87.8 | 81.8 | 100.0 | 100.0 | 90.0 |
| | ART | 0.27 | 0.50 | 0.09 | 0.11 | 0.51 | 0.18 | 0.35 | 0.47 | 0.13 | 0.41 |
| | AAPE | 0.25/0.48 | 0.20/0.64 | 0.32/1.66 | 0.31/0.28 | 0.27/0.70 | 0.22/0.57 | 0.23/0.31 | 0.35/0.74 | 0.34/0.45 | 0.25/0.55 |
| Hloc [5] | SR | 0.00 | 0.12 | 0.00 | 0.19 | 0.00 | 0.00 | 0.00 | 0.09 | 0.00 | 0.00 |
| | ART | 21.88 | 24.77 | 19.56 | 15.21 | 26.47 | 29.02 | 25.35 | 23.26 | 29.56 | 31.18 |
| | AAPE | — | 0.29/3.33 | — | 0.45/1.69 | — | — | — | 0.36/2.13 | — | — |
| CMRNet [7] (10 m, 20°) | SR | 0.00 | 0.00 | 0.00 | 0.00 | 0.00 | 0.00 | 0.00 | 0.00 | 0.00 | 0.00 |
| | ART | 0.37 | 0.26 | 0.31 | 0.73 | 0.56 | 0.39 | 0.37 | 0.48 | 0.29 | 0.18 |
| | AAPE | — | — | — | — | — | — | — | — | — | — |
| CMRNet [7] (2 m, 10°) | SR | 0.00 | 0.02 | 0.00 | 0.00 | 0.00 | 0.00 | 0.09 | 0.00 | 0.00 | 0.05 |
| | ART | 0.35 | 0.29 | 0.26 | 0.55 | 0.47 | 0.39 | 0.33 | 0.53 | 0.32 | 0.27 |
| | AAPE | — | 0.49/6.40 | — | — | — | — | 0.35/4.17 | — | — | 0.25/6.49 |
| LHMap [8] (10 m, 20°) | SR | 0.00 | 0.00 | 0.00 | 0.00 | 0.00 | 0.00 | 0.00 | 0.00 | 0.00 | 0.00 |
| | ART | 0.38 | 0.37 | 0.36 | 0.76 | 0.52 | 0.38 | 0.37 | 0.40 | 0.35 | 0.26 |
| | AAPE | — | — | — | — | — | — | — | — | — | — |
| LHMap [8] (2 m, 10°) | SR | 0.00 | 0.03 | 0.00 | 0.00 | 0.00 | 0.00 | 0.02 | 0.04 | 0.06 | 0.05 |
| | ART | 0.27 | 0.45 | 0.32 | 0.79 | 0.52 | 0.33 | 0.30 | 0.49 | 0.50 | 0.19 |
| | AAPE | — | 0.46/4.77 | — | — | — | — | 0.31/5.95 | 0.48/4.92 | 0.41/6.05 | 0.36/3.89 |

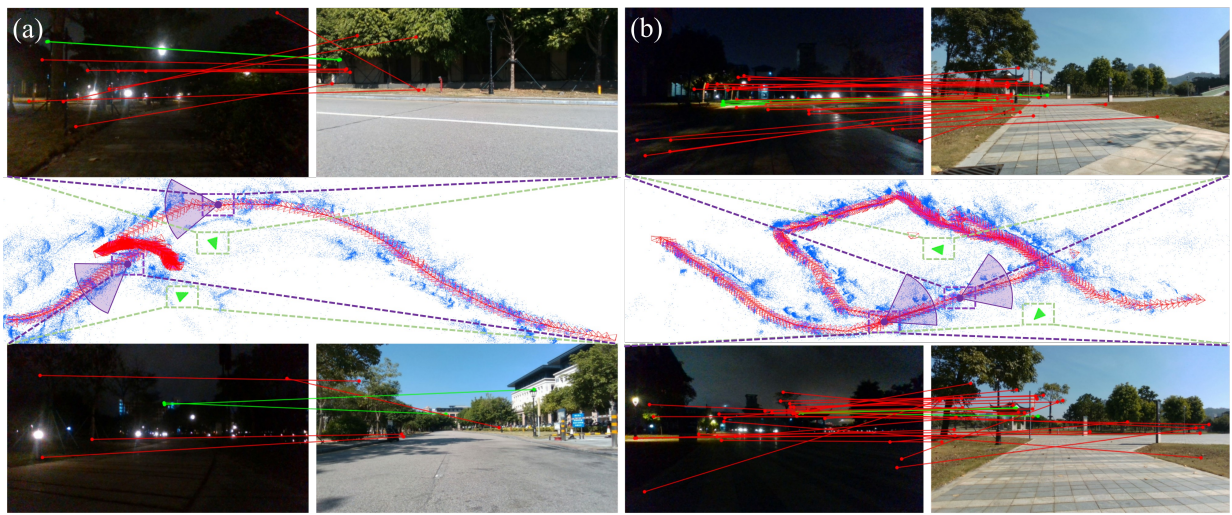

Fig. 11: Failure cases of Hloc. In each sequence, the left figures are the tested images and the right ones are the queried images from the mapping sequence. The middle figure shows the constructed feature map and mapping trajectory. The green triangulations indicate the predicted pose in the feature map while the purple circles with sectors indicate the ground truth pose. Note that all the labels denote the poses during the localization process (i.e., poses corresponding to the left figures in each scene.). The poses corresponding to the queried images of are not annotated, which is different from **Fig. 16** of **Section IX-D. Evaluation on the Collected Dataset** in the revised manuscript. For the poses of queried mapping images, **Fig. 16** in the revised manuscript provides a similar annotation style for reference, indicating large errors between the localization images and queried images. (a) *Scene_02*. (b) *Scene_08*.

and *inconsistency* issues of pixel-level features. Due to the limited information at night, the improvement of CMRNet and LHMap is much inferior to that of Night-Voyager, which demonstrates the substantial benefits brought by the object-level map and data association. We also demonstrate the trajectories generated by different algorithms in Fig. 9. The almost perfect alignment of the Night-Voyager trajectory with the ground truth proves the effectiveness of our method.

## APPENDIX J
## COMPARISON OF DIFFERENT INITIALIZATION METHODS

When the coarse pose is provided for initialization, the success rate can be significantly improved. The failure initialization caused by the overlapped streetlight detection and repetitive streetlight distribution patterns that are mentioned in **Section IX-E. Evaluation of Initialization** in the manuscript is corrected as shown in Fig. 10.

To the best of our knowledge, no prior work has explored global localization in sparse object maps. Additionally, most map-based localization methods [6], [8], [21], [24]–

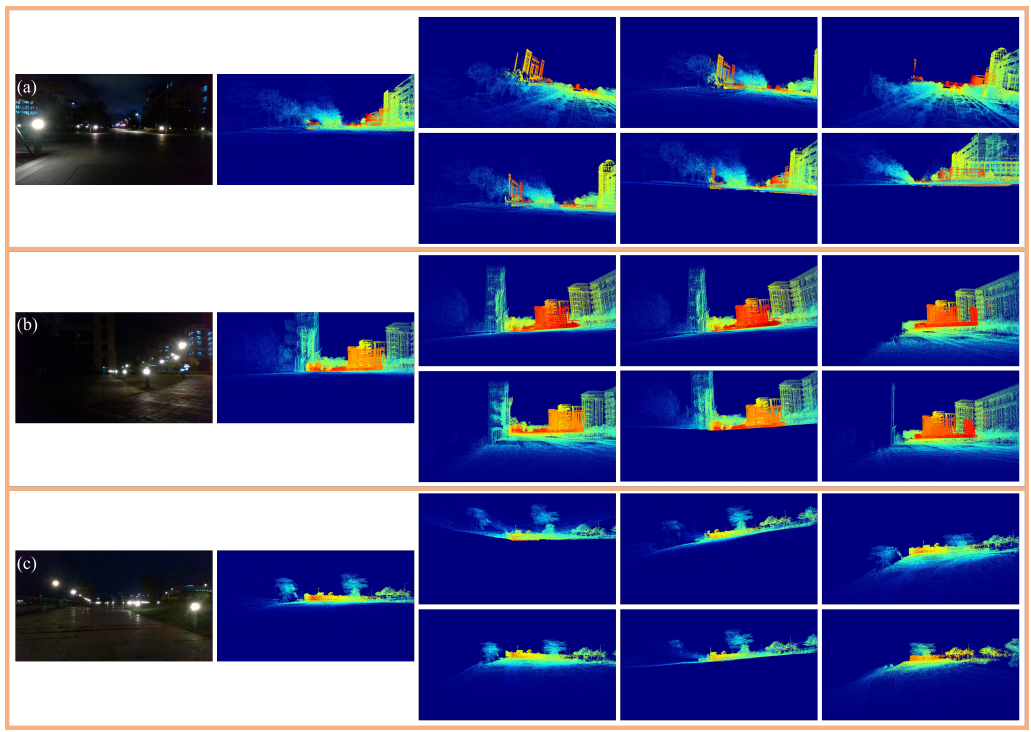

Fig. 12: Failure cases of CMRNet and LHMap. From the left to right: tested images, projections of point cloud maps using the ground truth poses, projections of point cloud maps using the perturbed poses, projections of point cloud maps using the poses refined by CMRNet, projections of point cloud maps using the poses refined by LHMap. For each scene, the top rows show results with 10 m translation and 20° rotation perturbations, while the bottom rows show results with 2 m translation and 10° rotation perturbations. (a) *Scene_01*. (b) *Scene_03*. (c) *Scene_05*

[27] assume that an accurate initial pose can be provided, not focusing on solving the global localization problem. Therefore, we select two representative kinds of methods to perform comparison experiments. The first one is Hloc [5]. We utilize finetuned SuperPoint [10] and SuperGlue [9] for feature map reconstruction and visual localization. Due to insufficient feature points at night, building a feature point map is infeasible. Therefore, we utilize the additionally collected daytime sequences for feature map construction. Additionally, we also compare our proposed initialization module with the two open-sourced approaches, CMRNet [7] and LHMap [8]. We provide the coarse pose by corrupting the ground truth with two additive Gaussian noises (10 m in the translation direction and 20° in the rotation, as well as 2 m in the translation direction and 10° in the rotation).

The results are shown in Table I. Compared with the proposed initialization module, Hloc, CMRNet, and LHMap almost fail across all sequences. Fig. 11 shows the initialization performance of Hloc. It is observed that the initialized poses are far from the true values. The essential reason lies in the poor information in nighttime images. The qualitative results of CMRNet and LHMap are presented in Fig. 12. Even the poses provided to CMRNet and LHMap are corrupted by a perturbation of 2 m translation and 10° rotation, they fail to refine the poses to the correct values. In contrast, utilizing streetlight information, our initialization module is able to calculate most initial poses correctly.

  
\end{document}